\documentclass[]{fairmeta}

\usepackage[english]{babel}

\usepackage{booktabs} %
\usepackage{nicefrac} %

\usepackage{xspace}

\usepackage{tabularx}
\usepackage{multirow}
\usepackage[normalem]{ulem}
\usepackage{float}
\usepackage[inline]{enumitem}
\usepackage{caption}
\usepackage{amsfonts, amsmath, amssymb}
\usepackage{xcolor}
\usepackage{pgfplots}
\usepackage{tikz}
\usetikzlibrary{arrows,automata,positioning}
\usepackage[inline]{enumitem}

\usepackage{pifont}

\definecolor{lightblue}{rgb}{0.88, 0.96, 1}
\definecolor{capblue}{HTML}{2962D9}
\definecolor{capyellow}{HTML}{E07200}
\definecolor{capteal}{HTML}{009B9B}
\definecolor{cappink}{HTML}{C42F8F}

\newcommand{\meanci}[2]{$#1_{\pm #2}$}

\definecolor{av_comment}{RGB}{255, 128, 0}

\usepackage[normalem]{ulem}
\usepackage{colortbl}

\newcommand{\mvga}{\OursAudio}

\newif\ifdraft  %
\drafttrue

\ifdraft

    \newcommand{\whr}[1]{\textcolor{magenta}{\sout{#1}}}
    \newcommand{\whc}[1]{\textcolor{magenta}{[WN: #1]}}
    \newcommand{\mlc}[1]{\textcolor{magenta}{[ML: #1}}
    \newcommand{\avc}[1]{\textcolor{av_comment}{[AV: #1]}}

    \newcommand{\avt}[1]{\textcolor{blue}{[AV: #1]}}
    \newcommand{\atj}[1]{{\color{violet}[AT: #1]}}
    
    \newcommand{\bsc}[1]{\textcolor{blue}{[BS: #1]}}

    \newcommand{\hleditc}[1]{\textcolor{red}{#1}}
\else

    \newcommand{\whr}[1]{}
    \newcommand{\whc}[1]{}
    \newcommand{\mlc}[1]{}
    \newcommand{\avc}[1]{}
    
    \newcommand{\avt}[1]{}
    \newcommand{\atj}[1]{}
    \newcommand{\bsc}[1]{}

    \newcommand{\hleditc}[1]{}
\fi

\newcommand{\OURS}{\textsc{Movie Gen}\xspace} %
\newcommand{\Ours}{\OURS} %
\newcommand{\OursAudio}{\textsc{Movie Gen Audio}\xspace} %
\newcommand{\OursVideo}{\textsc{Movie Gen Video}\xspace} %
\newcommand{\OursPTV}{\textsc{Personalized Movie Gen Video}\xspace} %
\newcommand{\OursVideoEdit}{\textsc{Movie Gen Edit}\xspace}
\newcommand{\OursVideoEditBench}{Movie Gen Edit Bench\xspace}
\newcommand{\textToVBenchmarkName}{Movie Gen Video Bench\xspace}
\newcommand{\textToVBenchmarkMiniName}{Movie Gen Video Bench-Mini\xspace}
\newcommand{\OursAudioBench}{Movie Gen Audio Bench\xspace}
\newcommand{\OursAudioBenchSGen}{SGen\xspace}
\newcommand{\OursAudioBenchSReal}{SReal\xspace}
\newcommand{\OursAudioBenchMGen}{MGen\xspace}
\newcommand{\CAVTPShort}{CAVTP\xspace}
\newcommand{\IB}{ImageBind\xspace}
\newcommand{\IBShort}{IB\xspace}
\newcommand{\Suno}{External API\xspace}

\definecolor{blue300}{HTML}{A4D8FF}

\newcommand{\TextToV}{Text-to-Video\xspace}
\newcommand{\textToV}{text-to-video\xspace}

\newcommand{\textToVShort}{T2V\xspace}
\newcommand{\TextToI}{Text-to-Image\xspace}
\newcommand{\textToI}{text-to-image\xspace}
\newcommand{\videoToA}{video-to-audio\xspace}
\newcommand{\textToIShort}{T2I\xspace}
\newcommand{\textToIVShort}{T2I/V\xspace}

\newcommand{\taeShort}{TAE\xspace}
\newcommand{\tae}{Temporal Autoencoder\xspace}
\newcommand{\upsampler}{Spatial Upsampler\xspace}

\newcommand{\flowmatching}{Flow Matching\xspace}
\newcommand{\timeschedule}{t-schedule\xspace}
\newcommand{\llamaNoVersion}{LLaMa\xspace}
\newcommand{\llama}{LLaMa3\xspace}

\newcommand{\llamaVideo}{LLaMa3-Video\xspace}
\newcommand{\llamaRewrite}{LLaMa3-FramesRewrite\xspace}

\newcommand{\quality}{Quality\xspace}
\newcommand{\qualityFull}{Visual quality\xspace}

\newcommand{\faithfulnessFull}{Text-alignment\xspace}
\newcommand{\qualityShort}{Q\xspace}
\newcommand{\faithfulnessShort}{A\xspace}
\newcommand{\timestep}{time-step\xspace}
\newcommand{\timesteps}{time-steps\xspace}
\newcommand{\pretraining}{pre-training\xspace}
\newcommand{\pretrained}{pre-trained\xspace}
\newcommand{\Pretraining}{Pre-training\xspace}

\newcommand{\Sora}{OpenAI Sora\xspace}
\newcommand{\LumaLabs}{LumaLabs\xspace}
\newcommand{\Kling}{Kling1.5\xspace}
\newcommand{\RunwayGen}{Runway Gen3\xspace}

\newcommand{\Flux}{Flux.1\xspace}
\newcommand{\Dalle}{OpenAI Dall-E 3\xspace}
\newcommand{\MJ}{Midjourney V6.1\xspace}
\newcommand{\Ideogram}{Ideogram V2\xspace}

\newcommand{\bigO}{$\mathcal{O}$}
\newcommand{\videoDataSize}{\bigO(100)M\xspace} %
\newcommand{\imageDataSize}{\bigO(1)B\xspace}

\newcommand{\bx}{\mathbf{X}}
\newcommand{\FMTime}{t}
\newcommand{\FMDiscreteMaxTimeSteps}{N}

\newcommand{\bv}{\mathbf{V}}

\newcommand{\bc}{\mathbf{c}}
\newcommand{\vTime}{T'}
\newcommand{\vHeight}{H'}
\newcommand{\vWidth}{W'}

\newcommand{\timeSmall}{t}
\newcommand{\heightSmall}{h}
\newcommand{\widthSmall}{w}
\newcommand{\zTime}{T}
\newcommand{\zHeight}{H}
\newcommand{\zWidth}{W}
\newcommand{\zChannels}{C}
\newcommand{\kTime}{k_{\timeSmall}}
\newcommand{\kHeight}{k_{\heightSmall}}
\newcommand{\kWidth}{k_{\widthSmall}}
\newcommand{\phiTime}{\phi_{\timeSmall}}
\newcommand{\phiHeight}{\phi_{\heightSmall}}
\newcommand{\phiWidth}{\phi_{\widthSmall}}
\newcommand{\prompt}{p} %
\newcommand{\bp}{\mathbf{P}} %

\makeatletter
\DeclareRobustCommand\onedot{\futurelet\@let@token\@onedot}
\def\@onedot{\ifx\@let@token.\else.\null\fi\xspace}

\def\eg{\emph{e.g}\onedot} 
\def\ie{\emph{i.e}\onedot} 
 
\def\etc{\emph{etc}\onedot} \def\vs{\emph{vs}\onedot}

\usepackage{amsmath}
\usepackage{amsfonts}
\usepackage{graphicx}
\usepackage{wrapfig}
\usepackage{algorithm}
\usepackage{algpseudocode}
\usepackage{algpseudocode}
\usepackage{adjustbox}
\usepackage{array}

\crefname{section}{Section}{Sections}
\Crefname{section}{Section}{Sections}
\crefformat{section}{Section~#2#1#3}
\crefformat{appendix}{Appendix~#2#1#3}
\crefformat{table}{Table~#2#1#3}
\crefformat{figure}{Figure~#2#1#3}
\crefformat{equation}{Eq~#2#1#3}

\newcolumntype{x}{>{\columncolor{Gray}}c}
\newcolumntype{H}{>{\setbox0=\hbox\bgroup}c<{\egroup}@{}}

\newcolumntype{L}[1]{>{\raggedright\let\newline\\\arraybackslash\hspace{0pt}}m{#1}}
\newcolumntype{C}[1]{>{\centering\let\newline\\\arraybackslash\hspace{0pt}}m{#1}}
\newcolumntype{R}[1]{>{\raggedleft\let\newline\\\arraybackslash\hspace{0pt}}m{#1}}

\title{\OURS: A Cast of Media Foundation Models}

\author[1]{The \OURS team @ Meta}
\affiliation[1]{A detailed contributor list can be found in the appendix of this paper.
}

\abstract{
We present \OURS, a cast of foundation models that generates high-quality, 1080p HD videos with different aspect ratios and synchronized audio.
We also show additional capabilities such as precise instruction-based video editing and  generation of personalized videos based on a user's image.
Our models set a new state-of-the-art on multiple tasks: text-to-video synthesis, video personalization, video editing, video-to-audio generation, and text-to-audio generation.
Our largest video generation model is a 30B parameter transformer trained with a maximum context length of 73K video tokens, corresponding to a generated video of 16 seconds at 16 frames-per-second.
We show multiple technical innovations and simplifications on the architecture, latent spaces, training objectives and recipes, data curation, evaluation protocols, parallelization techniques, and inference optimizations that allow us to reap the benefits of scaling \pretraining data, model size, and training compute for training large scale media generation models.
We hope this paper helps the research community to accelerate progress and innovation in media generation models.\\
All videos from this paper are available at \url{https://go.fb.me/MovieGenResearchVideos}.

}

\date{October 4, 2024}

\metadata[Blogpost]{\url{https://ai.meta.com/blog/movie-gen-media-foundation-models-generative-ai-video/}}

\begin{document}

\maketitle

\section{Introduction}

\begin{figure}
    \centering
    \setlength{\tabcolsep}{1pt}
\adjustbox{max width=0.93\textwidth}{%
\centering
\begin{tabular}{cccc}
    \rowcolor{blue300}
    \multicolumn{4}{c}{\textbf{Text-to-Video}} \\
    \multicolumn{4}{c}{\textit{Prompt}: A porcupine wearing a tutu, performing a ballet dance on a stage} \\
    \includegraphics[width=0.25\linewidth]{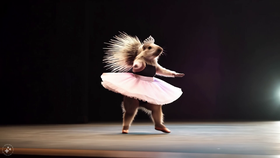} &
    \includegraphics[width=0.25\linewidth]{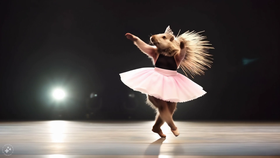} &
    \includegraphics[width=0.25\linewidth]{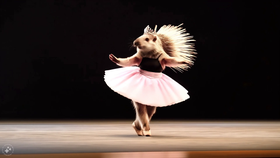} &
    \includegraphics[width=0.25\linewidth]{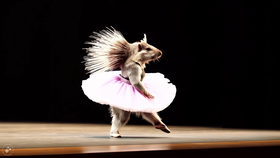} \\

    \multicolumn{4}{c}{\textit{Prompt}: Biker racing through the streets of Los Angeles. Camera tracking shot} \\
    \includegraphics[width=0.25\linewidth]{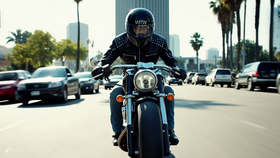} &
    \includegraphics[width=0.25\linewidth]{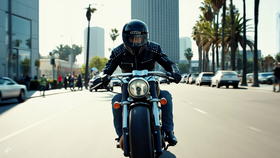} &
    \includegraphics[width=0.25\linewidth]{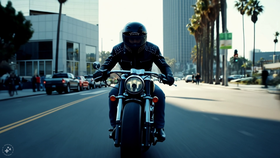} &
    \includegraphics[width=0.25\linewidth]{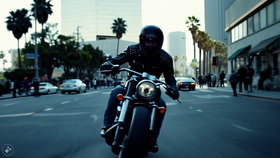} \\

    \rowcolor{blue300}
    \multicolumn{4}{c}{\textbf{Video Personalization and Consistency}} \\
    \multicolumn{1}{c}{\textbf{Reference Image}} &
    \multicolumn{3}{c}{\textit{Prompt}: A person as a scientist performing experiment with test tube} \\
    \includegraphics[width=0.14\linewidth]{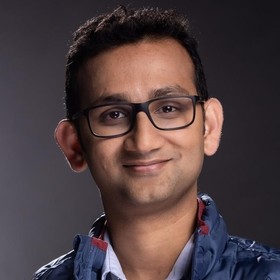} &
    \includegraphics[width=0.25\linewidth]{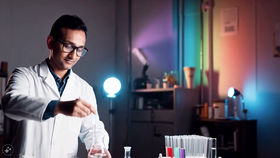} &
    \includegraphics[width=0.25\linewidth]{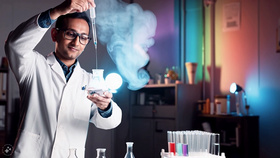} &
    \includegraphics[width=0.25\linewidth]{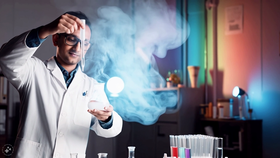} \\

    \multicolumn{1}{c}{\textbf{Reference Image}} &
    \multicolumn{3}{c}{\textit{Prompt}: A person releases a lantern into the sky} \\
    \includegraphics[width=0.14\linewidth]{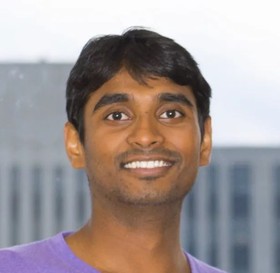} &
    \includegraphics[width=0.25\linewidth]{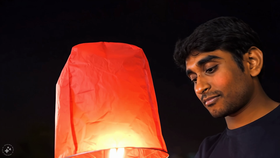} &
    \includegraphics[width=0.25\linewidth]{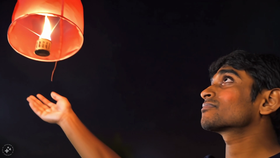} &
    \includegraphics[width=0.25\linewidth]{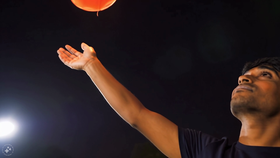} \\

    \rowcolor{blue300}
    \multicolumn{4}{c}{\textbf{Instruction-Guided Precise Video Editing}} \\
    \multicolumn{2}{c}{\textbf{Source Video}} &
    \multicolumn{2}{c}{\textit{Edit}: Add tinsel streamers to the lantern bottom} \\
    \includegraphics[width=0.25\linewidth]{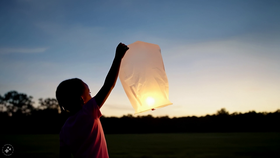} &
    \includegraphics[width=0.25\linewidth]{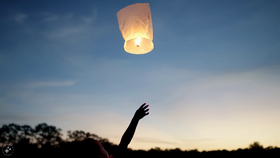} &
    \includegraphics[width=0.25\linewidth]{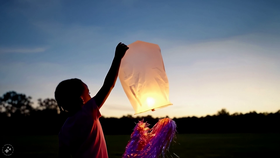} &
    \includegraphics[width=0.25\linewidth]{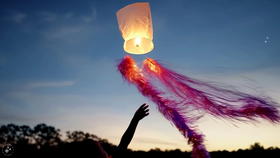} \\
    \multicolumn{2}{c}{\textit{Edit}: Transform the lantern into a soaring bubble} &
    \multicolumn{2}{c}{\textit{Edit}: Change background to a city park with a lake} \\
    \includegraphics[width=0.25\linewidth]{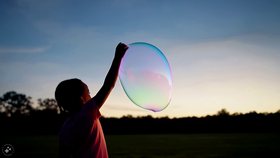} &
    \includegraphics[width=0.25\linewidth]{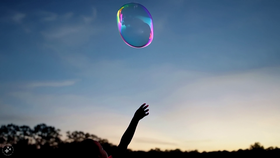} &
    \includegraphics[width=0.25\linewidth]{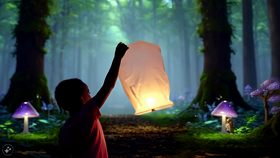} &
    \includegraphics[width=0.25\linewidth]{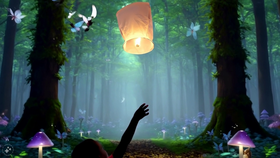} \\

    \rowcolor{blue300}
    \multicolumn{4}{c}{\textbf{Video-to-Audio}} \\
    \multicolumn{2}{c}{\multirow{2}{*}{\begin{tabular}[c]{@{}c@{}}\textit{Prompt}: splash of water and loud thud as the  \\ person hits the surface\end{tabular}}} & 
    \multicolumn{2}{c}{\multirow{2}{*}{\begin{tabular}[c]{@{}c@{}}\textit{Prompt}: thunder cracks loudly and shakes the ground  \\ and dark, intense music plays in the background\end{tabular}}} \\ \\
    \includegraphics[width=0.25\linewidth]{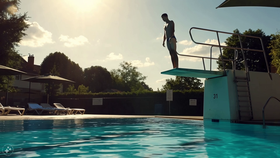} &
    \includegraphics[width=0.25\linewidth]{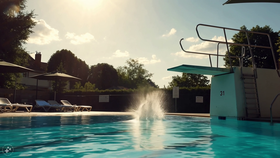} &
    \includegraphics[width=0.25\linewidth]{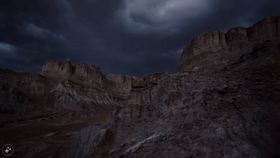} &
    \includegraphics[width=0.25\linewidth]{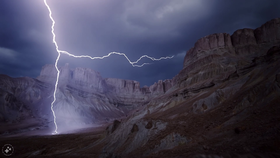} \\
    \multicolumn{4}{c}{\textit{Corresponding Spectrograms}} \\
    \multicolumn{2}{c}{\includegraphics[width=0.5\linewidth]{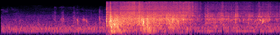}} &
    \multicolumn{2}{c}{\includegraphics[width=0.5\linewidth]{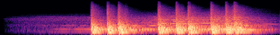}}  \\

\end{tabular}}

    \vspace{-3mm}
    \captionof{figure}{\textbf{Examples of the different capabilities of \OURS.}
    The \OURS cast of models generates videos from text prompts,
    supports the generation of videos consistent with characters in provided reference images,
    supports precise video editing given user provided instructions,
    and generates videos with synchronized audio.
    Videos in this Figure found at \url{https://go.fb.me/MovieGen-Figure1}.
    }
    \label{fig:main_figure}
\end{figure}

Imagine a blue emu swimming through the ocean.
Humans have the astonishing ability to imagine such a fictional scene in great detail.
Human imagination requires the ability to compose and predict various facets of the world.
Simply imagining a scene requires composing different concepts while predicting realistic properties about motion, scene, physics, geometry, audio \etc.
Equipping AI systems with such generative, compositional, and prediction capabilities is a core scientific challenge with broad applications.
While Large Language Models (LLMs)~\citep{llama3,llama2,brown2020language,team2023gemini} aim to learn such capabilities with a text output space, in this paper we focus on media -- image, video, audio -- as the output space.
We present \OURS, a cast of media generation foundation models.
\OURS models can natively generate high fidelity images, video, and audio while also possessing the abilities to edit and personalize the videos as we illustrate in~\cref{fig:main_figure}.

We find that scaling the training data, compute, and model parameters of a simple Transformer-based~\citep{vaswani2017attention} model trained with \flowmatching~\citep{flow-matching} yields high quality generative models for video or audio.
Our models are \pretrained on internet scale image, video, and audio data.
Our largest foundation \textToV generation model, \OursVideo, consists of 30B parameters, while our largest foundation \videoToA generation model, \OursAudio, consists of 13B parameters.
We further post-train the \OursVideo model to obtain \OursPTV that can generate personalized videos conditioned on a person's face.
Finally, we show a novel post-training procedure to produce \OursVideoEdit that can precisely edit videos.
In conjunction, these models can be used to create realistic personalized HD videos of up to 16 seconds (at 16 FPS) and 48kHz audio, and the ability to edit real or generated videos.

The \OURS cast of foundation models is state-of-the-art on multiple media generation tasks for video and audio.
On \textToV generation, we outperform prior state-of-the-art, including commercial systems such as \RunwayGen~\citep{gen3}, \LumaLabs~\citep{luma}, \Sora~\citep{sora} on overall video quality as shown in~\cref{tab:t2v_main_eval}.
Moreover, with \OursPTV and \OursVideoEdit we enable new capabilities on video personalization and precise video editing respectively, and both these capabilities are missing from current commercial systems.
On both these tasks too, we outperform all prior work (\cref{tab:pt2v_main} and~\cref{tab:eval_video_editing_tgve}).
Finally, \OursAudio, outperforms prior state-of-the-art, including commercial systems such as PikaLabs~\citep{pikalabs} and ElevenLabs~\citep{elevenlabs} for sound-effect generation (\cref{tab:eval_audio_main_ss_sfx}), music generation (\cref{tab:eval_audio_main_ms_sfx_music}), and audio extension.

To enable future benchmarking, we publicly release two benchmarks ---\textToVBenchmarkName (Section~\ref{sec:t2v_eval_benchmark}), \OursAudioBench (Section~\ref{subsubsec:audio_gen_bench}).
We also provide thorough details on model architectures, training, inference, and experimental settings which we hope will accelerate research in media generation models.

\section{Overview}

  The \OURS cast of models generates videos with synchronized audio, personalized characters, and supports video editing as illustrated in~\cref{fig:main_figure}.

  We achieve these wide capabilities using two foundation models:
  \begin{itemize}
      \item \textbf{\OursVideo}. A 30B parameter foundation model for joint text-to-image and text-to-video generation that generates high-quality HD videos of up to 16 seconds duration that follow the text prompt. The model naturally generates high-quality images and videos in multiple aspect ratios and variable resolutions and durations. The model is pre-trained jointly on \videoDataSize videos and \imageDataSize images and learns about the visual world by `watching' videos. We find that the pre-trained model can reason about object motion, subject-object interactions, geometry, camera motion, and physics, and learns plausible motions for a wide variety of concepts. %
      To improve the video generations, we perform supervised finetuning (SFT) on a small set of curated high-quality videos and text captions. We present the model architecture and training details in~\cref{sec:image_video_model}.

      \item \textbf{\OursAudio}. A 13B parameter foundation model for video- and text-to-audio generation that can generate 48kHz high-quality cinematic sound effects and music synchronized with the video input, and follow an input text prompt.
      The model naturally handles variable length audio generation and can produce long-form coherent audio for videos up to several minutes long via audio extension techniques.
      We pre-train the model on \bigO(1)M hours of audio and observe that it learns not only the physical association, but also the psychological associations between the visual and the audio world.
      The model can generate diegetic ambient sounds matching the visual scene even when the source is unseen, and also diegetic sound effects synchronized with the visual actions.
      Moreover, it can generate non-diegetic music that supports the mood and aligns with the actions of the visual scene, and blend sound effects and background music professionally.
      We further perform SFT on a small set of curated higher quality (text, audio) and (video, text, audio) data which improves the overall audio quality and aims for cinematic styles.
      The model and training recipe are outlined in~\cref{sec:audio_model}.
  \end{itemize}

  We add video personalization and video editing capabilities to our foundation \OursVideo model via post-training procedures:
  \begin{itemize}
    \item \textbf{Personalization} enables the video generation model to condition on text as well as an image of a person to generate a video featuring the chosen person. The generated  personalized video maintains the identity of the person while following the text prompt. We use a subset of videos containing humans, and automatically construct pairs of (image, text) inputs and video outputs to train the model. We outline the post training strategy for personalization in~\cref{sec:personalization}.
    \item \textbf{Precise Editing} allows users to effortlessly perform precise and imaginative edits on both real and generated videos using a textual instruction. Since large-scale supervised video editing data is harder to obtain, we show a novel approach to train such a video editing model without supervised video editing data (\cref{sec:video_editing}). We provide examples of our model's video editing capabilities in \url{https://go.fb.me/MovieGen-Figure24}.
  \end{itemize}

\section{Joint Image and Video Generation}
\label{sec:image_video_model}

\begin{figure}[!t]
  \centering
  \includegraphics[width=0.9\linewidth]{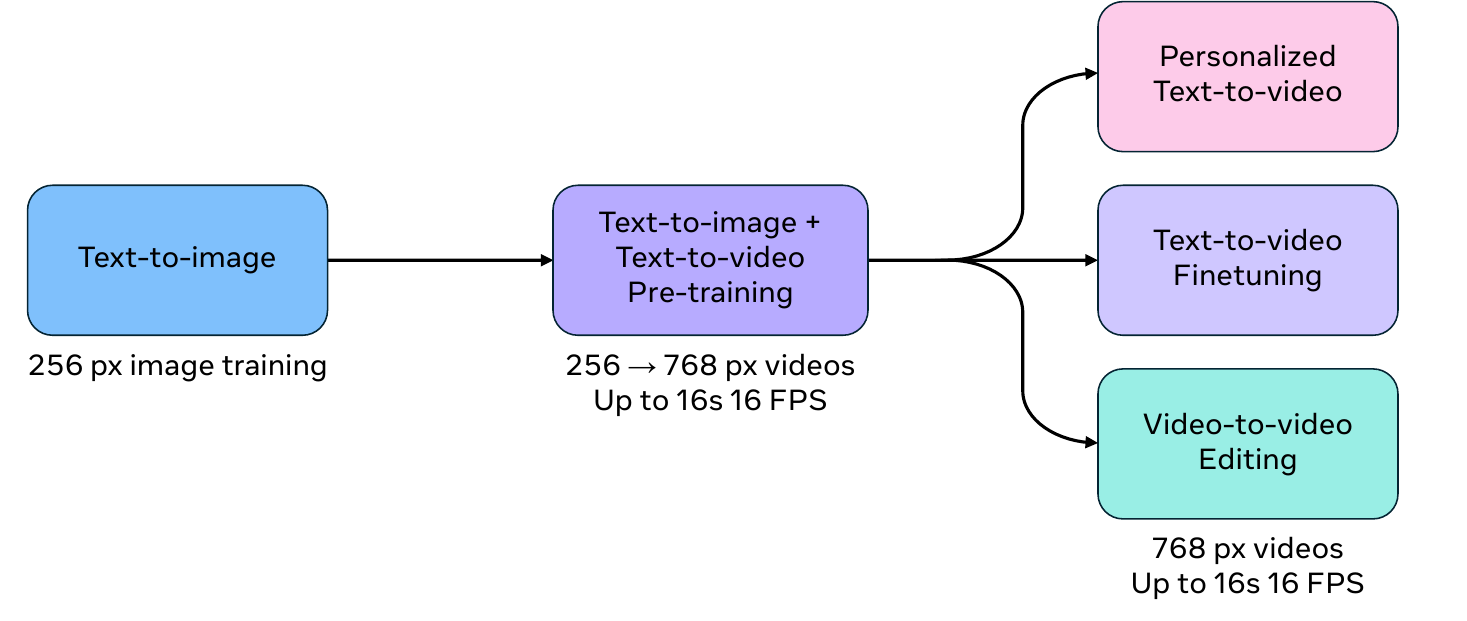}
  \caption{\textbf{Training recipe for \OursVideo.}
  We first pre-train our model for the \textToI task followed by joint \textToI and \textToV \pretraining at increasingly higher spatial resolutions~(\cref{sec:image_video_model}).
  We finetune the model on high aesthetic and motion quality videos to improve our video generations.
  We also add additional capabilities like personalization~(\cref{sec:personalization}) and video-to-video editing~(\cref{sec:video_editing}).
  }
  \label{fig:overview_t2v_recipe}
\end{figure}

We train a single joint foundation model, \OursVideo, for the text-to-image and the text-to-video tasks.
Given a text prompt as input, our foundation model generates a video consisting of multiple RGB frames as output.
We treat images as a single frame video, enabling us to use the same model to generate both images and videos.
Compared to video data, paired image-text datasets are easier to scale with diverse concepts and styles~\citep{ho2022imagen,emuvideo2023} and thus joint modeling of image and video leads to better generalization.
Our training recipe is illustrated in~\cref{fig:overview_t2v_recipe}.
We perform our training in multiple stages for training efficiency.
We first pretrain our model only on low-resolution 256 px images followed by joint \pretraining on low-resolution images and videos, and high-resolution joint training.
We finetune the model on high quality videos to improve the generations.
Additionally, we add capabilities such as personalization and editing by post-training.

\begin{figure}[t!]
  \includegraphics[width=\linewidth]{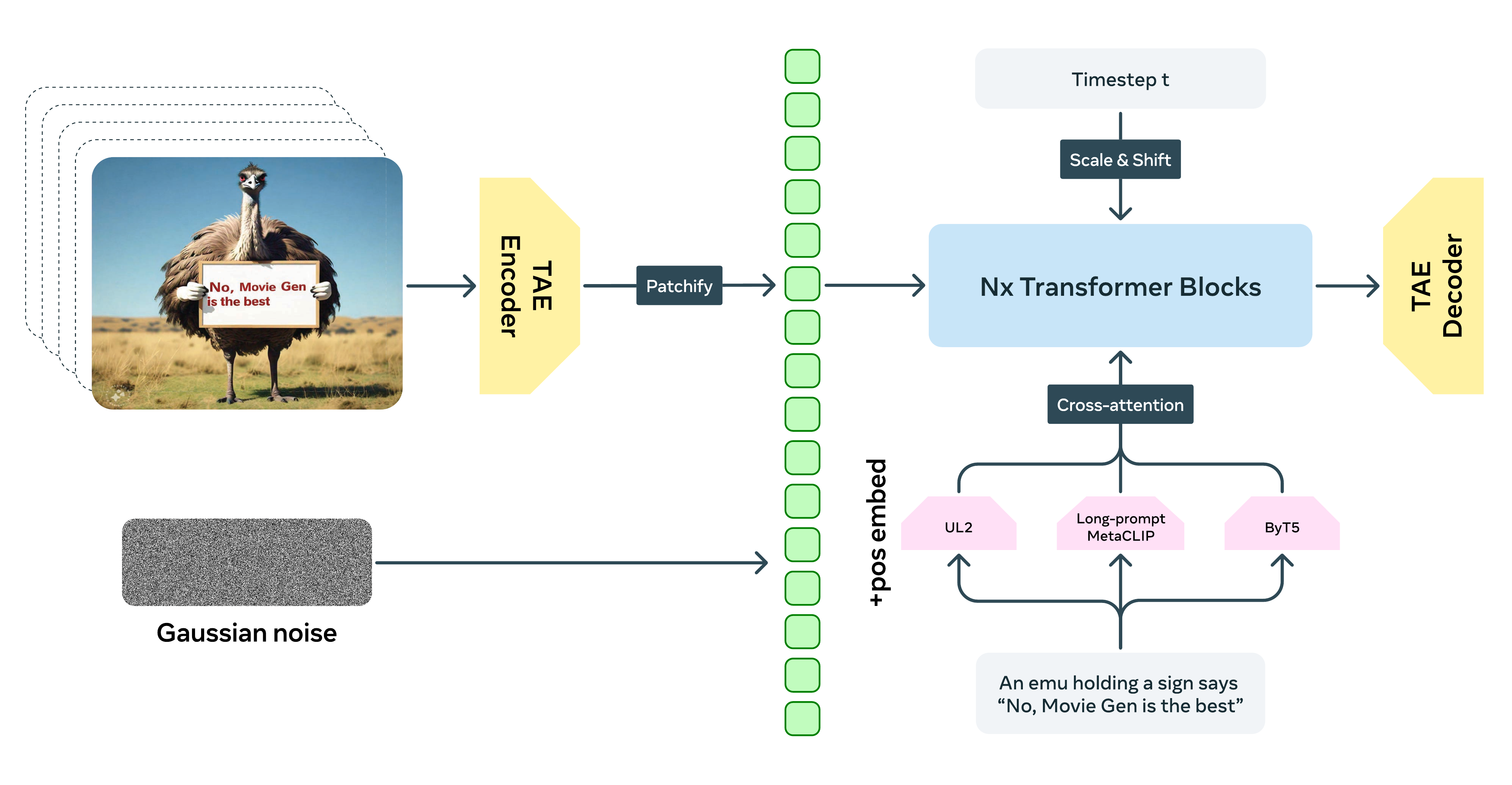}
  \vspace{-7mm}
  \caption{\textbf{Overview of the joint image and video generation pipeline.}
  We train our generative model on a spatio-temporally compressed latent space, which is learned via a temporal autoencoder model (\taeShort).
  User-provided text prompts are encoded using pre-trained text-encoders, and used as conditioning.
  Our generative model takes sampled Gaussian noise and all provided conditioning as input, and generates an output latent, which is decoded to an output image or video using the \taeShort decoder.
  }
  \label{fig:t2v_full}
\end{figure}

For improved training and inference efficiency, we perform generation in a spatio-temporally compressed latent space.
Towards this, we train a single temporal autoencoder model (\taeShort) to map both RGB images and videos into a spatio-temporally compressed latent space, and vice-versa.
We encode the user-provided text prompt using pre-trained text-encoders to obtain text prompt embeddings, which are used as conditioning for our model.
We use the \flowmatching training objective~\citep{flow-matching} to train our generative model.
Taking sampled noise and all provided conditioning as input, our generative model produces an output latent.
This is passed through the \taeShort decoder to map it back to the pixel space and produce an output image or video.
We illustrate the overview of the joint image and video generation pipeline in~\cref{fig:t2v_full}.

We focus on simplicity when making design choices for all components in our foundation model, including the training objective, backbone architecture, and spatio-temporal compression using the \taeShort.
These choices, which include using the \llama~\citep{llama3} backbone architecture for the joint image-video generation model, allow us to confidently scale the model size while allowing for efficient training.
Our largest $30$B parameter model can directly generate video at different aspect ratios (\eg, \texttt{1:1}, \texttt{9:16}, \texttt{16:9}), of multiple lengths (4 -- 16 seconds) at $768\times 768$ px resolution (scaled appropriately based on the aspect ratio).
Our \upsampler can further increase the spatial resolution to produce a video in full HD 1080p resolution.

Next, we describe the model architecture, pretraining and finetuning procedures for the foundation \OursVideo model.

\subsection{Image and Video Foundation Model}
\label{sec:t2v_model}
We describe the key components of the \OursVideo model---the spatio-temporal autoencoder (\taeShort), the training objective for image and video generation, model architecture, and the model scaling techniques we use in our work.

\subsubsection{\tae (\taeShort)}
\label{sec:tae}%

For the purposes of efficiency, we encode the RGB pixel-space videos and images into a learned spatio-temporally compressed latent space using a \tae (\taeShort), and learn to generate videos in this latent space.
Our \taeShort is based on a variational autoencoder~\citep{kingma2013auto} and compresses the input pixel space video $\bv$ of shape ${\vTime\times3\times\vHeight\times\vWidth}$ to a continuous-valued latent $\bx$ of shape ${\zTime\times\zChannels\times\zHeight\times\zWidth}$, where $\zTime<\vTime$, $\zHeight<\vHeight$, $\zWidth<\vWidth$.
In our implementation, we compress the input $8\times$ across each of the spatio-temporal dimensions, \ie, $\vTime/\zTime = \vHeight/\zHeight = \vWidth/\zWidth = 8$.
This compression reduces the overall sequence length of the input to the Transformer backbone, enabling the generation of long and high-resolution video at native frame rates.
This choice also allows us to forego frame-interpolation models commonly used in prior work~\citep{emuvideo2023,singer2023makeavideo,ho2022imagen}, thereby simplifying our model.

\par
\begin{figure}[t!]
\includegraphics[width=0.97\linewidth]{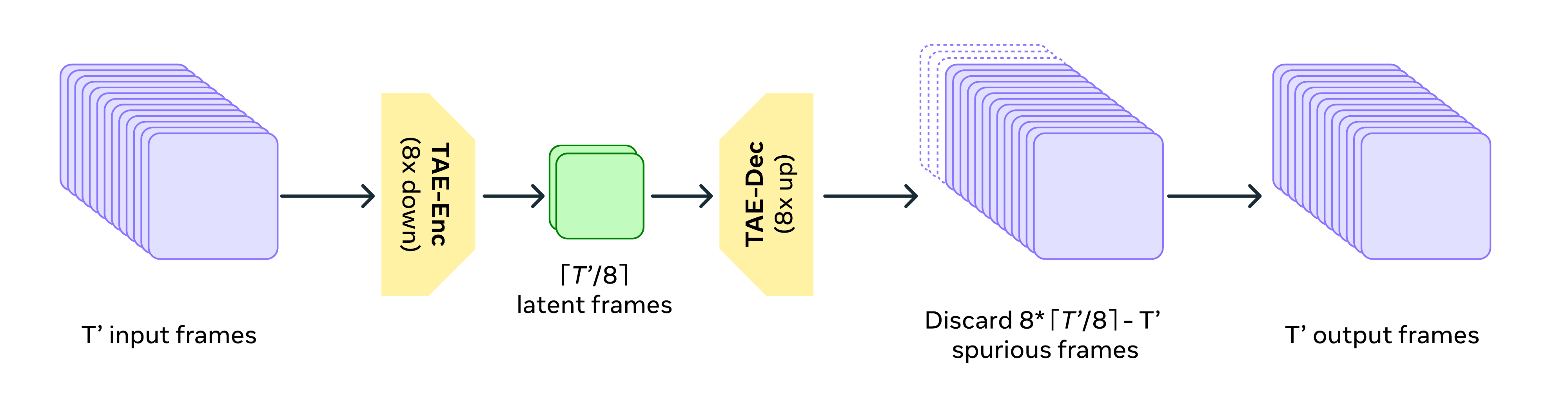}
\vspace{-2mm}
\caption{\textbf{Variable length video encoding and decoding using the \taeShort.}
The \taeShort encoder encodes $\vTime$ frames in the RGB pixel space to $\lceil\vTime/8\rceil$ latent frames.
The \taeShort decoder then decodes to $8 \times \lceil\vTime/8\rceil$ real frames, and any spurious frames ($8 \times \lceil\vTime/8\rceil - \vTime$) are discarded.
}
\label{fig:tae_up_down}
\end{figure}

 \noindent \textbf{\taeShort architecture.}
We adopt the architecture used for image autoencoders from~\citep{rombach2021highresolution} and `inflate' it by adding temporal parameters: a 1D temporal convolution after each 2D spatial convolution and a 1D temporal attention after each spatial attention.
All temporal convolutions use symmetrical replicate padding.
Temporal downsampling is performed via strided convolution with stride of 2, and upsampling by nearest-neighbour interpolation followed by convolution.
Downsampling via strided convolution means that videos of any length are able to be encoded (notably including images, which are treated as single-frame videos) by discarding spurious output frames as shown in~\cref{fig:tae_up_down}.
Similar to~\citep{dai2023emu}, we find that increasing the number of channels in the latent space $\bx$ improves both the reconstruction and the generation performance.
We use $\zChannels=16$ in this work.
We initialize the spatial parameters in the \taeShort using a pre-trained image autoencoder, and then add the temporal parameters to inflate the model as described above.
After inflation, we jointly train the \taeShort on both images and videos, in a ratio of 1 batch of images to 3 batches of videos.

\begin{figure}[b!]
  \centering
  \begin{tabular}{C{0.45\linewidth}C{0.45\linewidth}}
    \multicolumn{2}{c}{\includegraphics[width=0.9\linewidth,trim={0 2.5cm 0cm 0},clip]{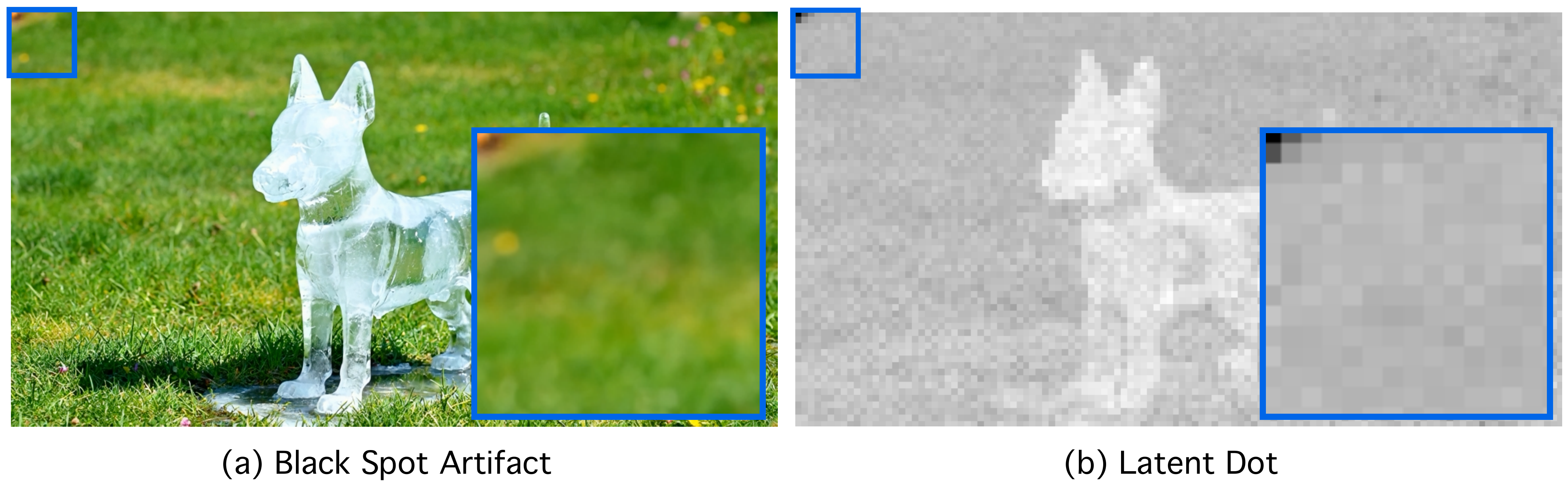}} \\
    (a) Spot artifact & (b) Latent dot
  \end{tabular}%
  \vspace{-2mm}
  \captionof{figure}{\textbf{Spot artifact and corresponding latent dot.} (a) Frame from a generated video displaying a spot artifact in the top left corner, (b) Visualization of a TAE feature channel, where the corresponding latent dot is visible.}
  \label{fig:tae_black_spot}
  \end{figure}

\noindent \textbf{Improvements to the training objective.}
We find that the standard training objective used in~\citep{rombach2021highresolution} leads to a `spot' artifact in the decoded pixel-space videos, as shown in~\cref{fig:tae_black_spot}.
On further inspection, we found that the model produced latent codes with high norms (`latent dots') in certain spatial locations, which when decoded led to `spots' in the pixel space.
We hypothesize that this is a form of shortcut learning, where the model learns to store crucial global information in these high-norm latent dots.
A similar phenomenon has been documented in~\citep{darcet2023vision}, where the authors discovered that vision Transformers can produce high-norm latent tokens, and also in~\citep{karras2024analyzing}, where they found that eliminating global operators such as group norms resolves the issue.

Rather than change the model architecture, we opt to add a term to the loss which penalizes the model for encoding latent values which are far from the mean.
Concretely, given an input latent $\bx$, our outlier penalty loss (OPL) is given by
\begin{equation} \label{eqn:tae_outlier_loss}
      \mathcal{L}_{\text{OPL}}\left(\bx, r\right) = \frac{1}{\zHeight\zWidth}\sum_{i=1}^H\sum_{j=1}^W \text{max}\left(\left\|\bx_{i,j} - \text{Mean}\left(\bx\right)\right\|-r\left\|\text{Std}\left(\bx\right)\right\|, 0\right),
\end{equation}
where $r$ is a scaling factor which denotes how far outside of the standard deviation a latent value needs to be to be penalized.
For images, equation \eqref{eqn:tae_outlier_loss} is used as-is; for videos, $\zTime$ is rolled into the batch dimension.
Adding $\mathcal{L}_{\text{OPL}}$ to the typical variational autoencoder losses (reconstruction, discriminator, and perceptual) removes the dot artifacts.
In practice, we set $r=3$ and a large loss weight ($1e5$) for the outlier loss.

\begin{figure}[t!]
\centering
\includegraphics[width=0.75\linewidth]{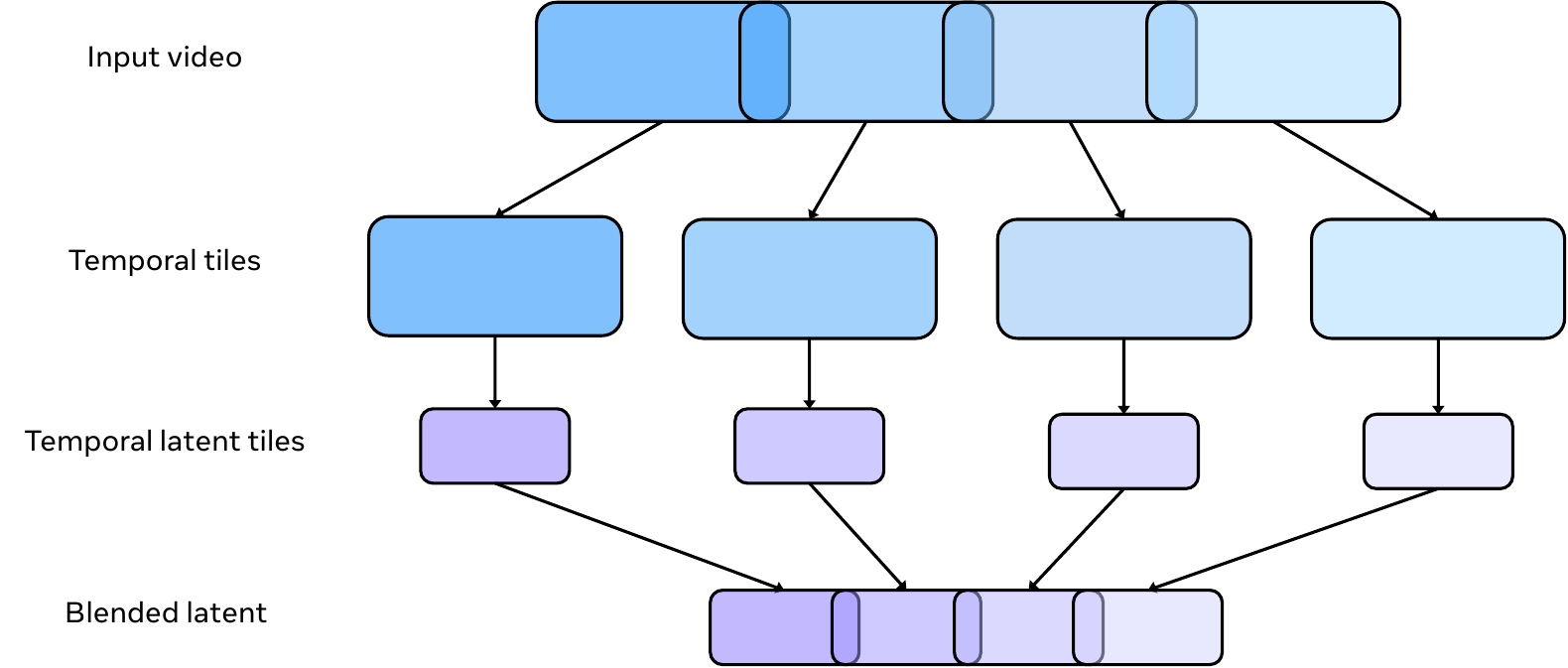}
\caption{\textbf{Tiled inference using the TAE.} An input video is split across the time dimension into uniform tiles, with optional overlap. Each tile is sent through the model forward pass. If overlap was used, a linearly weighted blend is performed during reconstruction.}
\label{fig:tae_tiling}
\end{figure}

\noindent \textbf{Efficient inference using temporal tiling.}
Encoding and decoding high resolution long videos, \eg, up to $1024 \times 1024$ px and $256$ frames na\"ively is not feasible due to memory requirements.
To facilitate inference with large videos, we divide both the input video and latent tensor into tiles along the temporal dimension, encode and/or decode each tile, and stitch the result together at the output.
When tiling, it is possible to include some overlap between tiles, with an additional weighted blend between adjacent tiles when stitching tiles back together.
Overlapping and blending can be applied to both the encoder and decoder, and has the effect of removing boundary artifacts at the cost of additional computation.
In practice, we use a tile size of 32 raw frames (or 4 latent frames), tile without overlap in the encoder, and tile with overlap of 16 raw frames (or 2 latent frames) in the decoder.
For blending, we use a linear combination between frames $i$ and $i+1$, $x^j_\text{blend}=\sum_j^N\left[w^jx_i^j+\left(1-w^j\right)x^j_{i+1}\right]$, where $j$ is indexed over the $N$ overlapping frames, and $w^j=j/N$.
~\cref{fig:tae_tiling} shows the basic flow of tiled inference.

\subsubsection {Training Objective for Video and Image Generation}
\label{sec:t2v_training_objective}
We use the Flow Matching~\citep{flow-matching,albergo2023building,liu2023flow} framework to train our joint image and video generation model.
Flow Matching generates a sample from the target data distribution by iteratively changing a sample from a prior distribution, \eg, Gaussian.
At training time, given a video sample in the latent space $\bx_{1}$, we sample a \timestep $\FMTime \in [0, 1]$, and a `noise' sample $\bx_{0} \sim \mathcal{N}(0, 1)$, and use them to construct a training sample $\bx_{\FMTime}$.
The model is trained to predict the velocity $\bv_{\FMTime} = \dfrac{d\bx_{\FMTime}}{d\FMTime}$ which teaches it to `move' the sample $\bx_{\FMTime}$ in the direction of the video sample $\bx_{1}$.

While there are numerous ways to construct $\bx_{\FMTime}$, in our work, we use simple linear interpolation or the optimal transport path~\citep{flow-matching}, \ie,
\begin{equation*}
  \bx_{\FMTime} = \FMTime~\bx_{1} + (1 - (1-\sigma_{\mathrm{min}})\FMTime)~\bx_{0},
\end{equation*}
where $\sigma_{\mathrm{min}}=10^{-5}$.
Thus, the ground truth velocity can be derived as
\begin{equation*}
  \begin{split}
    \bv_{\FMTime} &= \dfrac{d\bx_{\FMTime}}{d\FMTime} \\
    &= \bx_{1} - (1-\sigma_{\mathrm{min}}) \bx_{0}.
  \end{split}
\end{equation*}
Denoting the model parameters by $\theta$ and text prompt embedding $\bp$, we denote the predicted velocity as $u(\bx_{\FMTime}, \bp, \FMTime)$.
The model is trained by minimizing the mean squared error between the ground truth velocity and model prediction,
\begin{equation}
  \label{eq:fm}
  \mathbb{E}_{\FMTime, \bx_0, \bx_1, \bp}\|u(\bx_{\FMTime}, \bp, \FMTime; \theta) - \bv_t\|^2.
\end{equation}
As in prior work~\citep{sd3}, we sample $\FMTime$ from a logit-normal distribution where the underlying Gaussian distribution has zero mean and unit standard deviation.

\par \noindent \textbf{Inference.}
At inference, we first sample $\bx_{0}\sim \mathcal{N}(0, 1)$ and then use an ordinary differential equation (ODE) solver to compute $\bx_{1}$ using the model's estimated values for $\dfrac{d\bx_{\FMTime}}{d\FMTime}$.
In practice, there are multiple design choices in the exact ODE solver configuration, \eg, first or higher order solvers, step sizes, tolerance, \etc that affect the runtime and precision of the estimated $\bx_{1}$.
We use a simple first-order Euler ODE solver with a unique discrete set of $\FMDiscreteMaxTimeSteps$ \timesteps tailored to our model, as described in~\cref{sec:linear_quad_sampler}.

\par \noindent \textbf{Signal-to-noise ratio.}
The \timestep $\FMTime$ controls the signal-to-noise (SNR) ratio, and our simple interpolation scheme for constructing $\bx_{\FMTime}$ ensures zero SNR when $\FMTime=0$.
This ensures that, during training, the model receives pure Gaussian noise samples and is trained to predict the velocity for them.
Thus, at inference, when the model receives pure Gaussian noise at $\FMTime=0$ it can make a reasonable prediction.

Most video generation models~\citep{ho2022imagen,emuvideo2023,Blattmann2023AlignYL,singer2023makeavideo} are trained using the diffusion formulation~\citep{pmlr-v37-sohl-dickstein15,Ho2020DenoisingDP,ho2022imagen}.
Recent work~\citep{emuvideo2023,lin2024common} shows that choosing the right diffusion noise schedules with a zero terminal signal-to-noise ratio is particularly important for video generation.
Standard diffusion noise schedules do not ensure a zero terminal SNR, and thus need to be modified for video generation purposes.
As noted above, our \flowmatching implementation naturally ensures zero terminal SNR.
Empirically, we found that \flowmatching was more robust to the exact choice of noise schedules and it outperforms diffusion losses (see~\cref{sec:ablations_t2v}).
Thus, we adopt \flowmatching for its simplicity and high performance.

\subsubsection {Joint Image and Video Generation Backbone Architecture} \label{t2v_architecture_details}%
As discussed in~\cref{sec:tae}, we perform generation in a learned latent space representation of the video.
This latent code is of shape $\zTime\times\zChannels\times\zHeight\times\zWidth$.
To prepare inputs for the Transformer backbone, the video latent code is first `patchified' using a 3D convolutional layer~\citep{dosovitskiy2020image} and then flattened to yield a 1D sequence.
The 3D convolutional layer uses a kernel size of $\kTime\times\kHeight\times\kWidth$ with a stride equal to the kernel size and projects it into the same dimensions as needed by the Transformer backbone.
Thus, the total number of tokens input to the Transformer backbone is $\zTime\zHeight\zWidth/(\kTime\kHeight\kWidth)$.
We use $\kTime=1$ and $\kHeight=\kWidth=2$, \ie, we produce $2\times2$ spatial patches.

We use a factorized learnable positional embedding to enable arbitrary size, aspect ratio, and video length~\citep{dehghani2024patch} inputs to the Transformer.
Absolute embeddings of $D$ dimensions can be denoted as a mapping $\phi(i): [0, \mathrm{maxLen}] \rightarrow \mathbb{R}^{D}$
where $i$ denotes the absolute index of the patch.
We convert the `patchified' tokens, \ie, output of the 3D convolutional layer, into separate embeddings $\phiHeight$, $\phiWidth$ and $\phiTime$ of spatial $\heightSmall$, $\widthSmall$, and temporal $\timeSmall$ coordinates.
We define $\zHeight_{max}$, $\zWidth_{max}$, and $\zTime_{max}$ as the maximum sequence length ($\mathrm{maxLen}$) for each dimension, which correspond to the maximum spatial size and video length of the patchified inputs.
We calculate the final positional embeddings by adding all the factorized positional embeddings together.
Finally, we add the final positional embeddings to the input for all the Transformer layers.
Compared with adding the positional embeddings to the first layer only, adding to all layers can effectively reduce the distortion and morphing artifacts, especially in the temporal dimension.

We build our Transformer backbone by closely following the Transformer block used in the \llama~\citep{llama3} architecture.
We use RMSNorm~\citep{zhang2019root} and SwiGLU~\citep{shazeer2020glu} as in prior work.
We make three changes to the \llama Transformer block for our use case of video generation using \flowmatching:
\begin{enumerate}
  \item To incorporate text conditioning based on the text prompt embedding $\bp$, we add a cross-attention module between the self-attention module and the feed forward network (FFN) to each Transformer block. We leverage multiple different text encoders due to their complementary strengths, as explained in the following section, and simply concatenate their embeddings in a single sequence to construct $\bp$.
  \item We add adaptive layer norm blocks to incorporate the \timestep $\FMTime$ to the Transformer, as used in prior work~\citep{dit}.
  \item We use full bi-directional attention instead of causal attention used in language modeling.
\end{enumerate}

We intentionally keep the design of our backbone simple and similar to LLMs, specifically \llama.
This design choice allows us scale the model size and training, as discussed in~\cref{sec:t2v_impl_scaling}, using similar techniques as used in LLMs.
Empirically, we find that our architecture design performs on par or better than specialized blocks used in prior work~\citep{balaji2022ediffi,sd3} while being more stable to train across a range of hyperparameters such as model size, learning rate, and batch size.
We list the key hyperparameters for our largest model in~\cref{tab:t2v_hyperparameters} and illustrate the Transformer block in~\cref{fig:moviegen_sharding} with details of feature dimensions in a number of key places of our Transformer backbone.

\begin{table}[t]
  \centering
  \small
  \begin{tabular}{cccccc}
  \toprule
  Layers & Model Dimension & FFN Dimension & Attention Heads & Activation Function & Normalization	\\
  \midrule
  48 & 6144 & 16384 & 48 & SwiGLU & RMSNorm \\
  \bottomrule
  \end{tabular}%
  \caption{\textbf{Architecture hyperparameters for the \OursVideo 30B parameter foundation model.} Our model architecture is a Transformer~\citep{vaswani2017attention} and we closely follow the \llama~\citep{llama3} design space.
  Our model contains 30B parameters in the Transformer itself without including the the text embedding models, \taeShort, \etc.
  }
  \label{tab:t2v_hyperparameters}
  \end{table}

\subsubsection{Rich Text Embeddings and Visual-text Generation} %
\label{sec:t2v_text_encoder}
We use pre-trained text encoders to convert the input text prompt $\prompt$ into a text embedding $\bp$, which we use as conditioning input for the video generation backbone.
We use a combination of UL2~\citep{tay2022ul2}, ByT5~\citep{xue2022byt5}, and Long-prompt MetaCLIP as text encoders to provide both semantic-level and character-level text understanding for the backbone.
The Long-prompt MetaCLIP model is obtained by finetuning the MetaCLIP text encoder~\citep{xu2023demystifying} on longer text captions to increase the length of input text tokens from $77$ to $256$.
We concatenate the text embeddings from the three text encoders after adding separate linear projection and LayerNorm layers to project them into the same 6144 dimension space and normalize the embeddings.
The UL2 and Long-prompt MetaCLIP text encoders provide prompt-level embeddings with different properties---UL2 is trained using massive text-only data and potentially provides strong text reasoning abilities in its features;
Long-prompt MetaCLIP provides text representations that are aligned with visual representations that are beneficial for cross-modal generation.
The character-level ByT5 encoder is only used to encode visual text, \ie, the part of the text prompt that may explicitly ask for a character string to be generated in the output.

\noindent \textbf{Controlling the FPS.}
We use FPS conditioning to control the length of the generated videos by pre-appending the sampling FPS value of each training video to the input text prompt (\eg, ``FPS-16''). During pre-training, we sample video clips at their original FPS with minimum of 16 FPS. In finetuning, we sample clips at two fixed FPS values of 16 and 24.

\subsubsection{Spatial Upsampling} \label{sec:spatial_upsampler}
We use a separate \upsampler model to convert our $768$ px videos to full HD (1080p) resolution.
This lowers the overall computational cost for high resolution generation, since the base \textToV model processes fewer tokens.

\begin{figure}[!t]
\centering
\includegraphics[width=\linewidth]{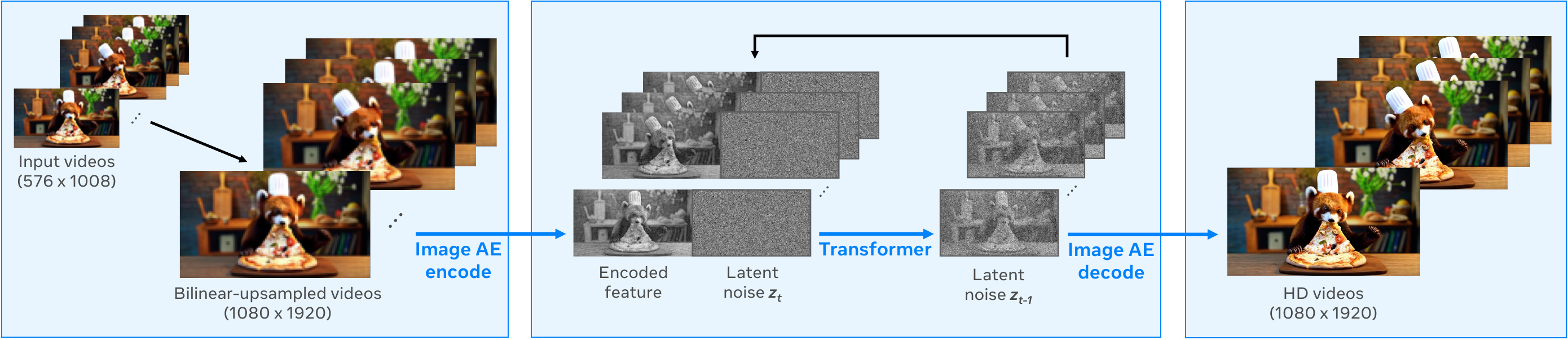}

\caption{\textbf{Overview of the \upsampler.} Our upsampler is a conditional video-to-video model that upsamples the $768$ px video to full HD $1080$p.
First, the input $768$ px video is bilinearly upsampled to HD and then encoded to the latent space of an image encoder.
The video latents are concatenated with noise, and denoised using a trained transfomer. Finally, the denoised latents are passed to the image decoder to produce the upsampled video.
}
\label{fig:t2v_upsampler}
\end{figure}

As shown in~\cref{fig:t2v_upsampler}, we formulate spatial upsampling as a video-to-video generation task, that generates a HD output video conditioned on a lower-resolution input video.
The low-resolution video is first spatially upsampled using bilinear interpolation in the pixel space to the desired output resolution.
Next, the video is converted to the latent space using a VAE. We use a frame-wise VAE for the upsampler to improve pixel sharpness.
Finally, a latent space model generates the latents of a HD video, conditioned on the latents of the corresponding low-resolution video.
The resulting HD video latents are subsequently decoded into pixel space frame-wise using the VAE decoder.

\noindent\textbf{Implementation details.}
Our \upsampler model architecture is a smaller variant (7B parameters) of the \textToV Transformer initialized from a \textToI model trained at 1024 px resolution, allowing for better utilization of high-resolution image data. %
The \upsampler is trained to predict the latents of a video which are then decoded frame-wise using the VAE's decoder.
Similar to~\citep{emuvideo2023}, the encoded video is concatenated channel-wise with the generation input and is fed to the \upsampler Transformer.
The additional parameters at the input, due to concatenation, are zero initialized~\citep{singer2023makeavideo,emuvideo2023}.
We train our \upsampler on clips of 14 frames at 24 FPS on $\sim$400K HD videos.
We apply a second-order degradation~\citep{wang2021real} process to simulate complex degradations in the input and train the model to produce HD output videos.
At inference time, we will use our \upsampler on videos that have been decoded with the \taeShort.
To minimize this potential train-test discrepancy, we randomly substitute the second-order degradation with artifacts produced by the \taeShort.
Due to the strong input conditioning, \ie, the low-resolution video, we observed that the model produces good outputs with as few as $20$ inference steps.
This simple architecture can be used for various multiples of super resolution; however, we train a $2\times$ spatial super-resolution model for our case.
Similar to \taeShort tiling (\cref{sec:tae}), we upsample videos using a sliding window approach with a window size of 14 and an overlap of 4 latent frames.

\noindent \textbf{Improved temporal consistency with Multi-Diffusion.}
Memory constraints prohibit us from training the \upsampler on longer video durations. As a result, during inference, we upsample videos in a sliding window fashion resulting in noticeable inconsistencies at the boundaries.
To prevent this, we leverage MultiDiffusion~\citep{bar2023multidiffusion}, a training-free optimization that ensures consistency across different generation processes bound by a common set of constraints.
Specifically, we use a weighted average of the latents from overlapping frames in each denoising step, facilitating the exchange of information across consecutive windows to enhance temporal consistency in the output.

\subsubsection{Model Scaling and Training Efficiency}
\label{sec:t2v_impl_scaling}

We describe the key details that allow us to scale and efficiently train the \OursVideo 30B parameter foundation model.
In the following section, we will (1) outline hardware and infrastructure details, (2) compare and contrast our training setup to state-of-the-art LLMs~\citep{llama2,llama3}, and (3) discuss model parallelism methods used for \OursVideo.

\noindent \textbf{Infrastructure.}
We trained the media generation models using up to 6,144 H100 GPUs, each running at 700W TDP and with 80GB HBM3, using Meta’s Grand Teton AI server platform~\citep{grand_teton}.
Within a server there are eight GPUs which are uniformly connected via NVSwitches. Across servers GPUs are connected via 400Gbps RoCE RDMA NICs.
Training jobs are scheduled using MAST~\citep{choudhury2024mast}, Meta's global-scale training scheduler.

\noindent \textbf{Comparison with Large Language Models.}
LLMs use structured causal attention masks to enforce token causality, unlike the full bi-directional attention used in \OursVideo. This causal masking can be leveraged to provide approximately a 2$\times$ speedup compared to attention without the causal mask while also reducing peak memory requirements~\citep{dao2023flashattention2}.

Secondly, state-of-the-art LLMs such as \llama~\citep{llama3} use Grouped-Query Attention (GQA) instead of Multi-head Attention (MHA), which reduces the number of $K$-, $V$-heads and thus the total dimension of the key and value projections. This results in a reduction in FLOPs and tensor memory size while also improving memory bandwidth utilization.
Furthermore, autoregressive LLMs gain additional inference time benefits through the use of GQA due to a reduction in their $K,V$-cache size.
In part due to the non-autoregressive design of \OursVideo, we do not explore this architectural design choice and leave it for future work.

Similar to current LLMs like \llama, our training is divided into stages of varying context lengths, where our context length varies depending on the spatial resolution (256 px or 768 px). For 768 px training this results in a context length of $\sim$73K tokens ($768\times768$ px video with 256 frames, compressed $8\times8\times8$ through the \taeShort, and $2\times2\times1$ through patchification).
But unlike LLMs which are trained at shorter context lengths for the majority of the training budget, the majority of our training FLOPs are expended on long-context 768 px training (see~\cref{tab:t2v_pt_recipe}). Due to the quadratic nature of self-attention, which is at the heart of a Transformer block, scaling to very large context lengths requires immense computation (FLOPs). This makes it even more important to optimize our training setup for long-context training.

\noindent \textbf{Model Parallelism.}
Our large model size and extremely long-context lengths necessitate the use of multiple parallelisms for efficient training. We employ 3D parallelism to support model-level scaling across three axes: number of parameters, input tokens, and dataset size, while also allowing horizontal scale-out to more GPUs. We utilize a combination of fully sharded data parallelism \citep{rajbhandari2020zeromemoryoptimizationstraining,ren2021zerooffloaddemocratizingbillionscalemodel,zhao2023pytorch}, tensor parallelism \citep{shoeybi2019megatron,narayanan2021efficient}, sequence parallelism \citep{li2021sequence,korthikanti2023reducing}, and context parallelism~\citep{liu2023ring,nvidia_cp}.

\begin{figure}[t]
  \centering
  \includegraphics[width=1.0\linewidth,]{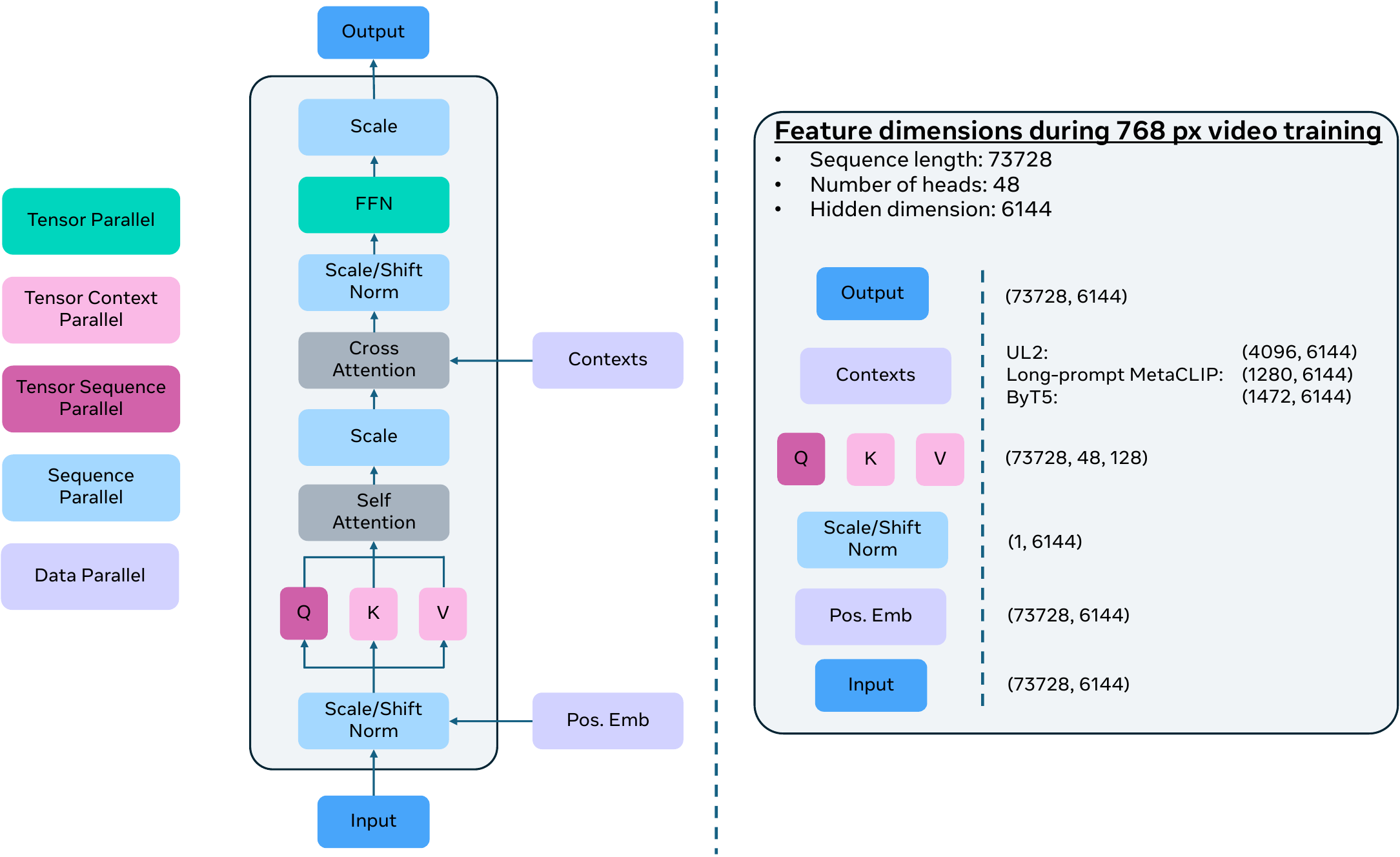}

  \caption{\textbf{\OursVideo Transformer backbone and model parallelism applied.}
  Left: we illustrate the Transformer backbone and color-code different model parallelizations used to shard our 30B model (described in~\cref{sec:t2v_impl_scaling}).
  Right: Feature dimensions in a number of key steps during the most expensive stage of \OursVideo training, processing 768 px video inputs with a per-sample sequence length of 73K tokens.
  }
  \label{fig:moviegen_sharding}
  \end{figure}

In the following, we describe different parallelisms and how they are utilized in different parts of our Transformer backbone (as illustrated in \cref{fig:moviegen_sharding}).

\begin{itemize}
  \item \textbf{Tensor-parallelism (TP)} shards the weights of linear layers either along columns or rows, and results in each GPU involved in the sharding performing \textit{tp-size} less work (FLOPs) and generating \textit{tp-size} less activations for column-parallel shards and consuming \textit{tp-size} less activations for row-parallel shards. The cost of performing such a sharding is the addition of all-reduce communication overheads in both the forward (row-parallel) and backward (column-parallel) passes.

  \item \textbf{Sequence-parallelism (SP)} builds upon TP to also allow the sharding of the input over the sequence dimension \textit{for layers which are replicated and in which each sequence element can be treated independently.} Such layers, \eg, LayerNorm, would otherwise perform duplicate compute and generate identical (and thus replicated) activations across the TP-group.

  \item \textbf{Context-parallelism (CP)} enables a \textit{partial} sharding over the sequence dimension for the \textit{sequence-dependent softmax-attention operation}. CP leverages the insight that for any given \mbox{\textit{(source (context), target (query))}} sequences pair, \textit{softmax-attention is only sequence-dependent over the context and not the query}. Therefore in the case of self-attention where the input source and target sequences are identical, CP allows the attention computation to be performed with only an all-gather for the $K$ and $V$ projections (instead of $Q$, $K$, and $V$) in the forward pass, and a reduce-scatter for their associated gradients in the backward.

  Additionally, due to the separation of behavior between the $Q$ and $K,V$ projections, the performance of CP is variable not only in the context length, but also the size of the context dimension.
  A consequence of this is the differentiation of scaling performance and overhead characterstics for CP between \OursVideo and state-of-the-art LLMs, such as \llama, which use GQA and thus generate smaller $K,V$ tensors to be communicated (\eg, 8$\times$ smaller for \llama 70B).

  \item \textbf{Fully Sharded Data Parallel (FSDP)} shards the model, optimizer and gradients across all data-parallel GPUs, synchronously gathering and scattering parameters and gradients throughout each training step.
\end{itemize}

\noindent \textbf{Overlapping Communication and Computation.}
While parallelism techniques can enable training large sequence Transformer models by partitioning FLOP and memory demands across GPUs, their direct implementation can introduce overheads and inefficiencies.
We build an analytical framework to model compute and communication times that allows us identify duplicated activations that require inter-GPU communication, and thus design a highly optimized model parallel solution.
Our new custom implementation of model parallelization, written in PyTorch and compiled into CUDAGraphs, achieves strong activation memory scaling and minimizes exposed communication time.
We provide more details on our optimized training setup in~\Cref{app:t2v_impl_scaling_additional}.

\subsection{\Pretraining}%
\subsubsection{\Pretraining Data}
\label{sec:pt-data} %

Our \pretraining dataset consists of \videoDataSize
video-text pairs and \imageDataSize image-text pairs.
We follow the \pretraining data curation strategy similar to~\citep{dai2023emu} for image-text data curation and focus on video data curation in this section.

\begin{figure}[b!]
    \centering
    \includegraphics[width=\linewidth]{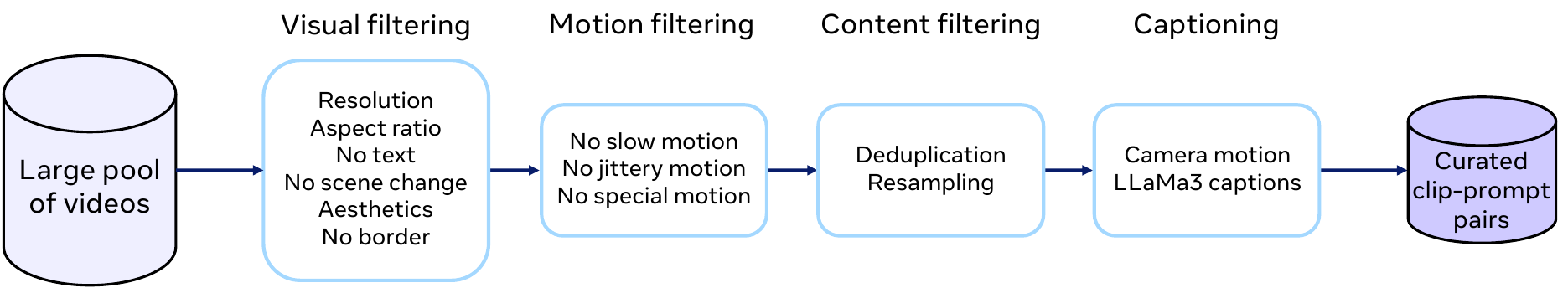}
    \caption{\textbf{\OursVideo \pretraining data curation pipeline.}
    We filter a large pool of videos through multiple stages to produce a small set of high-quality training clips with associated prompts.}
    \label{img:t2v_pt_data}
\end{figure}

Our original pool of data consists of videos that are 4 seconds to 2 minutes long, spanning concepts from different domains such as humans, nature, animals, and objects.
Our data curation pipeline yields our final \pretraining set of clip-prompt pairs, where each clip is 4s -- 16s long, with single-shot camera and non-trivial motion.
Our data curation pipeline is illustrated in~\cref{img:t2v_pt_data}.
It consists of three filtering stages: 1) visual filtering, 2) motion filtering, and 3) content filtering, and one captioning stage. The filtered clips are annotated with detailed generated captions containing 100 words on average. We describe each stage in detail below.

\textbf{Visual filtering.}
We apply a series of 6 filters to remove videos with low visual quality. We remove videos of size smaller than a minimum width or height of 720 px.
We filter on aspect ratio to achieve a mix of 60\% landscape and 40\% portrait videos.
We prefer landscape videos over portrait videos due to their longer duration, better aesthetics, and stable motion.
We use a video OCR model to remove videos with excessive text. We also perform scene boundary detection using FFmpeg~\citep{ffmpeg} to extract 4 to 16 second long clips from these videos.
We then train simple visual models to obtain prediction signals for filtering based on frame-level visual aesthetic, visual quality, large borders, and visual effects.
Following Panda-70M~\citep{chen2024panda70m}, we remove the first few seconds of clips whose start coincides with the beginning of a video, as the beginning of a video usually contains unstable camera movement or transition effects.

\textbf{Motion filtering.}
We follow prior work~\citep{emuvideo2023} to automatically filter out low motion videos.
First, we used an internal static video detection model to remove videos with no motion.
Next, we identified videos with `reasonable' motion based on their VMAF motion scores and motion vectors~\citep{ffmpeg}.
To remove videos with frequent jittery camera movements, we used Shot Boundary Detection from the PySceneDetect~\citep{pyscenedetect} library.
Finally, we removed videos with special motion effects, \eg, slideshow videos.

\textbf{Content filtering.}
To ensure diversity in our \pretraining set, we remove perceptually duplicate clips in our \pretraining set using similarity in a copy-detection embedding~\citep{pizzi2022self} space.
We also reduce the prevalence of dominant concepts by resampling to create our training set.
We cluster semantic embeddings from a video-text joint embedding model
to identify fine grained concept clusters.
Next, we merge duplicate clusters and sample clips from each merged cluster according to the inverse square root of the cluster size~\citep{mahajan2018exploring}.

\textbf{Captioning.}
We create accurate and detailed text prompts for the video clips by using the \llamaVideo~\citep{llama3} model.
We finetune the 8B and 70B variants of the model for the video captioning task and use these models to caption the entire training set of video clips.
Our training set consists of 70\% 8B captions and 30\% 70B captions.
To enable cinematic camera motion control, we train a camera motion classifier which predicts one of 16 classes of camera motion \eg, zoom-out, pan-left, \etc (see~\Cref{appendix:data_camera_motion} for more details).
We prefix high-confidence camera motion predictions to the previously generated text captions.
At inference, this allows a user to specify explicit camera control for video generation.

\textbf{Multi-stage data curation.}
We curate 3 subsets of \pretraining data with progressively stricter visual, motion, and content thresholds to meet the needs of different stages of \pretraining.
First, we curated a set of video clips with a minimum width and height of 720 px for low-resolution training.
Next, we filtered the set to provide videos with a minimum width and height of 768 px for high-resolution training.
Finally, we curated new videos augmenting our high-resolution training set.
Our high resolution set has 80\% landscape and 20\% portrait videos, with at least 60\% of them containing humans. During curation, we established a taxonomy of 600 human verbs and expressions and performed zero-shot text-to-video retrieval using the taxonomy to select videos with humans in them.
We preserved the frequency of these human videos during content resampling. See~\Cref{appendix:data_curation_thresholds} for details on thresholds for curation of these videos.

\begin{table}[!t]
    \centering
    \adjustbox{max width=\textwidth}{%
    \begin{tabular}{cccc}
        \toprule
        Duration & FPS & \#video frames & \#latent frames \\
        \midrule
        10.67s & 24 & 256 & 32 \\
        16s & 16 & 256 & 32 \\
        12s - 16s & 21 - 16 & 256 & 32 \\
        8s - 12s & 24 - 16 & 192 & 24 \\
        4s - 8s & 32 - 16 & 128 & 16 \\
        \bottomrule
    \end{tabular}}
    \caption{\textbf{Duration buckets for the \textToVShort pre-training datasets.}
    We split our training videos into five buckets based on their duration and FPS.
    Videos in each bucket have the same number of latent frames which allows for easy batching of training data.
    The buckets for the last three rows are based on the original duration of the video clips.
    The buckets for the first two rows are created based on middle clips sampled from videos of $10.67$-$12$ seconds and $\geq$ 16 seconds respectively.
    }
    \label{tab:t2v_pt_bucket}
\end{table}

\noindent \textbf{Bucketization for variable duration and size.}
To accommodate diverse video lengths and aspect ratios, we bucketize the training data according to aspect ratio and length.
The videos in each bucket lead to the exact same latent shape which allows for easy batching of training data.
We use five aspect ratio buckets for both image and video datasets.
Thus, our model can generate images and videos of different aspect ratios, \eg, $1024 \times 576$ for landscape and $576 \times 1024$ for portrait.
We define five duration buckets (4s -- 16s) and adjust the number of latent frames according to video length (see~\cref{tab:t2v_pt_bucket}).
As described in~\cref{sec:t2v_text_encoder}, we introduce FPS control by adding an FPS token to the text caption, allowing us to sample videos at different frame rates (16 -- 32 FPS).

\subsubsection{Training}
We describe the training details for our 30B parameter model.
To improve training efficiency and model scalability, we employ a multi-stage training procedure, similar to~\citep{emuvideo2023}.
This procedure consists of three main steps:
\begin{itemize}
    \item Initial training on the \textToI (\textToIShort) task, followed by joint training on both \textToI and \textToV (\textToVShort) tasks;
    \item Progressive resolution scaling from low-resolution $256$ px data to high-resolution $768$ px data;
    \item Continuous training using improved datasets and optimized training recipes while working with compute and time restrictions.
\end{itemize}
The training recipe is summarized in~\cref{tab:t2v_pt_recipe}.
We maintain a validation set of unseen videos and monitored the validation loss throughout training.
We observed that the validation loss for our model correlated well with visual quality judged by human evaluations.

\noindent \textbf{\TextToI Warm-up Training}.
Jointly training \textToIVShort models is significantly slower and more memory-intensive than training T2I models alone, primarily due to the substantially longer latent token length (\eg, 32$\times$).
Furthermore, we observed that directly training \textToIVShort models from scratch results in a slower convergence speed than initializing them from a \textToIShort model.
For instance, after training for the same number of GPU hours, we noticed significantly worse visual and temporal quality for both \textToIShort and T2V tasks compared to our proposed multi-stage training approach.
To address this, we begin with a T2I-only warm-up training stage.
Rather than training at the target 768 px resolution, we train this stage at a lower resolution (256 px) which allows us to train with a larger batch size and on more training data for the same training compute.

\begin{table}[!t]
    \centering
    \adjustbox{max width=\textwidth}{%
    \begin{tabular}{lcccccccrr}
        \toprule
        Training stage & Dataset & TP & CP & bs/node & \#GPUs & global bs & learning rate & \#iters & \#seen samples \\
        \midrule
        256 px T2I & \imageDataSize images & 1 & 1 & 48 & 1536 & 9216 & 1e-4 &
        210k & 1.94B \\
        \bottomrule
        \toprule
        \multirow{3}{*}{256 px T2I/V} & \multirow{2}{*}{\videoDataSize clips} & 4 & 1 & 4 & 3072 & 1536 & 6e-5 &
        123k & 173.6M \\
        & & 4 & 1 & 4 & 6144 & 3072 & 6e-5 & 72k & 221.2M \\
        \cmidrule{2-10}
        & {\bf Total} & & & & & & & {\bf 185k} & {\bf 394.8M} \\
        \bottomrule
        \toprule
        \multirow{6}{*}{768 px T2I/V} & \multirow{4}{*}{\shortstack{\videoDataSize clips}} & 4 & 1 & 2 & 6144 & 1536 & 6e-5 & 19.6k & 30.1M \\
        & & 4 & 1 & 2 & 6144 & 1536 & 3e-5 & 11k & 16.9M \\
        & & 4 & 2 & 1 & 6144 & 768 & 2e-5 & 15.9k & 12.2M \\
        & & 4 & 2 & 1 & 4096 & 512 & 1e-5 & 28k & 14.6M \\

        \cmidrule{2-10}
        & {\bf Total} & & & & & & & {\bf 74.5k} & {\bf 73.8M} \\
        \bottomrule
    \end{tabular}}
    \caption{\textbf{Progressive recipe and datasets for T2I/V pre-training.} Note that: (1) besides the listed video datasets, the same image dataset used for T2I training is also used in T2I/V training with a ratio of 1:10 over video data, and (2) the video datasets of difference sources are sampled with ratios respecting to the volume of the dataset
    }
    \label{tab:t2v_pt_recipe}
\end{table}

\noindent \textbf{T2I/V Joint Training}.
After the \textToIShort warmup training, we train the model jointly for \textToI and \textToV.
To enable the joint training, we double the spatial positional embedding (PE) layers to accommodate various aspect ratios,
add new temporal PE layers to support up to 32 latent frames,
and initialize spatial PE layers from the \textToIShort model with $2\times$ expansion.
We first use 256 px resolution images and videos for the \textToIVShort joint training.
For the 768 px stage, we expand the spatial PE layers by $3\times$.
\cref{tab:t2v_pt_recipe} summarizes the training recipe.
\begin{itemize}
    \item \textbf{256 px T2I/V stage.}
    We use a large batch size of 1536 samples and a larger learning rate $6e^{-5}$ that results in stable training.
    After 123k iterations, we double the number of GPUs, yielding the $2 \times$ bigger global batch size and a significant drop in the validation loss.
    We stop the training at 185k iterations after 395M (4+ epochs) video samples.
    \item \textbf{768 px T2I/V stage.}
    We observe that the validation loss decreases quickly in the first 10k iterations and then fluctuates, see~\cref{fig:t2v_pt_eval_losses_and_eval}.
    We reduce the learning rate by half at 19.6k iterations which further reduces the loss.
    We continue to train the model and decrease the learning rate whenever the validation loss plateaus.
\end{itemize}

\subsection{Finetuning}  %
\label{sec:t2v_post_training}

As in prior work~\citep{dai2023emu,emuvideo2023}, we improve the motion and aesthetic quality of the generated videos by finetuning the pre-trained model on a small finetuning set of selected videos.
The finetuning set videos and captions are manually curated, and thus we term this stage as supervised finetuning.
During this stage, we train multiple models and combine them to form the final model through a model averaging approach.
While our model can generate high quality images, we find that post-training specifically for images results in a significant boost in quality.
We describe the image-specific post-training recipe in~\Cref{sec:t2i_generation} and describe the video-specific post-training recipe next.

\noindent \textbf{Finetuning Video Data.}
We aim to collect a finetuning set of high quality videos with good motion, realness, aesthetics, with a wide range of concepts, and with high quality captions.
For finding such videos, we start with a large pool of videos and apply both automated and manual filtering steps (taking motivation from the curation recipe from~\citep{dai2023emu}).
There are four key stages that are run in sequence, each operating on the output of the previous stage:
(1) Establishing a set of candidate videos.
Here, we use automated filters that set strict thresholds on aesthetics, motion, scene change.
Additionally, we remove videos with small subjects using an object detection model~\citep{zhou2022detecting} on frames.
This stage results in a few million videos but with an unbalanced distribution of concepts.
(2) Balancing the concepts in set of videos.
The goal of this stage is to obtain a small enough subset of concept-balanced videos such that each can be manually filtered in the following steps.
We used our taxonomy of human verbs and expressions, defined in ~\cref{sec:pt-data}, to perform text $k$-NN methods
to retrieve videos for each concept from the candidate pool of videos.
We manually picked a few visually appealing seed videos per concept and performed video $k$-NN to get a concept-balanced subset of videos.
For $k$-NN, we used the video and text embeddings from a video-text joint embedding model.
(3) Manually identifying cinematic videos.
Many aspects to high quality finetuning data cannot be reliably captured by automated filters with high precision and recall.
At this stage, we instead rely on manual filtering.
Here we ensure that the remaining videos have angled (natural sunshine or studio) lighting, vivid (but not over-saturated) colors, no clutter, non-trivial motion, no camera shake, and no edited effects or overlay text.
During this stage, annotators additionally clip videos to the desired duration that will be trained on by selecting the best, most compelling clip of the video.
(4) Manually captioning the videos.
In detail, human annotators refine \llamaVideo generated captions by fixing incorrect details and ensuring the inclusion of certain key video details.
These include camera control, human expressions, subject and background info, detailed motion description and lighting information.
At this stage humans annotate six additional camera motion and position types (see~\Cref{appendix:data_camera_motion}).
Our video finetuning data is set to have duration between 10.6s and 16s.
In practice, 50\% of videos are 16s long, while the rest of 50\% videos are between 10.6s to 16s.

\noindent \textbf{Supervised Finetuning Recipe.} In video supervised finetuning (SFT), we use the same model architecture as the pre-training stage, and finetune the model with the pre-training checkpoints as initialization.
Different from pre-training that uses large-scale data, large batch sizes and training resources, we instead use relatively a small batch size and 64 nodes (512 H100 GPUs) to train the model, and use a cosine learning rate scheduler~\citep{loshchilov2016sgdr}.
Similar to the \pretraining stage, for videos that are at 16s, we train with 16 FPS, and for videos that are between 10.6s to 16s, we train with 24 FPS.
As a result, our model is trained to best support the generation of videos in both 10s and 16s.

\noindent \textbf{Model Averaging.}
Our experiments reveal that the choice of different sets of finetune data, hyperparameters as well as pre-train checkpoints significantly affects key aspects of the model's behavior, including motion, consistency, and camera control. To harness the diverse strengths of these models, we employ a model averaging approach.
Similar to \llama~\citep{llama3}, we average models obtained from SFT experiments that use various versions of finetune data, hyperparameters and pre-train checkpoints.

\subsection{Inference} \label{T2V_inference_details}
In this section, we describe the different hyper-parameters and settings used for sampling from \OursVideo.
For comparisons to prior work, we use a text classifier-free guidance scale of 7.5, and we use the linear-quadratic sampler described in~\cref{sec:linear_quad_sampler} with 50 steps (emulating 250 linear steps).
We also use an inference prompt rewrite on the input text prompt, as described below.
\subsubsection{Inference Prompt Rewrite} %
\label{sec:t2v_prompt_rewrite}

As mentioned in Section~\ref{sec:pt-data}, we train the model with high quality video/image-text pairs, and these training captions are characterized by their dense details and consistent paragraph structure.
However, the writing style and length of prompts in the inference stage vary widely.
For instance, most users typically type less than $10$ words, which is shorter than the average length of training captions.
To bridge the distribution gap between training captions and inference prompts, we utilize \llama~\citep{llama3} to transform the original input prompts into more detailed ones. The key details of the inference prompt rewrite model are:
\begin{itemize}
    \item We employ a standardized information architecture to rephrase the prompts, ensuring consistency in the visual composition.
    \item We refine the rewritten prompts by replacing complex vocabulary with more accessible and straightforward terminology, thereby enhancing their clarity and comprehensibility.
    \item We observe that excessively elaborate descriptions of motion details can result in the introduction of artifacts in the generated videos, highlighting the importance of striking a balance between descriptive richness and visual fidelity.
\end{itemize}

\textbf{Efficient Inference Rewrite Model.}
To improve the computation efficiency of the inference rewrite model, we developed a teacher-student distillation approach for this purpose.
Initially, we built a prompt rewrite teacher model based on the \llama 70B model, using detailed prompt instructions and in-context learning examples from the foundation model training set.
We then gathered human-in-the-loop (HITL) finetuning data.
This was achieved by using the \llama 70B prompt rewrite model as the teacher to conduct inference on a large prompt pool, and selecting high-quality rewrite pairs through human evaluations following the quality guideline.
Finally, we finetuned a 8B \llama model on the HITL prompt rewrite pairs to obtain the final prompt rewrite model to reduce the latency burden to the whole system.

\subsubsection{Improving Inference Efficiency}%
\label{sec:linear_quad_sampler}

\begin{figure}[t!]
  \centering
  \begin{tabular}{C{0.45\linewidth}C{0.45\linewidth}}
    \includegraphics[width=0.39\textwidth]{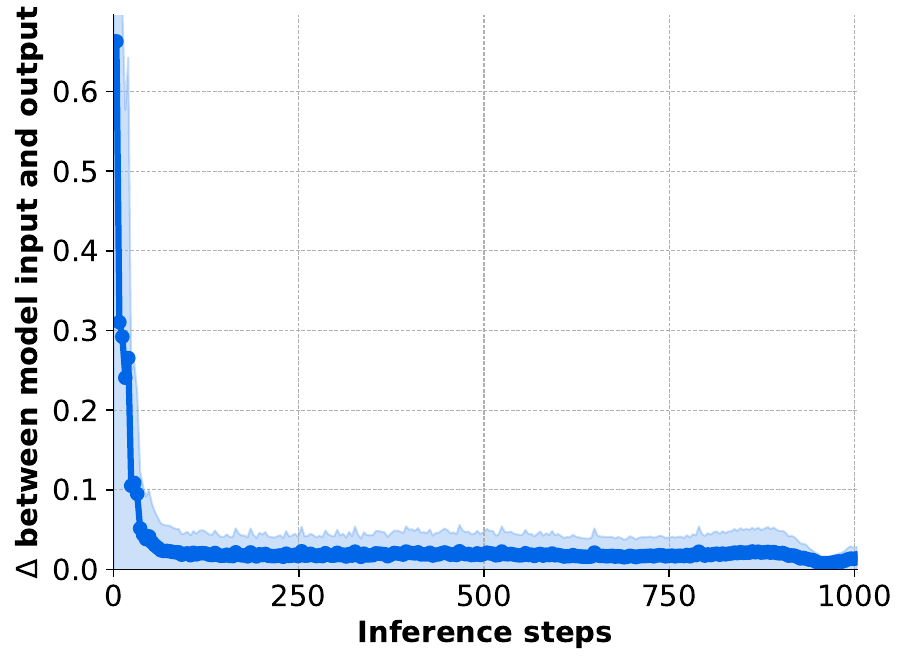} &
    \includegraphics[width=0.39\textwidth]{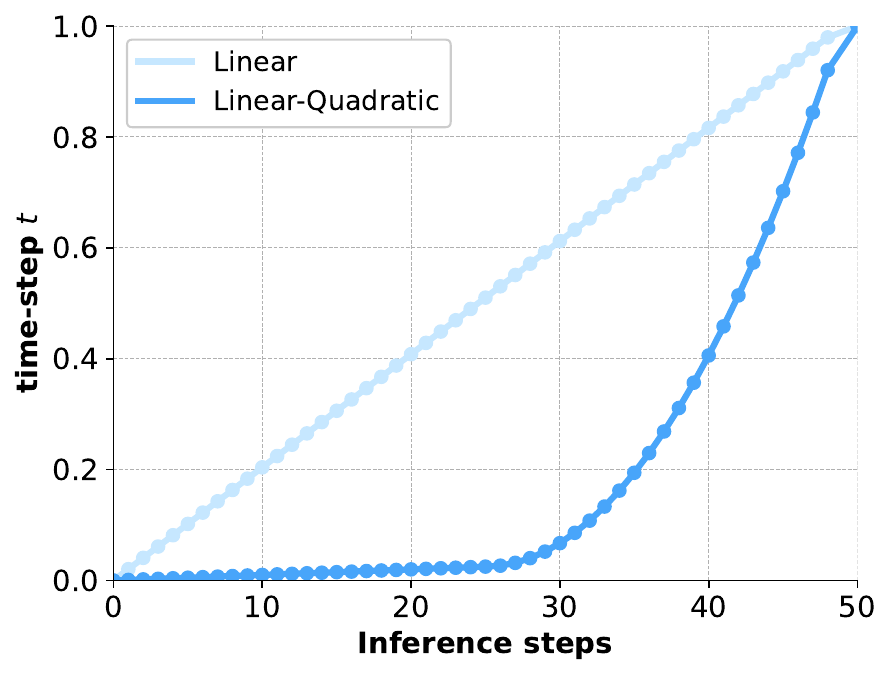} \\
 (a) Average change in model outputs between inference steps & (b) The linear-quadratic \timeschedule
  \end{tabular}%
  \captionof{figure}{\textbf{The linear-quadratic \timeschedule.}
 (a) Average change in transformer block input \vs output across inference steps, using a 1000-step linear \timeschedule.
 (b) A 50-step linear and linear-quadratic \timeschedule.
 Since the biggest changes in the model block inputs/outputs occur at early \timesteps,
 our scheduler copies the first 25 low \timesteps and bridges the remaining \timesteps
 with 25 quadratic steps.
 }
  \label{fig:deltas}
\end{figure}

To sample videos efficiently,
we use the Euler sampler with a unique \timeschedule tailored to our model.
Empirically, we found that Euler outperforms higher-order solvers like midpoint~\citep{atkinson1991introduction} or adaptive solvers like Dopri5~\citep{dormand1980family}.
We observed that reducing the number of inference steps for video generation is more challenging than for image generation due to the additional time dimension, \ie, the quality and prompt alignment of the generated motion are more sensitive to the number of inference steps compared to static images.
For example, videos generated with 250, 500, or 1000 linear steps show noticeable differences in scene composition and motion quality.
While techniques such as distillation~\citep{salimans2022progressive,kohler2024imagine} can be used to speed up the model inference, they require additional training.
Next, we show a simple inference-only technique that can lead up to $\sim20\times$ speed up with a few lines of code.

We found that we can closely approximate the quality of an $N$-step video generation process with merely 50 steps by implementing a linear-quadratic \timeschedule.
This approach follows the first 25 steps of an $N$-step linear schedule and
then approximates the remaining $N-25$ steps with 25 quadratically placed steps.
For example, a video generated with 1000 linear steps can be precisely emulated by 25 linear steps followed by 25 quadratic steps,
whereby the linear steps are identical to the first 25 linear steps of a 1000-step linear schedule.
The linear-quadratic strategy is predicated on the observation that the first inference steps are pivotal in setting up the scene and motion of the video.
This is visualized in~\cref{fig:deltas}, where we plot the average change between the input and output of each transformer block at every inference step.
Similar behavior is observed in the diffusion-based fast video model PAB~\citep{zhao2024pab},
where the average per-step difference of attention blocks follows a U-shaped pattern,
compared to the L-shaped curve in~\cref{fig:deltas}.
Since most changes occur in the first solver steps,
taking over the first linear steps of an $N$-step schedule followed by much bigger steps is enough to approximate the full $N$-step result.
The quadratic spacing of the latter steps is crucial, as it emphasizes the importance of the early stages in the flow-matching sequence.
In practice, we use a 50-step linear-quadratic schedule emulating $N=250$ linear steps for optimal results.

\subsection{Evaluation} \label{T2V_model_evaluation}
In this section, we explain how we evaluate the \textToV quality of \OursVideo and other models.
Our goal is to establish clear and effective evaluation metrics that identify a model's weaknesses and provide reliable feedback.
We explain the different \textToV evaluation axes and their design motivations in~\cref{sec:T2V_evaluation_axes}.
We introduce our new benchmark, \textToVBenchmarkName, in~\cref{sec:t2v_eval_benchmark}.
Throughout this work, we use human evaluation to assess the quality of generated videos across various evaluation axes.
When evaluating each axis, we conduct pairwise A/B tests where expert human evaluators assess two videos side-by-side.
Evaluators are instructed to choose a winner based on the axis being measured, or to declare a tie in case of no clear winner.
We include a discussion on the motivation and reliability of using human evaluations and on existing automated metrics in~\cref{sec:t2v_eval_discussion}.

\subsubsection{Evaluation Axes} \label{sec:T2V_evaluation_axes}

Evaluating \textToV generation presents unique challenges compared to \textToI tasks, primarily due to the added complexity of the temporal dimension.
For a video to be considered high quality, it must stay faithful to the provided text prompt, maintain a high visual quality across frames without noticeable flaws, and be visually appealing with a photorealistic style.
To assess these factors, we evaluate the quality of generated videos across three main axes: (1) \faithfulnessFull, (2) \qualityFull, and (3) realness and aesthetics. Each axis, along with their fine-grained sub-axes, is described in details below and summarized in~\cref{tab:eval_t2v_axes_list}.

\begin{table}[b]
    \centering
    \setlength{\tabcolsep}{4pt}
    \resizebox{\linewidth}{!}{%
    \begin{tabular}{cc|cccc|cc}
        \toprule
        \multicolumn{2}{c}{\faithfulnessFull} & \multicolumn{4}{c}{\qualityFull} & \multicolumn{2}{c}{Realness \& Aesthetics} \\
        \midrule
        Subject \& Motion alignment & & Overall & Frame consistency & Motion Completeness & Motion Naturalness & Realness & Aesthetics \\
         \bottomrule
    \end{tabular}
    }
    \caption{\textbf{Evaluation axes for \textToV generation.}
    We evaluate video generations across 3 axes, each of which is composed of multiple fine-grained sub-axes.
    \faithfulnessFull evaluates the `alignment' between the input text prompt and the video.
    \qualityFull, Realness \& Aesthetics evaluate the quality of the generated video independent of the input text prompt.
    }
    \label{tab:eval_t2v_axes_list}
\end{table}

\noindent \textbf{\faithfulnessFull.}
This axis measures how well a generated video aligns with the provided prompt.
An input prompt can include wide-ranging descriptions of subject appearance, motion, background, camera motion, lighting and style, visual text, \etc.
Human evaluators are asked to pay close attention to these specific aspects and select the video that aligns more closely with the prompt.
To provide more nuanced feedback, evaluators are also asked to specify their reasoning based on two orthogonal sub-axes: \textbf{Subject match}: This measures alignment of subject appearance, background, lighting and style; and \textbf{Motion match}: This measures the alignment of motion-related descriptions.

\noindent \textbf{Visual Quality.}
Compared to visual quality in \textToI generation, much of the perceived quality in generated videos stems from the quality of motion -- a video-specific dimension.
Therefore, in \textToV visual quality evaluation, we focus on measuring the model's ability to generate consistent, natural, and sufficient amounts of motion in the output videos.
To capture these critical aspects, we propose the following four sub-axes, which we outline below.

\begin{itemize}
    \item \textbf{Frame consistency}:
    This metric assesses the temporal consistency of generated content. Violations of frame consistency can manifest as morphing-like artifacts, blurred or distorted objects, or content that abruptly appears or disappears.
    We consider frame consistency a crucial measure of the model's ability to understand object framing and relationships in motion, as inconsistencies or distortions often arise when the model fails to accurately represent interactions between objects or their environment.
    Additionally, frame consistency reflects the model's capacity to handle challenging tasks, such as prompts that require fast-moving content, \eg, in sports scenarios, where maintaining consistent appearance is especially difficult; or reasoning about occlusions, \eg, objects re-appearing after being occluded.

    \item \textbf{Motion completeness}:
    This measures whether the output video contains enough motion. A lack of motion completeness may occur when the prompt involves out-of-distribution or unusual subjects (\eg, monsters, ghosts) or real-world objects performing unusual activities (\eg, people flying, pandas playing piano).
    Due to limited training data for such scenarios, the model may struggle to generate enough amount of motion, resulting in either static videos or those with only camera movement.
    Motion completeness evaluates the magnitude of motion in the video.
    A win on this axis indicates a greater amount of motion, even if it includes distortion, fast motion, or appears unnatural.

    \item \textbf{Motion naturalness}:
    This metric assesses the model's ability to generate natural and realistic motion, demonstrating a solid understanding of real-world physics.
    It covers aspects such as natural limb movements, facial expressions, and adherence to physical laws. Motion that appears unnatural or uncanny will be penalized.

    \item \textbf{Overall quality}:
    For a given pair of videos being compared, the above three metrics might not result in the same winner.
    To resolve this, we introduced the overall quality sub-axis, where human evaluators are asked to pick the winning video that has better ``overall'' quality given the previous three sub-axes.
    This is a holistic metric that asks the human annotators to use their perception and to balance the previous signals to capture overall how good a generated video is.

\end{itemize}

\noindent \textbf{Realness \& Aesthetics.}
Realness and aesthetics evaluate the model's ability to generate photorealistic videos with aesthetically pleasing content, lighting, color, style, \etc.
We ask human evaluators to evaluate along two sub-axes:

\begin{itemize}
    \item \textbf{Realness}:
    This measures which of the videos being compared most closely resembles a real video. For \textit{fantastical} prompts that are out of the training set distribution (\eg, depicting fantasy creatures or surreal scenes), we define realness as mimicking a clip from a movie following a realistic art-style.
    We additionally ask the evaluators to select a reason behind their choice \ie, ``subject appearance being more realistic'' or ``motion being more realistic''.
    \item \textbf{Aesthetics}:
    This measures which of the generated videos has more interesting and compelling content, lighting, color, and camera effects.
    Again, we ask the evaluators to provide details justifying their choice from ``content being more appealing/interesting'', and ``lighting/color/style being more pleasing''.
\end{itemize}

\subsubsection{Evaluation benchmark}
\label{sec:t2v_eval_benchmark}
\begin{figure}[t!]
    \centering
    \setlength{\tabcolsep}{10pt}
    \adjustbox{max width=\textwidth}{%
    \begin{tabular}{C{0.5\textwidth}C{0.4\textwidth}}
        \multirow{3}{*}{
        \includegraphics[width=0.5\textwidth]{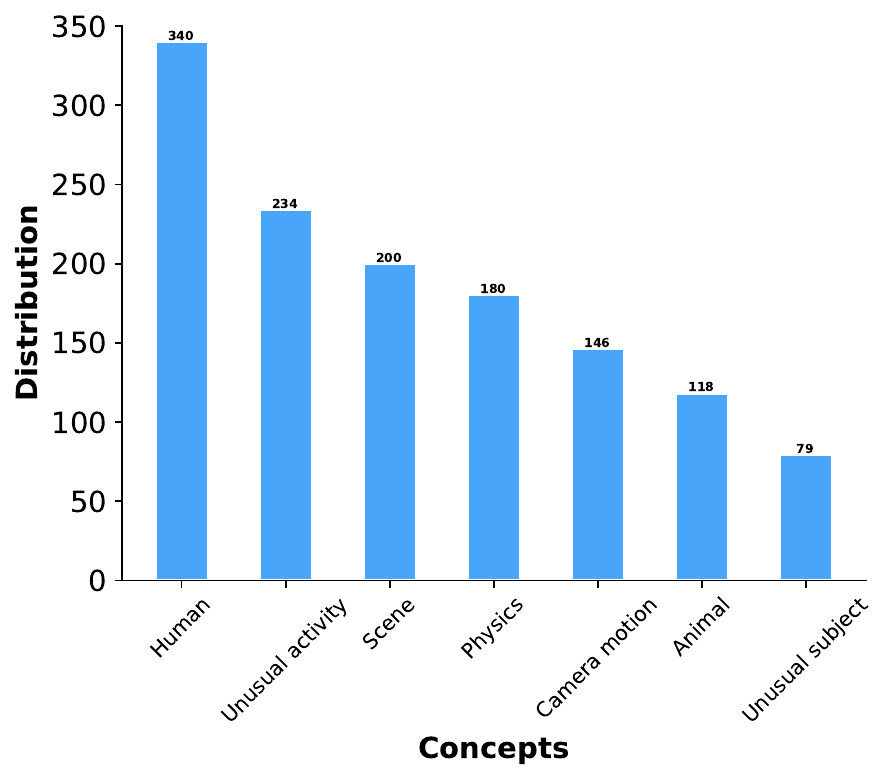}} & (b) Nouns in evaluation set \\
        & \includegraphics[width=0.36\textwidth,trim={3cm 1cm 2.5cm 1cm},clip]{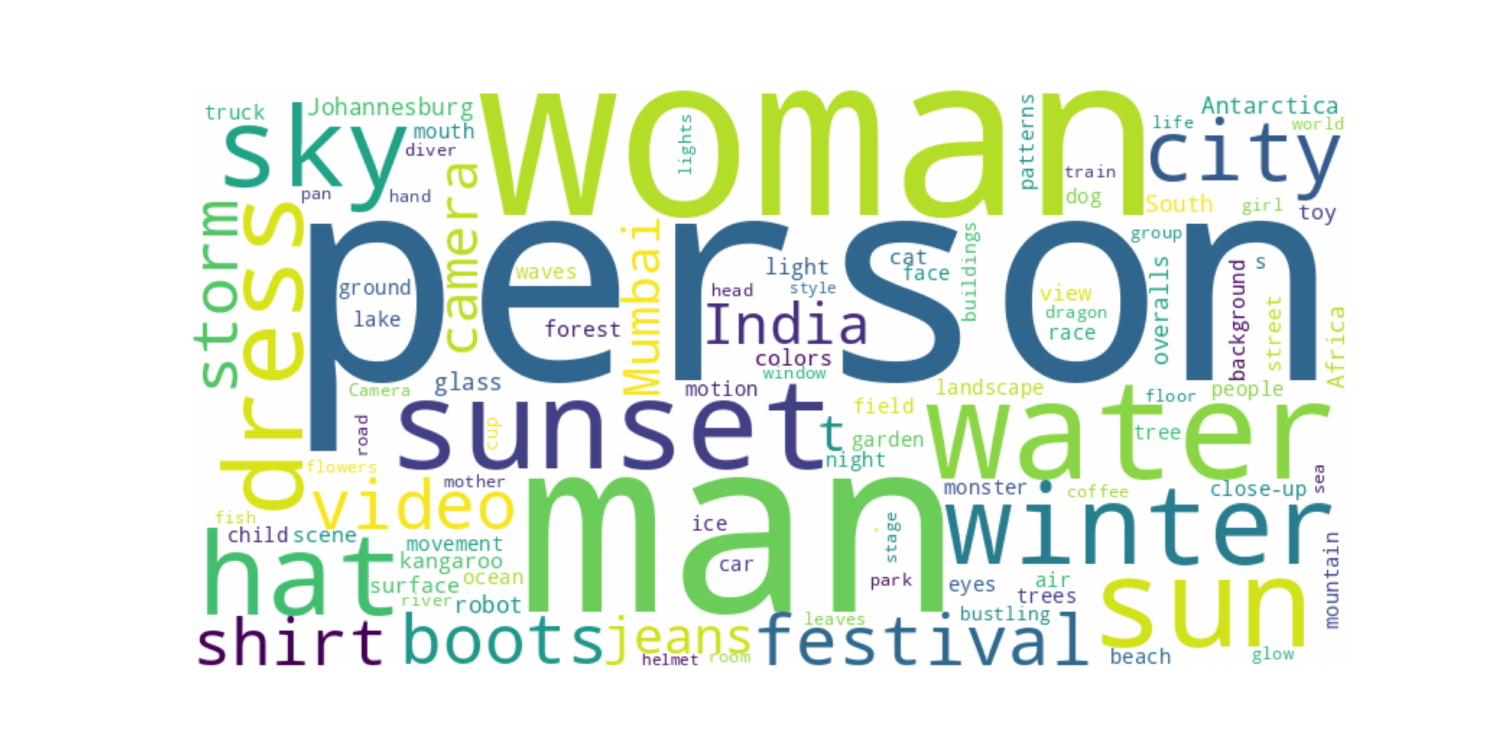} \\
        & \includegraphics[width=0.36\textwidth,trim={3cm 0 2.5cm 1cm},clip]{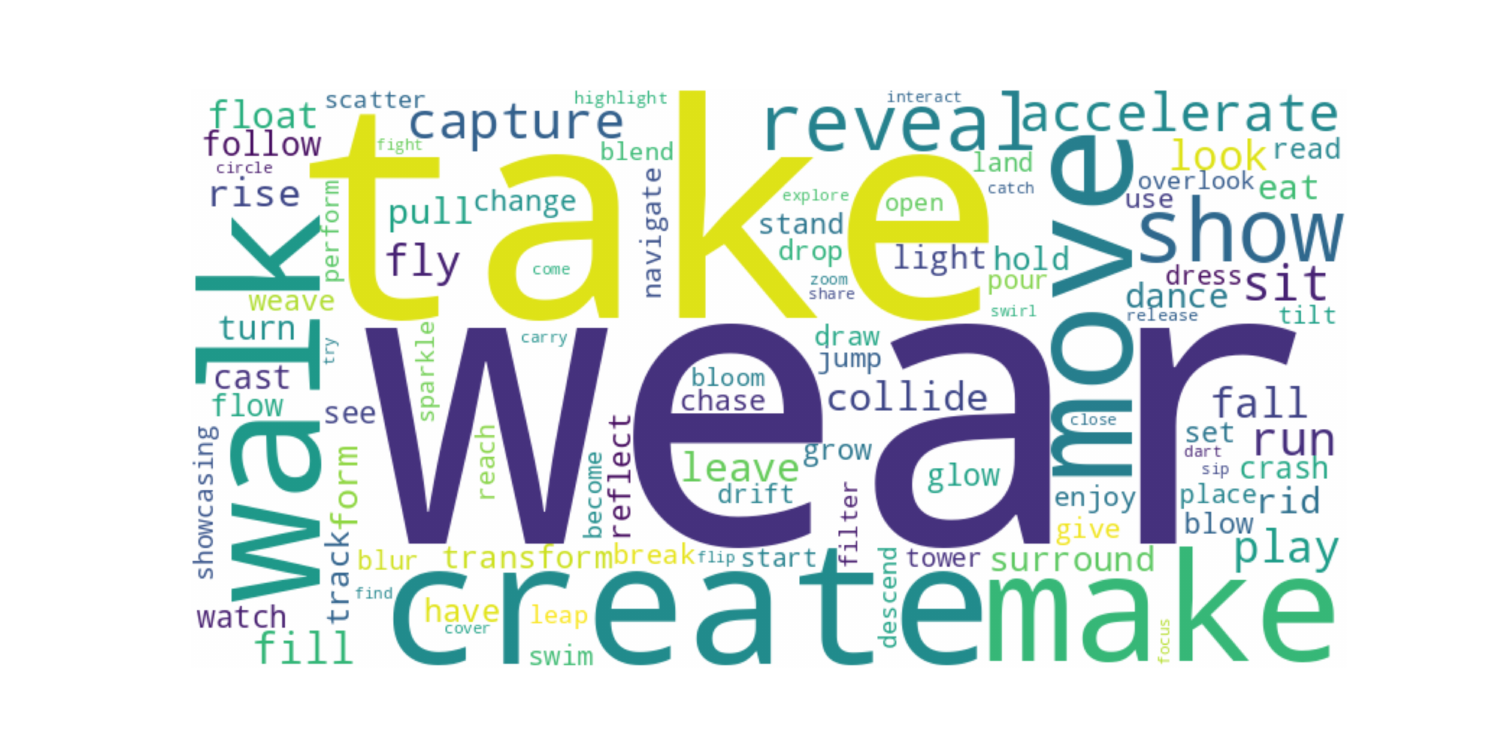} \\
        (a) Prompt concept distribution & (c) Verbs in evaluation set
    \end{tabular}}
    \vspace{-8pt}
    \caption{\textbf{Summary visualization of evaluation prompts in \textToVBenchmarkName.} (a) Since each prompt may encompass multiple testing concepts, we assign one or more concept tags to each prompt. As a result, the total count in the distribution exceeds 1003. (b), (c) Visualization of the common nouns and verbs in the evaluation set, respectively.}
    \label{fig:eval_prompt_distribution_main}
  \end{figure}

In order to thoroughly evaluate video generations, we publicly release a benchmark, \textToVBenchmarkName\footnote{\label{f_note_t2vb_1}\url{https://github.com/facebookresearch/MovieGenBench}}, which consists of 1003 prompts that cover all the different testing aspects summarized above.
In order to enable fair comparison to \OursVideo by future work, we also release non cherry picked generated videos from \OursVideo on \textToVBenchmarkName.
Our benchmark is more than $3\times$ larger than the prompt sets used in prior work~\citep{singer2023makeavideo,emuvideo2023} for evaluation.
We specifically include prompts capturing the following concepts of interest: 1) human activity (limb and mouth motion, emotions, \etc), 2) animals, 3) nature and scenery, 4) physics (fluid dynamics, gravity, acceleration, collisions, explosions, \etc), 5) unusual subjects and unusual activities.
To test the generation quality at different motion levels, we also tag each prompt with high/medium/low motion.
We show examples of evaluation prompts used in~\cref{tab:eval_prompts} and show the distribution of evaluation prompts across concepts in~\cref{fig:eval_prompt_distribution_main}.
\begin{table}[!t]
    \centering
    \adjustbox{max width=\textwidth}{%
    \begin{tabular}{ccc}
        \toprule
        Prompt & Concept & Motion level \\
        \midrule
        A marathon runner crossing the finish line after a grueling race. & human activity & high \\
        A curious cat peering out from a cozy hiding spot. & animals & medium \\
        A peaceful countryside with rolling hills and a setting sun. & nature \& scenery & low \\
        A high-speed video of champagne being poured into a glass, with bubbles rising rapidly. & physics -- fluid dynamics & high \\
        A cute golden dragon walking like a model on stage, as the audience claps for him. & unusual subjects & low \\
        \bottomrule
    \end{tabular}}%
    \caption{\textbf{Sample evaluation prompts from our proposed \textToVBenchmarkName benchmark.} We sample prompts that cover a wide range of concepts, with varying motion levels.}
    \label{tab:eval_prompts}
\end{table}
We evaluate the model quality on the entire evaluation prompt set as well as break down the quality by individual testing metrics.
Prompts involving unusual subjects and unusual motion help test model's ability to generalize to out-of-distribution content.

\subsubsection{Evaluation Discussion} \label{sec:t2v_eval_discussion}
Here, we motivate our decision for using human evaluation as opposed to automated metrics.

\noindent \textbf{Necessity of human evaluations for video generation.}
The motivation for choosing human evaluation stems from the complexity of evaluating video generation.
For \faithfulnessFull, evaluating the alignment of motion over time requires understanding how actions unfold and evolve in relation to the prompt.
Humans are particularly skilled at recognizing temporal coherence and handling ambiguities when the context is abstract or complex, whereas automated methods may only capture static frame-level correspondences.
When evaluating visual quality, such as motion naturalness or detecting inconsistencies in object appearance across frames, humans excel due to their innate understanding of real-world physics and object behavior.
Similarly, assessing realness and aesthetics heavily depends on human perception and preference.
Across all three tasks, we find that existing automated metrics struggle to provide reliable results, reinforcing the need for human evaluation.

\noindent \textbf{Reliability.}
An important aspect concerning reliability in evaluations is the randomness introduced both on the modeling side due to the probabilistic nature of the generative models, and the human evaluation side due to the annotation variance.
Defining objective criteria to measure generations remains challenging and humans can still be influenced by other factors such as personal biases or preferences.
We describe our efforts to reduce evaluation variance and increase the reliability of the human evaluations.
We take four key steps towards minimizing evaluation variance:
(1) We provide human evaluators with detailed evaluation guidelines and video examples, narrowing the definitions of evaluation axes and sub-axes to minimize subjectivity.
Additionally, inspired by the JUICE metric~\citep{emuvideo2023}, we found that asking evaluators to indicate the reason of their choices helps reduce annotation variance and improve agreement among evaluators.
(2) We evaluate the models over a large set of prompts (\eg, 1003 for \textToVBenchmarkName, $3\times$ larger than~\citep{singer2023makeavideo,emuvideo2023}) covering a wide variety of concepts.
(3) We use a majority vote system, with a majority vote from three annotations for each \faithfulnessFull and \qualityFull question, and a majority vote from six annotations for realness and aesthetic questions, as these are more subjective.
(4) We conduct thorough and frequent audits of human annotations to resolve edge cases and correct mislabelings.

\noindent \textbf{Automated Metrics for text-to-video evaluation.}
Prior works in text-to-video generation have relied upon automated metrics for assessing video quality.
Similar to recent studies~\citep{dai2023emu,podell2023sdxl,singer2023makeavideo,emuvideo2023,ho2022imagen,barratt2018note,chong2020effectively,ge2024content,huang2023vbench} we find that automated metrics such as FVD~\citep{unterthiner2019fvd} and IS~\citep{salimans2016improved} do not correlate with human evaluation scores for video quality, and hence do not provide useful signal for model development or comparison.
Some prior works utilize discriminative models for generated media evaluation axes, including text faithfulness with CLIP~\citep{radford2021learning}.
One key limitation of such automated metrics is that they are inherently limited by the performance of the underlying discriminative model~\citep{rambhatla2023selfevalleveragingdiscriminativenature}.
A key challenge in using discriminative models for evaluating text-to-video generation is the inavailability of sufficiently effective and expressive video-text discriminative models.
We note that other interesting automated metrics for generated video evaluation exist, such as those based on structure-from-motion~\citep{Li2024SoraGV}, that we did not explore the use of here.

\noindent \textbf{Enabling fair comparison to \OursVideo.} To enable fair and easy comparison to \OursVideo for future works, we publicly release our non cherry picked generated videos from \OursVideo on the \textToVBenchmarkName prompt set.

\subsection{Results}
In this section, we describe the experiments and results for \OursVideo.
We first include comparisons to prior work for \textToV generation in~\cref{sec:t2v_main_results}.
We ablate key design decisions for \OursVideo in~\cref{sec:ablations_t2v}.
We include key results and ablations for the \taeShort in~\cref{sec:tae_results}, and an evaluation of the \upsampler in~\cref{sec:spatial_upsampler_results}.
We include comparisons to prior work for \textToI generation in~\Cref{sec:t2i_generation}.

\subsubsection{Comparisons to Prior Work}
\label{sec:t2v_main_results}
\begin{table}[!t]
    \centering
    \begin{tabular}{cccccc}
        \toprule
         & \multicolumn{4}{c}{\OursVideo net win rate \vs. prior work} \\
         & \RunwayGen & \LumaLabs & \Sora & \Kling & $\sigma$ \\
         \midrule
         Overall Quality & 35.02 & 60.58 & 8.23 & 3.87 & $\pm$5.07 \\
         \midrule
         Consistency & 33.1 & 42.14 & 8.22 & 13.5 & $\pm$4.08 \\
         Motion Naturalness & 19.27 & 29.33 & 4.43 & 0.52 & $\pm$3.98 \\
         Motion Completeness & -1.72  & 23.59 & 8.86 & -10.04 & $\pm$1.68 \\
         \midrule
         \faithfulnessFull & 10.45 & 12.23 & 17.72 & -1.99 & $\pm$3.74 \\
         \midrule
         Realness & 48.49 & 61.83 & 11.62 & 37.09 & $\pm$2.52 \\
         Aesthetics & 38.55 & 48.19 & 6.45 & 26.88 & $\pm$4.84 \\
         \bottomrule
    \end{tabular}
    \caption{\textbf{\OursVideo \vs prior work}.
    The comparison uses either generated videos from the \textToVBenchmarkName prompt set (\RunwayGen, \LumaLabs, \Kling) or prompts from publicly released videos on their website (\Sora). A detailed summary of information from prior work is shown in~\cref{tab:t2v_competitors}. We measure the net win rate (win\% - loss\% of our model) which has a range of $[-100\%,100\%]$. To assess statistical significance, we perform an annotation variance analysis (\cref{appendix_sec:variance}), with the net win standard deviation, $\sigma$, indicated in the table above. A significant win/loss is identified when the net win rate is beyond 2$\sigma$ (95\% CI), a moderate win/loss within 1--2 $\sigma$ (68\% CI), and performance is considered on par within 1$\sigma$.
    }
    \label{tab:t2v_main_eval}
\end{table}

Where possible, we obtain non cherry picked generated videos from the \textToVBenchmarkName prompt set for the prior work methods, and compare to these using non cherry picked videos from \OursVideo for the same prompts.
This includes the black-box commercial models that offer API access through their website: \RunwayGen~\citep{gen3}, \LumaLabs~\citep{luma}, \Kling~\citep{kling}.
We also compare to closed source \textToV methods (\Sora), where our only option is to compare to them using the prompts and videos from their publicly released examples.
Note that the publicly released videos for closed source methods are likely to be `best' representative samples obtained through cherry picking.
Hence for fair comparison, we compare to \Sora by methodically manually choosing one video from five generated options from \OursVideo for each prompt.
One additional challenge when comparing to prior work is that each method generates videos at different resolutions and aspect-ratios.
We reduce annotator bias~\citep{emuvideo2023} by downsampling \OursVideo's videos for each comparison such that they match in these aspects.
Full details on this postprocessing and the \Sora comparison can be found in~\cref{appendix_prior_work_details}.

We compare \OursVideo to prior work for \textToV generation on different evaluation axes described in~\cref{T2V_model_evaluation}.
The results are shown in~\cref{tab:t2v_main_eval}.
We report the net win rate of our model, which can lie in the range $[-100,100]$.
On overall quality, \OursVideo strongly outperforms \RunwayGen (35.02\%) and \LumaLabs with the net win rate beyond $2\sigma$.
Our generations moderately net win over \Sora (8.23\%) (net win rate within 1-2 $\sigma$) and are on par with \Kling (3.87\%).
Against \RunwayGen, \LumaLabs and \Sora, we see that \OursVideo either outperforms or is on par across all quality breakdown axes, including large net wins against \RunwayGen on motion naturalness (19.27\%) and frame consistency (33.1\%), against Sora on frame consistency (8.22\%) and motion completeness (8.86\%).
These significant net wins demonstrate \OursVideo's ability to simulate the real world with generated videos that respect physics, with motion that is both reasonable in magnitude but consistent and without distortion.
Against \Kling, we see that \OursVideo significantly net wins in frame consistency (13.5\%) but loses on motion completeness (-10.04\%).
We note that this large motion completeness paired with poor frame consistency shows \Kling's tendency to occasionally generate unnaturally large motion with distortion.
As indicated in \cref{sec:T2V_evaluation_axes}, motion completeness only evaluates the magnitude of motion in the video, regardless of distortion, fast motion, or being unnatural.

On realness and aesthetics, \OursVideo significantly outperforms \RunwayGen, \LumaLabs and \Kling on both metrics, with 48.49\%, 61.83\% and 37.09\% net win rates on realness, respectively.
Compared to \Sora, \OursVideo has a significant win on realness with 11.62\% net win rate beyond $2\sigma$ and a moderate win over \Sora on aesthetics with 6.45\% net win rate within 1--2 $\sigma$.

This demonstrates \OursVideo's ability to generate photorealistic and visually compelling content.
For text faithfulness, \OursVideo outperforms \Sora, \RunwayGen, \LumaLabs and is on par with \Kling.

Several generated videos from \OursVideo are shown in~\cref{fig:t2v_qual_ours_figure}.
\OursVideo is able to generate high quality videos for both natural prompts (see~\cref{fig:t2v_qual_ours_figure}) and out-of-distribution prompts describing fantastical scenes from outside of the training set distribution (see~\cref{fig:main_figure}).
The generated videos contain complex motion, depicting detailed content over the video's duration \eg, a firefighter running into and then out of a burning forest or a puppy searching for, finding its owner, and continuing its quest (see~\cref{fig:t2v_qual_ours_figure}).

Qualitative comparisons between \OursVideo and prior work are shown in~\cref{fig:t2v_qual_prior_work_4way_figure} and~\cref{fig:t2v_qual_prior_work_sora_figure}.
As shown, \OursVideo generates realistic and high quality videos with natural-looking motion that is well aligned to the text prompt.
\OursVideo generates objects and identities that are consistent over the entire duration of the video, and that obey the laws of physics.
Differently, the prior work can struggle to generate videos that are simultaneously high quality and with good text-alignment.

\begin{center}
    \centering
    \captionsetup{type=figure}
    \setlength{\tabcolsep}{1pt}
\adjustbox{max width=\textwidth}{%
\centering
\begin{tabular}{cccc}
    \multicolumn{4}{c}{\textit{Prompt}: A child who discovers an ancient relic that allows them to talk to animals} \\
    \includegraphics[width=0.25\linewidth]{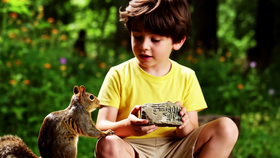} &
    \includegraphics[width=0.25\linewidth]{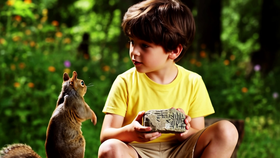} &
    \includegraphics[width=0.25\linewidth]{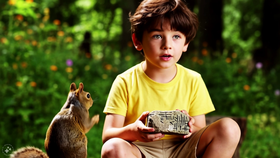} &
    \includegraphics[width=0.25\linewidth]{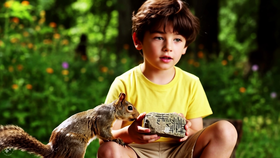} \\

    \multicolumn{4}{c}{\textit{Prompt}: Firefighter running through a burning forest} \\
    \includegraphics[width=0.25\linewidth]{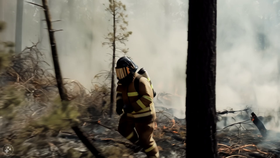} &
    \includegraphics[width=0.25\linewidth]{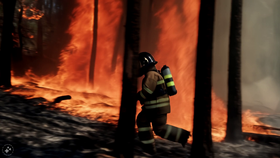} &
    \includegraphics[width=0.25\linewidth]{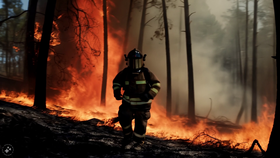} &
    \includegraphics[width=0.25\linewidth]{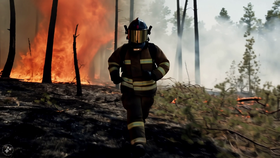} \\

    \multicolumn{4}{c}{\textit{Prompt}: A lost puppy that leads its finder on an epic quest} \\
    \includegraphics[width=0.25\linewidth]{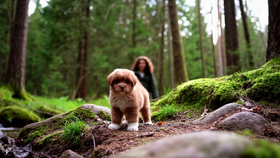} &
    \includegraphics[width=0.25\linewidth]{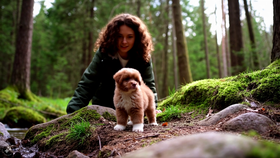} &
    \includegraphics[width=0.25\linewidth]{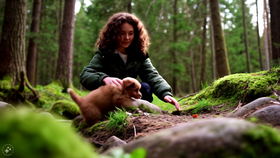} &
    \includegraphics[width=0.25\linewidth]{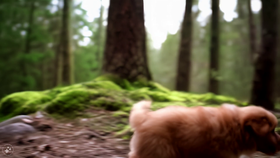} \\

\end{tabular}}

    \vspace{-3mm}
    \caption{\textbf{Generated videos from \OursVideo.} 
    \OursVideo generates high quality, visually compelling videos that are aligned to complex text prompts.
    Videos in this Figure found at \url{https://go.fb.me/MovieGen-Figure12}.}
    \label{fig:t2v_qual_ours_figure}
\end{center}%

\begin{center}
    \centering
    \captionsetup{type=figure}
    \setlength{\tabcolsep}{1pt}
\adjustbox{max width=\textwidth}{%
\centering
\begin{tabular}{cccc}
    \multicolumn{4}{c}{\textit{Prompt}: A computer mouse with legs running on a treadmill} \\
    \multicolumn{4}{c}{\textbf{\OursVideo }} \\
    \includegraphics[width=0.25\linewidth]{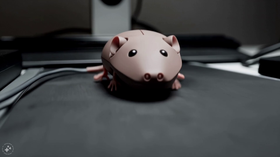} &
    \includegraphics[width=0.25\linewidth]{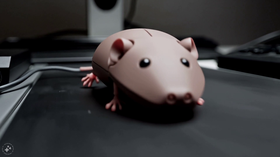} &
    \includegraphics[width=0.25\linewidth]{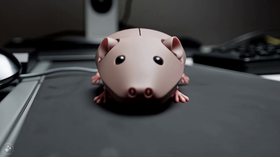} &
    \includegraphics[width=0.25\linewidth]{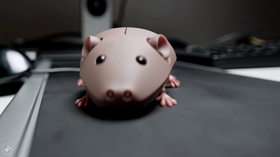} \\

    \multicolumn{4}{c}{\textbf{\RunwayGen }} \\
    \includegraphics[width=0.25\linewidth]{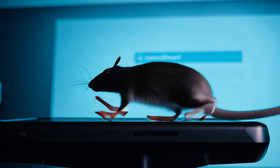} &
    \includegraphics[width=0.25\linewidth]{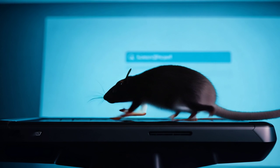} &
    \includegraphics[width=0.25\linewidth]{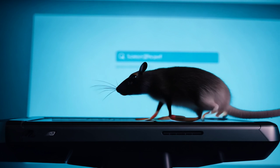} &
    \includegraphics[width=0.25\linewidth]{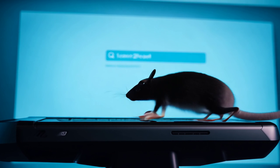} \\

    \multicolumn{4}{c}{\textbf{\LumaLabs }} \\
    \includegraphics[width=0.25\linewidth]{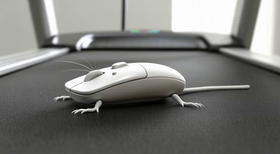} &
    \includegraphics[width=0.25\linewidth]{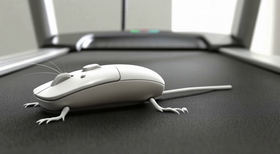} &
    \includegraphics[width=0.25\linewidth]{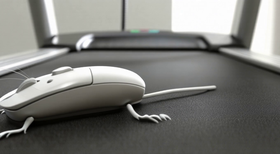} &
    \includegraphics[width=0.25\linewidth]{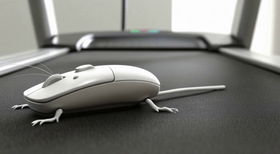} \\

    \multicolumn{4}{c}{\textbf{\Kling }} \\
    \includegraphics[width=0.25\linewidth]{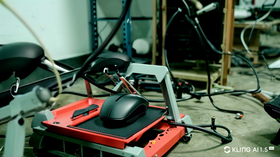} &
    \includegraphics[width=0.25\linewidth]{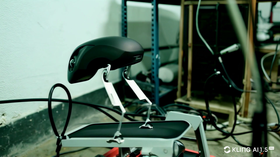} &
    \includegraphics[width=0.25\linewidth]{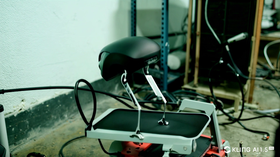} &
    \includegraphics[width=0.25\linewidth]{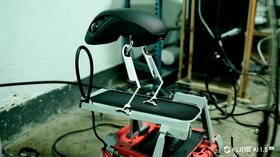} \\
\end{tabular}}

    \vspace{-3mm}
    \caption{\textbf{Qualitative comparisons between \OursVideo and prior work.}
    Here, we show generated videos for the same prompt for \OursVideo, \RunwayGen, \LumaLabs, and \Kling.
    Unlike prior work, \OursVideo generates videos that are high quality, with natural, realistic motion, and that are aligned to the text prompt.
    Videos in this Figure found at \url{https://go.fb.me/MovieGen-Figure13}.
    }
    \label{fig:t2v_qual_prior_work_4way_figure}
\end{center}%

\begin{center}
    \centering
    \captionsetup{type=figure}
    \setlength{\tabcolsep}{1pt}
\adjustbox{max width=\textwidth}{%
\centering
\begin{tabular}{cccc}
    \multicolumn{4}{c}{\textit{Prompt}: a kangaroo in purple overalls and boots walking in Johannesburg during sunset} \\
    \multicolumn{4}{c}{\textbf{\OursVideo }} \\
    \includegraphics[width=0.25\linewidth]{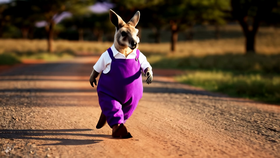} &
    \includegraphics[width=0.25\linewidth]{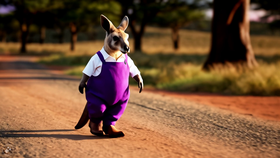} &
    \includegraphics[width=0.25\linewidth]{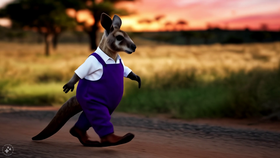} &
    \includegraphics[width=0.25\linewidth]{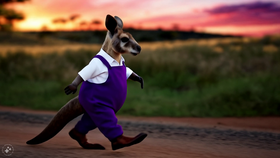} \\

    \multicolumn{4}{c}{\textbf{\Sora }} \\
    \includegraphics[width=0.25\linewidth]{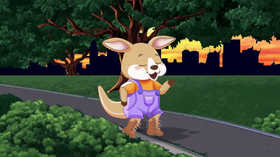} &
    \includegraphics[width=0.25\linewidth]{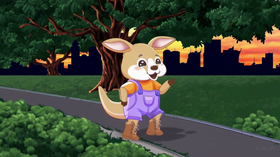} &
    \includegraphics[width=0.25\linewidth]{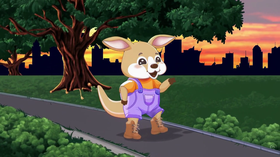} &
    \includegraphics[width=0.25\linewidth]{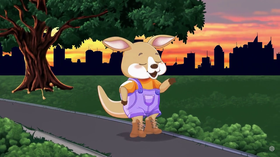} \\

    \midrule

    \multicolumn{4}{c}{\textit{Prompt}: a toy robot in a green dress and sun hat walking in Antarctica during a storm} \\
    \multicolumn{4}{c}{\textbf{\OursVideo }} \\
    \includegraphics[width=0.25\linewidth]{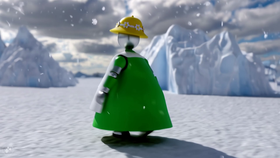} &
    \includegraphics[width=0.25\linewidth]{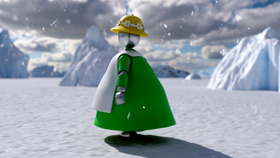} &
    \includegraphics[width=0.25\linewidth]{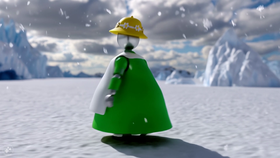} &
    \includegraphics[width=0.25\linewidth]{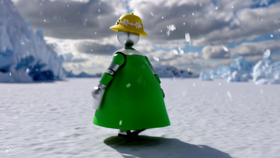} \\

    \multicolumn{4}{c}{\textbf{\Sora }} \\
    \includegraphics[width=0.25\linewidth]{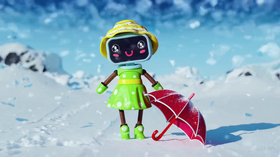} &
    \includegraphics[width=0.25\linewidth]{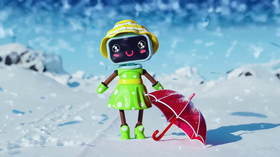} &
    \includegraphics[width=0.25\linewidth]{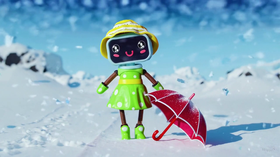} &
    \includegraphics[width=0.25\linewidth]{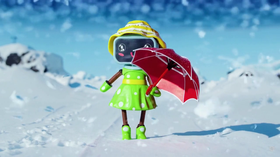} \\

\end{tabular}}

    \vspace{-3mm}
    \caption{\textbf{Qualitative comparisons between \OursVideo and prior work.}
    Here, we show two generated videos for the same prompt for \OursVideo and \Sora.
    \OursVideo generates natural-looking videos with realistic motion, even for the out-of-training-set-distribution prompts shown here.
    As shown here, for such prompts \Sora can tend to generate less realistic videos (\eg, the \textit{cartoonish} kangaroo in the second row), that can be missing the motion details described in the text prompt (\eg, the non-walking robot in the bottom row).
    Videos in this Figure found at \url{https://go.fb.me/MovieGen-Figure14}.
    }
    \label{fig:t2v_qual_prior_work_sora_figure}
\end{center}%

\par
\begin{figure}
    \begin{minipage}[hb]{.3\textwidth}
        \centering
        \includegraphics[width=\textwidth]{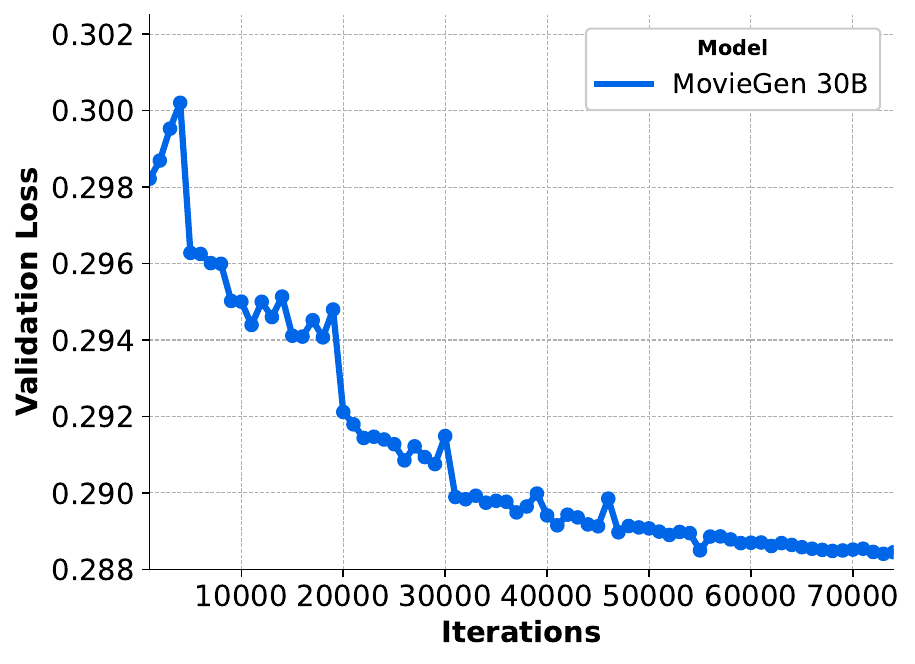}
    \end{minipage}\hfill
    \begin{minipage}[hb]{.69\textwidth}
        \centering
        \adjustbox{max width=\textwidth}{%
        \begin{tabular}{ccC{2.1cm}C{2.1cm}C{2.1cm}C{2.1cm}C{2.1cm}}
            \toprule
            \multirow{3}{*}{\shortstack{Model A\\iterations}} & \multirow{3}{*}{\shortstack{Model B\\iterations}} & \multicolumn{5}{c}{Model A net win rate \vs model B}\\
            & & Overall & Frame       & Motion       & Motion      & Text-\\
            & & Quality & Consistency & Completeness & Naturalness & alignment\\
            \midrule
            15.6k & 10.2k & 16.0 & 9.0 & 4.67 & 7.66 & 5.67\\
            19.2k & 15.6k & 13.66 & -0.67 & 5.34 & 5.66 & 3.34 \\
            37k & 26.4k & 0.39 & -4.3 & 0.78 & -1.95 & 3.13 \\
            59k & 54k & 6.33 & -1.33 & 8.34 & 3.0 & 8.0\\
            \bottomrule
        \end{tabular}}
    \end{minipage}
    \caption{\textbf{Validation loss correlates with human evaluations.}
        Left: We plot the validation loss for the 768 px stage \textToIVShort \pretraining.
        We decrease the learning rate whenever the validation loss plateaus.
        Right: We observe that the validation loss is well correlated with human evaluation results of the corresponding checkpoints, especially for the text alignment and overall quality.
        }
    \label{fig:t2v_pt_eval_losses_and_eval}
\end{figure}

 \noindent \textbf{Correlation between validation loss and human evaluation.}
In~\cref{fig:t2v_pt_eval_losses_and_eval}, we show the validation loss for \OursVideo as a function of pretraining steps and observe that it decreases smoothly.
We take \pretrained checkpoints after every few thousand iterations and evaluate them in a pairwise comparison.
We observe that the validation loss is well correlated with human evaluation results as the later checkpoints with lower validation loss perform better in the evaluations.
This suggests that the \flowmatching validation loss can serve as a useful proxy for human evaluations during model development.

\noindent\textbf{Effect of finetuning.}
We leverage supervised finetuning, described in~\cref{sec:t2v_post_training} to further improve the video generation quality.
In~\cref{tab:t2v_pretrain_vs_finetune_1}, we compare the evaluation metrics between pre-trained model and finetuned models at 24 FPS with 10.6s video duration.
We find that finetuning leads to a significant improvement on both the \qualityFull and \faithfulnessFull metrics.

\begin{table}[!t]
    \centering
    \resizebox{0.8\columnwidth}{!}{%
    \begin{tabular}{ccccc}
         \toprule
        \multicolumn{5}{c}{Finetune net win rate \vs \pretrained} \\
        Overall Quality & Consistency & Motion Completeness & Motion Naturalness & Text Alignment \\
         \midrule
         34.65 & 8.14 & 18.38 & 10.5 & 9.97 \\
         \bottomrule
    \end{tabular}
    }
    \caption{\textbf{Comparison of finetuned and \pretrained models.}
    We conduct A/B comparison between finetuned and \pretrained models, where the scores are winning rates of finetune model minus the winning rate of the pretrained model, which shows that finetuning significantly improves over the pretrained model.
    The videos are evaluated at 24 FPS with 10.6s video duration.
    }
    \label{tab:t2v_pretrain_vs_finetune_1}
\end{table}

\subsubsection{Ablations} %
\label{sec:ablations_t2v}

\begin{table*}[!t]
    \centering
    \adjustbox{max width = \linewidth}{%
    \subfloat[
    \label{tab:ablate_t2v_training_objective}
    ]{
        \centering
        \begin{tabular}{c  cc}
            \toprule
            Method & \qualityShort & \faithfulnessShort \\
            \midrule
            FM \vs Diffusion & 16.53 & 7.08\\

             \bottomrule
        \end{tabular}%
    }
    \hfill
    \subfloat[
    \label{tab:ablate_t2v_captions}
    ]{
        \centering
        \begin{tabular}{c  cc}
            \toprule
            Method & \qualityShort & \faithfulnessShort \\
            \midrule
            Video \vs Image caption & $-0.80$ & $10.80$ \\
             \bottomrule
        \end{tabular}%
    }
    \hfill
    \subfloat[
    \label{tab:ablate_t2v_architecture}
    ]{
        \centering
        \begin{tabular}{c  cc}
            \toprule
            Method & \qualityShort & \faithfulnessShort \\
            \midrule
            \llama-like \vs DiT & 18.63 & 12.60 \\
             \bottomrule
        \end{tabular}%
    }

    }
    \caption{\textbf{Key design decisions in \OursVideo}. Each table shows the net win rate, in terms of the overall \quality (\qualityShort) and \faithfulnessFull (\faithfulnessShort), on adopting a design decision \vs a model that does not have it.
    See~\cref{sec:ablations_t2v} for details.
    }
    \label{tab:ablate_t2v}

\end{table*}

Here, we ablate the critical design decisions for \OursVideo.
For all ablations described in this section, we use a simpler, smaller baseline training and model setup than used for the main results.
We analyze the effect of each design decision quantitatively via text-to-video human evaluation on a subset of \textToVBenchmarkName containing 381 prompts, termed \textToVBenchmarkMiniName, and report results on text faithfulness and overall quality (see~\cref{T2V_model_evaluation}).
For each ablation, every aspect of the model except for the design decision being tested is held constant for fair comparison.
Next, we describe the simpler baseline setup followed by each ablation result.
Unless described otherwise here, all other settings for the ablation experiments follow our $30$B model including text encoders, flow matching objective, image training set, \etc.

\noindent\textbf{Baseline model setup for ablations.}
We use a $5$B parameter version of \OursVideo trained to produce $352\times192$ videos of $4$ -- $8$s.
We use the \taeShort described in~\cref{sec:tae}, which does $8\times$ compression across every spatio-temporal dimension to produce latents of shape $16 \times 24\times 44$.
This smaller \OursVideo model has $32$ layers in the transformer with $3072$ embedding dimension and $24$ heads.

\noindent\textbf{Baseline training setup for ablations.}
We use a two-stage training pipeline: (1) \textToI pretraining; (2) \textToI and \textToV joint training.
For simplicity, we used a smaller dataset of 21M videos, captioned with \llamaVideo $8$B, that have a constant landscape aspect ratio for video training.
First, we train the model on the image dataset with a learning rate of $0.0003$, a global batch size of $9216$ on $512$ GPUs for $96$K iterations.
Next, we perform joint \textToI and \textToV training with an iteration ratio of $0.02:1$ where the global batch size is $4096$ for images and $256$ for videos.
We use a learning rate of 5e-5 and train for $100$K iterations.

\noindent\textbf{Ablation Result - Training objective.}
We compare the \flowmatching training objective to the diffusion training objective.
Following~\citep{emuvideo2023}, we use the v-pred and zero terminal-SNR formulation of diffusion for training which is effective for video generation.
As the human evaluation results in~\cref{tab:ablate_t2v_training_objective} show, \flowmatching leads to better generations both in terms of overall quality and text alignment while controlling for all other factors.
Empirically, we also found this result to also hold across a range of model sizes and thus use \flowmatching to train our models.

\noindent\textbf{Ablation Result - Effect of video captions.}
As described in \ref{sec:pt-data}, our video generation model is trained using clips from real videos and \llamaVideo generated video clip captions.
To assess the importance of video captions, we compare our \llamaVideo 8B video captioning model to an image-based captioning scheme also based on \llamaNoVersion.
This image-based captioning model captions the first, middle, and last frame of the video clip and then uses \llamaNoVersion to rewrite these three image-based captions into a single video caption.
We refer to this model as \llamaRewrite.
We first compare the quality of the two captioning schemes with human evaluations based A/B testing.
Human raters are asked to pick between two given captions for the same clip.
\llamaVideo generated captions are preferred $67\%$ of the time while \llamaRewrite captions are only preferred $15\%$ of the time.
We visually observe that the video captioning model is able to accurately describe more fine-grained details regarding movements in the video.
These fine-grained details provide a stronger supervision signal for training the video generation model,
significantly improving overall prompt alignment by $10.8\%$ (\cref{tab:ablate_t2v}),
with most of the increase coming from motion alignment ($+10.7\%$) particularly on prompts that require ask for a high degree of motion in the output video ($+16.1\%$).

\noindent\textbf{Ablation Result - Model architecture.}
In our work, we choose a transformer architecture based on \llama (see~\cref{t2v_architecture_details}).
We compare this to a Diffusion Transformer~\citep{dit} based model, which is commonly used in the media generation literature~\citep{dit,sora,ma2024latte}.
The architecture differences between these two models can be seen in~\cref{tab:ablation_architecture_details}.
As shown in~\cref{tab:ablate_t2v_architecture}, we find that our \llama based architecture significantly outperforms the Diffusion Transformer on both quality (18.6\%) and text-alignment (12.6\%).
This significant result shows that the \llama architecture has an advantage over the commonly used DiT for media generation.
Our goal with \OURS is to scale to large model sizes, and to the best of our knowledge we find no detailed examples in the literature of scaling the Diffusion Transformer to very large scales.
This result demonstrates that we can confidently transition from the commonly used Diffusion Transformer to architectures more commonly used in LLMs such as \llama, the scaling behavior of which has been well documented~\citep{llama2,llama3}.

\begin{figure}
    \centering
    \includegraphics[width=\textwidth]{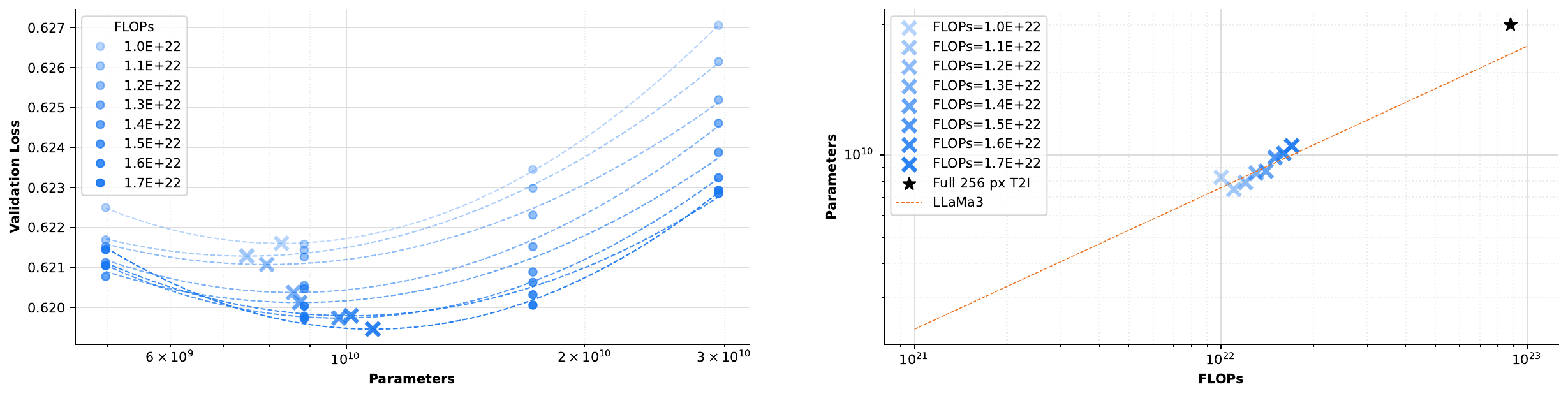}
    \captionof{table}{\label{fig:ablation_scaling}
    {\bf (Left) IsoFLOP curves} showing validation performance of different model sizes at different levels of compute. We compute the optimal model size for a given compute budget as the minima of the parabolas, represented using $\times$.
    {\bf (Right) Comparison to \llama scaling law.} We plot the optimal model size against the compute budget. We also overlay the scaling law as estimated in \llama~\citep{llama3}. Surprisingly, we find that our data points
    align very closely with the \llama scaling law, suggesting that it may serve as a reasonable predictor of model sizes and compute budgets, even for media generation models.
    }
\end{figure}

\noindent\textbf{Ablation Result - Model scaling behavior.}
To further understand the impact of our model architecture, we evaluate its scaling behavior. We experiment with four different instantiations of our model: 5B, 9B, 17B, and 30B. We train the 256px T2I stage for each of these models for varying amounts of compute, from $10^{22}$ to $1.7\times 10^{22}$ FLOPs. We use the same training data and hyperparameters as for the other ablation experiments for all models. We measure the validation loss for each compute budget for each model size, and plot it in~\cref{fig:ablation_scaling} (left). We fit the measured loss values using a second-degree polynomial, giving rise to the IsoFLOP curves shown in the figure. We identify the minimums of each parabola (represented in the figure using ``$\times$''), and refer to it as the compute-optimal model at the corresponding compute budget. We then plot these compute optimal models and corresponding compute budgets on~\cref{fig:ablation_scaling} (right). We overlay the scaling law for \llama~\citep{llama3} on this graph, and surprisingly, find that these optimal models align very closely with the \llama scaling law. We posit this scaling behavior of our model is likely due to the use of the \llama based transformer architecture. Most notably, this result suggests that \llama scaling laws may serve as a reasonable predictor of model sizes and compute budgets, even for media generation models.

\begin{table}[t]
    \centering
    \adjustbox{max width=\textwidth}{%
    \begin{tabular}{cccccc}
    \toprule
     & Normalization & Normalization eps & Normalization Affine & Activation Function & Bias in FC layers \\
    \midrule
    \OURS & RMSNorm  & 1e-5 &  True & SwiGLU & No	\\
    DiT & LayerNorm  & 1e-6 & False & SiLU & Yes  \\
    \bottomrule
    \end{tabular}}%
    \caption{\textbf{Architecture differences between \OURS's \llama architecture and DiT.} All other hyperparameters remain equal for the architecture ablation.}
    \label{tab:ablation_architecture_details}
\end{table}

\subsubsection{TAE Results}
\label{sec:tae_results}
We present here results and ablations from important design decisions for the temporal autoencoder (TAE).
For evaluation, we report the reconstructed peak signal-to-noise ratio (PSNR), structural similarity (SSIM)~\citep{wang2004image}, and Fr\'echet Inception distance (FID)~\citep{heusel2017gans} of video clips split from the training set, with 2s, 4s, 6s, and 8s duration, each with 200 examples.
We also measure the same metrics on a validation split of the image training set.
For video reconstruction evaluation, metrics are averaged over video frames.

\begin{figure}[t!]
  \centering
  \begin{subfigure}[b]{0.49\textwidth}
      \centering
      \includegraphics[width=1\linewidth]{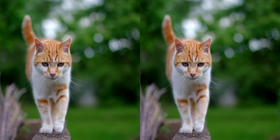}
  \end{subfigure}
  \begin{subfigure}[b]{0.49\textwidth}
      \centering
      \includegraphics[width=1\linewidth]{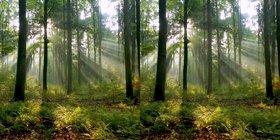}
  \end{subfigure}
  \caption{\textbf{Real (left) and \taeShort reconstructed (right) videos.} The \taeShort compresses the video by a factor of $8\times$ across each of the three spatiotemporal dimensions.
  We observe that the reconstructions from the \taeShort maintain visual detail present in the original videos. %
    }
  \label{fig:tae_examples}
\end{figure}

\noindent\textbf{Qualitative Results.}
We show sample reconstructions from our \taeShort in~\cref{fig:tae_examples} with frames from the original video and the reconstruction after the \taeShort encoder and decoder.
We observe that the \taeShort can reconstruct the video frames while preserving visual detail.
The \taeShort reconstruction quality decreases for high frequency spatial details in images and video frames, and fast motion in videos.
When both high frequency spatial details and large motion are present in a video, this can lead to a loss in detail, as can be seen in the examples in~\cref{fig:tae_examples}, where fine details are smoothed out in the reconstruction.

\noindent\textbf{Quantitative metrics.}
\cref{tab:tae_main_results} compares our \taeShort against a baseline frame-wise autoencoder that does not perform any temporal compression.
Our baseline also produces a 8 channel latent space as is standard for autoencoders used for frame-wise encoding in prior work~\citep{sdvideo,emuvideo2023}.
On video data, we observe that the \taeShort achieves a competitive performance to the frame-wise encoder while achieving a $8\times$ higher temporal compression.
On images, the \taeShort outperforms the frame-wise model, an improvement that can be attributed to the increased channel size of the latent (8 \vs 16)~\citep{dai2023emu}.

\begin{table}[t!]
    \centering
    \begin{tabular}{ccccccc}
        \toprule
         & \multicolumn{3}{c}{Video (512 px)} & \multicolumn{3}{c}{Image (512 px)} \\
         & $\text{SSIM}\uparrow$ & $\text{PSNR}\uparrow$ & $\text{FID}\downarrow$ & $\text{SSIM}\uparrow$ & $\text{PSNR}\uparrow$ & $\text{FID}\downarrow$  \\
         \midrule
         Frame-wise AE & 0.9348 & 34.11 & 0.9352 & 0.8877 & 30.83 & 1.7588 \\
        \taeShort
         & 0.9093 & 32.25 & 1.4872 & 0.9231 & 32.16 & 0.2873 \\
         \bottomrule
    \end{tabular}
    \caption{\textbf{\taeShort reconstruction metrics comparison} between an frame-wise autoencoder and our \taeShort model. We observe that the \taeShort achieves comparable performance to a frame-wise autoencoder for video reconstruction while achieving a $8\times$ higher compression.}
    \label{tab:tae_main_results}
\end{table}

\subsubsection{TAE Ablations}

We now perform a series of ablation experiments for the design choices in training our \taeShort model.

\noindent \textbf{Baseline setting for ablations.}
For simplicity, we use a \taeShort model with a smaller $4\times$ compression ratio that produces a 8-channel latent space.

\noindent\textbf{2.5D \vs 3D attention \& convolutions.}
We compare using 2.5D, \ie, 2D spatial attention/convolutions followed by 1D temporal attention/convolutions to using 3D spatiotemporal attention/convolutions in the \taeShort.
In~\cref{tab:tae_3d_ablation}, we observe that the 3D spatiotemporal attention leads to slightly better reconstruction metrics.
However, we found that this improvement was not large enough to justify the larger memory and compute costs associated with a fully 3D model compared to a 2.5D model.
Thus, we use 2.5D for our \taeShort.

\begin{table}[h!]
    \centering
    \begin{tabular}{ccccccc}
        \toprule
         & \multicolumn{3}{c}{Video (256 px)}\\
         & $\text{SSIM}\uparrow$ & $\text{PSNR}\uparrow$ & $\text{FID}\downarrow$   \\
         \midrule
         \taeShort 2.5D & 0.845 & 28.51 & 3.678\\
         \taeShort 3D & 0.852 & 28.80 & 4.715\\
         \bottomrule
    \end{tabular}
    \caption{\textbf{Comparing \taeShort models with 2.5D \vs 3D convolutions and attentions.} We observe that while the 3D models perform slightly better than the 2.5D models, the gap in performance between the two models is small.
    }
    \label{tab:tae_3d_ablation}
\end{table}

\noindent\textbf{Effect of outlier penalty loss.}
We ablate the effect of adding the outlier penalty loss (OPL) from~\cref{sec:tae}. The addition of this loss removes the artifacts from generated and reconstructed videos as seen in~\cref{fig:tae_black_spot}, and improves reconstruction performance.
We first train a baseline model without OPL for 50K iterations.
We then finetune this model with OPL for 10K iterations and compare it to a baseline finetuned without OPL for 20K iterations.
The results, summarized in~\cref{tab:tae_opl_ablation}, suggest that OPL finetuning improves the reconstruction for both images and videos.
\begin{table}[h!]
    \centering
    \begin{tabular}{ccccccc}
        \toprule
         & \multicolumn{3}{c}{Video (512 px)} & \multicolumn{3}{c}{Image (512 px)} \\
         & $\text{SSIM}\uparrow$ & $\text{PSNR}\uparrow$ & $\text{FID}\downarrow$ & $\text{SSIM}\uparrow$ & $\text{PSNR}\uparrow$ & $\text{FID}\downarrow$  \\
         \midrule
         \taeShort w/o OPL finetune & 0.897 & 31.11 & 1.389 & 0.845 & 28.25 & 0.568 \\
         \taeShort w/ OPL finetune & 0.910 & 31.93 & 1.241 & 0.862 & 29.26 & 0.614 \\
         \bottomrule
    \end{tabular}
    \caption{\textbf{Effect of OPL on \tae reconstructions}. We evaluate the effect of OPL finetuning on the reconstruction quality and observe that OPL improves both image and video metrics.}
    \label{tab:tae_opl_ablation}
\end{table}

\subsubsection{Spatial Upsampler Results}
\label{sec:spatial_upsampler_results}

\begin{figure}[h]
    \centering
    \includegraphics[width=0.97\linewidth]{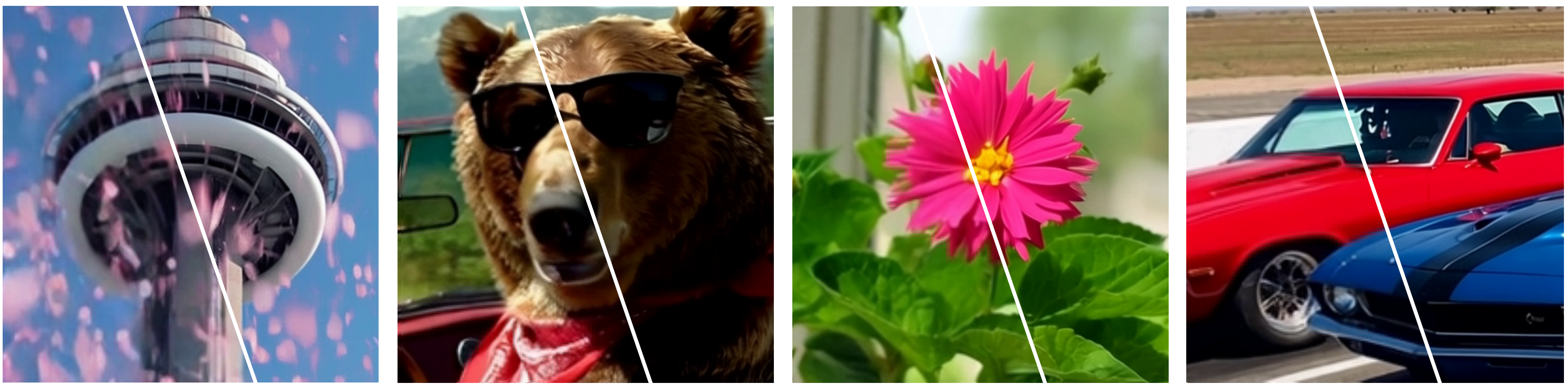}

    \caption{\textbf{Qualitative visualization of \upsampler:} A visual comparison of the 200 px\textbackslash 400 px crops before\textbackslash after upsampling.
    The \upsampler improves high frequency details in the output.
    }

    \label{img:upsampler_400_crop}
\end{figure}

Here, we include some results from the spatial upsampler described in~\cref{sec:spatial_upsampler}.
A visual comparison of the upsampling process is presented in~\cref{img:upsampler_400_crop}, which shows the 200 px and 400 px crops before and after upsampling.
The results demonstrate that the upsampler effectively sharpens and enhances visual details, producing a more refined and detailed output.

\subsection{\TextToI Generation}
\label{sec:t2i_generation}
The \Ours model is trained jointly on videos and images, and hence is able to generate both videos and images.
To further validate the model's image generation capabilities, we continued training it with an image autoencoder and compared its performance to prior work in image generation.
The following sections provide detailed experimental settings and evaluation results.

\subsubsection{Method}
\label{sec:post_train_images}
For the \TextToI model, our goal is to generate realistic images. We utilize the \Ours model as an initialization and replace the TAE with an image autoencoder. We then train the model on the text-to-image generation task, allowing it to generate images based on text descriptions. The final resolution is $1024$ px.  For post-training, we curated a total of \bigO(1000) images created by in-house artists for quality-tuning, following the approach outlined in~\citep{dai2023emu}.
We finetuned the model for 6k steps with a learning rate of 0.00001 and a batch size of 64. We used a constant learning rate scheduler with 2000 warm-up steps.

\subsubsection{Results}

To measure the quality of our \TextToI generation results, we use human evaluators to evaluate the following axes: (a) text faithfulness, and (b) visual quality.
For evaluating text faithfulness, we use a pairwise A/B comparison set up where evaluators select which image aligns better with a given generation prompt.
Evaluators are asked to choose which of the choices A or B, is better, or equal, in terms of text alignment.
For visual quality, we use a a similar pairwise A/B comparison set up and ask raters to help select the image that looks more realistic.
Evaluators are asked to look for flaws in the generation, such as errors in the number of fingers or arms, or visual text spelling errors, before making their decision.
For creating the benchmarking prompts, we analyzed typical \textToI user prompts and generated categories and distributions, and leverage LLMs to produce user prompts that mimic real users prompts.

\begin{figure}[h]
    \centering
    \includegraphics[width=.9\linewidth]{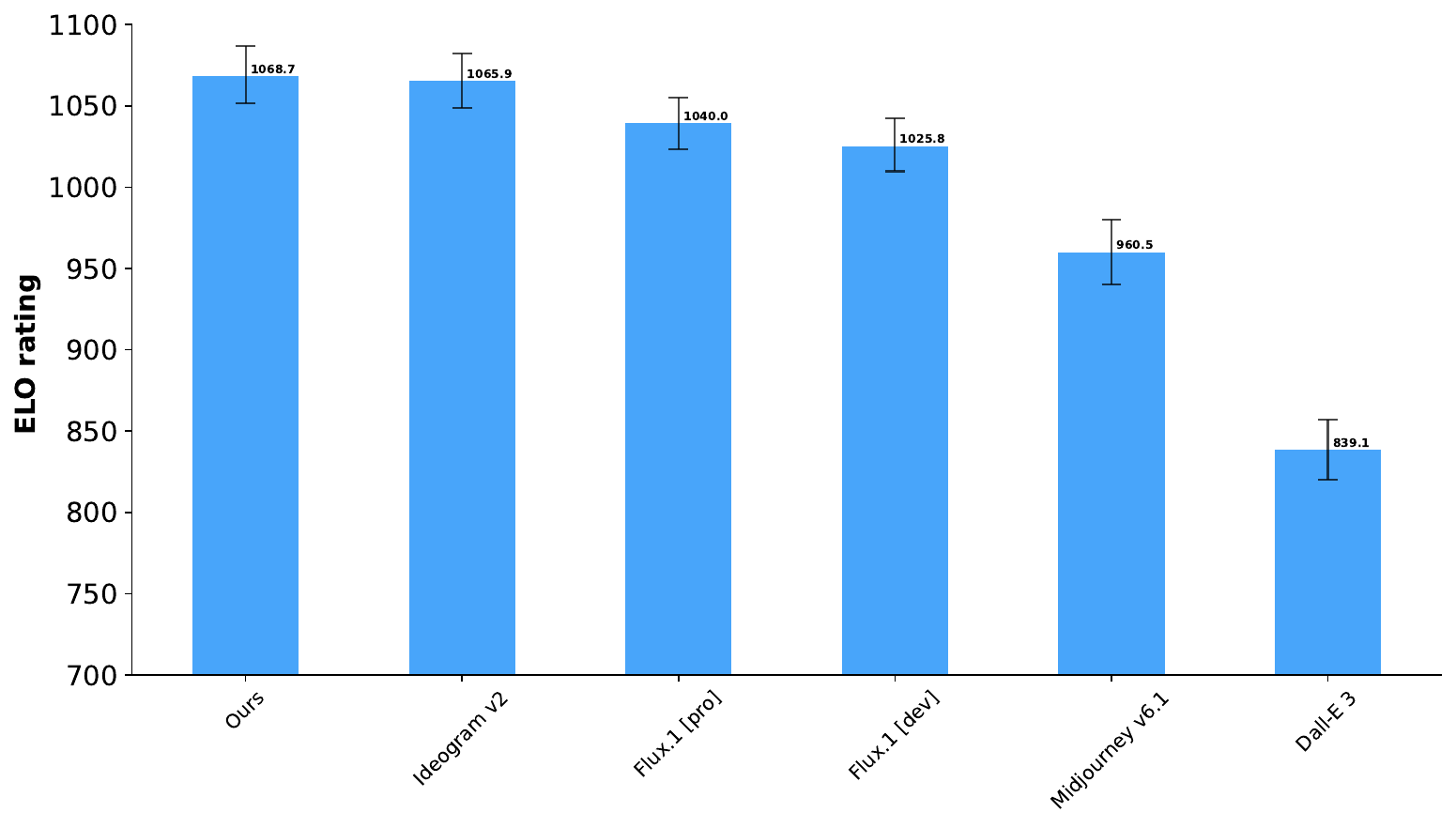}
    \caption{\textbf{ELO rating comparison of \textToI methods.} We compare our image generation model to state-of-the-art \textToI models and observe that it performs competitively with recent approaches.}
    \label{img:t2i_overall_elos_ranking}
\end{figure}

\begin{figure}[h]
    \centering
    \includegraphics[width=.24\linewidth]{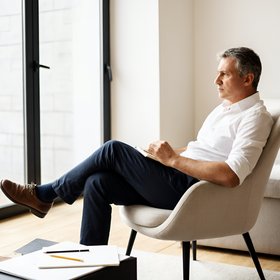}
    \includegraphics[width=.24\linewidth]{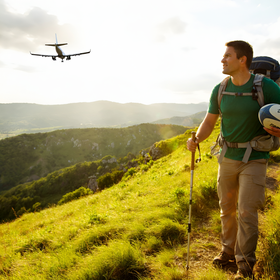}
    \includegraphics[width=.24\linewidth]{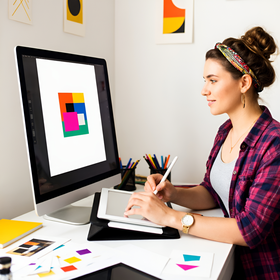}
    \includegraphics[width=.24\linewidth]{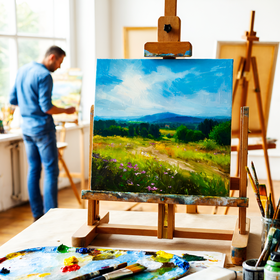}
    \caption{\textbf{Example T2I Images from \OURS.} Generation prompts from left to right:
    (1) A focused writer, with a trusty pencil by their side, sits in a modern, minimalist home office with a large window providing ample natural light.
    (2) A hiker wearing a backpack walks along a trail with a rugby ball in hand, looking up at a soaring airplane overhead, the sun shining down on a beautiful day.
    (3) A styled hair with a headband inspired a graphic designer using a pen tablet and stylus to create vibrant digital art in a cubism style.
    (4) A vivid landscape painting on an easel stands in a bright, natural light-filled studio, showcasing its vibrant colors and textures, while a painter works in the background.
    }
    \label{img:t2i_qualitative_results}
\end{figure}

We compare with the best contemporary \TextToI models including \Flux~\citep{flux}, \Dalle~\citep{dalle3}, \MJ~\citep{midjourney}, and \Ideogram~\citep{ideogram} available at time of benchmark
These are however black-box commercial solutions, which makes fair comparison a challenge.
Similar to \TextToV evaluation, we obtain non cherry picked generated images from the benchmark prompts for the prior work methods, and compare to them using non cherry picked images from \OURS for the same prompts.
To ensure consistent comparison across all models and evaluation axes, we utilize the ELO rating system to establish rankings based on battle records converted from raw human evaluation results.
For A/B comparison evaluations, the ``win/tie/lose'' on a given prompt between two models were directly interpreted as one battle record.
This approach allowed us to combine ratings on all evaluation axes to generate an overall performance.
The comparison results are summarized in~\cref{img:t2i_overall_elos_ranking}, where we see that our model achieves the highest ELO rating compared to all recent state-of-the-art \textToI methods available at the time of benchmarking.
In~\cref{img:t2i_qualitative_results}, we show some qualitative results of our generations.

\section{Video Personalization}%
\label{sec:personalization}
Generating personalized high quality videos that accurately capture an individual's identity is an important research area with significant practical applications.
We integrate personalization into video generation, yielding state-of-the-art outcomes as detailed in this section.
We describe our novel model architecture in~\cref{sec:pt2v_model_arch} followed by the training recipe in~\cref{sec:pt2v_pretrain} and~\cref{sec:pt2v_finetune}.
We explain the evaluation criteria for personalization in~\cref{sec:pt2v_eval} and show quantitative results in~\cref{sec:pt2v_results}.
\subsection{Model}
\label{sec:pt2v_model_arch}

\begin{figure}[ht!]
\includegraphics[width=0.97\linewidth]{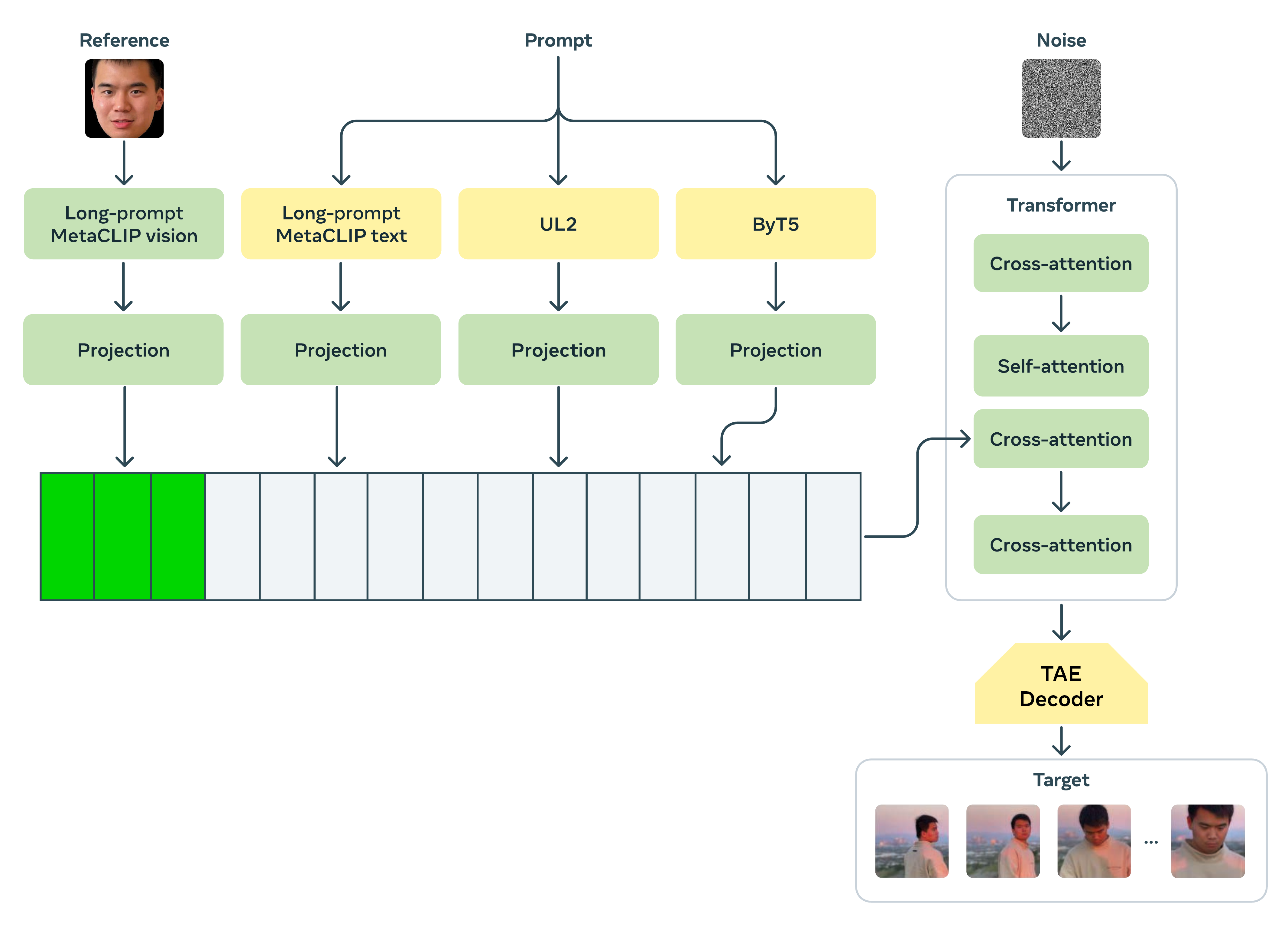}
\vspace{-4mm}
\caption{\textbf{Architecture and inference pipeline of the Personalized \OursVideo (PT2V) model}.
We initialize the model from \OursVideo's weights and add additional learnable parameters to enable conditioning on a reference image.
We encode the reference image using a trainable vision encoder initialized from Long-prompt MetaCLIP ~\citep{xu2023demystifying}, and concatenate the embedding with the text prompt embedding. Frozen layers are illustrated in yellow while trainable ones in green.
}
\label{img:pt2v}
\end{figure}

We extend our 30B \OursVideo model for Personalized Text-to-Video generation, PT2V, by conditioning the model on the identity information extracted from an input reference image in addition to the text prompt. Figure~\ref{img:pt2v} illustrates the architecture of our PT2V model initialized from the T2V \OursVideo weights.
We use vision token concatenation in the condition, enabling integration into a unified framework, which allows to scale up the model size.
Similar to~\citep{meta24memu}, we extract identity features from a masked face image using a trainable Long-prompt MetaCLIP vision encoder~\citep{xu2023demystifying}, followed by a projection layer to align them with the text feature dimension.
Our training strategy includes a PT2V \pretraining phase followed by PT2V high quality finetuning.

\subsection{\Pretraining}

\subsubsection{\Pretraining Data}
\label{sec:pt2v_pretrain}

For the PT2V training sets, our focus is exclusively on videos where the same person appears across all frames.
We curate this training set from the \OursVideo \pretraining datasets described in~\cref{sec:pt-data}.
To achieve this, we first filter the raw \textToVShort videos based on captions by selecting those with human-related concepts.
We extract frames at one-second intervals and apply a face detector to keep videos that contain a single face and where the ArcFace cosine similarity score~\citep{deng2019arcface} between consecutive frames exceeds 0.5.
This processing provides us with \bigO(1)M text-video pairs where a single person appears, with durations from 4s to 16s.
Based on the source reference face, our PT2V training dataset can be categorized into ``paired'' and ``cross-paired'' data.
We define ``paired'' data as cases where the reference image is taken from the same video clip,
while ``cross-paired'' data refers to cases where the reference image originates from a different video but features the same subject.

\textbf{Paired Data.}
For each selected text-video pair, we uniformly sample 5 frames from the video clip, yielding \bigO(10)M paired training samples.
For each frame, we crop the face area and segment the face region to prevent the model attending to non-critical areas such as the background.

\textbf{Cross-Paired Data.}
We observed that training solely on the above paired data makes the model easily learn a copy-paste shortcut solution,
\ie, the generated video always follows the expression or the head pose from the reference face.
To address this issue, we collect training pairs where the reference image comes from a different video of the same person.

We collected both real and synthetic cross-paired data samples. \bigO(10)K real cross-pairs from a subset of our \pretraining data that contains different camera views of the same scene.
For the synthetic cross-paired data, we use a \pretrained personalization image generation model~\citep{meta24memu} to create synthetic reference images.
Specifically, we apply the model to the first frame of each video from the paired data, generating images with diverse prompts to vary expressions, head poses, and lighting conditions, \etc.
To maintain identity consistency, we discard any generated images with an ArcFace similarity score below 0.7 compared to the reference image.
In total, this process yields \bigO(1)M synthetic cross-paired data samples.

\subsubsection{\Pretraining recipe}
\label{subsec:pt2v_recipe}
\begin{figure}
\begin{minipage}[hb]{.73\textwidth}
    \centering
    \adjustbox{max width=\textwidth}{%
    \begin{tabular}{lccccccccr}
        \toprule
        Training stage & TP & CP & bs/Node & \#GPUs & global bs & learning rate & frame length & \#iters & \#seen videos \\
        \midrule
        Stage-I & 4 & 2 & 1 & 4096 & 512 & 2e-5
        & 64 & 18k & 9.21M \\

        Stage-II & 4 & 2 & 1 & 4096 & 512 & 2e-5
        & 128/192/256 & 8k & 4.1M \\

        \multirow{2}{*}{Stage-III} & \multirow{2}{*}{4} & \multirow{2}{*}{2} & \multirow{2}{*}{1} & \multirow{2}{*}{4096} & \multirow{2}{*}{512} & 2e-5 Transformer
        & \multirow{2}{*}{128/192/256} & \multirow{2}{*}{4k} & \multirow{2}{*}{2.05M} \\
         & & & &  &  & 1e-7 Vision Encoder
        & & & \\
        \midrule
        {\bf Total} & & & & & & & & {\bf 30k} & {\bf 15.36M} \\
        \bottomrule
    \end{tabular}}
\end{minipage} \hfill
\begin{minipage}[hb]{.26\textwidth}
    \centering
    \includegraphics[width=\textwidth]{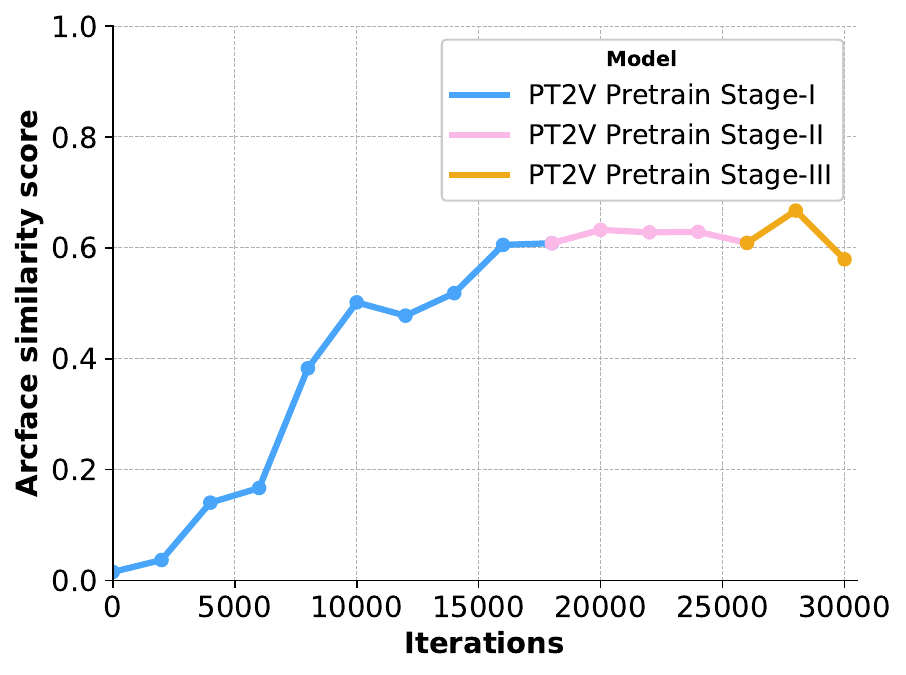}
\end{minipage}

\caption{\textbf{Personalized \OursVideo (PT2V) \pretraining.}
    Left: PT2V \pretraining recipe. Our training recipe consists of three stages as described in~\cref{subsec:pt2v_recipe}.
    Right: ArcFace similarity score of the PT2V Stage I-III \pretraining. The model gradually follows the identity as the training going on in Stage-I. Stage-II maintains the identity similarity of Stage-I. In Stage-III, identity similarity tends to fluctuate due to training with cross-paired data.
    }
\label{fig:pt2v_pt_recipe}
\end{figure}

 There are three goals in PT2V \pretraining: 1) train the model to condition on a reference image and preserve the identity, 2) generate long personalized videos, and 3) improve generated human expressions and motion naturalness. We found that directly training the model on long videos is inefficient and often leads to slow identity injection to the personalized model since  (1) training speed is nearly proportional to the square of the number of latent frames (tokens), and (2) the weak reference image-to-video correspondence in long videos makes the task more challenging.
 More details on the \pretraining recipe is shared in~\cref{fig:pt2v_pt_recipe}.

\noindent \textbf{Stage-I: Identity injection.}
In the first stage of PT2V \pretraining, we simplify the problem by conditioning the model on the reference image and training on short videos.
Specifically, we truncate the TAE embedding to 8 latent frames (corresponding to 64 RGB video frames) to accelerate identity injection using the paired training samples. We freeze the vision encoder and only train the transformer backbone.
We observe that the model can quickly learn to follow the reference image during this stage, as measured by the average ArcFace similarity score in~\cref{fig:pt2v_pt_recipe}.

\noindent \textbf{Stage-II: Long video generation.}
To recover the model's capability to generate long videos, we continue training the PT2V model from Stage-I with a larger number of latent frames, similar to the pre-trained T2V model in~\cref{tab:t2v_pt_bucket}. This stage substantially enhances the consistency of long video generation, particularly in terms of background and motion coherence.

\noindent \textbf{Stage-III: Improve naturalness.} Since the model in stage-I and stage-II has been trained on the paired image-video samples, it often demonstrates a strong copy-paste effect. For instance, in the generated video frames, the person tends to gaze directly at the camera, resulting in an unnatural-looking facial expression. We improve video naturalness and facial expression in stage-III by training on the cross-paired samples where the reference image is not from the corresponding target video. We leverage both real cross-paired data and synthetic cross-paired data in this stage as discussed in~\cref{sec:pt2v_pretrain}.
We also finetune the vision encoder to extract more detailed identity features from the reference image.

\subsection{Supervised Finetuning}
\label{sec:pt2v_finetune}
Similar to \textToVShort, we further improve the video aesthetics in a high-quality finetuning stage by leveraging high quality aesthetic data.

\subsubsection{Finetuning Dataset}
The large scale \pretraining data enables the model to generate videos following the identity from the reference face image.
Similar to the post-training of \OursVideo (see~\cref{sec:t2v_post_training}), we collect a small set of high-quality finetuning data, with the goal of generating highly aesthetic videos with good quality motion.
To match the visual quality and aesthetics of \OursVideo, we started from the T2V finetuning set and collected videos with a single person.
Subsequently, we manually selected videos with diverse human actions, ensuring that the dataset captured a variety of movements and behaviors.
In total, our final finetuning set contains \bigO(1000) high-quality videos with both paired and real cross-paired reference images used with a 1:1 ratio.

\subsection{Evaluation}
\label{sec:pt2v_eval}

We evaluate the quality of our PT2V models across three axes: identity preservation, video quality, and video-text alignment. The latter two axes are similar to T2V A/B evaluations in Section~\ref{T2V_model_evaluation}, where video quality can be further broken down into overall quality, frame consistency, motion completeness, and motion naturalness. To measure identity preservation, given an identity reference image and the generated video clip, the annotators are asked to rate on how well the generated character's face captures the reference person likeness in both the best and the worst frame (identity score), as well as how visually consistent the faces are among the generated frames containing the reference person (face consistency score). These two scores are measured in an absolute sense with ratings as ``really similar'', ``somewhat similar'', and ``not similar'' for the identity question and ``really consistent'', ``somewhat consistent'', and ``not consistent'' for the face consistency question. Annotators were trained to follow specific guidelines for the labeling on these axes and are constantly audited for quality.

\noindent\textbf{Evaluation Dataset.}
We selected 50 subjects who were not seen during training as the reference face in the evaluation data.
These reference face images include both frontal and side views.
For each image, we pair it with 5-7 unique prompts, and curate 330 image-prompt pairs for evaluation.
Similar to the T2V evaluation datasets, these prompts cover different human activities and facial expressions. We follow the same prompt rewrite as Section~\ref{sec:t2v_prompt_rewrite} to bridge the gap between our training and inference captions.

\subsection{Results}
\label{sec:pt2v_results}

\begin{table}[!t]
    \centering
        \begin{tabular}{c ccc}
        \toprule
			Method & $\text{Identity}_{best}$ ($\uparrow$) & $\text{Identity}_{worst}$ ($\uparrow$) & Face Consistency ($\uparrow$) \\
			\midrule
			ID-Animator & 3.69\% & 3.08\% & 79.69\% \\
			PT2V Pre-train & 71.91\% & 66.36\% & 97.53\% \\
			PT2V Finetune & 65.52\% & 60.19\% & 95.61\% \\
		\bottomrule
		\end{tabular}
    \caption{\textbf{Personalized \OursVideo (PT2V) evaluation.} We compare our model after the \pretraining and supervised high-quality finetuning stages against ID-Animator~\citep{he2024id} on Identity score on the best similar frame, the worst similar frame, and face consistency across frames.
    }
    \label{tab:pt2v_main_identity}
\end{table}

\begin{table}[!t]
    \centering
        \begin{tabular}{c ccc}
        \toprule
			Method & $\text{Identity}_{best}$ ($\uparrow$) & $\text{Identity}_{worst}$ ($\uparrow$) & Face Consistency ($\uparrow$) \\
			\midrule
			Frozen Vision Encoder & 73.93\% & 66.67\% & 83.50\% \\
			Trainable Vision Encoder & 90.10\% & 87.22\% & 99.69\% \\
			Cross-Paired Training & 71.79\% & 66.36\% & 97.53\% \\
		\bottomrule
		\end{tabular}
    \caption{\textbf{Ablation study for Personalized \OursVideo in terms of identity preservation metrics.} We observed that the trainable vision encoder better preserves identity than the frozen vision encoder. Note that although cross-paired training decrease the identity similarity, it leads to more diverse head poses and more natural expressions.}
    \label{tab:pt2v_ablation_face}
\end{table}

In~\cref{tab:pt2v_main_identity} and~\cref{tab:pt2v_vs_baseline}, we compare our \OursPTV after supervised finetuning with ID-Animator~\citep{he2024id}. For the Identity score, we aggregate the ``really similar'' and ``somewhat similar'' scores in the best frame, and for the consistency score, we aggregate the ``really consistent'' and ``somewhat consistent'' scores. As evident, our method significantly outperforms the baseline by a large margin in all axes of identity preservation, video quality, and text alignment.
We also compare it with \OursVideo without the visual conditioning in terms of video quality and text alignment in~\cref{tab:pt2v_vs_t2v}.

\begin{table}[!t]
    \centering
    \adjustbox{max width=\linewidth}{%
    \subfloat[
    \label{tab:pt2v_vs_baseline}
    ]{
       \centering
       \begin{tabular}{c c}
        \toprule
         & \multicolumn{1}{c}{PT2V-Finetune net win rate} \\
         &  \vs ID-Animator \\
         \midrule
          Overall Quality & 64.74 \\
          Consistency & 22.18 \\
          Motion Naturalness & 37.38\\\
          Motion Completeness & 5.17 \\
          \midrule
         Text Alignment & 53.20 \\
         \bottomrule
    \end{tabular}
    }
    \hfill
    \subfloat[
    \label{tab:pt2v_vs_t2v}
    ]{
        \centering
        \begin{tabular}{c c}
        \toprule
         & \multicolumn{1}{c}{PT2V-Pretrain net win rate} \\
         &  \vs T2V-Pretrain \\
         \midrule
          & 3.95 \\
          & 10.33 \\
          & -1.82 \\
          & -5.16 \\
         \midrule
          & -11.25 \\
         \bottomrule
    \end{tabular}

    }
    }
    \caption{\textbf{Personalized \OursVideo (PT2V) evaluation on video quality and text alignment.} (a) Net win rate (win\% - loss\%) of our PT2V after supervised finetuning \vs SOTA (ID-Animator~\citep{he2024id}). PT2V significantly outperforms ID-Animator in all metrics. (b) PT2V \vs \OursVideo (T2V) without the visual conditioning. We observed that PT2V wins consistency and performs on par in overal quality accounting for statistical significance, but loses in motion completeness and prompt alignment due to the narrow concept distribution (activities, objects, \etc) of PT2V.
    }
    \label{tab:pt2v_main}
\end{table}

We present generated videos from \OursPTV in ~\cref{fig:pt2v_qual_ours_figure}.
The first four videos are generated  with the same prompt but different identities, 
and the latter four are generated with the same identity but different prompts. 
The generated videos follow the identity with diverse motion and camera views. 
Qualitative comparisons between \OursPTV and ID-Animator~\citep{he2024id} are shown in ~\cref{fig:pt2v_qual_ida_vs_ours_figure}. 
\OursPTV consistently outperforms ID-Animator in terms of identity consistency and video quality.

\subsubsection{Ablations}
\label{sec:pt2v_ablation}
We ablate the impact of key design choices in our 30B Personalized \TextToV training pipeline.

\par \noindent \textbf{Effect of training visual conditioning embedding.}
Our models use an embedding from a visual encoder of the face as the visual embedding to condition the generation.
We study whether training this embedding jointly during the video generation task improves performance.
We re-train the third stage of our model with either a fixed or trainable vision encoder model and report the evaluation results in~\Cref{tab:pt2v_ablation_face,tab:pt2v_ablation_vid}.
We observe that using a fixed vision encoder compromises identity preservation significantly, $-16\%$ as seen in~\cref{tab:pt2v_ablation_face}.

\par \noindent \textbf{Effect of cross-paired data.}
Our training pipeline uses cross-paired data, \ie, where the image of the face used to condition the generation comes from a different video clip than the video clip to be generated.
We observe in~\cref{tab:pt2v_ablation_face} that cross-paired training leads to a decrease in identity metrics, however, it is crucial in improving facial expressions and natural movement in the generated videos. Human annotation in~\cref{tab:pt2v_ablation_vid} reveals that cross-pair trained model improves text alignment by 27.36\% and overall quality by 13.68\%, especially 26.14\% in motion naturalness.

\par \noindent \textbf{Effect of high quality finetuning.}
We show the impact of a final high-quality finetuning stage on all axes of video quality and text alignment in Table~\ref{tab:pt2v_ablation_vid} and on identity preservation in Table~\ref{tab:pt2v_main_identity}. Similarly, since our high-quality finetuning set includes cross-paired data, identity drops slightly while video quality and naturalness is improved significantly.

\begin{table}[!t]
    \centering
    \adjustbox{max width=\linewidth}{%
        \begin{tabular}{c ccc}
        \toprule
         & Trainable \vs Frozen Vision Encoder & Cross-Paired \vs Paired & Finetune \vs Pretrain \\
         \midrule
         Overall Quality & 0.91 & 13.68 & 26.53\\
         Consistency & 6.38 & -5.16 & 9.15\\
         Motion Naturalness & -2.43 & 26.14 & 9.45\\
         Motion Completeness & -2.72 & 7.6 & 5.49\\
         \midrule
         Text Alignment & -4.56 & 27.36 & -1.82 \\
         \bottomrule
    \end{tabular}}
    \caption{\textbf{Ablation study for Personalized \OursVideo in terms of \TextToV evaluation metrics.} Each column shows the net win rate on adopting a design decision \vs a model that does not have it. Using a trainable vision encoder is comparable to the frozen one in terms of quality and alignment metrics while boosting the identity preservation significantly as in Table~\ref{tab:pt2v_ablation_face}. The results also confirm the importance of cross-paired training data and supervised finetuning for video naturalness, higher quality and text alignment.}
    \label{tab:pt2v_ablation_vid}
\end{table}

\begin{center}
    \centering
    \captionsetup{type=figure}
    \begin{table}[H]

\setlength{\tabcolsep}{1pt}
\adjustbox{max width=\textwidth}{%
\centering
\begin{tabular}{cccc}
    \multicolumn{1}{c}{\textbf{Reference Image}} &
    \multicolumn{3}{c}{\textit{Prompt}: A person dressed in elegant attire is seen checking the table settings} \\
    
    \multicolumn{4}{c}{\textbf{\OursPTV}} \\
    \includegraphics[width=0.14\linewidth]{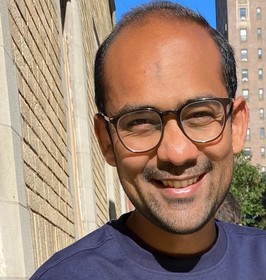} &
    \includegraphics[width=0.25\linewidth]{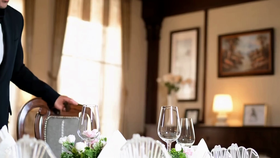} &
    \includegraphics[width=0.25\linewidth]{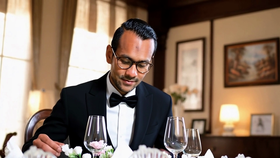} &
    \includegraphics[width=0.25\linewidth]{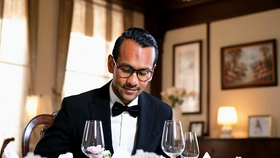} \\
\end{tabular}}
\begin{tabular}{ccccccc}
    \multicolumn{7}{c}{\textbf{ID-Animator}} \\
    \includegraphics[width=0.14\linewidth]{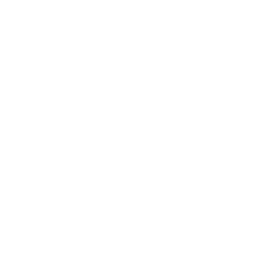} &
    & 
    \includegraphics[width=0.14\linewidth]{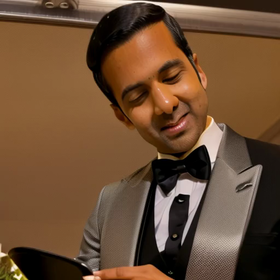} &
    \includegraphics[width=0.14\linewidth]{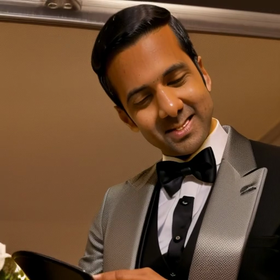} &
    \includegraphics[width=0.14\linewidth]{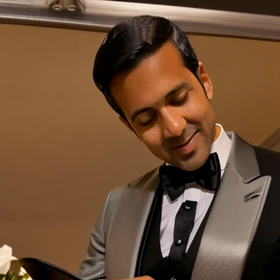} &
    \includegraphics[width=0.14\linewidth]{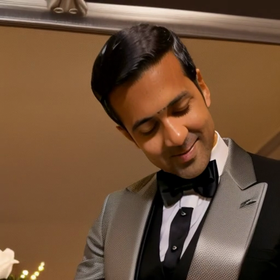} &
    \includegraphics[width=0.14\linewidth]{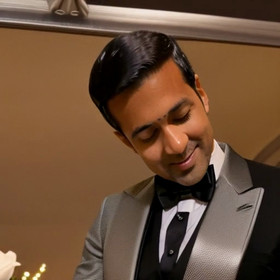} \\
\end{tabular}

\setlength{\tabcolsep}{1pt}
\adjustbox{max width=\textwidth}{%
\centering
\begin{tabular}{cccc}
    \multicolumn{1}{c}{\textbf{Reference Image}} &
    \multicolumn{3}{c}{\textit{Prompt}: A person wearing a fur-lined hat rides a llama} \\
    
    \multicolumn{4}{c}{\textbf{\OursPTV}} \\
    \includegraphics[width=0.14\linewidth]{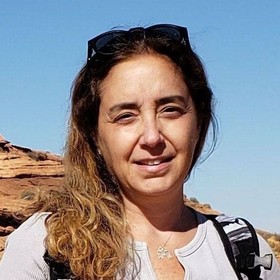} &
    \includegraphics[width=0.25\linewidth]{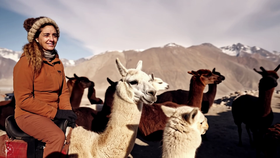} &
    \includegraphics[width=0.25\linewidth]{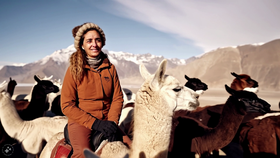} &
    \includegraphics[width=0.25\linewidth]{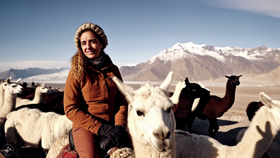} \\
\end{tabular}}

\begin{tabular}{ccccccc}
    \multicolumn{7}{c}{\textbf{ID-Animator}} \\
    \includegraphics[width=0.14\linewidth]{figures/pt2v_qualitative_comparison/white.png} &
    & 
    \includegraphics[width=0.14\linewidth]{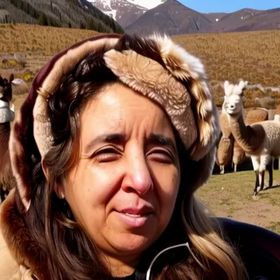} &
    \includegraphics[width=0.14\linewidth]{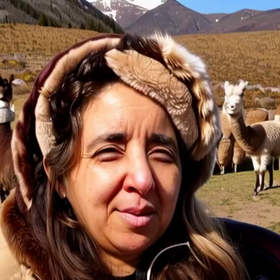} &
    \includegraphics[width=0.14\linewidth]{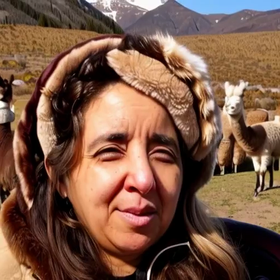} &
    \includegraphics[width=0.14\linewidth]{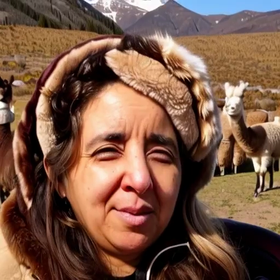} &
    \includegraphics[width=0.14\linewidth]{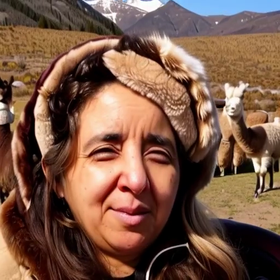} \\
\end{tabular}

\setlength{\tabcolsep}{1pt}
\adjustbox{max width=\textwidth}{%
\centering
\begin{tabular}{cccc}

    \multicolumn{1}{c}{\textbf{Reference Image}} &
    \multicolumn{3}{c}{\textit{Prompt}: A person pauses to wipe sweat from the brow with a white handkerchief} \\
    
    \multicolumn{4}{c}{\textbf{\OursPTV}} \\
    \includegraphics[width=0.14\linewidth]{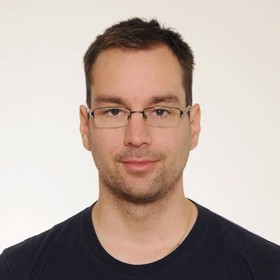} &
    \includegraphics[width=0.25\linewidth]{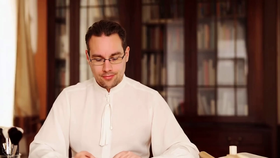} &
    \includegraphics[width=0.25\linewidth]{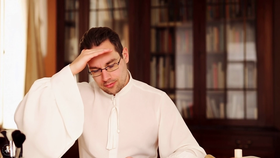} &
    \includegraphics[width=0.25\linewidth]{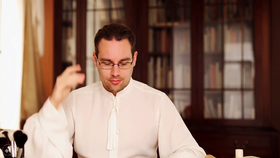} \\
\end{tabular}}
\begin{tabular}{ccccccc}
    \multicolumn{7}{c}{\textbf{ID-Animator}} \\
    \includegraphics[width=0.14\linewidth]{figures/pt2v_qualitative_comparison/white.png} &
    & 
    \includegraphics[width=0.14\linewidth]{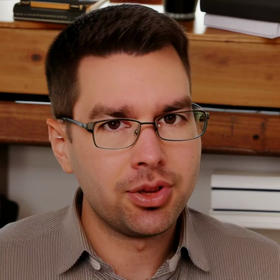} &
    \includegraphics[width=0.14\linewidth]{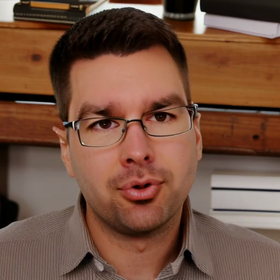} &
    \includegraphics[width=0.14\linewidth]{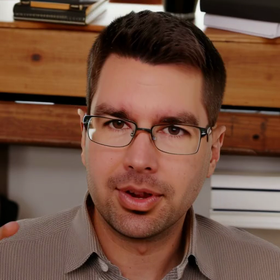} &
    \includegraphics[width=0.14\linewidth]{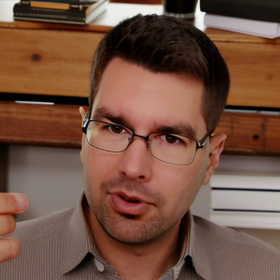} &
    \includegraphics[width=0.14\linewidth]{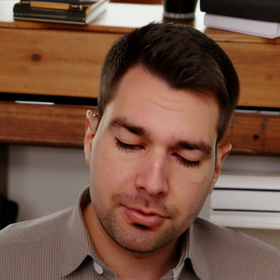} \\
\end{tabular}

\end{table}

    \vspace{-3mm}

    \caption{\textbf{Qualitative comparisons between \OursPTV and prior work.}
    Here, we show two generated videos for the same prompt, showcasing the output of \OursPTV (first row), and ID-Animator (second row) for comparison. 
    Videos in this Figure found at \url{https://go.fb.me/MovieGen-Figure22}.
    }

    \label{fig:pt2v_qual_ida_vs_ours_figure}
\end{center}%

\begin{center}
    \centering
    \captionsetup{type=figure}
    \setlength{\tabcolsep}{1pt}
\adjustbox{max width=\textwidth}{%
\centering
\begin{tabular}{cccc}
    \multicolumn{1}{c}{\textbf{Reference Image}} &
    \multicolumn{3}{c}{\textit{Prompt}: A person feeding a llama in a zoo} \\

    \includegraphics[width=0.14\linewidth]{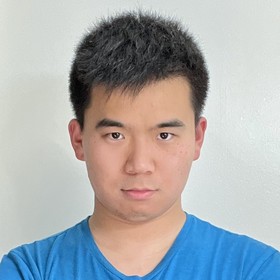} &
    \includegraphics[width=0.25\linewidth]{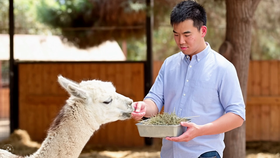} &
    \includegraphics[width=0.25\linewidth]{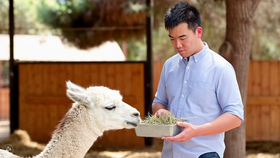} &
    \includegraphics[width=0.25\linewidth]{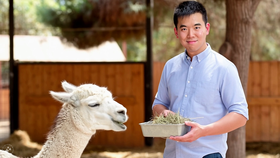} \\
    
    \includegraphics[width=0.14\linewidth]{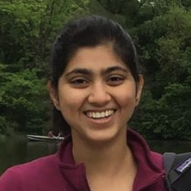} &
    \includegraphics[width=0.25\linewidth]{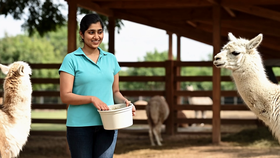} &
    \includegraphics[width=0.25\linewidth]{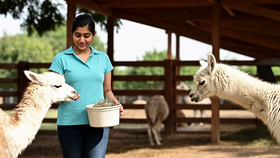} &
    \includegraphics[width=0.25\linewidth]{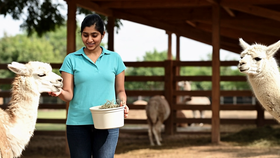} \\

    \includegraphics[width=0.14\linewidth]{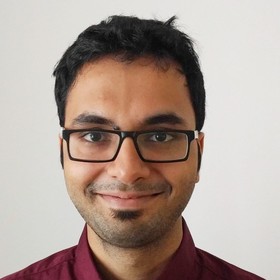} &
    \includegraphics[width=0.25\linewidth]{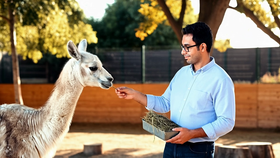} &
    \includegraphics[width=0.25\linewidth]{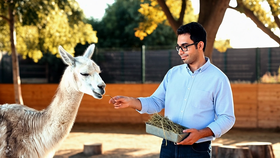} &
    \includegraphics[width=0.25\linewidth]{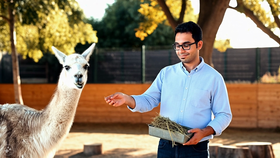} \\

    \includegraphics[width=0.14\linewidth]{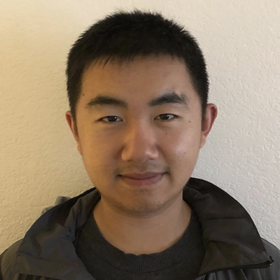} &
    \includegraphics[width=0.25\linewidth]{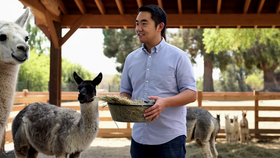} &
    \includegraphics[width=0.25\linewidth]{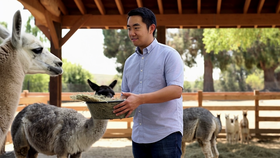} &
    \includegraphics[width=0.25\linewidth]{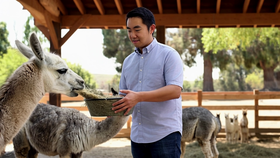} \\

    \multicolumn{1}{c}{\textbf{Reference Image}} &
    \multicolumn{3}{c}{\textit{Prompt}: A person is walking on a crowded city street} \\
    \includegraphics[width=0.14\linewidth]{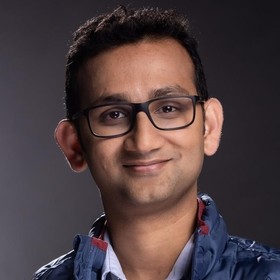} &
    \includegraphics[width=0.25\linewidth]{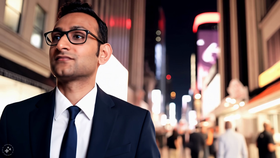} &
    \includegraphics[width=0.25\linewidth]{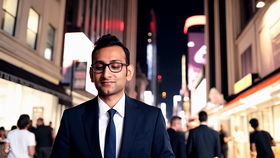} &
    \includegraphics[width=0.25\linewidth]{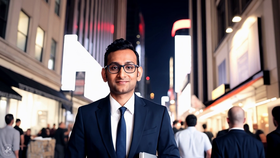} \\

    \multicolumn{1}{c}{\textbf{ }} &
    \multicolumn{3}{c}{\textit{Prompt}: A person talking with someone on a laptop} \\
    \includegraphics[width=0.14\linewidth]{figures/pt2v_qualitative/videos/roshan-laptop/frames/roshan.jpg} &
    \includegraphics[width=0.25\linewidth]{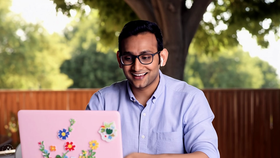} &
    \includegraphics[width=0.25\linewidth]{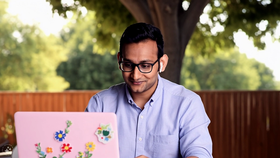} &
    \includegraphics[width=0.25\linewidth]{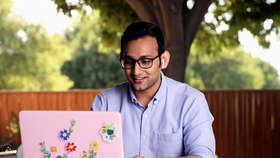} \\

    \multicolumn{1}{c}{\textbf{}} &
    \multicolumn{3}{c}{\textit{Prompt}: A person dressed in a suit is leaning against the parked car} \\
    \includegraphics[width=0.14\linewidth]{figures/pt2v_qualitative/videos/roshan-laptop/frames/roshan.jpg} &
    \includegraphics[width=0.25\linewidth]{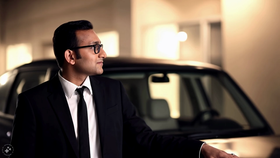} &
    \includegraphics[width=0.25\linewidth]{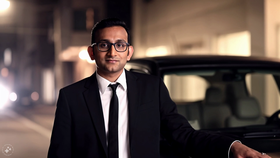} &
    \includegraphics[width=0.25\linewidth]{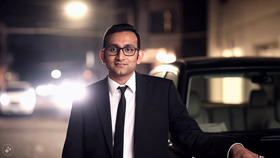} \\

    \multicolumn{1}{c}{\textbf{}} &
    \multicolumn{3}{c}{\textit{Prompt}: A person riding a bike infront of an erupting volcano} \\
    \includegraphics[width=0.14\linewidth]{figures/pt2v_qualitative/videos/roshan-laptop/frames/roshan.jpg} &
    \includegraphics[width=0.25\linewidth]{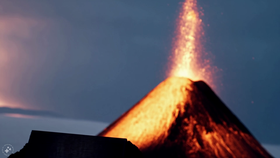} &
    \includegraphics[width=0.25\linewidth]{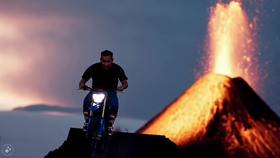} &
    \includegraphics[width=0.25\linewidth]{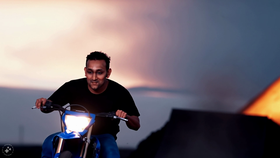} \\

\end{tabular}}

    \vspace{-3mm}
    \caption{\textbf{Generated videos from \OursPTV.} 
    \OursPTV generates high quality videos that follows the reference identity. Videos in this Figure found at \url{https://go.fb.me/MovieGen-Figure23}.}
    \label{fig:pt2v_qual_ours_figure}
\end{center}%

\section{Instruction-Guided Precise Video Editing}
\label{sec:video_editing}
\newcommand{\highlight}[1]{\textbf{#1}}
As video content continues to dominate across various platforms, the demand for accessible, controllable, and precise video editing tools is rapidly increasing.
In particular, there is a growing interest in developing text-guided video editing models.
This interest arises from the limitations of more traditional software, which is inaccessible to most users and time-consuming for expert users.
In contrast, text-guided video editing models aim to enable \textit{any} user to edit a video (whether real or generated) easily, quickly, and precisely through natural language.
However, the development of high-performing video editing models is hindered by the scarcity of supervised video editing data.
In this section we introduce \OursVideoEdit, a model that achieves state-of-the-art results in video editing, and outline our approach for training it without any supervised video editing data.\footnote{We provide examples of our model's video editing capabilities in \url{https://go.fb.me/MovieGen-Figure24}.}

Our approach for training \OursVideoEdit is guided by two main assumptions.
The first is that explicitly training the model for video editing offers significantly greater potential compared to training-free methods~\citep{meng2021sdedit,Geyer2023TokenFlowCD}.
Moreover, to fully control all aspects of the input video, we must train the model to process the entire video input rather than limited proxy features of the input video (\eg, depth maps)~\citep{esser2023structure,Liang2023FlowVidTI,Yan2023MotionConditionedIA}.
Second, unlike tasks where abundant supervised data can be collected (\eg, \textToV), it is far less practical to gather supervised video editing data.
Consequently, any large-scale training for video editing is expected to suffer from discrepancies between training and test-time scenarios.
Therefore, the second assumption is that minimizing train-test discrepancies is crucial to unlocking the model’s full potential.
Accordingly, our approach involves several training stages that aim to gradually reduce such train-test discrepancies.

When considering the \textToV model from \Cref{sec:image_video_model}, a clear discrepancy is that it was never trained to alter a media input based on an editing instruction.
Therefore, in the first stage we train the \textToV model with a multi-tasking objective that alternates between image editing, which we treat as \textit{single-frame video editing}, and video generation (\Cref{subsec:edit_stage_1}).
While the model demonstrates some generalization to video editing after this stage, it often produces blurry videos.
We attribute these artifacts to the distribution shift between training the model on single-frame video editing and testing it on multi-frame video editing.
Thus, in the second stage we introduce two new synthetic tasks that more closely resemble \textit{multi-frame video editing} and finetune the model on them (\Cref{subsec:edit_stage_2}).
The first task creates a synthetic video editing example by animating image editing examples using random affine augmentations.
The second task casts video segmentation as a video editing task, by requiring the model to mark a specific object in the video using a specific color.
After this stage, the main observed artifacts are lack of natural motion and oversaturation of newly generated elements.
To address these issues, in the third and final stage, we introduce an adaptation of backtranslation for video editing, enabling us to train the model on multi-frame, high-quality output videos.
We demonstrate that human annotators prefer \OursVideoEdit more than 74\% of the time when compared to the previous state-of-the-art~\citep{EVE} on the TGVE+ benchmark~\citep{wu2023cvpr,EVE}.

Finally, to facilitate the proper evaluation of the next generation of video editing models we collect a new comprehensive video editing benchmark, which we call \OursVideoEditBench~(\cref{subsec:editing_eval}).
This benchmark spans six different video editing tasks, each containing diverse editing instructions and corresponding videos.
Unlike previous benchmarks, which assume models are limited to square, short, low resolution, and low FPS videos, \OursVideoEditBench includes videos with varied aspect ratios, resolutions, FPS, and more.

\subsection{Model}\label{sec:edit_method}
\begin{figure}[h]
    \centering
    \includegraphics[width=1.0\textwidth]{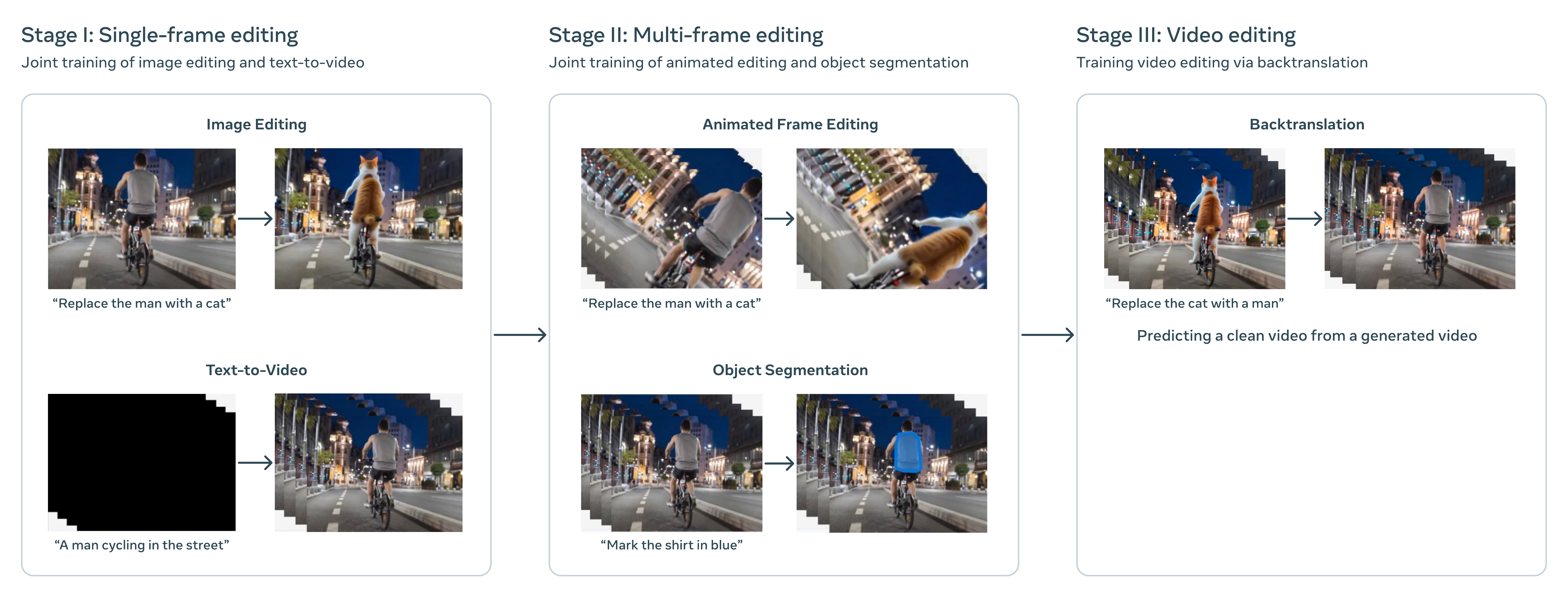}
    \caption{\textbf{Extending the \textToV model to video editing.} We add video editing capabilities to the \textToV model using three training stages: single-frame editing (\Cref{subsec:edit_stage_1}), multi-frame editing (\Cref{subsec:edit_stage_2}), and video editing via backtranslation (\Cref{subsec:edit_stage_3}). We provide examples of our model's video editing capabilities in \url{https://go.fb.me/MovieGen-Figure24}.}
    \label{fig:edit}
\end{figure}

Given the scarcity of supervised video editing data, methods for training models to perform video editing are prone to train-test discrepancies, resulting in suboptimal quality.
To address this challenge, we introduce a multi-stage approach that progressively minimizes these discrepancies.
We explain below the architecture modifications made to support video editing and then detail each step of our approach.
The process is visualized in~\Cref{fig:edit}.

\subsubsection{Model Architecture and Initialization}
\label{subsec:edit_init}
To support video editing, we introduce several adaptations to the architecture described in \Cref{sec:image_video_model}.
First, we enable input video conditioning by adding additional input channels to the patch embedder.
This allows us to concatenate the latent video input with the noisy output latent video along the channels dimension, and provide the concatenated latent videos to the model.
Additionally, following Emu Edit~\citep{emuedit}, we incorporate support for conditioning the model on specific editing tasks (\eg, adding an object, changing the background, \etc).
Specifically, our model has a learned task embedding vector for each task.
For a given task, the model applies a linear transformation on the corresponding task embedding, producing four embeddings that are concatenated to the text encoders' hidden representations.
We also apply a second linear transformation to the task embedding, and add the resulting vector to the \timestep embedding.
Crucially, to fully preserve the model's video generation capabilities, we set all newly added weights to zero and initialize the remaining weights from the pre-trained \textToV model.

Formally, the video editing architecture is conditioned on the following triplet $\bc = (\text{TAE}(\bc_{vid}), \bc_{instruct}, j)$, where $\bc_{vid}$ is the input video, $\text{TAE}$ is the temporal auto-encoder, $\bc_{instruct}$ is the editing instruction prompt, and $j$ is the task-id of the relevant editing operation.
We update the flow step in Eq.~\ref{eq:fm} to be $u(\bx_t, \bc, \FMTime; \theta)$, where $\bx_{\FMTime}$ are the latents of the output video $x_{vid}$ at step flow $\FMTime$, and $\theta$ are the model parameters.
For brevity, we omit the task-id, $j$, and the activation of the temporal autoencoder, $\text{TAE}$, in the rest of the section.

\subsubsection{Stage~I: Single-frame Video Editing}
\label{subsec:edit_stage_1}
We begin by training the model to utilize an editing instruction and a video input during the denoising process of the output video.
However, since we lack supervised video editing data, we leverage an image editing dataset, treating image editing as single-frame video editing.
Concretely, the image editing dataset is composed of triplets of $\bc_{img-edit}=(\bc_{img}, \bc_{instruct}, x_{img})$, where $\bc_{img},x_{img}$ are the input and output images which we treat as single-frame videos.
Clearly, high-quality \textit{video} editing demands more than just precise editing of individual frames.
For example, it is essential to ensure that the output video maintains temporal consistency and that any newly generated elements appear natural.
Therefore, we aim to preserve the temporal consistency and generation quality of our model by simultaneously training it on both image editing and \textToV generation.

As the new model architecture expects a video input as an additional condition, we condition the model on a black video during video generation training.
Formally, given a \textToV dataset with pairs $(\bc_{txt},x_{vid})$ of caption and target video, we create the following triplet, $\bc_{text-to-video} = (\bc_{\emptyset}, \bc_{instruct}, x_{vid})$, where $\bc_{\emptyset}$ is a black video, and $\bc_{instruct}$ is the video output caption with $\bc_{txt}$ rephrased as an instruction.

Due to difference in the sequence length between image editing and video generation, an image editing step requires significantly fewer operations than a video generation step.
Therefore, we accelerate training by alternating between image editing and video generation batches, instead of mixing both tasks within each batch.
Additionally, because our model is already trained on \textToV generation, we further accelerate training by sampling image editing batches five times more frequently than video generation batches.
Hence, we update Eq.~\ref{eq:fm} as follows:
\begin{equation}
\label{eq:fm_v2v_stage1}
\mathbb{E}_{\FMTime, \bx_0, \bx_1, \bc}\|u(\bx_{\FMTime}, \bc, \FMTime; \theta) - \bv_t\|^2, \quad \text{where } \bc \sim \text{Categorical}\left( \left\{ \bc_{text-to-video}: \frac{1}{6}, \bc_{img-edit}: \frac{5}{6} \right\} \right).
\end{equation}

Interestingly, in preliminary experiments we found that na\"ively using the first frame's positional embedding during image editing training leads to completely distorted outputs when testing the model on video editing.
We resolve this issue by instead using a randomly sampled temporal positional embedding as the positional embedding for the image
We train the model using this objective for thirty thousand steps.

\subsubsection{Stage~II: Multi-frame Video Editing}
\label{subsec:edit_stage_2}
\begin{figure}[h]
    \centering
    \includegraphics[width=1.0\textwidth]{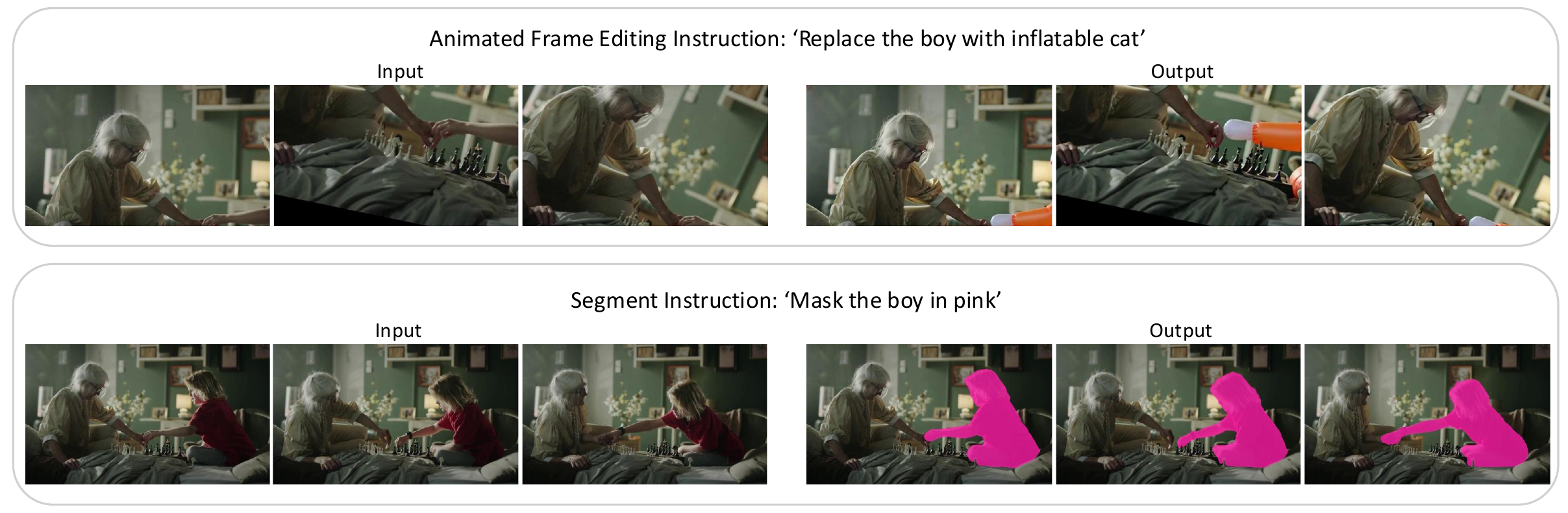}
    \caption{\textbf{Multi-Frame Editing Stage.} The model is trained on two synthetic multi-frame editing tasks: Animated Frame Editing and Generative Instruction-Guided Video Segmentation.}
    \label{fig:edit_stage2}
\end{figure}

The trained model from Stage~I (\ref{subsec:edit_stage_1}) is capable of both precisely editing images and generating high-quality videos from text.
However, it produces very blurry edited videos when tasked with video editing.
We hypothesize that these artifacts are due to train-test discrepancies between Stage~I training and video editing.
The most significant discrepancy that we identify is that the model is not conditioned on multi-frame video inputs during Stage~I training.
We try to mitigate the blurriness artifacts by creating two complementary datasets that do include multi-frame videos inputs and outputs.
We describe each of these datasets below and discuss the model's performance after training on them.
Additionally, we visualize the two tasks in \Cref{fig:edit_stage2}.

\textbf{Animated Frame Editing.}
We create an animated frame editing dataset by leveraging a video-caption pair dataset  $(\bc_{txt}, x_{vid})$.
The process begins by prompting a language model (\eg, \llama) with the caption  $\bc_{txt}$~(\eg, ``A person walking down the street'') to generate an editing instruction $\bc_{instruct}$~(\eg, ``Put the person at the beach'') and an output caption for the desired edited image $\hat{\bc}_{txt}$ (\eg, ``A person walking at the beach'').
Next, a random frame  $x_{frame}$  is selected from  $x_{vid}$, and we apply a single-frame editing model, $p_\theta$~(introduced in Stage~I, \Cref{subsec:edit_stage_1}) to generate an edited frame $\hat{x}_{frame} \sim p_\theta(x_{frame}, \bc_{instruct})$.
We filter the resulting data points, $(\bc_{txt}, \bc_{instruct}, \hat{\bc}_{txt}, x_{frame}, \hat{x}_{frame})$ using automated image editing metrics in a similar method to the one described in~\citep{emuedit}.
To animate both the input and edited frames, we use an iterative process.
In each iteration $i<n$, a random affine transformation $\mathcal{F}_i$ is applied to both the input frame $x_{frame}^{(i-1)}$ and the edited frame $\hat{x}_{frame}^{(i-1)}$, producing the next frames $x_{frame}^{(i)}$ and $\hat{x}_{frame}^{(i)}$.
This process results in an animated sequence of input frames $\hat{\bc}_{vid} = \{x_{frame}^{(i)}\}_{i=0}^n$ and edited frames $\hat{x}_{vid} = \{\hat{x}_{frame}^{(i)}\}_{i=0}^n$.
Finally, combining the animated frames with the editing instruction forms a multi-frame editing example $\bc_{animated}=(\hat{\bc}_{vid}, \bc_{instruct}, \hat{x}_{vid})$.
The full process is outlined in Algorithm~\ref{alg:augmentation}, and a visual example of animated frame editing is provided in \Cref{fig:edit_stage2}.
\begin{algorithm}[H]
    \caption{Animated Frame Editing Dataset Creation}
    \label{alg:augmentation}
    \begin{algorithmic}[1]
        \Require Video-caption dataset $ (c_{txt}, x_{vid}) $, editing model $ p_\theta $
        \State \textbf{Input:} Video $ x_{vid} $, Caption $ c_{txt} $
        \State \textbf{Output:} Animated frames $ \hat{c}_{vid} $, Editing instruction $ c_{instruct} $, Animated edited frames $ \hat{x}_{vid} $

        \State Generate editing instruction: $ c_{instruct} \sim \text{\llama}(c_{txt}) $
        \State Sample random frame: $ x_{frame} \sim x_{vid} $
        \State Generate edited frame: $ \hat{x}_{frame} \sim p_\theta(x_{frame}, c_{instruct}) $

        \State Initialize animated sequences: $ \hat{c}_{vid} \gets \emptyset, \hat{x}_{vid} \gets \emptyset $
        \State Set initial frames: $ x_{frame}^{(0)} \gets x_{frame}, \hat{x}_{frame}^{(0)} \gets \hat{x}_{frame} $

        \For{$i = 1$ to $n$}
            \State Sample random affine augmentation: \( \mathcal{F}_i \)
            \State Apply augmentation: \( x_{frame}^{(i)} \gets \mathcal{F}_i(x_{frame}^{(i-1)}) \), \( \hat{x}_{frame}^{(i)} \gets \mathcal{F}_i(\hat{x}_{frame}^{(i-1)}) \)
            \State Append frames to sequences: \( \hat{c}_{vid} \gets \hat{c}_{vid} \cup \{x_{frame}^{(i)}\} \), \( \hat{x}_{vid} \gets \hat{x}_{vid} \cup \{\hat{x}_{frame}^{(i)}\} \)
        \EndFor
        \State \textbf{Return:} $ (\hat{c}_{vid}, c_{instruct}, \hat{x}_{vid}) $
    \end{algorithmic}
\end{algorithm}

\textbf{Generative Instruction-Guided Video Segmentation.}
The lack of natural motion in animated frame editing examples poses a clear discrepancy between animated frame editing and video editing.
To address this, we complement the animated frame editing task with the task of generative instruction-guided video segmentation, which extends the Segment task from Emu Edit~\citep{emuedit} from images to videos.
In this task, the model is required to edit a video by marking a specific object in a particular color based on the given instruction.

We begin by collecting editing instructions using a procedure similar to the one employed while collecting animated frame editing examples.
However, we prompt the language model to generate an instruction, $\bc_{instruct}$, to \textit{mark} a particular subject or object in the video in a specific color, and to output the name of the edited object (\eg, ``apple'').
We then use DINO~\citep{groundingdino} and SAM 2~\citep{ravi2024sam2} to extract the segmentation mask for the object in the video.
Finally, we create the target video, $\hat{x}_{vid}$, by marking the object in the relevant color using the extracted segmentation mask.
Following the notation described above, the paired data, $\bc_{segmentation}=(\bc_{vid}, \bc_{instruct}, \hat{x}_{vid})$, then consists of a real input video, $\bc_{vid}$, an instruction to mark a specific object in a certain color, $\bc_{instruct}$, and a corresponding edited video, $\hat{x}_{vid}$.

\textbf{Training.}
We finetune the model from Stage~I on these datasets, alongside \textToV generation using multi-task training for one thousands steps.
During training we sample animated frame editing examples three times more frequently than generative instruction-guided video segmentation and \textToV generation.
To put it formally, we update the sampling in Eq.~\ref{eq:fm_v2v_stage1} as follows
\begin{equation}
\label{eq:fm_v2v_stage2}
\bc \sim \text{Categorical}\left( \left\{ \bc_{text-to-video}: \frac{1}{5}, \bc_{animated}: \frac{3}{5}, \bc_{segmentation}: \frac{1}{5} \right\} \right).
\end{equation}
We observe that this stage mitigates the blurriness artifacts from Stage~I; however, newly generated elements in the edited video exhibit less motion than desired, and at times appear oversaturated.

\subsubsection{Stage~III: Video Editing via Backtranslation}
\label{subsec:edit_stage_3}
\begin{figure}[h]
    \centering
    \includegraphics[width=1.0\textwidth]{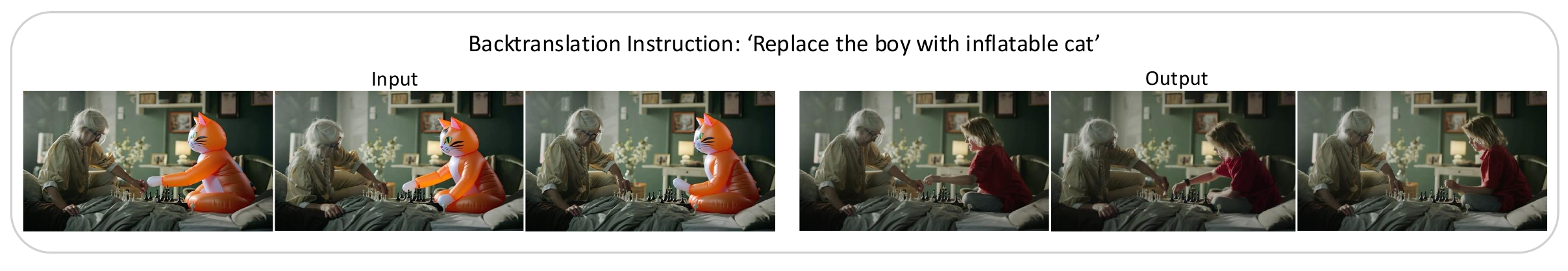}
    \caption{\textbf{Backtranslation Stage.} The model is trained to denoise the clean input video while conditioning on the edited generated video.}
    \label{fig:edit_stage3}
\end{figure}
While Stage~II training (\Cref{subsec:edit_stage_2}) mitigates most of the artifacts observed in Stage~I (\Cref{subsec:edit_stage_1}), we notice that newly generated elements often lack motion and sometimes appear oversaturated.
These artifacts are likely due to the output videos in the animated frame editing dataset, which lack natural motion and are model-generated.
Therefore, in Stage~III, we create video editing data with \textit{real} output videos.
Similarly to Stage~II~(\Cref{subsec:edit_stage_2}), we assume access to a dataset of videos $x_{vid}$ and corresponding captions $\bc_{txt}$~(\eg, ``\textit{Apples on a table}'').
We employ \llama to create an editing instruction $\bc_{instruct}$~(\eg, ``\textit{Put the apples in a small basket}''), and an output caption $\hat{\bc}_{txt}$~(\eg, ``\textit{Apples in a small basket on the table}'').
Then, we use the model from Stage~II to generate an edited video $\hat{x}_{vid} \sim p_\theta(x_{vid},\bc_{instruct})$ based on the input video $x_{vid}$ and editing instruction $\bc_{instruct}$.
Afterward, we utilize $\bc_{txt}, \hat{\bc}_{txt}, x_{vid}, \hat{x}_{vid}$ to filter the generated examples based on automatic ViCLIP scores, following a similar filtering process as in Stage~II.

A na\"ive approach would be to tune the model on the resulting dataset, $(x_{vid}, \bc_{instruct}, \hat{x}_{vid})$, teaching the model to predict its own generations.
However, in this case, the output videos are likely to contain the very same artifacts we aim to mitigate.
Therefore, we adapt the backtranslation technique from natural language processing~\citep{edunov2018understandingbacktranslationscale} to video editing.
Specifically, we prompt \llama using $(\bc_{txt}, \hat{\bc}_{txt}, \bc_{instruct})$ to generate an editing instruction, $\bc_{instruct-bwd}$, that should alter the generated video $\hat{x}_{vid}$ into the original video $x_{vid}$~(\eg, ``\textit{Remove the small basket and put the apples on the table}'').
Then, we build a synthetic paired dataset, $\bc_{backstranslation}=(\hat{x}_{vid}, \bc_{instruct-bwd}, x_{vid})$, and use it to train the model to denoise the clean video $x_{vid}$ while conditioning on the potentially noisy video $\hat{x}_{vid}$ and editing instruction $\bc_{instruct-bwd}$.
In this manner, we construct a weakly-supervised video editing dataset, with real output videos.

\subsection{Evaluation}
\label{subsec:editing_eval}
We evaluate the capabilities of our model against two main video editing benchmarks.
The first benchmark, TGVE+~\citep{EVE}, is a recently proposed extension of the TGVE benchmark~\citep{wu2023cvpr}.
While this benchmark is comprehensive, it features low-resolution, low-FPS, short, and square videos.
This is in contrast to state-of-the-art video generation models and most media content, which typically feature higher resolution, longer videos with higher FPS, and varied aspect ratios.
Therefore, to enable proper evaluation of next-generation video editing models with more relevant video inputs, we introduce a new benchmark, called \OursVideoEditBench.
This benchmark consists of videos with varying resolutions, FPS, lengths, and aspect ratios.
We compare our approach against several baselines and measure its effectiveness across multiple axes, including fidelity to the user instructions and input video, and overall visual quality.

\subsubsection{Video Editing Benchmarks}
\label{edit:benchmark}
The TGVE+ benchmark~\citep{EVE,wu2023cvpr} consists of seventy-six videos, each accompanied by seven editing instructions for the following tasks: (i) local object modification, (ii) style change, (iii) background change, (iv) simultaneous execution of multiple editing tasks, (v) object removal, (vi) object addition, and (vii) texture modification.
While the benchmark offers a comprehensive evaluation across a diverse set of editing tasks, the videos in the benchmark are of 480 $\times$ 480 px resolution, and 3.20 seconds length at 10 FPS, or 8.00 seconds at 16 FPS.
In contrast, real user videos are expected to have a higher resolution, higher FPS, and may contain various aspect ratios.
Hence, it is unclear whether evaluation against TGVE+ will accurately reflect video editing performance on real user videos.
Moreover, current foundational video generation models~\citep{sora,gen2,gen3} can operate at high resolution~(\eg, 768p or 1080p), 16 or more FPS, multiple aspect ratios, and can process much longer videos than those from TGVE.

Thus, to enable the evaluation of video editing using more practical videos, we collect a new benchmark, \OursVideoEditBench, that aims to evaluate the video editing capabilities of the next generation of video editing models.
To build \OursVideoEditBench, we rely on videos from the publicly released Segment-Anything-V2~\citep{ravi2024sam2} dataset.
For each video out of the 51,000 videos found in the dataset, we generate a caption using a similar approach to the one in \Cref{sec:pt-data}, calculate its motion score~\citep{motion,opencv_library}, and calculate its aesthetics score~\citep{schuhmann2022laion}.
We then filter all videos with an aesthetics score lower than the median score of the dataset.
For each category, we bin the videos based on their motion score and sample videos uniformly from the bins for each category.
Overall, the benchmark validation set has 64 videos, whereas the test set has 128 videos.

To facilitate a realistic benchmark with editing instructions written by humans, we employ crowd workers.
For each video and for each editing operation, we assign crowd workers the task of writing down a creative editing instruction.
Finally, to support the use of CLIP-based image editing evaluation metrics (similar to those used in~\citep{emuedit}), we additionally collect an input caption and output caption.
Thus, the benchmark has altogether 1,152 examples, and spans six different editing tasks.

\subsubsection{Video Editing Measures}\label{subsubec:edit_measures}
Our experiments evaluate the ability of video editing models to modify an input video while accurately following the provided instructions and preserving the structure and elements that should remain unchanged.
We assess the video editing performance of our model and the baselines using both Human evaluation and automated metrics.
For automated evaluation, we use the main automatic metrics reported by \citep{EVE}, which account for both temporal and spatial coherence.
Specifically, we measure (i) ViCLIP text-image direction similarity~($\text{ViCLIP}_{dir}$), which evaluates the alignment between changes in captions and corresponding changes in the videos, and (ii) ViCLIP output similarity~($\text{ViCLIP}_{out}$), which measures the similarity between the edited video and the output caption.

For Human evaluation, we follow the standard evaluation protocol of TGVE+~\citep{EVE,wu2023cvpr}.
Human annotators are presented with an input video, the editing instruction, and a pair of edited videos.
We then ask the raters to respond to the following questions: (i) Text Alignment: Which edited video more accurately reflects the given caption, (ii) Structure: which edited video better maintains the structural integrity of the original input, and (iii) Quality: which edited video is visually more appealing and aesthetically superior.
Additionally, we extend this protocol with a fourth question: (iv) Overall: considering quality, structure, and text alignment, which edited video is better.

\subsection{Results}
In this section, we compare our model with leading video editing baselines.
We then analyze the importance and impact of the main design and implementation choices in our approach (\Cref{edit:abl}).

\subsubsection{Comparisons to Prior Work}
\label{edit:eval}
We evaluate our model against different video editing baselines, including both training-free methods, and methods that require prior training such as our method.
A common training-free method for video editing is Stochastic Differential Editing~(SDEdit)~\citep{meng2021sdedit} which performs image editing by adding noise to the input video and then denoising it while conditioning the model on a descriptive caption.
Recent video foundation models~\citep{BarTal2024LumiereAS,videoworldsimulators2024}, have used SDEdit for video editing, and demonstrated its ability to maintain the overall structure of the input video.
However, this approach can lead to the loss of important details, such as subject identity and texture, making it less effective for precise editing.
In our experiments, we utilize SDEdit with the base T2V model, \OursVideo, and perform the denoising process for 60\% of the total iterations.
Another prominent approach for video editing is to inject information about the input or generated video from key frames via cross-attention interactions~\citep{Wu2023FairyFP,yatim2023space}.

On the other hand, the current top performing methods for video editing utilize prior training while overcoming the lack of supervised datasets for video editing.
For example, InsV2V~\citep{Cheng2023ConsistentVT} extends the general approach of InstructPix2Pix~\citep{brooks2022instructpix2pix} to video editing, enabling the creation and training of a video editing model using synthetic data.
EVE~\citep{EVE}, relies on unsupervised training by employing knowledege distillation from two expert models, one for image editing and the other for \textToV generation (see \Cref{subsec:edit_background} for more details).
Finally, we compare to Tune-A-Video (TAV)~\citep{wu2023tune} which served as the baseline in the TGVE contest.
TAV tunes a \textToI model to a specific video, followed by inverting the input video and using the inverted noise to generate the output video.
We compare with all of the baselines described above on the TGVE+ benchmark.

In addition, we evaluate our method versus SDEdit and Runway Gen3 Video-to-Video\footnote{Runway Gen3 Video-to-Video videos were collected on September 24th, 2024.}~(Runway Gen3 V2V)~\citep{gen3} on \OursVideoEditBench~(Sec.~\ref{edit:benchmark}).
We compare to Runway Gen3 V2V in two settings.
The first, employs Runway Gen3 V2V on all tasks comprising the benchmark -- (i) local object modification, (ii) style change, (iii) background change, (iv) object removal, (v) object addition, and (vi) texture modification.
However, as we observe that Runway Gen3 V2V struggles to preserve fine details in the input video (in contrast to general structure), in the second setting we focus on the style editing task, denoted as Runway Gen3 V2V Style.
We omit comparison with other baselines, as they are mostly limited to operating on short videos with 32 frames and do not fully utilize \OursVideoEditBench's videos duration, resolution, or varied aspect ratios.

Results of our evaluation versus all baselines are presented in~\Cref{tab:eval_video_editing_tgve}.
Throughout this section we report `win rates', which can lie in the range $[0,100]$, where 50 indicates a tie between two models.
Human raters prefer \OursVideoEdit over all baselines by a significant margin on both benchmarks.
On the TGVE+ benchmark, our model is preferred 74\% more often than the current state-of-the-art EVE in the overall Human evaluation criterion.
In terms of automated metrics, \OursVideoEdit presents state-of-the-art results on the $\text{ViCLIP}_{dir}$ metric.
On the $\text{ViCLIP}_{out}$ metric, \OursVideoEdit performance is comparable to EVE.
However, unlike \OursVideoEdit, EVE has access to the video output caption which is used for calculating the $\text{ViCLIP}_{out}$ score.

On the \OursVideoEditBench, our method is preferred over the Runway Gen3 V2V and Runway Gen3 V2V Style settings.
Interestingly, when compared to Runway Gen3 V2V Style, Human evaluation metrics highlight our advantage in maintaining the structure of the input videos.
Compared to SDEdit, \OursVideoEdit is preferred by human raters in Human evaluation criterions by a significant margin, despite a lower $\text{ViCLIP}_{out}$ score.
Similarly to EVE, SDEdit has an advantage in the $\text{ViCLIP}_{out}$ automatic metric as it has access to the same output caption that $\text{ViCLIP}_{out}$ uses.

\begin{table*}[t]
\centering
\adjustbox{max width=\textwidth}{%
\begin{tabular}{llcccccc}
\toprule
\multirow{2}{*}{Dataset} & \multicolumn{1}{c}{\multirow{2}{*}{Method}}  & \multicolumn{4}{c}{Human Evaluation} & \multicolumn{2}{c}{Automated} \\
    & & Text & Struct. &  Quality & Overall &  $\text{ViCLIP}_{dir}\uparrow$ & $\text{ViCLIP}_{out}\!\uparrow$ \\
\midrule
\multirow{6}{*}{\shortstack[c]{TGVE+}} & TAV~\citep{wu2023tune} & 85.00 & 81.94 & 91.57 & 89.70 & 0.131 & 0.242 \\
& STDF~\citep{yatim2023space}   & 84.43 & 61.60 & 73.21 & 74.43 & 0.093 & 0.227 \\
& Fairy~\citep{Wu2023FairyFP}   & 84.15  & 77.52 & 84.20 & 84.91 & 0.140 & 0.197   \\
& InsV2V~\citep{Cheng2023ConsistentVT} & 73.75 & 66.60 & 70.73 & 70.85 & 0.174 & 0.236 \\
& SDEdit~\citep{meng2021sdedit} & 85.51 & 90.07 & 76.19 & 80.59 & 0.131 & 0.241 \\
& EVE~\citep{EVE} & 69.48 & 70.05 & 75.18 & 74.38 & 0.198 & 0.251  \\
& \OursVideoEdit (Ours) & -- & -- & -- & -- & 0.225 & 0.248 \\
\midrule
\multirow{4}{*}{\OursVideoEditBench}
& Runway Gen3 V2V~\citep{gen3} & 88.14 & 98.33 & 83.14 & 93.33 & 0.068 & 0.188 \\
& Runway Gen3 V2V Style~\citep{gen3} & 55.55 & 73.61 & 58.33 & 59.72 & 0.124 & 0.214 \\
& SDEdit~\citep{meng2021sdedit} & 94.37 & 86.34 & 85.14 & 91.96 & 0.124 & 0.239 \\
& \OursVideoEdit (Ours) & -- & -- & -- & -- & 0.209 & 0.224 \\
\bottomrule
\end{tabular}}
\caption{\textbf{Comparison with video editing baselines on the TGVE+ and \OursVideoEditBench benchmarks.}
We report ViCLIP metrics and human ratings.
Human evaluation shows the win rate of \OursVideoEdit~(Ours) against the baselines.
Runway Gen3 videos were collected on September 24th, 2024.
For the human evaluations we report `win rates', which can lie in the range $[0,100]$, where 50 indicates a tie between two models.}
\label{tab:eval_video_editing_tgve}
\end{table*}

\subsubsection{Ablations}
\label{edit:abl}
In this section, we aim to assess and quantify the importance and impact of the main design and implementation choices in our approach.
Unless stated otherwise, ablations are conducted on the validation set of \OursVideoEditBench (\Cref{edit:benchmark}) using the Human evaluation metrics described in \Cref{subsubec:edit_measures}.

\textbf{Stage~I: Multi-tasking versus Adapter.}
As mentioned in \Cref{subsec:edit_stage_1}, the first stage of our approach involves training the model using a multi-tasking objective that alternates between image editing and video generation.
However, an alternative approach would be to train an image editing adapter on top of the \textToV generation model.
The advantage of this approach is that by freezing the weights of the \TextToV model, one can ensure that the model's video generation capabilities are preserved.
However, this approach is more memory demanding and typically requires providing the model with two text inputs for video editing: (i) a video caption for the frozen text-to-video model, and (ii) an editing instruction for the trained adapter.

To ablate this design choice, we implement a variant of a ControlNet adapter~\citep{zhang2023adding} that aligns with the model described in \Cref{subsec:edit_stage_1}.
Specifically, we freeze the original \textToV model and clone a trainable copy of it.
We apply the same adaptations as described in \Cref{subsec:edit_init} to the trainable model.
Finally, we follow \citep{zhang2023adding} by introducing a zero-initialized convolutional layer after each layer of the trainable model.
During the forward pass, the frozen model gets a caption that describes the output video, and the trainable model gets similar inputs as described in \Cref{subsec:edit_stage_1}.
After each layer we add the hidden states of the trainable model to the hidden states of the frozen model.

We train the adapter on image editing using the same image editing data as used in Stage~I (\Cref{subsec:edit_stage_1}) for 10K iterations and compare it to the Stage~I model trained for the same number of iterations.

We evaluate the models' image editing capabilities on the Emu Edit benchmark~\citep{emuedit} and measure performance using the $L_1$ distance between the input and output images, distance between DINO~\citep{groundingdino} features of the input and output images, and several CLIP-based image editing evaluation metrics:
\noindent{$\text{CLIP}_{im}$} estimates whether the model preserved elements from the input image by measuring the CLIP-space distance between the input image and the edited image.
\noindent{$\text{CLIP}_{out}$} estimates whether the model followed the editing instruction by measuring the CLIP-space distance between a caption describing the desired edited image and edited image itself.
\noindent{$\text{CLIP}_{dir}$} estimates if the elements that were supposed to change were edited correctly, while ensuring that elements intended to remain unchanged were preserved.
Furthermore, we conduct a human evaluation in which human raters assess text alignment and image faithfulness.
During this assessment the human raters see the original image and instruction alongside two modified images and are asked: (i) which edited image better preserves the required elements from the input image, and (ii) which edited image best follows the editing instruction.

As can be seen, the Stage~I variant achieves comparable results to the ControlNet variant on {$\text{CLIP}_{im}$}, {$\text{CLIP}_{out}$}, and DINO metrics.
However, it achieves a significantly better performance on both the {$\text{CLIP}_{dir}$} and L1 metrics.
Furthermore, Human evaluation indicates that full model training results in edits that are better aligned with the editing instructions and more faithful to the input images.
This indicates that full model training can better support high quality editing than a ControlNet adapter.

\begin{table}[h!]
\centering

\adjustbox{max width=\textwidth}{%
\begin{tabular}{lccccccc}
\toprule
 \multicolumn{1}{c}{\multirow{2}{*}{Method}} & \multicolumn{2}{c}{Human Evaluation} & \multicolumn{5}{c}{Automated} \\
 & Text align. & Image faith. & $\text{CLIP}_{dir}\!\uparrow$ &  $\text{CLIP}_{im}\!\uparrow$ & $\text{CLIP}_{out}\!\uparrow$ &  $\text{L1}\!\downarrow$  & DINO$\uparrow$ \\
\midrule
ControlNet adapter & 69.325 & 66.235 & 0.115 & 0.843 & 0.235 & 0.099 & 0.794 \\
Stage~I (\Cref{subsec:edit_stage_1}) & -- & -- & 0.128 & 0.840 & 0.238 & 0.088 & 0.789 \\
\bottomrule
\end{tabular}}
\caption{\textbf{Ablation on the best way to incorporate image editing information to a \textToV model.} We compare two variants: (i) training a ControlNet on a frozen \textToV model, and (ii), multitask learning on \textToV and image editing.
Human evaluation shows the win rate of the Stage~I model against the ControlNet adapter.}
\label{tab:eval_image_editing}
\end{table}

\begin{table}[h!]
    \centering

    \adjustbox{max width=\textwidth}{%
    \centering
    \begin{tabular}{l@{\hspace{1cm}}cccc}
        \toprule
        Method & Text & Structure &  Quality & Overall \\
        \midrule
        Animated Image Editing             & 70.67 & 53.4 & 61.31 & 61.16 \\
        \bottomrule
    \end{tabular}}
    \caption{\textbf{Contribution of training the second stage using animated \textit{frames} compared to animated \textit{images}.}  Human evaluation shows the win rate of the Stage~II model (\Cref{subsec:edit_stage_2}) versus its animated \textit{image} editing counterpart.}
    \label{tab:abl_study_stage2}
\end{table}

\begin{table}[h!]
    \centering

    \adjustbox{max width=\textwidth}{%
    \centering
    \begin{tabular}{l@{\hspace{1cm}}cccc}
        \toprule
        Method & Text & Structure &  Quality & Overall \\
        \midrule
            Standard  & 44.66 & 72.56 & 73.61 & 70.23 \\
        \bottomrule
    \end{tabular}}
    \caption{\textbf{Contribution of training with backtranslation rather than standard fine-tuning.} Human evaluation shows the win rate of \OursVideoEdit versus performing Stage~III (\Cref{subsec:edit_stage_3}) with standard fine-tuning rather than backtranslation.}
    \label{tab:abl_study_stage3}
\end{table}

\begin{table}[h!]
    \centering

    \adjustbox{max width=\textwidth}{%
    \centering
    \begin{tabular}{l@{\hspace{1cm}}cccc}
        \toprule
        Method & Text & Structure &  Quality & Overall \\
        \midrule
        Stage~II (vs Stage~I)   & 61.65 & 90.86 & 91.9 & 89.29 \\
        \OursVideoEdit (vs Stage~II)   & 49.36 & 59.78 & 61.86 & 60.82 \\
        \bottomrule
    \end{tabular}}
    \caption{\textbf{Contribution of each stage in our approach.} In each row, we show the win rate of the model from a certain stage when compared to its previous stage counterpart.}
    \label{tab:abl_study}
\end{table}

\textbf{Stage~II: Animated Frame/Image Editing}
As mentioned in \Cref{subsec:edit_stage_2}, the second stage of our approach involves finetuning the model from Stage~I (\Cref{subsec:edit_stage_1}) on an animated \textit{frame} editing dataset.
However, a more straightforward alternative would have been to animate the \textit{image} editing dataset from Stage I, thereby avoiding the need to collect a new single-frame editing dataset.
To explore this choice, we train the model from Stage~I using a similar approach to the one described in \Cref{subsec:edit_stage_2}, but animate the image editing dataset from Stage~I rather than the frame-editing dataset.
As shown in \Cref{tab:abl_study_stage2}, human raters consistently prefer the outputs of the model from Stage~II over its animated image editing counterpart.
Specifically, they find the model to be more text faithful in over 70\% of the time, and rate its quality higher over 61\% of the time.

\textbf{Stage~III: Backtranslation versus Standard Fine-tuning.}
During Stage~III (\Cref{subsec:edit_stage_3}), we generate edited videos using the model from Stage~II, apply filtering, and then perform backtranslation training.
In this ablation, we assess whether backtranslation is necessary or if standard fine-tuning on model generated outputs suffices.
We follow the same training protocol as in Stage~III, but instead of backtranslation, we train the model to predict the generated video, $\hat{x}_{vid}$ from the input video $x_{vid}$ and original editing instruction $\bc_{instruct}$.
As shown in~\Cref{tab:abl_study_stage3}, while training with backtranslation slightly degrades text faithfulness when compared to standard finetuning, it provides very significant improvements in structure, quality, and crucially, overall preference.

\textbf{Evaluating the Contribution of Each Stage.}
To assess the contribution of each training stage, we compare the model from each stage with the model from the previous stage.
As shown in \Cref{tab:abl_study}, Stage~II (\Cref{subsec:edit_stage_2}) demonstrates significant improvements over the Stage~I model, with human evaluators preferring it more than 89\% of the time.
The benefits of Stage~III (\Cref{subsec:edit_stage_3}) are more subtle, with human evaluators preferring \OursVideoEdit over the Stage~II model in more than 60\% of cases.
Importantly, most of the contributions from Stage~III are reflected in the improved quality of the edited videos, with only a very minor trade-off in text faithfulness.

\section{Joint Sound Effect and Music Generation}
\label{sec:audio_model}
Our goal with \OursAudio is to generate soundtracks for both video clips and short films~\citep{holman2012sound},
which may range from a few seconds to a few minutes.
The soundtrack considered in this work includes ambient sound, sound effects (Foley), and instrumental music,
but does not include speech or music with vocals.
In particular, the ambient sound should match the visual environment,
the sound effects should be temporally aligned with the actions and plausible with respect to the visual objects,
music should express the mood and sentiment of the video,
blend properly with sound effects and ambient,
and align with scenes as what one would expect when watching a movie.

\begin{figure}[h]
    \centering
    \includegraphics[width=\linewidth]{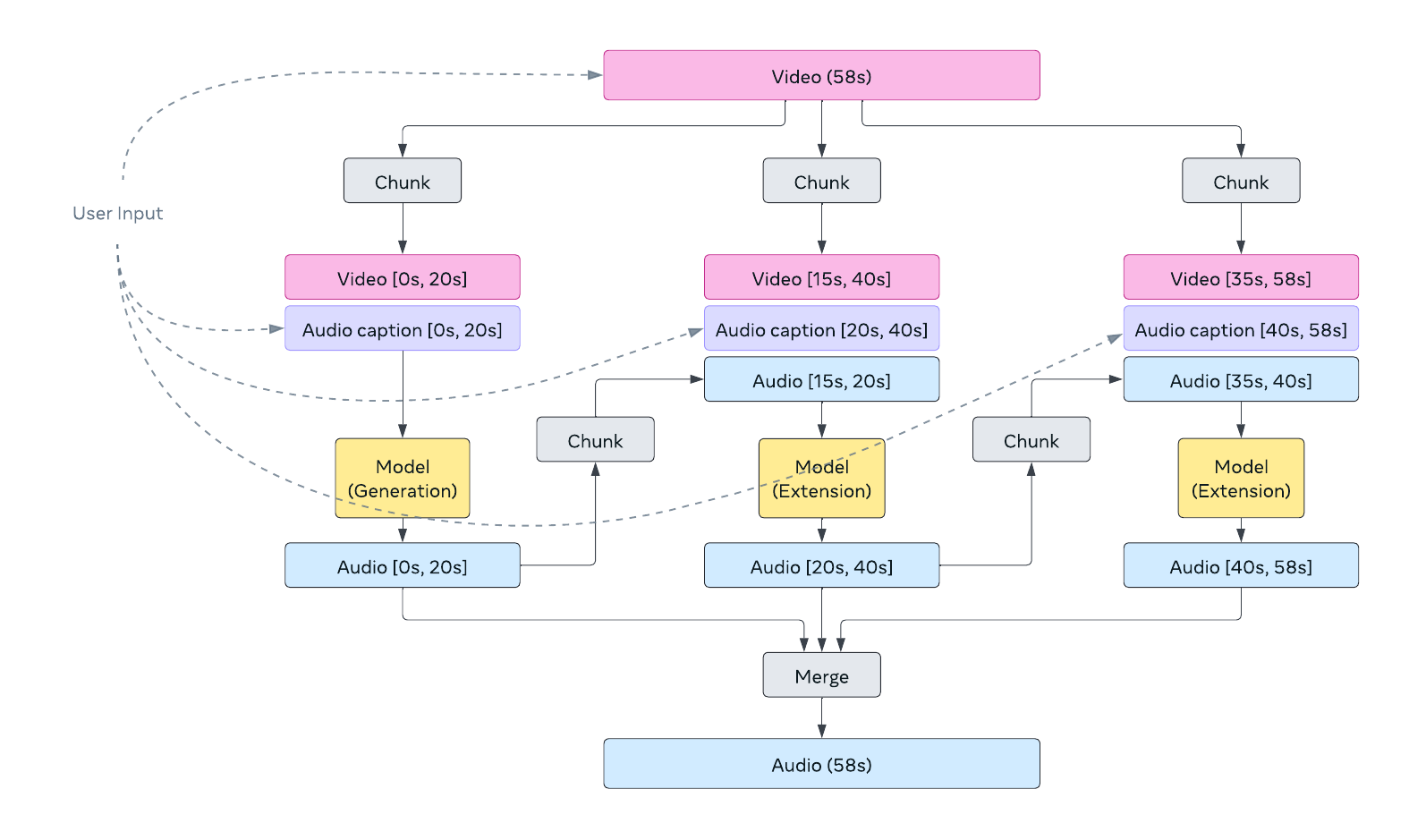}
    \caption{\textbf{\mvga{} extension diagram.} A user provides a video (\eg, 58s), and audio caption for each video chunk (\eg, 20s). Starting from the second chunk, the model takes not only the video chunk and the caption, but also a segment from the previously generated audio (\eg, the last 5s) in order to generate a new chunk that is coherent with the previous one.
    }
    \label{fig:audio_sys}
\end{figure}

In order to generate soundtracks for variable duration of videos, we build
a single model that can perform both \textit{audio generation} given a video,
and \textit{audio extension} given a video with partially generated audio.
We aim to generate up to 30 seconds of audio in a single shot,
and allow the model to utilize extension to generate audio of arbitrary lengths.
\cref{fig:audio_sys} illustrates the process for long-form video generation.

We enable audio extension by training the model to perform masked audio prediction,
where the model predicts the audio target given the whole video and its surrounding audio.
The surrounding audio can be empty (\ie, audio generation),
before or after the target audio (\ie, audio extension in either direction),
or around the target (\ie, audio infilling).
Audio infilling is useful for fixing small segments that contains artifacts or unwanted sound effects.

Lastly, for sound design purposes, users would often want to specify what and how acoustic events should be added to the video,
such as deciding what on-screen sounds to emphasize, what off-screen sounds to add, whether there is background music, and what style to generate for the music.
To provide users more control, we enable text prompting.

\subsection{Model}
\begin{figure}[h]
    \centering
    \includegraphics[width=\linewidth]{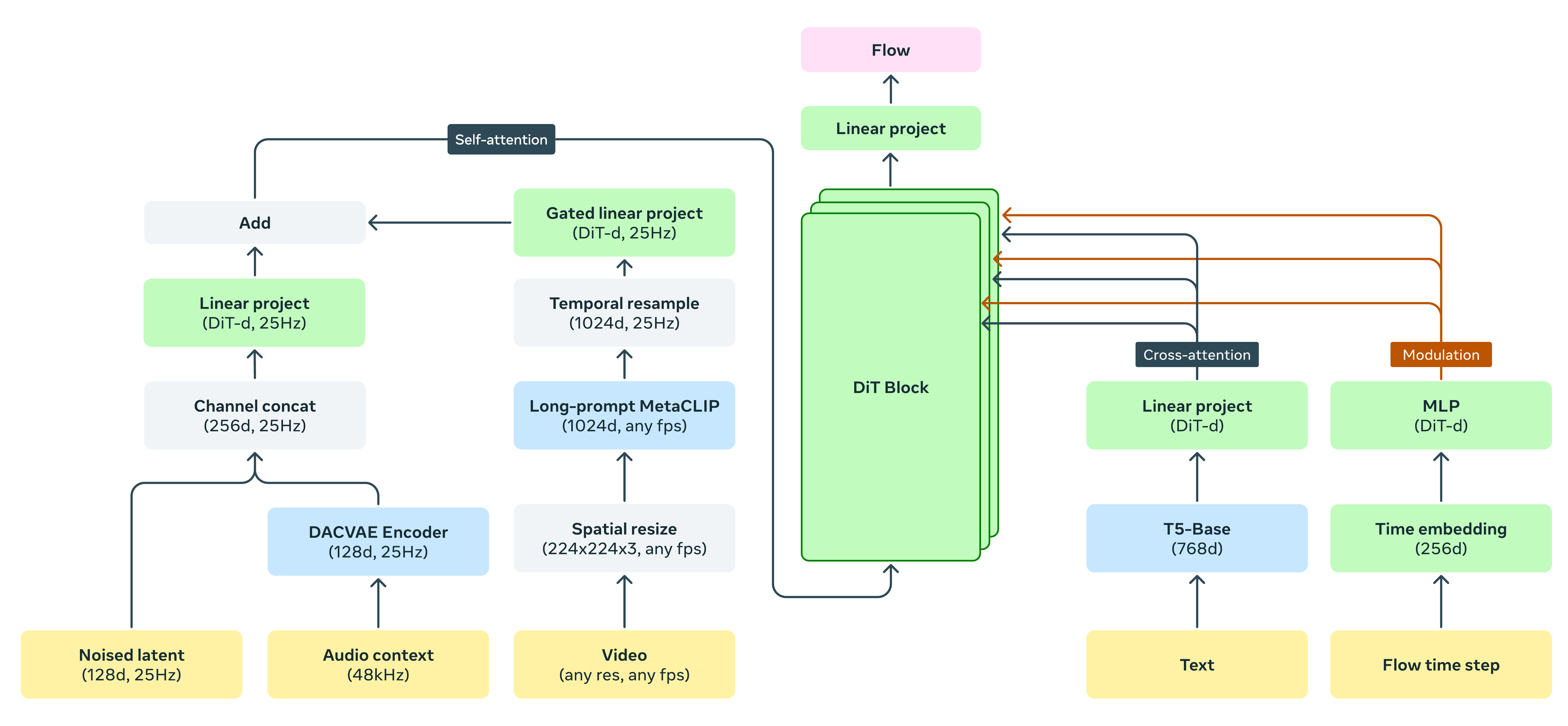}
    \caption{\textbf{\mvga{} model diagram.} Yellow blocks denote input, blue blocks denote pre-trained and frozen modules, gray blocks denote operations without learnable parameters, green blocks denote learnable modules, and the pink block shows the output velocity $u(\bx_t, \bc, t; \theta)$. Conditioning input $\bc$ includes masked audio context, video, and text. $\bx_t$ is a sample from $p_t$, and $t$ is the flow time step. For audio context, we replace the masked frames with zeros for DAC-VAE output.
    }
    \label{fig:audio_model}
\end{figure}

We adopt the flow-matching~\citep{flow-matching} based generative models and the diffusion transformer (DiT)~\citep{dit} model architecture.
Additional conditioning modules are added to provide control.
\cref{fig:audio_model} illustrates the model architecture.

\subsubsection{Flow-matching}
We also use the Flow Matching~\citep{flow-matching} objective as described in~\cref{sec:t2v_training_objective} to train the \mvga{} model.
The same optimal transport path is used for constructing $\bx_{t}$ which is now an audio sample in the latent space that we will describe in \cref{sec:audio_repr_and_cond}, and the same logit-normal distribution is used for sampling flow-step $t$.
Instead of conditioning only on text prompt embedding $\bp$, \mvga{} is conditioned on multimodal prompt $\bc$ which will be described in \cref{sec:audio_repr_and_cond}.

We choose diffusion-style models~\citep{Ho2020DenoisingDP, song2020denoising} over discrete token-based language models~\citep{kreuk2022audiogen} because
(1) it shows strong empirical performance in sound, music, and speech generation~\citep{liu2023audioldm, ghosal2023text, majumder2024tango, shen2023naturalspeech, huang2023noise2music},
(2) its non-autoregressive nature permits flexible generation direction, and can be used for both infilling or out-filling in both directions,
(3) modeling audio in continuous space enables applications of techniques such as SDEdit~\citep{meng2021sdedit} for editing and multi-diffusion~\citep{bar2023multidiffusion} for infinite-length audio generation,
(4) it enables users to flexibly trade-off quality for runtime through configuring ODE parameters,
and enjoys recent advancement in distillation or consistency training techniques that boost quality significantly at a much lower runtime.
On the other hand, we choose flow-matching over diffusion because we found it achieves better training efficiency, inference efficiency, and performance compared to diffusion models as shown in recent works~\citep{lan2024high, le2023voicebox, vyas2023audiobox, prajwalmusicflow, mehta2024matcha, sd3}.

\subsubsection{Diffusion Transformer}
\mvga{} adopts the diffusion transformer (DiT) architecture~\citep{dit},
which modulates the  outputs of normalization layers with scale and bias,
and outputs of self-attention and feed-forward layers with scale in each transformer block~\citep{vaswani2017attention}.
A multi-layer perceptron (MLP) takes the flow time embedding as input and predicts the six modulation parameters (four scales and two biases).
The MLP is shared across all layers, different from the original DiT, and only layer-dependent biases are added to the MLP outputs.
This saves parameters without sacrificing performance. The next section describes how other inputs are conditioned.

\subsubsection{Audio Representation and Conditioning Modules}\label{sec:audio_repr_and_cond}

\textbf{Audio.}
The latent diffusion framework~\citep{rombach2021highresolution} is adopted,
where data (48kHz) is represented as compact 1D latent features of shape $T \times C$ at a much lower frame rate (25Hz) and $C=128$ extracted from a separately trained DAC-VAE (Descript Audio Codec~\citep{kumar2024high} with variational autoencoder~\citep{kingma2013auto} formulation) model.
Compared to the commonly used Encodec~\citep{Defossez2022HighFN} features (75Hz, 128-d) for 24kHz audio in audio diffusion models~\citep{shen2023naturalspeech, vyas2023audiobox},
our DAC-VAE offers a lower frame rate (75Hz$\rightarrow$25Hz), a higher audio sampling rate (24kHz$\rightarrow$48kHz), and much higher audio reconstruction quality.
Specifically, to outperform Encodec under a similar bitrate, DAC adopts the multi-scale STFT discriminators to reduce the periodicity artifacts and adds the Snake~\citep{ziyin2020neural} activation function to introduce periodic inductive biases inspired by the BigVGAN~\citep{lee2022bigvgan} architecture.
Although the code factorization technique of DAC also greatly reduces the quantization errors for much better reconstruction, discrete tokens are not necessary for diffusion-style models.
Therefore, we remove the residual vector quantizer (RVQ)~\citep{van2017neural, gray1984vector} from the DAC and trained with the variational autoencoder (VAE)~\citep{kingma2013auto} objective (which adds a KL-regularization to encourage latents to be normally distributed).
This significantly boosts the reconstruction performance especially at more compressed frame rates (25Hz).

\textbf{Video.}
Long-prompt MetaCLIP fine-tuned from MetaCLIP~\citep{xu2023demystifying} is used to extract a 1024-dimension embedding for each frame in a video.
Since the frame rate of the video might not match that of the audio, we take the nearest visual frame for each audio frame.
The resampled sequence is then projected to the DiT model dimension with a gated linear projection layer and added to the audio features frame by frame.
Adding visual and audio features frame by frame improves video-audio alignment compared to concatenating features along the time dimension,
because the former provides direct supervision of video-audio frame alignment.
We have also explored reconstruction-based features extracted from a video-autoencoder for conditioning,
which is expected to preserve more video details compared to contrastive features.
However, the results were significantly worse and also slows down training due to the large feature dimension per frame.
We concluded that the Long-prompt MetaCLIP features trained with a contrastive objective~\citep{oord2018representation, radford2021learning} encodes higher level semantic information that eases learning while keeping sufficient low-level details to capture the timing of each motion for the model to produce motion-aligned sound effects.

\textbf{Audio context.}
We follow the Voicebox~\citep{le2023voicebox} and Audiobox~\citep{vyas2023audiobox} frameworks and condition on partially masked target audio, which we coined \textit{audio context}.
This enables the model to infill (or out-fill, depending on where the mask is) audio that is coherent with the context.
Without conditioning on context, the audio would sound incoherent and change abruptly when stitching together audio segments generated independently given only the video, especially when audio contains heavy ambient sound or music.
Audio context is also represented as a DAC-VAE feature sequence, and is concatenated with the noised audio latent along the channel dimension frame by frame.
For masked frames, we replace it with zero-vectors.
To perform audio generation without any audio context, we simply input a zero-vector sequence for audio context.

\textbf{Text.}
We use text to provide additional guidance on target audio quality, sound events, and music style if music is present, which we collectively refer to as the \textit{audio caption}, details of which is described in \cref{sec:audio_cap}.
T5-base~\citep{raffel2020exploring} is used to encode an audio caption into a sequence of 768-dimensional features, where sequence length is capped at 512 tokens.
We insert a cross-attention layer right after the self-attention layer and before the feed-forward layer in each DiT transformer block for conditioning.\\

Putting it together, we denote an $N_{aud}$ frame-long sample at flow-step $t$ as $\bx_t \in \mathbb{R}^{N_{aud} \times 128}$,
and conditioning input as $\bc = \{\bc_{vid}, \bc_{ctx}, \bc_{txt}\}$,
where $\bc_{vid} \in \mathbb{R}^{N_{aud} \times 1024}$ is the resampled Long-prompt MetaCLIP features,
$\bc_{ctx} \in \mathbb{R}^{N_{aud} \times 128}$ is the masked DAC-VAE features that serves as audio context,
and $\bc_{txt} \in \mathbb{R}^{N_{txt} \times 768}$ is the T5 text feature sequence of $N_{txt}$ tokens long.
At each step, the model predicts a velocity $u(\bx_t, \bc, t; \theta) \in \mathbb{R}^{N_{aud} \times 128}$ that can be used to estimate $x_{t+\Delta t}$ during inference.
We use an additional subscript to index a subsequence when necessary. For example, $\bc_{vid,15:25}$ denotes a 10-frame segment starting on the 15th frame (0-based).

\subsubsection{Inference: One-shot Generation}
During training, each conditioning input (video, audio context, text) is dropped-out independently with some probabilities.
This enables the model to perform
\begin{enumerate*}[label=(\arabic*)]
    \item video-to-audio (V2A) generation (dropping out text and audio context),
    \item text-instructed video-to-audio (TV2A) generation (dropping out audio context),
    \item video-to-audio infilling or extension (dropping out text), and
    \item text-instructed video-to-audio infilling or extension,
\end{enumerate*}
with a single model by simply changing the conditioned inputs.

\subsubsection{Inference: Audio Extension}\label{sec:audio_ext}
Due to memory constraint and training efficiency considerations, training data is capped at a predetermined length.
To generate high quality and coherent long-form audio for videos whose lengths are beyond the cap,
we consider two algorithms: \textit{segment-level autoregressive generation} and \textit{multi-diffusion}~\citep{bar2023multidiffusion}.

Given a long video, we first split the video into \textit{overlapping} segments, and assume text captions are available for all segments.
At a high level, both algorithms run inference on individual segments first, and then consolidate the prediction.
Information can propagate across segments through overlapping frames.
Without loss of generality, we assume each segment is $n_{win}$ frames long,
and the end times of consecutive segments differ by $n_{hop}$ frames.
Two consecutive segments overlap by $n_{ctx} = n_{win} - n_{hop}$ frames.
For a video $\bc_{vid}$ of $N$ frames,
there will be $J = \lceil N / n_{hop} \rceil$ segments,
and the $j$-th segment spans $[n_{start}^{(j)}, n_{end}^{(j)}] = [max(0, (j-1)n_{hop} - n_{ctx}), min(N, jn_{hop})]$ where $j \in [J]$.
We denote the text caption for $j$-th segment as $\bc_{txt}^{(j)}$,
and the consolidated prediction at flow step $t_i$ as $\bx_{t_i}^{(j)}$,
assuming the same $T$-step flow time schedule $\{t_i\}_{i=1}^T$ are used for all segments ($t_1 = 0$, $t_T = 1$, and $t_i < t_{i+1}\;\forall i$).
Note that $\bx_0^{(j)}$ is the noise drawn from the prior $p_0$ and $\bx_1^{(j)}$ is the predicted audio for the $j$-th segment.

We introduce two extension methods below: segment-level autoregressive generation and multi-diffusion. Multi-diffusion achieves slightly better results empirically, which we use as the default method.

\textbf{Segment-level autoregressive generation.}
This algorithm emulates autoregressive generation of language models but at the segment level.
It generates one segment at a time conditioned on the information from the last $n_{ctx}$ frames of the previous segment.
Given the trajectory $\bx_{t_i}^{(j)}$ from the $j$-th segment,
we consider passing information through two routes when generating $\bx_{t_i}^{(j+1)}$: \textit{context conditioning} and \textit{trajectory regularization}.

The first route, context conditioning, is to simply update the audio context $\bc_{ctx}^{(j+1)}$ using the prediction from the previous segment.
Specifically, we set $\bc_{ctx}^{(j+1)} = [\bx_{1,-n_{ctx}:}^{(j)}; \mathbf{0}]$,
where $\bx_{1,-n_{ctx}:}^{(j)}$ denotes the last $n_{ctx}$ frames of $\bx_1^{(j)}$, and $\mathbf{0}$ is a zero matrix of shape $n_{hop} \times C$.

The second route, trajectory regularization, is to pass information through the flow trajectory $\bx_{t_i}^{(j+1)}$.
We \textit{blend} the trajectory from the previous segment with the current one at each flow step $t_i$ with a weighting function $w \in \mathbb{R}_{\ge 0}^{n_{ctx}}$,
such that the trajectory of the current segment does not diverge too much from the previous one for the overlapping segment.
Algorithmically, let $\hat{\bx}_{t_{i+1}}^{(j+1)} = \text{ODE-Solve}(u, \bx_{t_i}^{(j+1)}, \bc^{(j+1)}, t_i, t_{i+1})$ be the solution from the ODE solver at $t_{i+1}$ taking the initial condition $\bx_{t_i}^{(j+1)}$ at $t_i$ with conditioning input $\bc^{(j+1)} = \{\bc_{vid}^{(j+1)}, \bc_{ctx}^{(j+1)}, \bc_{txt}^{(j+1)}\}$ and flow model $u$.
We update the state to $\bx_{t_{i+1},0:n_{ctx}}^{(j+1)} = w \odot \hat{\bx}_{t_{i+1},0:n_{ctx}}^{(j+1)} + (1-w) \odot \bx_{t_{i+1},-n_{ctx}:}^{(j)}$ for the overlapping frames and keep other frames intact $\bx_{t_{i+1},n_{ctx}:n_{win}}^{(j+1)} = \hat{\bx}_{t_{i+1},n_{ctx}:n_{win}}^{(j+1)}$,
where $\odot$ denotes point-wise product.
In the end, we also update the audio of the overlapping frames with $\bx_{1,0:n_{ctx}}^{(j+1)}$, the consolidated prediction from the later segment.
The blending function $w$ dictates how trajectories are mixed for each frame.
To encourage a smooth transition,
we consider a linear ramping function $w_n = n / n_{ctx}$ for $n \in [n_{ctx}]$,
such that weights are higher on the previous segment for frames closer to it and vice versa.

To further boost the performance of segment-level autoregressive generation, we also explore \textit{segment-level beam search}.
At each segment, multiple candidates are generated and the resulting partial generations are ranked and pruned by a scoring model.
Those top candidates are used as prefixes to generate multiple candidates for the next segment.

\textbf{Multi-diffusion.}
Inspired by its success on leveraging a diffusion model trained on $512 \times 512$ images to generate panorama images that are 9 times wider ($512 \times 4608$) and application to video upsampling described in~\cref{sec:spatial_upsampler},
we explore multi-diffusion for audio extension, which is virtually the audio version of the panorama generation problem.
At a high level, multi-diffusion solves the ODE for each step (\eg, from $t_i$ to $t_{i+1}$) in parallel for all segments,
consolidate the prediction, and then continues with the next ODE step.
This is in contrast to segment-level autoregressive generation,
which solves the ODE for one segment completely (\ie, from $t=0$ to $t=1$, potentially with multiple steps) before solving it for the next segment.

We next describe the algorithm in detail.
To initialize, $\bx_0$ is first drawn from a prior distribution $p_0$.
At each step $t_i$, $\bx_{t_i}$ is first split into chunks $\{\bx_{t_i}^{(j)}\}_j$ based on the predefined segmentation.
$\hat{\bx}_{t_{i+1}}^{(j)}$ is computed as $\text{ODE-Solve}(u, \bx_{t_i}^{(j)}, \bc^{(j)}, t_i, t_{i+1})$ for all $j$.
Then we merge the prediction as $\bx_{t_{i+1}} = \sum_j \text{zero-pad}( m^{(j)} \odot \hat{\bx}_{t_{i+1}}^{(j)}, j)$.
The function $\text{zero-pad}(\bx^{(j)}, j)$ maps a segment of shape $n_{win} \times C$ back to a sequence of shape $N \times C$ (shape of the $\bx$),
by padding zeros based on the segment's position (\ie, the $j$-th segment spans from $n_{start}^{(j)}$ to $n_{end}^{(j)}$;
hence $n_{start}^{(j)}$ zeros are padded at the beginning and $N - n_{end}^{(j)}$ zeros at the end). 
The soft-masking function $m^{(j)} \in \mathbb{R}_{\ge> 0}^{n_{win}}$ defines how much the $j$-th segment contributes to each frame it spans.
We note that $\sum_j \text{zero-pad}( w^{(j)}, j ) = 1$, where for each frame the contribution over all segments sums to 1.
This completes one step of the flow.
The same process repeats until we reach $t_T = 1$.
Note that the weights $w$ used in segment-level autoregressive generation is applied to the overlapping frames ($n_{ctx}$ frames), 
while the soft-masking functions $\{ m^{(j)} \}_j$ are defined for all the frames each segment spans.

\begin{figure}
  \centering
  \begin{tabular}{cc}
    \includegraphics[width=0.48\textwidth]{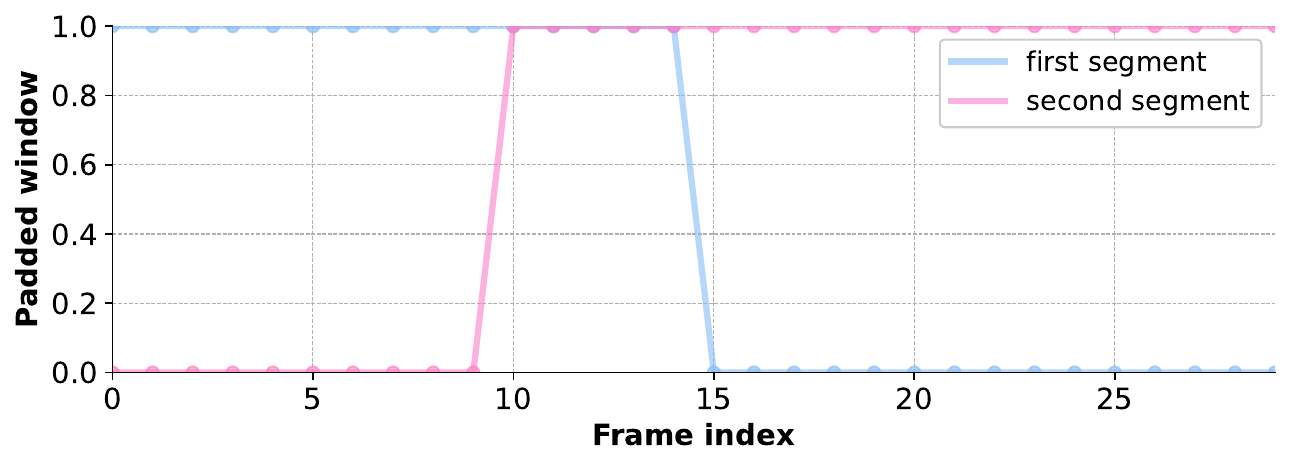} &
    \includegraphics[width=0.48\textwidth]{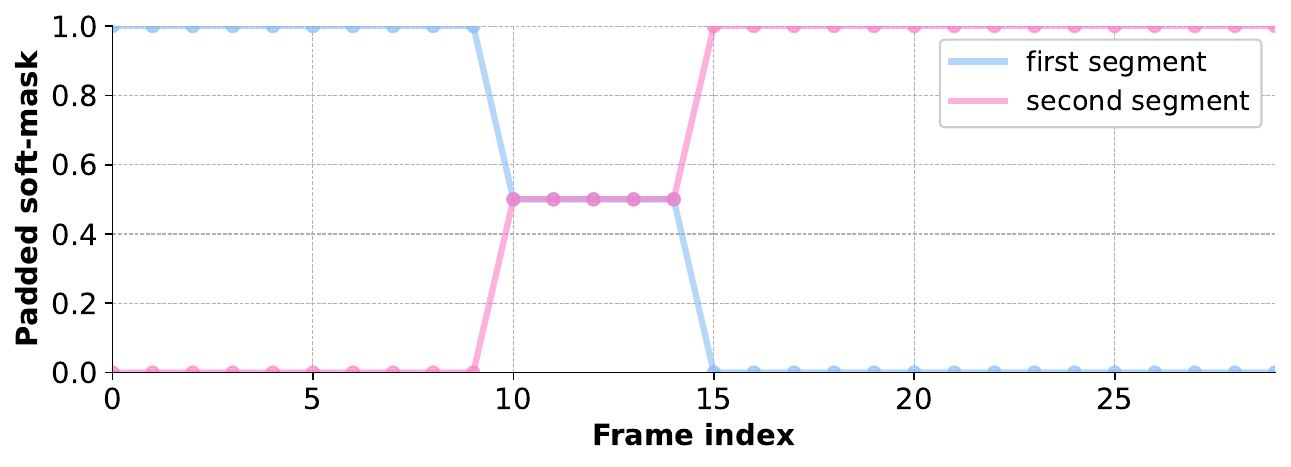} \\
    \includegraphics[width=0.48\textwidth]{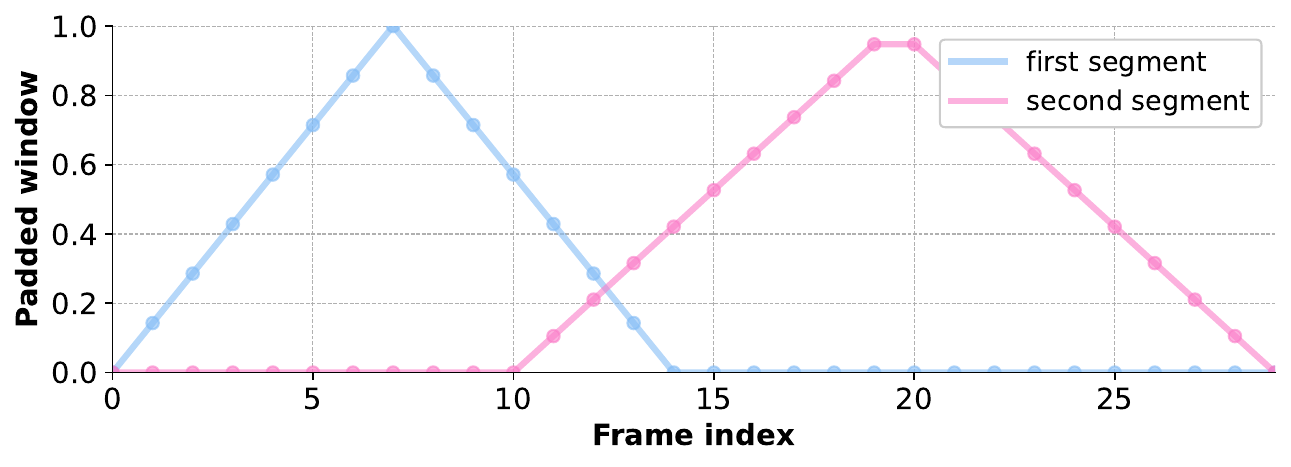} &
    \includegraphics[width=0.48\textwidth]{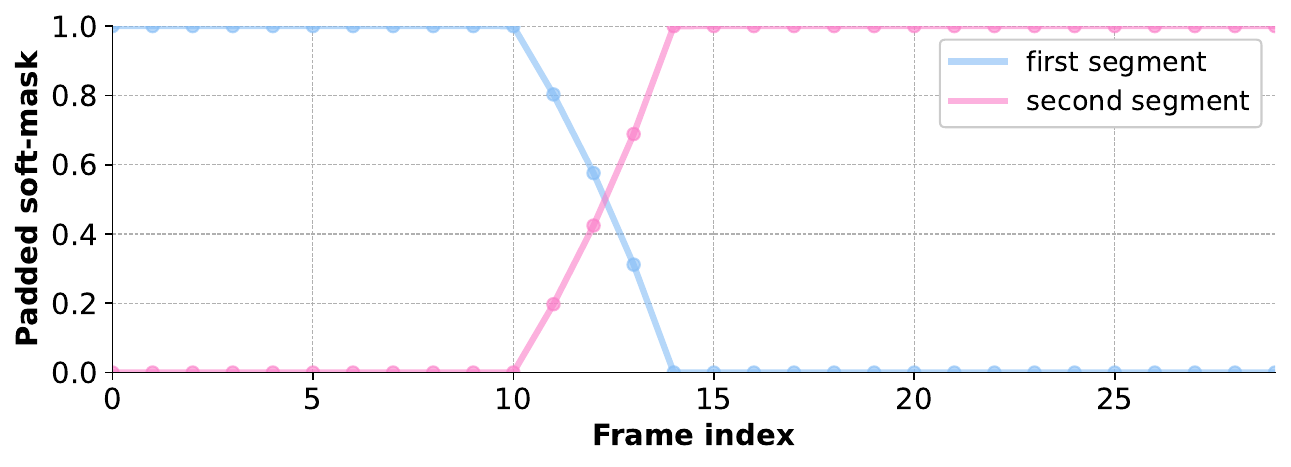}
  \end{tabular}
  \caption{\textbf{Comparing uniform weighting (top) and triangle window based weighting (bottom)} for multi-diffusion merging with an example of $l_{win}=20$ and $l_{ctx}=5$. The left column shows the unnormalized weights contributed from the first segment ([0, 15]) and the second ([10, 30]), and the right column shows the normalized weights. With triangle window, the transition in the overlapping region ([10, 15]) is much smoother.
  }
  \label{fig:audio_md_weights}
\end{figure}

\cref{fig:audio_md_weights} shows two variants of soft-masking functions (zero-padded) for two segments on the right column.
The soft-masking functions $m$ are derived from window functions $\hat{m} \in \mathbb{R}_{\ge 0}^{n_{win}}$
(the left column of the same figure, also zero-padded).
Specifically, we have $\text{zero-pad}( m^{(j)}, j ) = \left( \text{zero-pad}( \hat{m}^{(j)}, j ) / \sum_{j'} \text{zero-pad}( \hat{m}^{(j')}, j' ) \right)$ given window functions $\{ \hat{m}^{(j)} \}_j$.
The uniform window function ($\hat{m}^{(j)} = \mathbf{1}$) in the top row is equivalent to the method proposed in \citet{bar2023multidiffusion}.
However, we observe that this sometimes results in abrupt transition at the boundaries (especially for smaller models),
because the prediction from two neighboring segments might still be considerably different on overlapping frames at $t=1$.
Hence the transition from taking 100\% of the first segment to 50-50 from both segments leads to discontinuity,
as shown on~\cref{fig:audio_md_weights} top row.
Using the triangular window function (\ie, Barlette window, $\hat{m}_n^{(j)} = \dfrac{2}{n_{win}-1} \left( \dfrac{n_{win}-1}{2} - \left| n - \dfrac{n_{win}-1}{2} \right| \right)$) from the bottom row results in smoother transition,
similar to what we observed in chunk-level autoregressive generation with the linear ramping function.

\subsubsection{Model, Training, and Inference Configurations}
The DiT model has 36 layers and attention/feed-forward dimension of 4,608/18,432,
which has 13B parameters in total (excluding Long-prompt MetaCLIP, T5, and DAC-VAE).
In pre-training, videos are capped at 30 seconds (750 frames) and randomly chunked if lengths exceed.
In finetuning, we randomly sample 10s and 30s segments from the video.
Fully-sharded data parallelism is used to fit the model size.

The model is trained in two stages: \textit{pre-training} and \textit{fine-tuning}, using the same objective, but different data (described in \cref{sec:audio_pt_data} and \cref{sec:audio_ft_data}) and optimization configurations.
For pre-training, the effective batch size is 1,536 sequences, and each sequence is capped at 30 seconds (750 tokens).
The model is pre-trained for 500K updates, taking 14 days on 384 GPUs, using a constant learning rate of 1e-4 with a linear ramp-up for the first 5K steps.

For fine-tuning, the effective batch size is 256 sequences, also capped at 30 seconds.
The model fine-tunes the pre-trained checkpoint for 50K updates on 64 GPUs, which takes one day.
The learning rate linearly ramps up to 1e-4 for the first 5K steps, and then linearly decay to 1e-8 for the remaining steps.
An exponential moving average (EMA) checkpoint with a decay of 0.999 is accumulated during finetuning and used for inference.
AdamW optimizer with a weight decay of 0.1 and bf16 precision are used for both pre-training and fine-tuning.

To leverage classifier-free guidance (CFG) for inference, during training we drop conditioning inputs altogether (video, text, audio context) with a probability of 0.2.
To enable both audio generation and audio extension, the masked audio is dropped (\ie, completely masked) with a probability of 0.5, and otherwise masked between 75\% to 100\%.
To reduce the reliance on either modality, text and video input are dropped independently with a probability of 0.1 for each.

For inference, the midpoint solver with 64 steps is used.
We did not find using adaptive dopri5 solver or increasing the number of steps to boost the performance.
We use CFG with a weight of 3 on unconditional vector fields and further conduct reranking with 8 candidates per sample.
Quality score of 7.0 is used for sound effect generation, and 8.0 for joint sound effect and music generation.
For audio extension, we use dynamic guidance~\citep{wang2024analysisclassifierfreeguidanceweight} with a weight of 3 and multi-diffusion with a triangle window of $n_{win} = 40$, $n_{hop} = 30$ and $n_{ctx} = 10$ by default.

\subsection{Data}

\subsubsection{Audio-Visual Data Categorization}
The relationship between audio and visual components are complex.
For example, some sound effects like footsteps correlate with the low-level motion of objects in the scene,
while others like canned laughter in sitcoms correlates with high-level semantics (\eg, when something funny happens).
Similarly, music in the video can either be played by someone in the scene,
or be added during post production to enhance the storytelling power.

We classify audio by two axes and show an overview in~\cref{tab:audio_data_class}.
The first axis is audio type, which is divided into voice (speech and singing), non-vocal music, and general sound.
An audio event detection (AED) model~\citep{gemmeke2017audio} is used for automatic classification,
where each sample may contain multiple types of audio. This axis solely considers the audio.

The second axis is diegetic/non-diegetic~\citep{dykhoff2012non, stilwell200711}.
Diegetic audio components refer to those that \textit{can be heard at the scene} (crowd talking, newscasters speaking, music of a live band performing, birds chirping) and have a causal relationship with the video,
while non-diegetic, such as narrations in documentaries, background music in movies, or canned laughter in sitcoms, do not.
Note that diegetic sounds can be either on-screen or off-screen (birds chirping in the forest is diegetic even when birds are not seen), real or created in post-production (\ie, Foley sound).
A video can also contain both diegetic and non-diegetic sounds, which is especially common in professionally-produced videos like movies~\citep{tan2017effects}.
We leverage a contrastive audio-video-text pre-training model (\CAVTPShort) to determine how likely an audio sample is diegetic given the corresponding video.
Because this model is trained on data that contain mostly diegetic sounds,
the audio and the video embeddings are closer and have higher cosine similarity if the audio is diegetic and matches the video content.

\begin{table}[h]
    \centering
    \begin{tabular}{cp{100px}p{100px}p{100px}}
        \toprule
         & Voice & Non-vocal music & General sound \\
         \midrule
         \multirow{2}{*}{Diegetic} & on-screen or off-screen people speaking & on-screen or off-screen live music performance & on-screen wave splashing, off-screen barking \\
         \midrule
         Non-diegetic & narration & background music & canned laughter, riser \\
         \bottomrule
    \end{tabular}
    \caption{\textbf{Audio data classification and examples from each category.}
    We focus primarily on diegetic general sounds, non-diegetic sound effects, and instrumental music.
    }
    \label{tab:audio_data_class}
\end{table}

To generate each class of sounds correctly, a model would learn different levels of relationships between audio and conditioning input.
\begin{enumerate}
    \item Diegetic on-screen sounds have very strong correspondence between video and audio,
    where what sound should be heard when is deterministic.
    This demands stronger video understanding and dense action recognition capabilities from a model.
    The difficulty depends on how dense and structured the events are,
    where general sounds are overall easier than music and speech
    (\eg, generating golf club hitting the ball is easier than generating a person playing a guitar matching the chords one presses).
    \item Generating diegetic off-screen audio requires understanding what sounds may occur in what environments
    (\eg, birds chirping are possible in a forest scene) and logical orders between events (\eg, crowd cheering is likely to occur after, rather than before one performs a difficult trick).
    Hence, compared to on-screen sounds, it demands stronger reasoning capabilities.
    \item Non-diegetic audio is correlated with the video at the semantic level.
    For example, background music needs to match the mood,
    and risers are often used to create a sense of tension or anticipation.
    This demands the deepest level of understanding beyond understanding the physics of the world and requires reasoning and modeling human emotions.
\end{enumerate}

In this work, we focus primarily on \textit{diegetic general sounds, non-diegetic sound effects, and instrumental music}.
Generating diegetic speech is challenging when transcripts are not provided and when there are artifacts from the generated video.
We also omit non-diegetic speech as it can be created with text-to-speech synthesis systems if scripts are given.
As there are not only correlations between video and audio, but also between different classes of audio,
we choose to build a model that generates \textit{all classes of audio jointly},
instead of having separate models for diegetic/non-diegetic vocal/music/sound effects.

\subsubsection{Pre-training Data}\label{sec:audio_pt_data}
Pre-training aims to learn the structure of audio and alignment between audio and video/text from large quantities of data,
including both low quality and high quality audio samples. Below we describe filtering criteria based on AED tags and \CAVTPShort scores for pre-training data selection.

We start by sourcing data from a large volume and using the AED model to tag audio events for each sample based on the Audioset~\citep{gemmeke2017audio} ontology that has $527$ classes.
We then drop any videos where silence is the dominant class.

Next, we map the remaining events to one of three categories: speech, non-vocal music (music), or general sound (sound).
An event is mapped to ``voice'' if any of the ``speech'' and ``singing'' subclasses in the AudioSet ontology is tagged.
Similarly, an event is mapped to ``music'' if any of the music subclasses in the ontology is detected.
If any other subclass not mentioned above is deteced, the event is mapped to ``sound''.
This means that an utterance can contain any combination of these three classes.
After grouping samples by audio types, we use \CAVTPShort score, which is the cosine similarity between audio and video emeddings from \CAVTPShort, to categorize an utterance into one of the buckets as described in~\cref{tab:audio_ib_categories}.
The thresholds are determined based on manual inspection.

\begin{table}[h!]
    \centering
    \begin{tabular}{c@{\hspace{0.75cm}}@{\hspace{0.25cm}}c@{\hspace{0.75cm}}c}
    \toprule
    Category & AED detected & \CAVTPShort cosine similarity threshold \\
    \midrule
    \multirow{6}{*}{Diegetic} & sound & \\
     & voice & $> 0.2$ \\
     & voice + sound & \\
     \cmidrule(l){2-3}
     & music & \\
     & music + sound &  $> 0.3$ \\
     & voice + music + sound & \\
    \midrule
    \multirow{2}{*}{Non-diegetic} & music & $< 0.1$ \\
    \cmidrule(l){2-3}
     & sound + music & $> 0.1$ and $< 0.25$ \\
     \midrule
     Mixed & sound + voice + music & $> 0.1$ and $< 0.25$ \\
    \bottomrule
    \end{tabular}
    \caption{\textbf{\Pretraining data categorization.} We show the categories, AED tags, and \CAVTPShort thresholds used for grouping.}
    \label{tab:audio_ib_categories}
\end{table}

\cref{tab:audio_data_pt} shows the statistics for each category used for pre-training.
We consider the diegetic-or-mixed audio (filtered by \CAVTPShort score) along with a small proportion of non-diegetic background music. We prioritize general sound (filtered by AED tags),
as learning low-level physics is challenging and the errors of which are most noticeable.

To reduce noises from the visual modality,
we applied a series of quality filters to remove videos that contain text with OCR (Optical Character Recognition)~\citep{liao2020mask},
are static or are of low resolution($\mathit<480$ px).
The length of the videos have been constrained to be between 4s and 120s.
Additionally, we leveraged copy detection embeddings~\citep{pizzi2022self} for visual deduplication.

\begin{table}[h]
    \centering
    \begin{tabular}{crr}
        \toprule
         Type&  \#samples (M) & \#hours (K)\\
         \midrule
         Sound              &  \bigO(100)   & \bigO(1,000)   \\
         Music              &  \bigO(10)    & \bigO(100)     \\
         Sound+Music        &  \bigO(10)    & \bigO(100)     \\
         Sound+Voice        &  \bigO(10)    & \bigO(100)     \\
         Sound+Music+Voice  &  \bigO(10)    & \bigO(100)     \\
         \midrule
         Total              & \bigO(100)    & \bigO(1,000)   \\
         \bottomrule
    \end{tabular}
    \caption{\textbf{Pre-training data for \mvga{} by audio type.} Only diegetic and mixed diegetic/non-diegetic videos are used in pre-training.}
    \label{tab:audio_data_pt}
\end{table}

\subsubsection{Finetuning Data}\label{sec:audio_ft_data}
Once pre-training learns the foundational knowledge of audio structure and cross-modal alignment,
finetuning aims to align the model output with what we expect in \textit{cinematic soundtracks for videos},
which is very different from general recordings like those directly dumped from low-end devices (\eg, cellphones or security cameras).
Concretely, cinematic soundtracks are expected to be recorded with professional microphones and undergo post-production like mixing and mastering,
in order to reduce unwanted noises (\eg, pop noise, wind blowing to microphone)
and balance the presence of various audio events as well as the level of background music
(\eg, suppressing ambient noise and irrelevant off-screen sounds,
enhancing audio events like explosion or conversation relevant to storytelling,
and mixing ambient music with fade-in/fade-out).

Broadly speaking, cinematic soundtracks differ from low-quality recordings in two aspects:
audio quality (how it sounds) and sound design (what sounds to include).
To bridge the gap, we include two sources of finetuning data (summarized in~\cref{tab:audio_data_ft}).
First is the \textit{cinematic split}, which includes clips that are professionally produced that often contains both diegetic and non-diegetic sounds (ambient and theme music).
Clips with vocals are excluded.
An audio-visual cinematic classifier and an AED model are used for automatic data filtering, followed by human annotation for selection.
The second is the \textit{high quality audio split},
which includes high quality music (O(10)K hours) and sound effects (O(10)K hours) without videos.
Such data are available in larger quantities compared to the first split,
and can be used to boost the audio quality.
During fine-tuning, cinematic videos and high-quality audio are mixed with a 10 batches to 1 batch ratio.

\begin{table}[h]
    \centering
    \begin{tabular}{crr}
         \toprule
         Split&  \#samples (K)&  \#hours (K)\\
         \midrule
         Cinematic video (video+audio)      &  \bigO(100)   &  \bigO(1) \\
         High-quality audio (audio-only)    &  \bigO(1,000) &  \bigO(10)\\
         \midrule
         Total&  \bigO(1,000) &  \bigO(10) \\
         \bottomrule
    \end{tabular}
    \caption{\textbf{Finetuning data for \mvga{}.} We show the split and statistics of finetuning data used to make the audio generations closer to cinematic soundtracks.}
    \label{tab:audio_data_ft}
\end{table}

\subsubsection{Caption Structure and Synthetic Caption}\label{sec:audio_cap}
The caption is composed of four parts: audio quality, voice and music presence, sound caption~\citep{kim2019audiocaps}, and music style caption~\citep{manco2021muscaps}.
We use several models to create synthetic captions for all training data.~\cref{tab:audio_caption} shows two examples.

Given the scale, we leverage several models to build synthetic captions for all training samples.
Audio quality is a real-value number between 1 and 10 labeled by an audio quality prediction model
(annotations are collected in a similar way to LAION aesthetic~\citep{schuhmann2022laion}, where 10 means the highest quality and 1 means the lowest).
Voice and music presence are determined by the previously described AED model,
where the former takes the binary output using a predetermined posterior threshold,
and the latter is represented with AED posterior probability given the ambiguity
(certain cinematic sound effects like risers may also be considered music).
Sound caption is derived from a general audio caption model that provides free-form description about the sound.
To boost the controllability on music, we further deploy a music caption model to add more details such as mood and genre.
Note that music caption is appeneded regardless of whether the audio includes music or not.
We find using both music probability from AED and music caption from music caption model provide the best control, because music caption model is trained on mainly music samples and tend to hallucinate even when music is absent.

Each sample is split into both 10-second chunks and 30-second chunks, and then captioned.
Note that they are still segments of different lengths, which are from the last chunk of a sequence.
During training, the 10-second and 30-second chunks are sampled with a 5 batches to 1 batch ratio.

\begin{table}[H]
    \centering
    \begin{tabular}{p{15cm}}
    \toprule
    \textbf{Example captions for \mvga{}} \\
    \midrule
    \textcolor{capblue}{This audio has quality: 8.0}.
    \textcolor{capyellow}{This audio does not contain speech. This audio does not contain vocal singing.}
    \textcolor{capteal}{This audio has a description: "gentle waves lapping against the shore, and music plays in the background.".}
    \textcolor{capyellow}{This audio contains music with a 0.90 likelihood.}
    \textcolor{cappink}{This audio has a music description (if applicable): "A beautiful, romantic, and sentimental jazz piano solo.".} \\ \midrule
    \textcolor{capblue}{This audio has quality: 7.0}.
    \textcolor{capyellow}{This audio does not contain speech. This audio does not contain vocal singing.}
    \textcolor{capteal}{This audio has a description: "fireworks exploding with loud booms and crackles.".}
    \textcolor{capyellow}{This audio contains music with a 0.01 likelihood.}
    \textcolor{cappink}{This audio has a music description (if applicable): "A grand, majestic, and thrilling orchestral piece featuring a massive symphony orchestra with a soaring melody and pounding percussion, evoking a sense of awe and wonder.".}
    \\
    \bottomrule
    \end{tabular}
    \caption{\textbf{Two example captions for \mvga{}.} The blue part describes audio quality, the orange part controls presence of speech, vocal, and music, the teal part controls general sounds, and the pink part provides fine-grained control of music styles.}
    \label{tab:audio_caption}
\end{table}

\subsection{Evaluation}

We evaluate soundtrack generation mainly on audio quality, audio-video alignment, and audio-text alignment.
We prioritize alignment to video over alignment to text, because text is used as a supplement and may not capture all the details in the video.
Moreover, text input is not presented to the viewers in the final output.
\cref{tab:audio_metric} summarizes the metrics. Correlations between subjective and objective metrics are studied in~\Cref{sec:app_audio_metric_corr}.

\begin{table}[h]
    \centering
    \adjustbox{max width=\textwidth}{%
    \begin{tabular}{l ll}
        \toprule
        &  \textbf{Subjective} & \textbf{Objective}\\
        \midrule
        \multirow{3}{*}{Audio quality}
        & Overall quality (Ovr.) & \multirow{3}{*}{Audio quality score (AQual)} \\
        & Naturalness (Nat.) \\
        & Professionalness (Pro.) \\
        \midrule
        \multirow{4}{*}{Video alignment}
        & Diegetic sound correctness (Corr.)& \multirow{4}{*}{\IB score (\IBShort)} \\
        & Diegetic sound synchronization (Sync.)\\
        & Non-diegetic music mood alignment (Mood) \\
        & Non-diegetic music motion/scene alignment (Action) \\
        \midrule
        \multirow{2}{*}{Text alignment}
        & Precision & \multirow{2}{*}{CLAP score} \\
        & Recall \\
        \bottomrule
    \end{tabular}
    }
    \caption{\textbf{\OursAudio evaluation metrics}. Audio quality measures the standalone quality of the generated audio, while video and text alignment measure how well the generated audio aligns with the input video and text respectively.}
    \label{tab:audio_metric}
\end{table}

\subsubsection{Metrics\label{sec:audiobox_metrics}}
\noindent\textbf{Audio quality.}
We aim to evaluate how \textit{natural} (free of artifacts) and \textit{professional} (\eg, volume balance, crispness) an audio sample sounds.
For subjective tests, a pairwise protocol is adopted which asks raters to choose which audio has better overall quality and on those two axes,
where the pair of audio samples are generated conditioned on the same video and text prompts when applicable.
We report the Net Win Rate (NWT), defined as ``win\% - lose\%'' for pairwise comparisons.
NWT ranges from $-100\%$  to $100\%$. Details are described at the end of the section.

For objective metric, the audio quality score (AQual) predicted by the model described in~\cref{sec:audio_cap} is used as an automatic metric.
We note that the model tends to assign higher scores to samples with music;
hence the metric should not be used when comparing samples with music and those without music.

Note that we do not adopt Frech\'et audio distance (FAD)~\citep{kilgour2018fr} or KL-divergence (KLD) metrics that are often reported in text-to-audio generation~\citep{liu2023audioldm, vyas2023audiobox}
because these metrics are not applicable to generated videos that do not have corresponding audio, which we mainly evaluate on in this paper.

\noindent\textbf{Video alignment.}
This measures how well the audio is aligned with the video. %
For diegetic sound, we measure \textit{correctness} and \textit{synchronization}.
Correctness reflects whether the \textit{right type} of audio is generated with respect to the scene and the objects in the video (\eg, dog barking versus cat meowing).
Synchronicity on the other hand focuses on whether the audio is generated at the \textit{right time} matching the motions in the video.
For non-diegetic background music, we measure how well it supports the \textit{mood} of the scene and how well the \textit{score synchronizes} with the on-screen actions and scene changes (\ie, action scoring).
Similarly, a pairwise protocol is adopted for subjective tests and net win rate is reported.

For automatic metrics, the \IB (\IBShort) score is used for measuring the alignment between video and diegetic sounds, as used in \citet{mei2023foleygen}.
As mentioned earlier, when non-diegetic music is present in the video,
the score usually decreases regardless whether the mood and the score matches because the model is trained on mostly diegetic data without non-diegetic music.

\noindent\textbf{Text alignment.}
Finally, we measure \textit{precision} (percentage of generated audio events that are in the text caption) and \textit{recall} (percentage of audio events from the caption that are generated in the audio).
We note that we focus more on recall than on precision, as the caption might not include all the acoustic events that are supposed to be heard in a video.
In terms of the subjective tests, raters are asked to rate on a scale from 1 to 5 for precision and recall.
We adopt the standalone protocol instead of a pairwise one because text alignment is more objective and is easier to rate in absolute scale.
For objective test, we measure this with CLAP score~\citep{wu2023clap},
which is commonly used for text-to-audio generation and does not distinguish hallucination and missing errors.

\noindent\textbf{Computing net win rate}
The reported subjective preference metric is Net Win Rate.
For a given model pair, NWT is computed as follows: each item is evaluated by three raters.
For each item evaluated, we take the mean of the preference between model $A$ and model $B$ (+1 if the model $A$ is preferred, 0 if a tie and -1 if the model $B$ is preferred).
These are the \textit{consensus} scores for each item.
We then average these consensus scores across all items to obtain a net win rate of $A$ (this is the expected fraction of items where model $A$ is preferred minus the fraction of items where model $B$ is preferred).
To obtain 95\% confidence intervals around Net Win Rate, we bootstrap resample the item-level consensus scores 1,000 times,
compute Net Win Rate for each, and take difference between the 2.5\%-ile and 97.5\%-ile of the Net Win Rate as the 95\% confidence interval.
The net win rate of $A$ \vs $B$ ranges from -100\% to 100\%.

\subsubsection{Audio Generation Benchmarks}
\label{subsubsec:audio_gen_bench}
To thoroughly evaluate audio generation, we consider multiple existing video sources including both real and generated videos,
and propose to release a benchmark \OursAudioBench\footnote{\url{https://github.com/facebookresearch/MovieGenBench}} which contains high quality videos generated by \OursVideo that cover a wide spectrum of audio events, and human reviewed sound and music captions for those videos.
In order to enable fair comparison to \OursAudio by future work, we also release non cherry picked generated audio from \OursAudio on \OursAudioBench.

We group videos into two categories: \textit{single-shot} and \textit{multi-shot}.
Single-shot videos are available in larger quantities and cover a wider spectrum of sound effects,
which are suitable for testing robustness and generalization.
Multi-shot videos, extracted from short films, contain scene transitions and have stronger sentiment and deeper narratives than single-shot videos.
Hence, they are suitable for evaluating video-music alignment and sound design perspectives,
such as when music enters, how music evolves with the story and aligns with the cuts, and whether music and sound effects are mixed harmonically.
We describe the composition of single-shot and multi-shot benchmarks next and \cref{tab:audio_eval_set} provides a summary.

\begin{table}[h]
    \centering
    \adjustbox{max width=\textwidth}{%
    \begin{tabular}{cc cc cc}
    \toprule
    Benchmark & Video sources & Type & Shot & Num. of samples &  Max length \\
    \midrule
    \OursAudioBenchSReal  & VGGSound            & Real & Mostly single  & 51    & 10s \\
    \OursAudioBenchSGen   & \RunwayGen, \Sora   & Gen  & Single         & 151   & 10s \\
    \OursAudioBench       & \OursVideo          & Gen  & Single         & 527   & 10s \\
    \OursAudioBenchMGen   & \OursVideo, \Sora   & Gen  & Multi          & 107   & 15s \\
    \bottomrule
    \end{tabular}
    }
\caption{\textbf{\OursAudio evaluation datasets}. We use "SFX" or "SFX+music" suffix to indicate what version of text captions is used.}
\label{tab:audio_eval_set}
\end{table}

\noindent\textbf{Single-shot.}
This includes VGGSound~\citep{vggsound}, \Sora~\citep{sora}, \RunwayGen~\citep{gen3}, and our proposed \OursAudioBench.
\begin{itemize}
    \item VGGSound contains real videos and is widely used for training and evaluating video-to-audio generation models~\citep{mei2023foleygen, luo2024diff, xing2024seeing}.
    However, we discovered that there are many duplicates or near duplicates (\eg, with added static watermark or text) between the training and evaluation split,
    and some testing videos are static.
    We perform deduplication based on video embeddings, and manually review test sets to select 51 samples that are not static, do not contain diegetic speech, and mostly have motion synchronized sounds.
    \item \Sora has been used to demonstrate video-to-audio generation for generated videos~\citep{xing2024seeing,mei2023foleygen}, but the number of available videos is small and of limited domain.
    Hence, it is not suitable for being used alone as a benchmark.
    We review those used in text-to-video comparison (\Cref{appendix_prior_work_details}) and selected 43 samples for audio generation evaluation.
    \item \OursAudioBench is the new benchmark dataset we create using \OursVideo.
    It includes 527 videos and is designed to cover various ambient environments (\eg, indoor, urban, nature, transportation) and sound effects (\eg, human, animal, objects).
    This is the first large scale synthetic benchmark for evaluating video-to-audio generation.
    To create this benchmark, we first define an ontology with 36 audio categories and audio concepts for each category (\eg, expression $\rightarrow$ \{cry, laugh, yell\}), with a total of 434 concepts.
    Llama3 is next used to propose video prompts for each audio concept,
    and \OursVideo is used to generate videos given these prompts.
    We next review the generated videos to exclude those with artifacts that would severely impact the judgement of whether an audio fits the video, resulting in the final 527 videos.
    \item \RunwayGen contains 108 synthetic videos.
    It is created with a similar process as \OursAudioBench but with a subset of prompts.
    The goal is to include synthetic videos from different models that may contain different types of artifacts and artistic styles, in order to test the robustness of video-to-audio generation models.
\end{itemize}

We group them into three sets based on video properties:
``\OursAudioBenchSReal'' is the real single-shot videos which includes VGGSound;
``\OursAudioBenchSGen'' is the generated single-shot videos from prior video generation models which include \Sora and \RunwayGen;
``\OursAudioBench'' is the new generated single-shot benchmark we create, which we hope will facilitate future work for thorough text and video-to-audio generation benchmarking.

\noindent\textbf{Multi-shot.}
We source 26 short films generated by \Sora and by \OursVideo to create this set.
These videos are 30 second to 2 minute long, and are composed of multiple related shots.
To compare our model with baseline methods, many of which have length limitations and do not support audio extension, we chunk these videos into 15-second segments and discard last segments shorter than 10 seconds, which results in 107 segments in total.
The combined set is referred to as \OursAudioBenchMGen.

\noindent\textbf{Text prompt creation for SFX and SFX+music generation.}
For all the video samples, we use Llama3~\citep{llama3} to propose 5 sound and music captions for each video given its video caption,
and manually select the best sound and music caption for each video.
To generate non-diegetic music and sound effects jointly,
we set the prompt to ``This audio contains music with a 0.90 likelihood.'' for the music presence part of the text caption (see \cref{tab:audio_caption}).
In contrast, the likelihood is set to 0.01 if music is undesired.
For baseline models that take text prompt as input, we use sound caption as input when generating sound effects only,
and concatenate sound and music caption when generating sound effect and music jointly.

\subsection{Results}

\subsubsection{Comparisons to Prior Work}
\label{sec:main_res}

We present qualitative and quantitative results of sound effect generation, joint sound effect and music generation, and long-form generation with audio extension in this section.
More audio samples can be found in \cref{app:audio_samples}.

\begin{table}[h]
    \centering
    \captionsetup{type=figure}
    \setlength{\tabcolsep}{1pt}
    \adjustbox{max width=0.95\textwidth}{%
    \centering
    \begin{tabular}{cccccc}
        \rowcolor{blue300}
        \multicolumn{6}{c}{\textbf{Motion synchronized sound effects and ambient sounds}} \\
        \multicolumn{6}{c}{{(a) Boxing}}\\
        \includegraphics[width=0.17\linewidth]{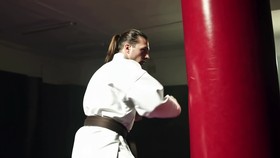}& 
        \includegraphics[width=0.17\linewidth]{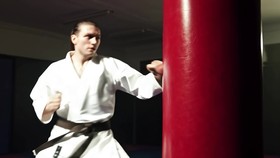}& 
        \includegraphics[width=0.17\linewidth]{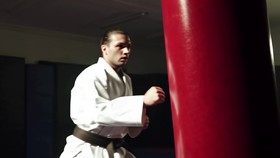}& 
        \includegraphics[width=0.17\linewidth]{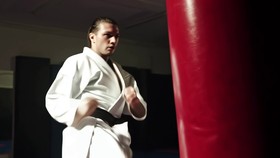}& 
        \includegraphics[width=0.17\linewidth]{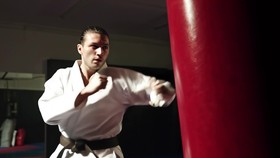}& 
        \includegraphics[width=0.17\linewidth]{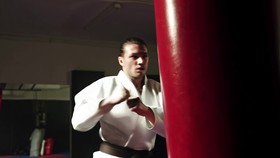}\\
        \multicolumn{6}{c}{
            \includegraphics[width=\linewidth, height=0.05\linewidth]
            {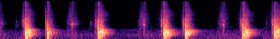}
        } \\
        \multicolumn{6}{c}{{(b) Golf swing}}\\
        \includegraphics[width=0.17\linewidth]{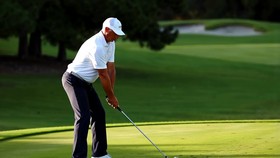}& 
        \includegraphics[width=0.17\linewidth]{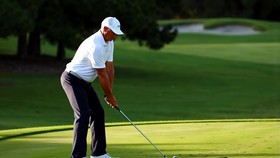}& 
        \includegraphics[width=0.17\linewidth]{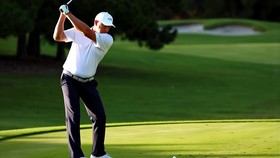}& 
        \includegraphics[width=0.17\linewidth]{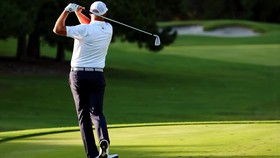}& 
        \includegraphics[width=0.17\linewidth]{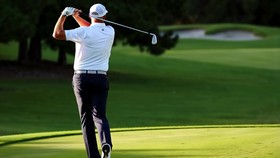}& 
        \includegraphics[width=0.17\linewidth]{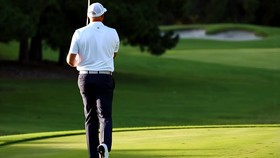}\\
        \multicolumn{6}{c}{
            \includegraphics[width=\linewidth, height=0.05\linewidth]
            {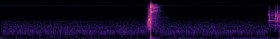}
        } \\
        \multicolumn{6}{c}{{(c) Geese honking, waddling and flapping}}\\
        \includegraphics[width=0.17\linewidth]{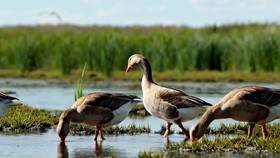}& 
        \includegraphics[width=0.17\linewidth]{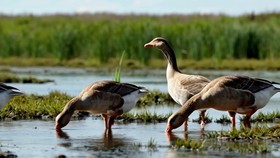}& 
        \includegraphics[width=0.17\linewidth]{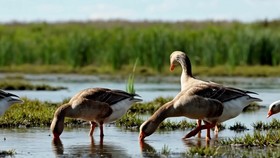}& 
        \includegraphics[width=0.17\linewidth]{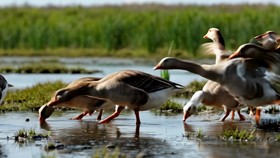}& 
        \includegraphics[width=0.17\linewidth]{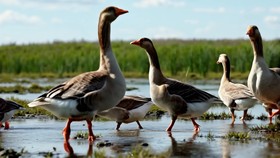}& 
        \includegraphics[width=0.17\linewidth]{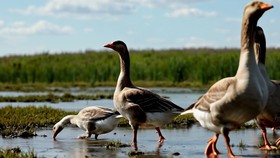}\\
        \multicolumn{6}{c}{
            \includegraphics[width=\linewidth, height=0.05\linewidth]
            {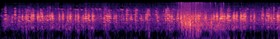}
        } \\
        \multicolumn{6}{c}{{(d) Waves crashing}}\\
        \includegraphics[width=0.17\linewidth]{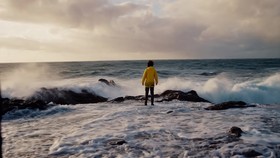}& 
        \includegraphics[width=0.17\linewidth]{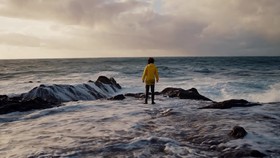}& 
        \includegraphics[width=0.17\linewidth]{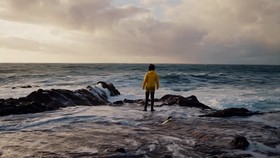}& 
        \includegraphics[width=0.17\linewidth]{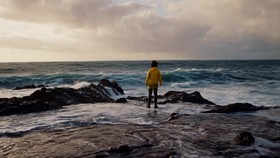}& 
        \includegraphics[width=0.17\linewidth]{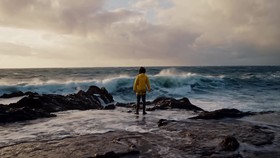}& 
        \includegraphics[width=0.17\linewidth]{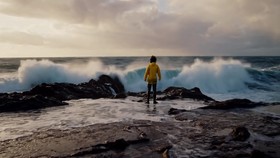}\\
        \multicolumn{6}{c}{
            \includegraphics[width=\linewidth, height=0.05\linewidth]
            {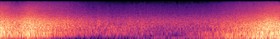}
        } \\
        \multicolumn{6}{c}{{(e) Explosion}}\\
        \includegraphics[width=0.17\linewidth]{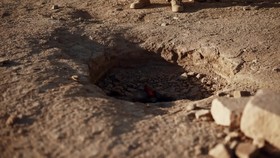}& 
        \includegraphics[width=0.17\linewidth]{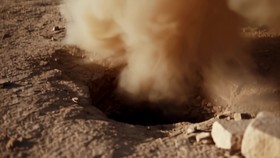}& 
        \includegraphics[width=0.17\linewidth]{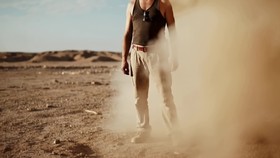}& 
        \includegraphics[width=0.17\linewidth]{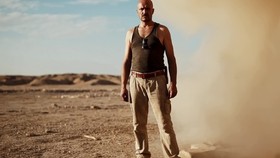}& 
        \includegraphics[width=0.17\linewidth]{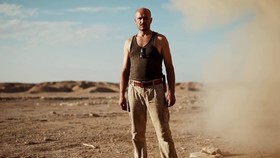}& 
        \includegraphics[width=0.17\linewidth]{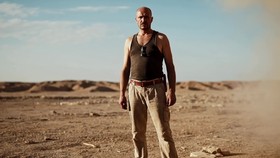}\\
        \multicolumn{6}{c}{
            \includegraphics[width=\linewidth, height=0.05\linewidth]
            {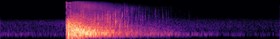}
        } \\
    \end{tabular}}
    \vspace{-3mm}
    \caption{\textbf{Video to SFX generation samples of \OursAudio.} It generates sounds that are tightly synchronized with the motions. Videos are from \OursAudioBench.
    Videos in this Figure found at \url{https://go.fb.me/MovieGen-Figure30}.
    }
    \label{fig:audio_main_sfx}
    \end{table}

\noindent\textbf{Sound effect generation.}
We first compare \OursAudio with 4 open-sourced models (Diff-Foley~\citep{luo2024diff}, FoleyCraft~\citep{zhang2024foleycrafter}, VTA-LDM~\citep{xu2024video}, Seeing\&Hearing~\citep{xing2024seeing}) and 2 blackbox commercial models (PikaLabs~\citep{pikalabs}, ElevenLabs~\citep{elevenlabs}) on sound effect generation for single shot videos.
Among these, Seeing\&Hearing and PikaLabs support video and optional text input (denoted as TV2A when text is used and V2A otherwise), ElevenLabs supports text input (T2A), and others support video input (V2A). More details about the baselines can be found in \cref{sec:related_audio}.

\begin{table}[h]
    \centering
    \adjustbox{max width=\textwidth}{%
    \begin{tabular}{lll ccc cc}
    \toprule
    \multirow{4}{*}{Dataset} & \multirow{4}{*}{Baseline} & \multirow{4}{*}{Type} & \multicolumn{5}{c}{\OursAudio net win rate \vs baseline}\\
    \cmidrule(lr){4-8}
    & & & \multicolumn{3}{c}{Quality} & \multicolumn{2}{c}{Video-SFX Alignment} \\
    \cmidrule(lr){4-6} \cmidrule(lr){7-8}
    & & & Ovr. & Nat. & Pro. & Corr. & Sync. \\
    \midrule
    \multirow{8}{*}{\shortstack[c]{\OursAudioBenchSReal SFX}}
    & Diff-Foley~\citep{luo2024diff} & V2A & \meanci{76.6}{12.6} & \meanci{48.1}{15.6} & \meanci{79.5}{11.1} & \meanci{61.6}{13.0} & \meanci{46.1}{14.3} \\
    & FoleyCraft~\citep{zhang2024foleycrafter} & V2A & \meanci{69.2}{14.1} & \meanci{57.2}{16.3} & \meanci{69.2}{14.1} & \meanci{50.4}{13.4} & \meanci{49.7}{17.0} \\
    & VTA-LDM~\citep{xu2024video} & V2A & \meanci{32.9}{18.5} & \meanci{31.5}{18.5} & \meanci{38.2}{18.9} & \meanci{47.4}{16.7} & \meanci{50.4}{16.3} \\
    & Seeing\&Hearing~\citep{xing2024seeing} & V2A & \meanci{85.8}{9.3} & \meanci{83.6}{11.1} & \meanci{85.8}{9.3} & \meanci{63.6}{14.8} & \meanci{63.7}{14.1} \\
    & Seeing\&Hearing~\citep{xing2024seeing} & TV2A & \meanci{76.8}{11.1} & \meanci{67.9}{15.2} & \meanci{76.8}{11.1} & \meanci{56.1}{17.4} & \meanci{51.3}{18.7} \\
    & PikaLabs~\citep{pikalabs} & V2A & \meanci{58.6}{15.2} & \meanci{49.7}{16.3} & \meanci{60.0}{14.1} & \meanci{56.9}{14.1} & \meanci{48.8}{18.1} \\
    & PikaLabs~\citep{pikalabs} & TV2A & \meanci{41.9}{20.4} & \meanci{31.9}{23.0} & \meanci{41.9}{20.4} & \meanci{35.8}{18.5} & \meanci{34.2}{18.4} \\
    & ElevenLabs~\citep{elevenlabs} & T2A & \meanci{13.2}{21.5} & \meanci{8.7}{21.5} & \meanci{13.2}{21.5} & \meanci{27.5}{18.9} & \meanci{35.0}{19.3} \\
    \midrule
    \multirow{8}{*}{\shortstack[c]{\OursAudioBenchSGen SFX}}
    & Diff-Foley~\citep{luo2024diff} & V2A & \meanci{78.7}{6.8} & \meanci{76.2}{6.6} & \meanci{78.5}{6.6} & \meanci{82.2}{5.4} & \meanci{70.4}{8.7}\\
    & FoleyCraft~\citep{zhang2024foleycrafter} & V2A & \meanci{65.0}{8.7} & \meanci{59.5}{8.5} & \meanci{65.0}{8.6} & \meanci{57.2}{7.7} & \meanci{49.6}{10.0} \\
    & VTA-LDM~\citep{xu2024video} & V2A & \meanci{77.7}{7.0} & \meanci{63.8}{7.7} & \meanci{76.8}{7.1} & \meanci{61.7}{8.2} & \meanci{58.2}{9.0} \\
    & Seeing\&Hearing~\citep{xing2024seeing} & V2A & \meanci{82.1}{7.4} & \meanci{76.9}{8.0} & \meanci{82.6}{7.3} & \meanci{63.6}{8.6} & \meanci{33.8}{10.1} \\
    & Seeing\&Hearing~\citep{xing2024seeing} & TV2A & \meanci{76.2}{7.1} & \meanci{75.4}{7.1} & \meanci{76.1}{7.3} & \meanci{64.1}{7.9} & \meanci{33.8}{10.1} \\
    & PikaLabs~\citep{pikalabs} & V2A & \meanci{61.2}{10.6} & \meanci{55.5}{10.7} & \meanci{62.6}{9.6} & \meanci{56.2}{12.5} & \meanci{52.1}{12.7} \\
    & PikaLabs~\citep{pikalabs} & TV2A & \meanci{53.6}{11.6} & \meanci{46.0}{11.6} & \meanci{54.5}{11.4} & \meanci{44.6}{12.9} & \meanci{39.4}{11.7} \\
    & ElevenLabs~\citep{elevenlabs} & T2A & \meanci{49.7}{9.8} & \meanci{45.3}{9.9} & \meanci{47.3}{9.8} & \meanci{31.8}{8.1} & \meanci{35.5}{9.5} \\
    \midrule
    \multirow{6}{*}{\shortstack[c]{\OursAudioBench SFX}}
    & Diff-Foley~\citep{luo2024diff}           & V2A & \meanci{91.0}{2.3} & \meanci{78.1}{3.0} & \meanci{90.7}{2.3} & \meanci{81.8}{3.0} & \meanci{70.9}{4.3}  \\
    & FoleyCraft~\citep{zhang2024foleycrafter} & V2A & \meanci{71.4}{4.0} & \meanci{60.7}{4.2} & \meanci{71.9}{4.0} & \meanci{57.4}{4.3} & \meanci{53.3}{5.1} \\
    & VTA-LDM~\citep{xu2024video}              & V2A & \meanci{71.7}{4.0} & \meanci{65.3}{4.2} & \meanci{72.0}{4.0} & \meanci{76.5}{3.6} & \meanci{72.8}{4.4}  \\
    & Seeing\&Hearing~\citep{xing2024seeing}   & V2A & \meanci{83.9}{3.0} & \meanci{72.3}{3.6} & \meanci{83.9}{3.0} & \meanci{66.6}{3.9} & \meanci{56.7}{4.9}  \\
    & Seeing\&Hearing~\citep{xing2024seeing}   & TV2A & \meanci{71.5}{4.0} & \meanci{70.0}{3.9} & \meanci{71.4}{3.9} & \meanci{59.4}{4.4} & \meanci{51.4}{5.3} \\
    & ElevenLabs~\citep{elevenlabs}            & T2A  & \meanci{31.3}{5.6} & \meanci{27.4}{5.4} & \meanci{31.1}{5.5} & \meanci{38.3}{5.1} & \meanci{36.0}{6.0} \\
    \bottomrule
    \end{tabular}}
    \caption{\textbf{Sound effect generation pairwise subject evaluation.}
    This table compares \OursAudio with prior work on audio quality and video alignment.
    We report net win rate, which has a range [-100\%, 100\%], and its 95\% confidence intervals.
    Positive values indicate \OursAudio outperforms the baseline on the metric.
    }
    \label{tab:eval_audio_main_ss_sfx}
\end{table}

\cref{fig:audio_main_sfx} shows 5 samples on \OursAudioBench. \cref{tab:eval_audio_main_ss_sfx} presents the pairwise subjective evaluation results on audio quality and video alignment. Additional metrics can be found in~\Cref{sec:app_audio_main_sfx}.
At a high level, \OursAudio outperforms all baselines on all metrics by a large margin: with 33.8\% to 72.8\% on synchronization, 27.5\% to 82.2\% on correctness for all videos, and 31.3\% to 91.0\% on overall quality for generated videos.
Compared to the commercial baselines, \OursAudio wins even more on generated videos, demonstrating its robustness.
In terms of audio quality, we highlight that \OursAudio wins more on \textit{professionalness} than on \textit{naturalness}, indicating that \OursAudio learns to not only generate realistic sounds but also professionally produced sounds, which leads to higher overall quality.

Among the baselines, commercial models generally outperforms open-sourced models in audio quality, while remaining similar in audio-video alignment.
Surprisingly text-based ElevenLabs model achieves similar performance to other baselines on video synchronization.
This is likely because text-based model can still perform well for videos that contain mostly ambient sounds, and for challenging videos with dense actions, none of the baseline can generate well-aligned audio.
Additionally, we observe that models using text prompts generally achieve better performance compared to their original forms.
This shows text captions provides complimentary information for guiding generation.

\noindent\textbf{Sound effect and music generation.}
We next showcase \OursAudio's ability to generate cinematic soundtracks for short films that also include non-diegetic music supporting the mood and synchronized with the visual actions. We evaluate \OursAudio on \OursAudioBenchMGen SFX+music, where music likelihood is set to 0.90 in text prompts.
As for the baselines, we need models that support text input to prompt them for joint SFX and music generation, since V2A models only produce diegetic SFX most of the time.
Seeing\&Hearing (S\&H)~\citep{xing2024seeing} and PikaLabs~\citep{pikalabs} are the only two options, while ElevenLabs is another option with only text input. From preliminary testing, we find neither Pika nor ElevenLabs can generate SFX and music jointly, so we do not consider them in this experiment.

\begin{table}[H]
    \centering
    \adjustbox{max width=\textwidth}{%
    \begin{tabular}{lll ccc cc cc}
    \toprule
    & & & \multicolumn{7}{c}{\OursAudio net win rate \vs baseline} \\
    Dataset & \multicolumn{2}{c}{Baseline method} & \multicolumn{3}{c}{Quality} & \multicolumn{2}{c}{Video-SFX Alignment} & \multicolumn{2}{c}{Video-Music Alignment} \\
    \cmidrule(lr){2-3} \cmidrule(lr){4-6} \cmidrule(lr){7-8} \cmidrule(lr){9-10} 
    & SFX Model & Music Model & Ovr. & Nat. & Pro. & Corr. & Sync. & Mood & Action \\
    \midrule
    \multicolumn{10}{c}{\textit{joint sound effect and music generation}} \\
    \multirow{8}{*}{\OursAudioBenchMGen SFX+music}
    & \multicolumn{2}{c}{S\&H TV2A} & \meanci{89.9}{5.0} & \meanci{82.4}{5.9} & \meanci{89.9}{5.0} & \meanci{76.1}{6.7} & \meanci{81.3}{6.5} & \meanci{67.5}{8.8} & \meanci{71.8}{8.3} \\
    \cmidrule(lr){2-10}
    \multicolumn{10}{c}{\textit{separate sound effect and music generation}} \\
    & S\&H TV2A & S\&H TV2A & \meanci{67.7}{8.6} & \meanci{69.3}{8.4} & \meanci{66.4}{8.7} & \meanci{48.6}{9.4} & \meanci{42.3}{7.6} & \meanci{59.3}{8.4} & \meanci{65.1}{9.6} \\
    & S\&H TV2A & \Suno T2A & \meanci{12.5}{11.8} & \meanci{11.3}{11.4} & \meanci{11.3}{11.9} & \meanci{55.2}{9.8} & \meanci{57.4}{9.0} & \meanci{32.5}{12.3} & \meanci{26.8}{9.3}  \\
    & Diff-Foley V2A & \Suno T2A & \meanci{27.4}{11.1} & \meanci{20.6}{11.1} & \meanci{28.0}{10.9} & \meanci{25.2}{11.1} & \meanci{22.7}{8.4} & \meanci{49.2}{9.3} & \meanci{42.2}{11.0} \\
    & FoleyCraft V2A & \Suno T2A & \meanci{38.9}{10.0} & \meanci{39.8}{10.3} & \meanci{32.6}{10.6} & \meanci{60.8}{8.7} & \meanci{60.9}{10.2} & \meanci{18.0}{11.2} & \meanci{23.9}{10.7} \\
    & VTA-LDM V2A & \Suno T2A & \meanci{38.7}{11.5} & \meanci{28.5}{12.8} & \meanci{35.3}{11.4} & \meanci{80.4}{6.1} & \meanci{77.0}{7.9} & \meanci{52.9}{9.2} & \meanci{45.2}{7.2}\\
    \bottomrule
    \end{tabular}}
    \caption{\textbf{Sound effect and music generation pairwise subject evaluation.}
    This table compares \OursAudio with prior work on audio quality and video alignment.
    We report net win rate, which has a range [-100\%, 100\%], and its 95\% confidence intervals.
    Positive values indicate \OursAudio outperforms the baseline on the metric.}
    \label{tab:eval_audio_main_ms_sfx_music}
\end{table}

\begin{table}[h]
\centering
\captionsetup{type=figure}
\setlength{\tabcolsep}{1pt}
\adjustbox{max width=0.95\textwidth}{%
\centering
\begin{tabular}{cccccc}
    \rowcolor{blue300}
    \multicolumn{6}{c}{\textbf{SFX + music generation for single-shot videos}} \\
    \multicolumn{6}{c}{{(a) ATV trick / high-energy, action-packed electronic rock track}} \\
    \includegraphics[width=0.17\linewidth]{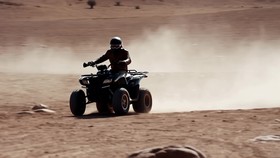}& 
    \includegraphics[width=0.17\linewidth]{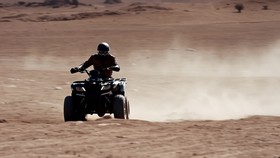}& 
    \includegraphics[width=0.17\linewidth]{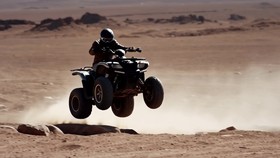}& 
    \includegraphics[width=0.17\linewidth]{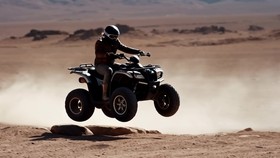}& 
    \includegraphics[width=0.17\linewidth]{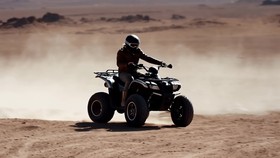}& 
    \includegraphics[width=0.17\linewidth]{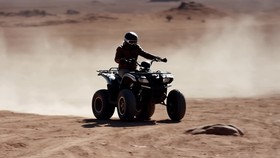}\\
    \multicolumn{6}{c}{
        \includegraphics[width=\linewidth, height=0.05\linewidth]
        {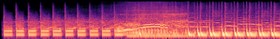}
    } \\
    \multicolumn{6}{c}{{(b) Waterfall / dramatic and intense orchestral piece}} \\
    \includegraphics[width=0.17\linewidth]{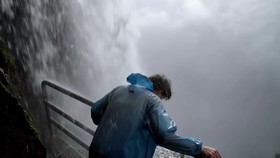}& 
    \includegraphics[width=0.17\linewidth]{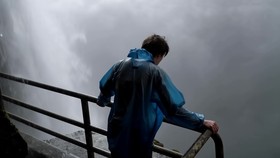}& 
    \includegraphics[width=0.17\linewidth]{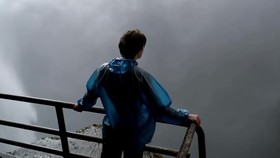}& 
    \includegraphics[width=0.17\linewidth]{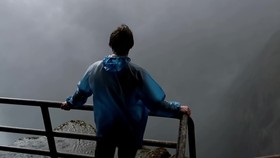}& 
    \includegraphics[width=0.17\linewidth]{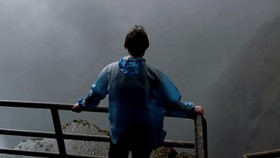}& 
    \includegraphics[width=0.17\linewidth]{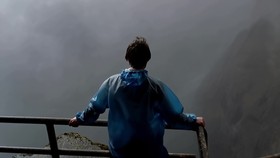}\\
    \multicolumn{6}{c}{
        \includegraphics[width=\linewidth, height=0.05\linewidth]
        {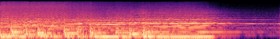}
    } \\
    \multicolumn{6}{c}{{(c) Penguin / A fun, upbeat, and quirky jazz piano track}} \\
    \includegraphics[width=0.17\linewidth]{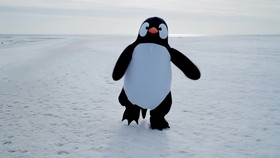}& 
    \includegraphics[width=0.17\linewidth]{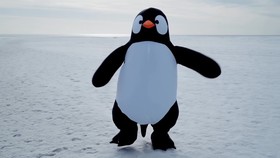}& 
    \includegraphics[width=0.17\linewidth]{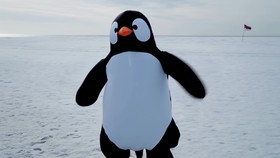}& 
    \includegraphics[width=0.17\linewidth]{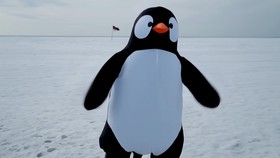}& 
    \includegraphics[width=0.17\linewidth]{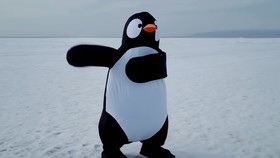}& 
    \includegraphics[width=0.17\linewidth]{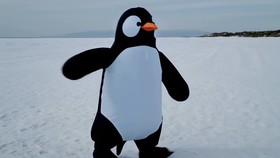}\\
    \multicolumn{6}{c}{
        \includegraphics[width=\linewidth, height=0.05\linewidth]
        {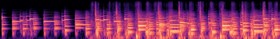}
    } \\
    \rowcolor{blue300}
    \multicolumn{6}{c}{\textbf{SFX + music generation for multi-shot videos}} \\
    \multicolumn{6}{c}{{(d) Astronaut (\Sora) / driving, energetic, and intense orchestral track}} \\
    \includegraphics[width=0.17\linewidth]{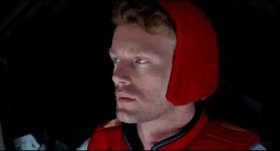}& 
    \includegraphics[width=0.17\linewidth]{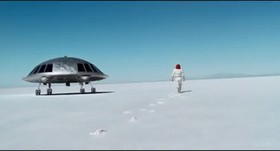}& 
    \includegraphics[width=0.17\linewidth]{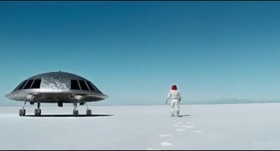}& 
    \includegraphics[width=0.17\linewidth]{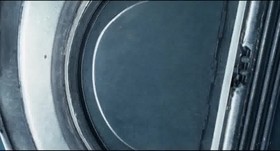}& 
    \includegraphics[width=0.17\linewidth]{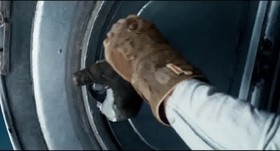}& 
    \includegraphics[width=0.17\linewidth]{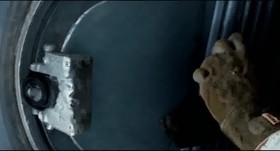}\\
    \multicolumn{6}{c}{
        \includegraphics[width=\linewidth, height=0.05\linewidth]
        {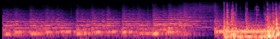}
    } \\
    \multicolumn{6}{c}{{(e) Movement (\Sora \& Paul Trillo) / fast-paced, energetic, and mesmerizing electronic track}} \\
    \includegraphics[width=0.17\linewidth]{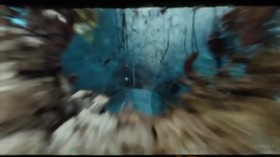}& 
    \includegraphics[width=0.17\linewidth]{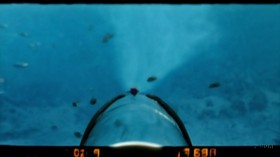}& 
    \includegraphics[width=0.17\linewidth]{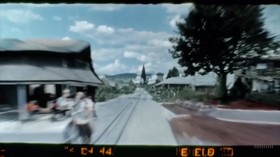}& 
    \includegraphics[width=0.17\linewidth]{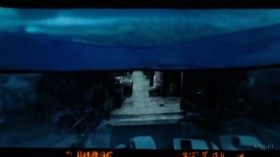}& 
    \includegraphics[width=0.17\linewidth]{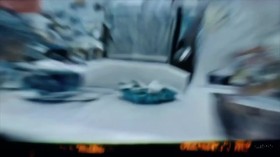}& 
    \includegraphics[width=0.17\linewidth]{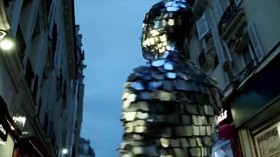}\\
    \multicolumn{6}{c}{
        \includegraphics[width=\linewidth, height=0.05\linewidth]
        {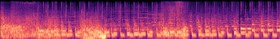}
    } \\
\end{tabular}}
\vspace{-3mm}
\caption{\textbf{Video to SFX and music generation samples of \OursAudio.}
\OursAudio can generate cinematic music that supports the mood, aligns with visual scene transitions, and blends harmonically with sound effects.
Videos in this Figure found at \url{https://go.fb.me/MovieGen-Figure31}.
}
\label{fig:audio_main_sfx_music}
\end{table}

In addition to joint generation, we include baselines that mix separately generated sound effects and music with an SNR sampled uniformly from [-5, 5]dB.
For sound effect generation, we include the open-sourced baselines used in the previous section.
For music generation, we consider the open-sourced S\&H using both video and text input, and an external text-to-music generation API that accepts only text input.
Sound captions (quoted text of the orange part in \cref{tab:audio_caption}) are used for sound effect generation if text prompt is supported, and music captions (quoted text of the pink part in \cref{tab:audio_caption}) are used for both music generation models.
We show qualitative samples on both single- and multi-shot videos in \cref{fig:audio_main_sfx_music}.

\cref{tab:eval_audio_main_ms_sfx_music} presents the pairwise subjective evaluation results on audio quality and video alignment for both sound effects and music.
Similar to the single-shot scenario, we outperform all baselines significantly across all aspects of alignment and quality.
Notably, the margin by which we surpass the joint generation baseline S\&H TV2A is even larger than in the sound effect-only case.
Separately generating sound effects and music with S\&H improves from joint generation, but it stills falls behind \OursAudio.
This highlights the limitations of existing public V2A models in cinematic content creation.

Although incorporating the high quality music generated by the external API greatly improves music quality,
this approach still falls short compared to our proposed model, especially on the alignment metrics.
There are two main reasons.
Since music and sound effects are generated separately, the correlation between them (\eg, music volume should be lowered when there are prominent sound effects) cannot be modeled.
Moreover, because the external API is a text-to-music model that is entirely unaware of video, it cannot generate music capturing the scene and mood changes in the video.

\noindent\textbf{Audio extension.}
Lastly, we evaluate \OursAudio's ability to generate long-form audio using the audio extension methods described in \cref{sec:audio_ext}.
\cref{fig:audio_main_long} shows three long videos with cinematic soundtracks generated by \OursAudio with audio extension.

\begin{table}[h]
    \centering
    \captionsetup{type=figure}
    \setlength{\tabcolsep}{1pt}
    \adjustbox{max width=0.95\textwidth}{%
    \centering
    \begin{tabular}{cccccc}
        \rowcolor{blue300}
        \multicolumn{6}{c}{\textbf{Long-form video-to-cinematic audio generation}} \\
        \multicolumn{6}{c}{{(a) Airhead (57s, \Sora \& shy kids)}} \\
        \includegraphics[width=0.17\linewidth]{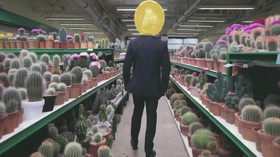}&
        \includegraphics[width=0.17\linewidth]{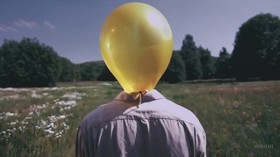}&
        \includegraphics[width=0.17\linewidth]{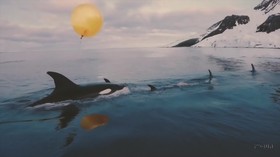}&
        \includegraphics[width=0.17\linewidth]{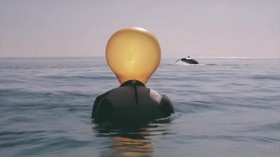}&
        \includegraphics[width=0.17\linewidth]{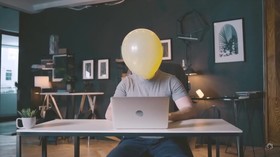}&
        \includegraphics[width=0.17\linewidth]{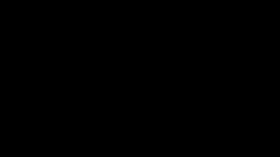}\\
        \multicolumn{6}{c}{
            \includegraphics[width=\linewidth, height=0.05\linewidth]
            {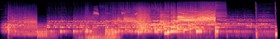}
        } \\
        \multicolumn{6}{c}{{(b) Ocean, cloud, galaxy (2m09s, \Sora \& Tammy Lovin)}} \\
        \includegraphics[width=0.17\linewidth]{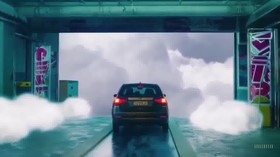}&
        \includegraphics[width=0.17\linewidth]{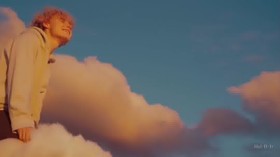}&
        \includegraphics[width=0.17\linewidth]{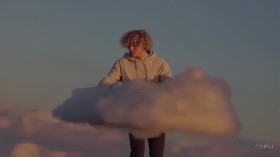}&
        \includegraphics[width=0.17\linewidth]{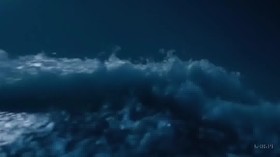}&
        \includegraphics[width=0.17\linewidth]{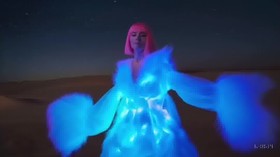}&
        \includegraphics[width=0.17\linewidth]{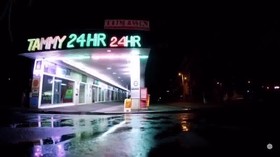}\\
        \multicolumn{6}{c}{
            \includegraphics[width=\linewidth, height=0.05\linewidth]
            {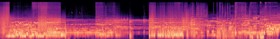}
        } \\
    \end{tabular}}
\vspace{-3mm}
\caption{\textbf{Examples of sound track generation for long videos with \OursAudio.}
\OursAudio can generate long-form coherent and cinematic soundtracks for videos up to a few minutes long.
It learns to compose music that progresses with the story.
For example, in Airhead, verse 1 starts at 0:10 that aligns with the protagonist walking,
verse 2 starts at 0:22 when the ballon starts floating around the world and the music intensity builds up,
and finally music calms down and fades out in the last scene at 0:42.
Videos in this Figure found at \url{https://go.fb.me/MovieGen-Figure32}.
}
\label{fig:audio_main_long}
\end{table}

For quantitative comparison, since there is no prior work on audio extension for video-to-audio generation,
we compare against a simple stitching method where audio is generated independently for each segment and then stitched together.
We use \OursAudio as the base model (denoted as ``\OursAudio stitch'').
Ideally, audio extension should be evaluated on long-form generated videos.
However, there are limited sources (26 full videos from \OursAudioBenchMGen).
We also found raters having trouble staying focused comparing long videos.
Both factors contribute to large variance on subjective tests.
As a workaround, we probe whether audio extension leads to \textit{smooth transitions} across segment boundaries,
using shorter videos and generating both sound effect and music  (\OursAudioBenchSGen SFX+music).
We set the segment size to 5.5 seconds, where each video from \OursAudioBenchSGen is split into two segments.
Single-shot generated video evaluation also enables us to use Seeing\&Hearing as
an additional baseline without stitching.
On this evaluation set, we set $n_{hop} = 5.5$ seconds and $n_{ctx} = 5.5$ seconds.
Comparison of different extension methods and configurations are presented in ablation studies in \cref{sec:audio_ablation}.

\cref{tab:eval_audio_main_ext} presents the pairwise subjective evaluation results on audio quality and video alignment. \OursAudio with extension outperforms both baselines as expected.
In particular, we note that \OursAudio-extension and \OursAudio-stitch use the same base model, and hence they should have similar quality and alignment \textit{within} each segment.
However, we can observe from the table that the audio quality and the video-music alignment metrics are significantly worse for \OursAudio-stitch, because stitching independently generated audio would lead to abrupt transition and incoherent music.

\begin{table}[h]
    \centering
    \adjustbox{max width=\textwidth}{%
    \begin{tabular}{ll ccc cc cc}
    \toprule
    & & \multicolumn{7}{c}{Subjective: \OursAudio extension net win rate \vs baseline} \\
    Dataset & Baseline method & \multicolumn{3}{c}{Quality} & \multicolumn{2}{c}{Video-SFX Alignment} & \multicolumn{2}{c}{Video-Music Alignment} \\
    \cmidrule(lr){3-5} \cmidrule(lr){6-7} \cmidrule(lr){8-9} 
    & & Ovr. & Nat. & Pro. & Corr. & Sync. & Mood & Action \\
    \midrule
    \multirow{2}{*}{\OursAudioBenchSGen SFX+music}
    & \OursAudio stitch & \meanci{34.5}{11.4} & \meanci{33.7}{11.1} & \meanci{34.5}{11.6} & \meanci{19.6}{10.0} & \meanci{5.6}{11.9} & \meanci{20.3}{10.8} & \meanci{26.5}{9.9} \\
    & Seeing\&Hearing & \meanci{85.1}{5.6} & \meanci{79.5}{6.5} & \meanci{85.1}{5.6} & \meanci{66.2}{7.9} & \meanci{61.0}{9.8} & \meanci{54.6}{10.5} & \meanci{26.2}{8.8} \\
    \bottomrule
    \end{tabular}}
    \caption{\textbf{Audio extension \vs simple stitching.}
    This table compares \OursAudio with extension to simple stitch-based baseline and Seeing\&Hearing on audio quality and video alignment.
    We report net win rate, which has a range [-100\%, 100\%], and its 95\% confidence intervals.
    Positive values indicate \OursAudio with extension outperforms the baseline on the metric.}
    \label{tab:eval_audio_main_ext}
\end{table}

\subsubsection{Ablations}\label{sec:audio_ablation}
We ablate critical design decisions for \OursAudio in this section, including text prompts, scaling, data, and extension methods. Unless otherwise described, we use the 13B parameter model and evaluate on \OursAudioBenchSGen SFX for sound effect generation.

\begin{figure}[h]
    \begin{subfigure}[b]{.48\linewidth}
        \centering
        \includegraphics[width=.8\linewidth]{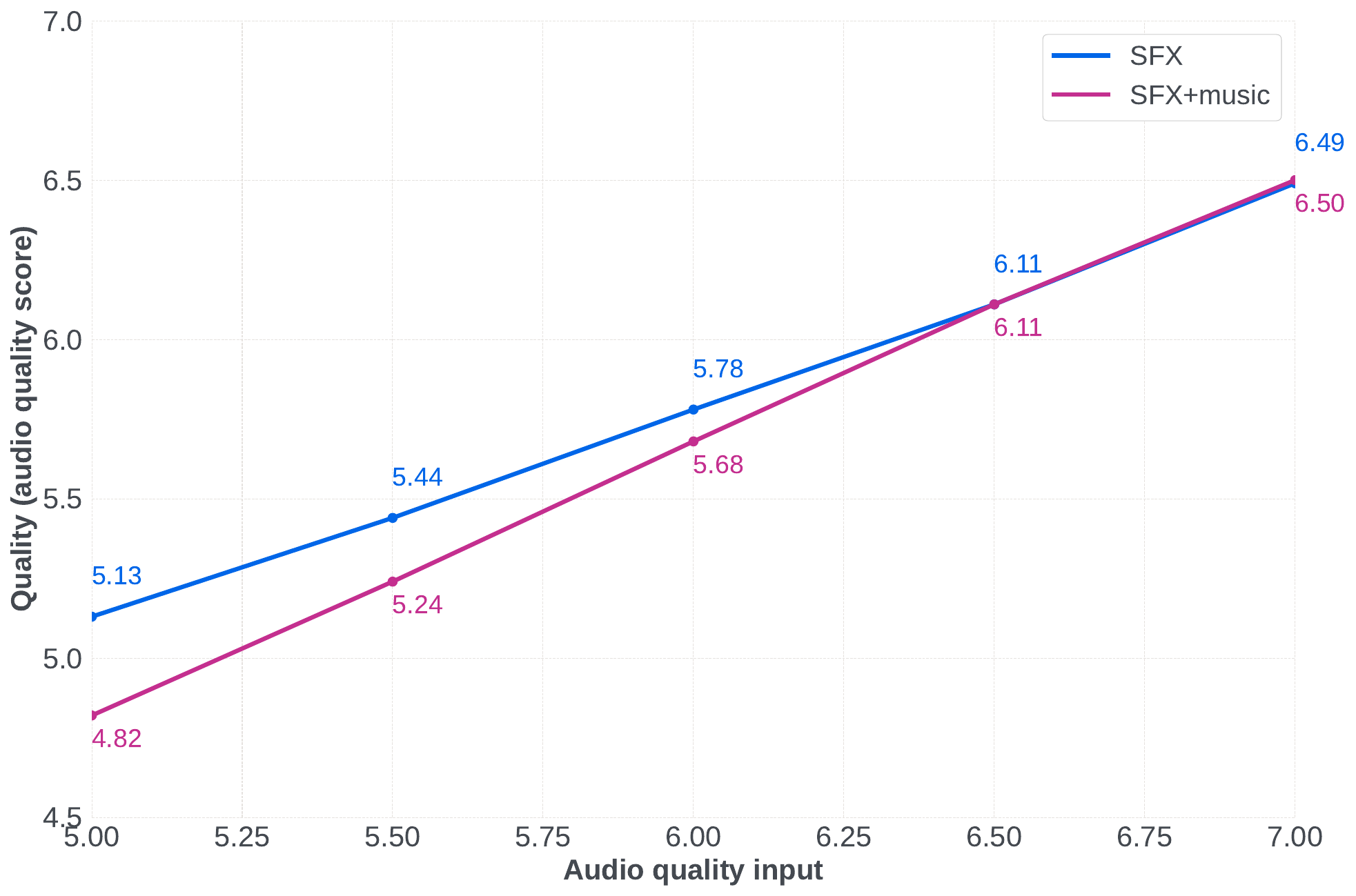}
        \subcaption{Audio quality score \vs prompted audio quality.}
    \end{subfigure}
    \begin{subfigure}[b]{.48\linewidth}
        \centering
        \includegraphics[width=.8\linewidth]{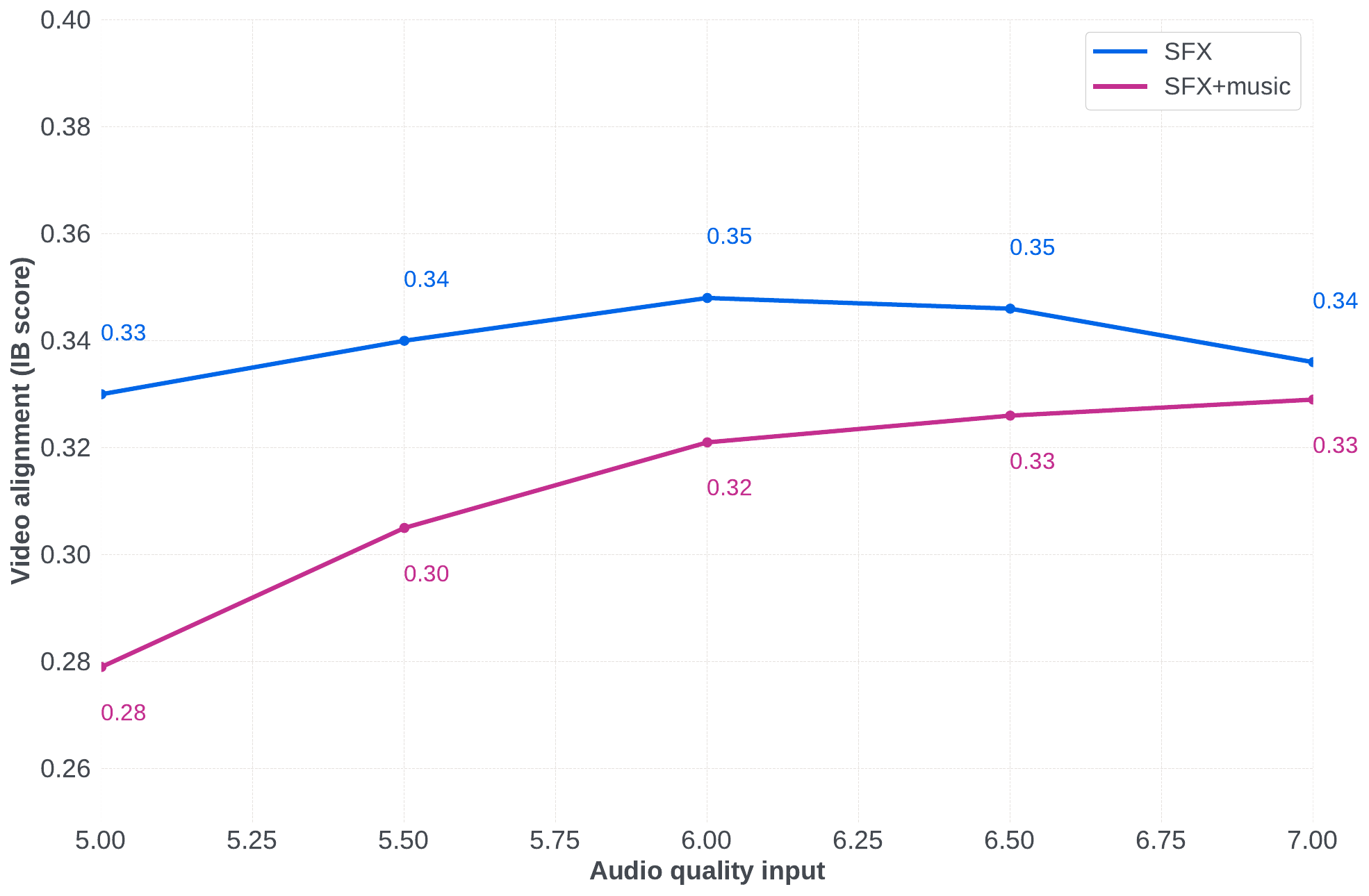}
        \subcaption{\IB score \vs prompted audio quality.}
    \end{subfigure}
    
    \caption{\textbf{\OursAudio audio quality control ablations with objective tests}}
    \label{fig:audio_abl_qual_obj}
\end{figure}

\textbf{Text prompt: audio quality control.}
We vary the audio quality specified in the text prompt (blue part in \cref{tab:audio_caption}) and demonstrate it can effectively control the audio quality.
We evaluate both SFX and joint SFX+music generation,
and present object and subjective metrics on both audio quality and video-SFX alignment in \cref{fig:audio_abl_qual_obj} and \cref{tab:audio_abl_qual_sbj}, respectively.
Qualitative samples are shown in \cref{app:qual_control}.

\begin{table}[h]
    \centering
    \adjustbox{max width=0.7\textwidth}{%
    \begin{tabular}{ccc ccc cc}
        \toprule
        \multirow{3}{*}{Dataset} & \multicolumn{2}{c}{\multirow{2}{*}{Prompted audio qual.}} & \multicolumn{5}{c}{A net win rate \vs B} \\
        & & & \multicolumn{3}{c}{Quality} & \multicolumn{2}{c}{Video-SFX Alignment} \\
        \cmidrule(lr){4-6} \cmidrule(lr){7-8} 
        & Model A & Model B & Ovr. & Nat. & Pro. & Corr. & Sync. \\
        \midrule
        \multirow{4}{*}{SFX}
        & 5.5 & 5.0 & \meanci{54.6}{20.0} & \meanci{50.0}{21.7} & \meanci{54.6}{20.0} & \meanci{3.3}{16.7} & \meanci{5.1}{22.2}\\
        & 6.0 & 5.5 & \meanci{32.4}{21.5} & \meanci{37.7}{17.6} & \meanci{32.3}{21.7} & \meanci{5.9}{15.0} & \meanci{3.1}{19.4} \\
        & 6.5 & 6.0 & \meanci{29.8}{23.4} & \meanci{27.9}{22.8} & \meanci{30.8}{22.8} & \meanci{-9.8}{18.6} & \meanci{-15.9}{23.5} \\
        & 7.0 & 6.5 & \meanci{-6.6}{22.8} & \meanci{-4.1}{22.8} & \meanci{-8.8}{21.1} & \meanci{11.5}{13.8} & \meanci{-3.1}{19.2} \\
        \midrule
        \multirow{4}{*}{SFX+music}
        & 5.5 & 5.0 & \meanci{58.6}{19.0} & \meanci{49.9}{19.1} & \meanci{58.6}{19.0} & \meanci{16.1}{16.1} & \meanci{20.9}{15.0} \\
        & 6.0 & 5.5 & \meanci{27.0}{22.8} & \meanci{33.9}{20.6} & \meanci{28.2}{22.8} & \meanci{23.1}{19.5} & \meanci{21.9}{21.4} \\
        & 6.5 & 6.0 & \meanci{24.5}{20.6} & \meanci{13.5}{21.7} & \meanci{25.6}{20.6} & \meanci{6.6}{15.6} & \meanci{15.1}{17.5} \\
        & 7.0 & 6.5 & \meanci{16.2}{20.6} & \meanci{22.9}{21.7} & \meanci{15.0}{22.2} & \meanci{-2.2}{12.8} & \meanci{-11.1}{13.3}\\
        \bottomrule
    \end{tabular}
    }
    \caption{\textbf{\OursAudio audio quality control ablations with subjective tests}}
    \label{tab:audio_abl_qual_sbj}
\end{table}

As we increase the conditioned audio quality, the predicted audio quality scores consistently improve on both datasets, and aligns with the subjective tests mostly up to 6.5.
Human raters show similar preference to 6.5 and 7.0 for SFX generation, where quality is harder to differentiate at that level.
These results validate the correlation between the subjective quality metric and the objective proxy metric (AQual), and demonstrate the effectiveness of quality control.
In terms of the impact on video-SFX alignment, the impact is not significant on SFX generation (IB score is between 0.33 and 0.35, and subjective preference is not significant for any pair). In contrast, we observe significant improvement on SFX+music generation from 5.0 to 6, where IB score improves from 0.28 to 0.32 and raters show significant preference to higher conditioned quality.
More details on metric correlation can be found in~\Cref{sec:app-audio-corr}.

\textbf{Text prompt: control SFX and music styles}
\cref{fig:audio_ablation_text_sfx} and \cref{fig:audio_ablation_text_music} in the appendix present examples of audio event and music style control through text prompts.
We observe in \cref{fig:audio_ablation_text_sfx} that text is particularly useful for dictating what \textit{unseen} sound events should be generated.
On the other hand, \cref{fig:audio_ablation_text_music} shows that text prompts can effectively control the music style, rendering different emotions for the same video.

\begin{table}[h]
    \centering
    \captionsetup{type=figure}
    \setlength{\tabcolsep}{1pt}
    \adjustbox{max width=0.95\textwidth}{%
    \centering
    \begin{tabular}{cccccc}
        \rowcolor{blue300}
        \multicolumn{6}{c}{\textbf{Ablation: SFX audio generation with different text prompts (10s, \OursVideo)}} \\
        \includegraphics[width=0.17\linewidth]{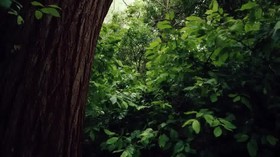}& 
        \includegraphics[width=0.17\linewidth]{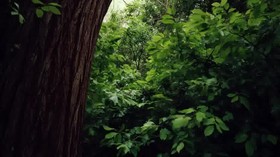}& 
        \includegraphics[width=0.17\linewidth]{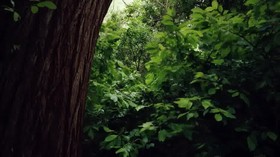}& 
        \includegraphics[width=0.17\linewidth]{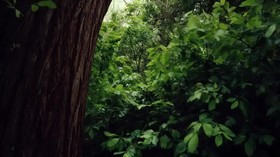}& 
        \includegraphics[width=0.17\linewidth]{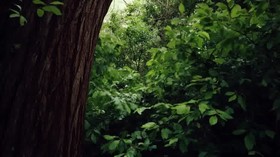}& 
        \includegraphics[width=0.17\linewidth]{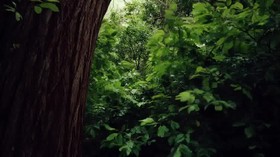}\\
        \multicolumn{6}{c}{{(a) Sound caption: footsteps crunching on leaves.)}} \\
        \multicolumn{6}{c}{
            \includegraphics[width=\linewidth, height=0.05\linewidth]
            {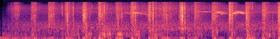}
        } \\
        \multicolumn{6}{c}{{(b) Sound caption: an owl hooting in the distance.}} \\
        \multicolumn{6}{c}{
            \includegraphics[width=\linewidth, height=0.05\linewidth]
            {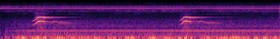}
        } \\
        \multicolumn{6}{c}{{(c) Sound caption: insects buzzing near the ground}} \\
        \multicolumn{6}{c}{
            \includegraphics[width=\linewidth, height=0.05\linewidth]
            {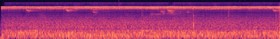}
        } \\
    \end{tabular}}
\vspace{-3mm}
\caption{\textbf{Examples of generated audio with different audio description in the text prompt with \OursAudio.}
Videos in this Figure found at \url{https://go.fb.me/MovieGen-Figure34}.
}
\label{fig:audio_ablation_text_sfx}
\end{table}

\textbf{Text prompt: with \vs without prompts.}
We study the impact of using text prompts during training and generation.
Here we pre-train and fine-tune two 1B parameter models, one with text input and other without, denoted as TV2A and V2A, respectively.
Text dropout is used for training for TV2A, so we can generate samples using only video input with that model as well.
We denote model trained with caption and generating without caption as ``TV2A $\rightarrow$ V2A'', and similarly for other setups.

Results are presented in \cref{tab:audio_abl_text_cap}.
We first note that on subjective tests, using text captions slightly improves quality, and significantly improves most alignment metrics.
Second, when running inference without text prompt, the model trained with text prompts (TV2A $\rightarrow$ V2A) outperforms the one without (V2A $\rightarrow$ V2A) especially on the subjective alignment metrics, showing that text can facilitate learning audio-visual correspondence.
Third, using text prompts can effectively guide model to generate the desired sound effects as shown by the higher CLAP score (0.37 \vs 0.23),
since there is still a high level of ambiguity on what sound events should present given a video.

\begin{table}[h]
    \begin{subfigure}[b]{.68\linewidth}
        \centering
        \adjustbox{max width=\textwidth}{%
        \begin{tabular}{cc ccc cc}
            \toprule
            \multicolumn{2}{c}{\multirow{2}{*}{Model and inference setup}} 
            & \multicolumn{5}{c}{A net win rate \vs B} \\
            & & \multicolumn{3}{c}{Quality} & \multicolumn{2}{c}{Video-SFX Alignment} \\
            \cmidrule(lr){3-5} \cmidrule(lr){6-7} 
            Model A & Model B & Ovr. & Nat. & Pro. & Corr. & Sync. \\
            \midrule
            TV2A $\rightarrow$ TV2A & TV2A $\rightarrow$ V2A & \meanci{12.5}{20.0} & \meanci{12.5}{20.0} & \meanci{11.1}{20.7} & \meanci{20.4}{20.3} & \meanci{17.5}{21.6} \\
            TV2A $\rightarrow$ TV2A &  V2A $\rightarrow$ V2A & \meanci{15.9}{17.4} & \meanci{10.3}{17.1} & \meanci{18.8}{18.0} & \meanci{26.7}{18.2} & \meanci{32.1}{19.7} \\
            \bottomrule
        \end{tabular}
        }
        \subcaption{Subjective metrics.}
    \end{subfigure}
    \begin{subfigure}[b]{.30\linewidth}
        \centering
        \adjustbox{max width=\textwidth}{%
        \begin{tabular}{c c}
            \toprule
            Model & CLAP \\
            \midrule
            TV2A $\rightarrow$ TV2A & 0.37 \\
            TV2A $\rightarrow$  V2A & 0.23* \\
            V2A  $\rightarrow$  V2A & 0.21* \\
            \bottomrule
        \end{tabular}
        }
        \subcaption{Objective metrics.}
    \end{subfigure}
    
    \caption{\textbf{\OursAudio text prompt ablations.} TV2A and V2A are models trained with and without text caption, ``w/ cap'' and ``w/o cap'' indicate whether text caption is used or not during inference. We put ``*'' on CLAP results when text is not used at inference.}
    \label{tab:audio_abl_text_cap}
\end{table}

\textbf{Model: scaling}.
We study the benefit of scaling and compares models of four different sizes: 300M, 3B, 9B, and 13B parameters.
The 300M model adopts the same architecture and configuration as \citet{vyas2023audiobox},
while the remaining ones use the DiT architecture described in \cref{sec:audio_model}.
Performance generally improves across all metrics as the model scales up, as shown in \cref{tab:audio_abl_model_size}.

\begin{table}[h]
    \begin{subfigure}[b]{.65\linewidth}
        \centering
        \adjustbox{max width=\textwidth}{%
        \begin{tabular}{cc ccc cc}
            \toprule
            \multicolumn{2}{c}{\multirow{2}{*}{Model parameter}}
            & \multicolumn{5}{c}{A net win rate \vs B} \\
            & & \multicolumn{3}{c}{Quality} & \multicolumn{2}{c}{Video-SFX Alignment} \\
            \cmidrule(lr){3-5} \cmidrule(lr){6-7}
            Model A & Model B & Ovr. & Nat. & Pro. & Corr. & Sync. \\
            \midrule
            3B & 300M & \meanci{29.9}{19.0} & \meanci{25.1}{18.7} & \meanci{20.2}{19.1} & \meanci{35.2}{14.3} & \meanci{34.9}{19.5} \\
            9B & 3B   & \meanci{34.6}{18.8} & \meanci{36.7}{18.8} & \meanci{36.7}{18.4} & \meanci{19.4}{13.4} & \meanci{20.1}{17.3}\\
            13B & 9B  & \meanci{11.0}{21.3} & \meanci{10.3}{21.3} & \meanci{11.7}{21.3} & \meanci{18.8}{19.3} & \meanci{23.1}{16.8} \\
            \bottomrule
        \end{tabular}
        }
        \subcaption{Subjective metrics.}
        \label{tab:audio_abl_model_size_sbj}
    \end{subfigure}
    \begin{subfigure}[b]{.33\linewidth}
        \centering
        \adjustbox{max width=0.9\textwidth}{%
        \begin{tabular}{c c}
            \toprule
            Model & CLAP \\
            \midrule
            300M & 0.23 \\
            3B  & 0.35 \\
            9B & 0.35 \\
            13B  & 0.38 \\
            \bottomrule
        \end{tabular}
        }
        \subcaption{Objective metrics.}
        \label{tab:audio_abl_model_size_obj}
    \end{subfigure}

    \caption{\textbf{Scaling the \OursAudio model size.}
    We observe that scaling the model size generally improves performance across all metrics.
    }
    \label{tab:audio_abl_model_size}
\end{table}

\textbf{Data: effectiveness of fine-tuning}.
We compare performance before and after fine-tuning in \cref{tab:audio_abl_pt_vs_ft}.
Fine-tuning significantly enhances both audio quality and video alignment.
Qualitatively, the generated videos exhibit a much more cinematic feel after fine-tuning, which highlights the importance of high-quality data curation for the fine-tuning process.

\begin{table}[h]
    \centering
    \adjustbox{max width=0.7\textwidth}{%
    \begin{tabular}{cc ccc cc}
        \toprule
        & & \multicolumn{5}{c}{A net win rate \vs B} \\
        & & \multicolumn{3}{c}{Quality} & \multicolumn{2}{c}{Video-SFX Alignment} \\
        \cmidrule(lr){3-5} \cmidrule(lr){6-7}
        Model A & Model B & Ovr. & Nat. & Pro. & Corr. & Sync. \\
        \midrule
        FT & PT & \meanci{41.7}{15.3} & \meanci{37.8}{16.3} & \meanci{43.0}{14.7} & \meanci{31.0}{16.0} & \meanci{24.9}{17.7} \\
        \bottomrule
    \end{tabular}
    }
    \caption{\textbf{Effectiveness of fine-tuning \OursAudio.}
    When comparing the \pretrained (PT) model to the finetuned (FT) model, we observe that generations from the finetuned model are significantly better.
    }
    \label{tab:audio_abl_pt_vs_ft}
\end{table}

\textbf{Data: effectiveness of high-quality audio-only data for fine-tuning}.
During fine-tuning, we supplement the cinematic audio-video data (Cin-AV), with an additional high-quality audio data (HQ-A) including both music and sound effects.
We show in \cref{tab:audio_abl_hq_data} that inclusion of high-quality audios yields significant improvement on quality and even slightly improves video alignment for SFX-only generation.
For joint sound effect and music generation, it leads to significant improvement on video-music alignment.
The inclusion of large-scale text-sound effect and text-music pairs enables the model to effectively disentangle different audio types.
The alignment between audio and video thus also improves, along with the overall quality of the generated sound.

\begin{table}[h]
    \centering
    \adjustbox{max width=\textwidth}{%
    \begin{tabular}{ccc ccc cc cc}
        \toprule
        & \multicolumn{2}{c}{\multirow{2}{*}{FT data}} & \multicolumn{7}{c}{A net win rate \vs B} \\
        & & & \multicolumn{3}{c}{Quality} & \multicolumn{2}{c}{Video-SFX Alignment} & \multicolumn{2}{c}{Video-Music Alignment}  \\
        \cmidrule(lr){4-6} \cmidrule(lr){7-8} \cmidrule(lr){9-10}
        Dataset & Model A & Model B & Ovr. & Nat. & Pro. & Corr. & Sync. & Mood & Action \\
        \midrule
        SFX & Cin-AV + HQ-A & Cin-AV        & \meanci{21.5}{18.7} & \meanci{24.3}{17.3} & \meanci{22.8}{18.7} & \meanci{11.7}{17.4} & \meanci{ 9.6}{18.1} & n/a & n/a \\
        SFX+music & Cin-AV + HQ-A & Cin-AV  & \meanci{ 1.7}{18.0} & \meanci{ 3.0}{18.0} & \meanci{ 0.4}{18.3} & \meanci{ 0.0}{14.7} & \meanci{ 8.6}{16.0} & \meanci{16.4}{12.8} & \meanci{14.9}{11.8} \\
        \bottomrule
    \end{tabular}
    }
    \caption{\textbf{Using high quality audio-only data in fine-tuning \OursAudio.}
    Including high quality audio-only data significantly improves audio quality and even video alignment for SFX-only generation.
    }
    \label{tab:audio_abl_hq_data}
\end{table}

\textbf{Extension: autoregressive \vs multi-diffusion}.
We compare the default extension method used in the main results (multi-diffusion, MD, with triangle window) with the other method (segment-level autoregressive generation, AR) and other configurations (windowing function for MD, use of beam search and conditioning methods for AR) described in \cref{sec:audio_ext}.
A one-shot generation topline that generates audio for the entire video without extension is also included.
We evaluate on the \OursAudioBenchSGen SFX+music set, following the setup described in \cref{sec:main_res},
and study models in two scale: 3B and 13B.
Results are shown in \cref{tab:audio_abl_ext}. We show qualitative samples for extension methods from the 13B model in \cref{fig:audio_ablation_extn}.

\begin{table}[h]
\centering
\adjustbox{max width=\textwidth}{%
\begin{tabular}{ll ccc cc cc}
    \toprule
    & & \multicolumn{5}{c}{A net win rate \vs B} \\
    & & \multicolumn{3}{c}{Quality} & \multicolumn{2}{c}{Video-SFX Alignment} & \multicolumn{2}{c}{Video-Music Alignment} \\
    \cmidrule(lr){3-5} \cmidrule(lr){6-7} \cmidrule(lr){8-9}
    Model & Extension method B & Ovr. & Nat. & Pro. & Corr. & Sync. & Mood & Action \\
    \midrule
    \multirow{6}{*}{3B}
    & \multicolumn{8}{c}{\textit{Extension method A: MD with triangle window}} \\
    & AR w/ context cond. \& beam            & \meanci{26.7}{12.5} & \meanci{24.7}{11.1} & \meanci{28.6}{13.1} & \meanci{ 8.2}{11.4} & \meanci{ 7.9}{11.3} & \meanci{7.2}{10.2} & \meanci{5.4}{8.4} \\
    & AR w/ traj. reg. \& tri. win.          & \meanci{18.5}{10.2} & \meanci{16.5}{10.3} & \meanci{18.4}{10.0} & \meanci{ 4.1}{10.5} & \meanci{10.6}{11.8} & \meanci{2.4}{10.4} & \meanci{-4.0}{8.3} \\
    & AR w/ traj. reg. \& tri. win. \& beam  & \meanci{10.6}{11.0} & \meanci{11.7}{ 9.7} & \meanci{10.6}{11.0} & \meanci{-6.4}{10.8} & \meanci{-1.7}{11.6} & \meanci{10.8}{9.7} & \meanci{10.8}{9.4} \\
    \cmidrule{2-9}
    & MD w/ uni. win.                        & \meanci{11.5}{ 7.9} & \meanci{12.4}{ 7.8} & \meanci{10.7}{ 8.2} & \meanci{ 1.1}{ 8.5} & \meanci{ 2.3}{ 9.0} & \meanci{5.2}{8.5} & \meanci{7.2}{8.7} \\
    \cmidrule{2-9}
    & One-shot generation (topline)          & \meanci{ 4.9}{12.7} & \meanci{ 9.6}{11.7} & \meanci{ 4.6}{12.8} & \meanci{ 2.4}{ 8.8} & \meanci{-4.6}{10.4} & \meanci{2.7}{9.6} & \meanci{3.7}{7.7} \\
    \midrule
    \multirow{6}{*}{13B}
    & \multicolumn{8}{c}{\textit{Extension method A: MD with triangle window}} \\
    & AR w/ context cond. \& beam            & \meanci{ 3.4}{11.1} & \meanci{ 4.1}{11.4} & \meanci{ 4.0}{11.0} & \meanci{10.0}{11.1} & \meanci{12.5}{11.3} & \meanci{ 9.8}{10.1} & \meanci{13.9}{ 8.6} \\
    & AR w/ traj. reg. \& tri. win.          & \meanci{ 3.6}{11.4} & \meanci{ 0.7}{10.3} & \meanci{ 3.0}{11.4} & \meanci{ 7.6}{10.0} & \meanci{ 3.1}{11.6} & \meanci{16.5}{11.4} & \meanci{ 3.7}{ 8.7} \\
    & AR w/ traj. reg. \& tri. win. \& beam  & \meanci{ 1.5}{10.2} & \meanci{ 6.0}{10.3} & \meanci{ 3.6}{10.3} & \meanci{-2.0}{ 9.0} & \meanci{ 3.0}{10.8} & \meanci{13.9}{ 9.5} & \meanci{15.4}{10.1} \\
    \cmidrule{2-9}
    & MD w/ uni. win.                        & \meanci{ 0.7}{11.1} & \meanci{-3.2}{10.2} & \meanci{ 0.7}{11.1} & \meanci{-1.6}{ 9.6} & \meanci{-2.5}{10.7} & \meanci{-1.5}{10.5} & \meanci{ 0.6}{ 8.6} \\
    \cmidrule{2-9}
    & One-shot generation (topline)          & \meanci{11.7}{11.6} & \meanci{ 9.3}{10.6} & \meanci{12.3}{11.7} & \meanci{ 0.3}{ 9.9} & \meanci{-1.6}{11.3} & \meanci{-0.5}{10.3} & \meanci{-3.0}{ 8.9} \\
    \bottomrule
\end{tabular}
}
\caption{\textbf{Ablation study on audio extension methods and configurations.}
We compare different audio extension methods across two model scales and evaluate the generations.
}
\label{tab:audio_abl_ext}
\end{table}

\begin{table}[h]
    \centering
    \captionsetup{type=figure}
    \setlength{\tabcolsep}{1pt}
    \adjustbox{max width=0.95\textwidth}{%
    \centering
    \begin{tabular}{cccccc}
        \rowcolor{blue300}
        \multicolumn{6}{c}{\textbf{Sound effects: crackling and popping of the fire, Music: dark, intense, and suspenseful orchestral track}} \\
\includegraphics[width=0.17\linewidth]{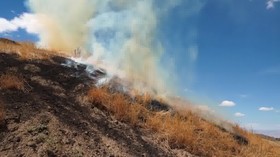}&
\includegraphics[width=0.17\linewidth]{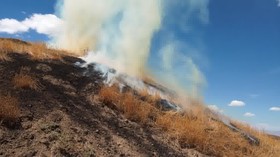}&
\includegraphics[width=0.17\linewidth]{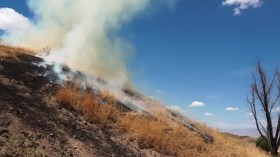}&
\includegraphics[width=0.17\linewidth]{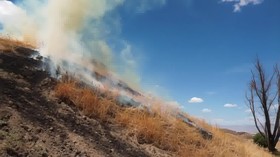}&
\includegraphics[width=0.17\linewidth]{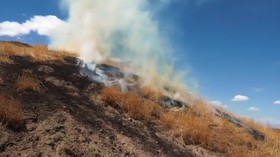}&
\includegraphics[width=0.17\linewidth]{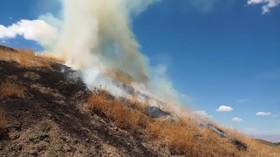}\\
        \multicolumn{6}{c}{{(a) multi-diffusion}} \\
\multicolumn{6}{c}{
    \includegraphics[width=\linewidth, height=0.05\linewidth]
    {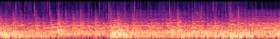}
} \\
        \multicolumn{6}{c}{{(b) stitch}} \\
\multicolumn{6}{c}{
    \includegraphics[width=\linewidth, height=0.05\linewidth]
    {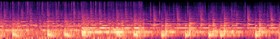}
} \\

        \multicolumn{6}{c}{{(c) autoregressive}} \\
\multicolumn{6}{c}{
    \includegraphics[width=\linewidth, height=0.05\linewidth]
    {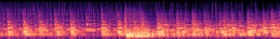}
} \\
        \rowcolor{blue300}
        \multicolumn{6}{c}{\textbf{Sound effects: water splashes against the rocky cliff, Music: classical piano music piece}} \\
\includegraphics[width=0.17\linewidth]{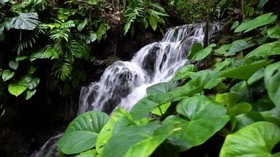}&
\includegraphics[width=0.17\linewidth]{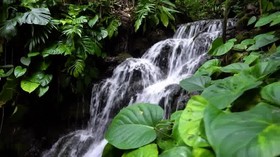}&
\includegraphics[width=0.17\linewidth]{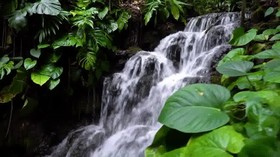}&
\includegraphics[width=0.17\linewidth]{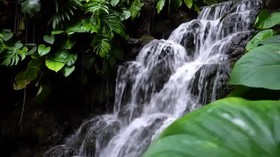}&
\includegraphics[width=0.17\linewidth]{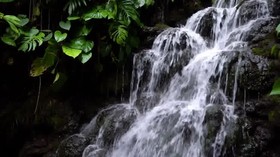}&
\includegraphics[width=0.17\linewidth]{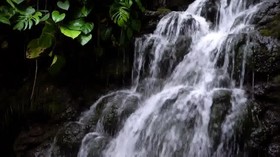}\\
        \multicolumn{6}{c}{{(a) multi-diffusion}} \\
\multicolumn{6}{c}{
    \includegraphics[width=\linewidth, height=0.05\linewidth]
    {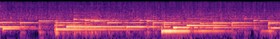}
} \\
        \multicolumn{6}{c}{{(b) stitch}} \\
\multicolumn{6}{c}{
    \includegraphics[width=\linewidth, height=0.05\linewidth]
    {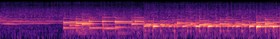}
} \\

        \multicolumn{6}{c}{{(c) autoregressive}} \\
\multicolumn{6}{c}{
    \includegraphics[width=\linewidth, height=0.05\linewidth]
    {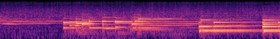}
} \\

    \end{tabular}}
\vspace{-3mm}
\caption{\textbf{Examples of sound track generation using different audio extension methods with \OursAudio.}
Each video comprises 2 segments of about 5 seconds each. Both multi-diffusion and autoregressive generate coherent samples with smooth transitions at the boundary. Stitching results in noticible music artifacts at the boundary.
Videos in this Figure found at \url{https://go.fb.me/MovieGen-Figure35}.
}
\label{fig:audio_ablation_extn}
\end{table}

We observed
\begin{enumerate*}[label=(\arabic*)]
    \item most methods are statistically similar on video-SFX alignment metrics,
    \item multi-diffusion outperforms most alternative extension methods on quality significantly on 3B, but the gap disappears after scaling to 13B,
    \item multi-diffusion is on par with the one-shot generation topline at 3B and even marginally better at 13B,
    \item the proposed triangle window leads to smoother transitions compared to the uniform window proposed in~\citep{bar2023multidiffusion}
    and results in higher audio quality (\vs ``MD w/ uni. win'') at 3B, but the gain again disappears at the larger scale.
    \item beam search improves autoregressive generation (``AR w/ traj. reg. \& tri. win.'' \vs ``AR w/ traj. reg. \& tri. win. \& beam'') at 3B,
    likely because the sample quality varies more for different seeds at smaller scales.
\end{enumerate*}

\section{Related work}
\subsection{\TextToI Generation}

Diffusion models have revolutionized the field of text-to-image generation. While a comprehensive review of all text-to-image models is beyond the scope of this paper, we will focus on the most relevant ones that have been published, productionized or open-sourced, and thus widely used by a large user base.

The seminal work of latent diffusion models~\citep{rombach2021highresolution} proposes compressing the original image space to latent space using a variational autoencoder, which improves training and inference efficiency, thus popularizing latent-space-based diffusion models. Dalle-3~\citep{dalle3} proposes using GPT to rewrite image captions, reducing noise in curated internet-scale text-image pairs for more effective training. Emu~\citep{dai2023emu} proposes using a higher latent dimension and fine-tuning a pre-trained model with a small, high-quality dataset to exclusively generate high-quality, professional-looking images. Stable Diffusion 3~\citep{sd3} proposes using rectified flow transformers with a multimodal diffusion backbone to improve generation quality.

In MovieGen, we also use a 16-channel variational autoencoder and flow transformers with prompt rewrite to achieve both high visual quality and text alignment.

\subsection{\TextToV Generation}

The swift progress in text-to-image generation has led to substantial improvements in temporally coherent high quality video generation. After the success of diffusion models for image generation~\citep{dhariwal2021diffusion, ramesh2022hierarchical}, they have been vastly used to improve video synthesis~\citep{NEURIPS2022_39235c56}.
Several works introduce zero-shot video generation by enriching the pre-trained text-to-image generation models with motion dynamics~\citep{text2video-zero, wu2023tune}. DirecT2V~\citep{hong2023direct2v} leverages instruction-tuned large language models for zero-shot video creation by dividing user inputs into separate prompts for each frame.

Several other papers propose a cascaded or factorized approach for text-to-video generation.  Imagen-Video~\citep{ho2022imagen} and Make-A-Video~\citep{singer2023makeavideo} trained a deep cascade of spatial and temporal layers via pixel diffusion modeling while many other works focus more on applying diffusion to the latent space of an auto-encoder for more efficiency~\citep{Blattmann2023AlignYL, an2023latentshift, wang2023lavie, wang2023videofactory, modelscope}.
AnimateDiff~\citep{guo2023animatediff} introduces a pre-trained motion module into a pre-trained T2I model. Emu-Video~\citep{emuvideo2023}, Stable Video Diffusion~\citep{sdvideo}, I2VGen-XL~\citep{i2vgen}, Dynamicrafter~\citep{xing2023dynamicrafter}, VideoGen~\citep{li2023videogen}, and VideoCrafter1~\citep{chen2023videocrafter1} add an image as an extra conditioning to the T2V model. Lumiere~\citep{BarTal2024LumiereAS} uses a Space-Time U-Net to generate the full temporal duration of the video at once. SEINE~\citep{chen2023seine} facilitates the smooth integration of shots from diverse scenes and generates videos of various lengths through auto-regressive prediction.

A few papers have studied the role of noise scheduling for more coherent~\citep{ge2023preserve, qiu2023freenoise, luo2023videofusion} and longer~\citep{kim2024fifo} video generation.
While most of the above text-to-video generation models use a U-Net based architecture, Snap-Video~\citep{snapvideo} and \Sora~\citep{sora} show the scalability and out-performance of transformer architectures for diffusion-based video generation.
Latte~\citep{ma2024latte} also uses a DiT instead of the U-Net backbone for text-to-video generation. On the other hand, a couple of works have been focused on transformer models within an auto-regressive framework~\citep{yan2021videogpt, kondratyuk2023videopoet, hong2022cogvideo, wu2022nuwa, villegas2023phenaki, ge2022long}. RIVER~\citep{davtyan2023efficient} uses flow matching for efficient video prediction by conditioning on a small set of past frames in the latent space of a pre-trained VQGAN. In this work, we leverage a Llama3 transformer architecture~\citep{llama3} and train a text-to-video generation model within a flow matching framework~\citep{flow-matching}.

\textbf{Encoding videos into a latent space.} Since their inception~\citep{rombach2021highresolution,Esser2021TamingTF}, encoder/decoder models have been a core part of latent generative architectures, and serve to compress raw media (images, video, audio) into a lower-dimensional latent space. Latent diffusion models~\citep{rombach2021highresolution} typically use either a normal variational autoencoder (VAE)~\citep{kingma2013auto}, a quantized VAE such as a VQVAE~\citep{van2017neural} or VQGAN~\citep{Esser2021TamingTF} and its variants~\citep{lee2022autoregressive}, which adds a GAN discriminator loss~\citep{goodfellow_GAN} to achieve improved reconstruction quality with greater compression.  Most image and video generation models use convolutional autoencoders, though transformer-based encoding models such as Efficient-VQGAN~\citep{cao2023efficient}, ViT-VQGAN~\citep{yu2021vector}, and TiTok~\citep{yu2024image} show promising results using vision transformers. For the \OURS autoencoder, we found best results using a continuous convolutional VAE with discriminator loss.

Of the methods using convolutional autoencoders, the models have been split between image (2D) and video (3D) models. First are those which use an image VAE (with or without quantization) and encode frame-by-frame — for example Stable Video Diffusion~\citep{sdvideo}, Latent Shift~\citep{an2023latentshift}, VideoLDM~\citep{Blattmann2023AlignYL}, Emu-Video~\citep{emuvideo2023}, and CogVideo~\citep{hong2022cogvideo}. Models which use image encoders are typically unable to directly generate long, high FPS videos due to lack of temporal compression. A more recent alternative has been to use 3D or mixed 2D-3D models. For example, MAGViT~\citep{52431} uses a 3D VQGAN with both 3D and 2D downsampling layers, with average pooling for downsampling, and W.A.L.T.~\citep{gupta2023photorealistic} and MAGViT-V2~\citep{yu2023language} use fully 3D convolutional encoders with strided downsampling. In our work, we chose to use an interleaved 2D-1D (\eg, 2.5D) convolutional encoder, where we trade off a slight improvement to reconstruction quality from a fully 3D model for lower memory and computational costs.

A feature of W.A.L.T., MAGViT-V2 and also some ViT-based encoders such as C-ViViT~\citep{villegas2023phenaki} is the inclusion of causality, which is typically implemented for convolutions through an asymmetrical padding. Causality is usually implemented because the first frame is always encoded independently, allowing images to be explicitly encoded for joint image and video generation. However, we have found that causal encoding is not necessary to encode images and videos jointly — symmetrical padding functions for joint image and video generation, and encoded images with symmetrically-padded convolutions are able to be used as conditioning for image-to-video models. Symmetrical padding works for different video lengths as long as replicate padding is used; a similar result is reported by TATS~\citep{ge2022long}.

\subsection{Image and Video Personalization}

\textbf{Personalized Image Generation.}
Prior work in personalized image generation has primarily focused on two technical directions: 1) identity-specific tuning and 2) tuning-free methods. Identity-specific tuning trains a text-to-image model to incorporate the identity by finetuning on a specific identity.
Textual Inversion~\citep{gal2022image} finetunes special text tokens for the new identity. DreamBooth~\citep{ruiz2023dreambooth} selects a few images from the same identity as reference as well as a special text token to represent the identity concept. LoRA techniques~\citep{hu2021lora} have been explored to tune a light-weight low-rank adapter
to accelerate the training process. HyperDreamBooth~\citep{ruiz2023hyperdreambooth} further reduces the training latency by directly predicting the initial weights of LoRA from the reference images. A major drawback of identity-specific tuning personalization methods is that the final model has parameters that are trained for and associated with a specific identity, and this process does not scale well to multiple users.

To overcome the limitations of the identity-specific tuning methods, another line of research extracts vision embeddings from the reference image and directly injects it into the diffusion process. This direction is more scalable as all users can share the same base model.
ELITE~\citep{wei2023elite} extracts vision features from the reference image and converts it to the text-embedding space through a local and a global mapping. PhotoMaker~\citep{li2023photomaker} merges the vision and text tokens and replaces the original text tokens for cross-attention.
PhotoVerse~\citep{chen2023photoverse} incorporates an image adapter and a text adapter to merge the vision and language tokens respectively. IP-Adapter-FaceID-Plus~\citep{ye2023ip} leverages face embedding and clip
vision encoder for identity preservation. InstantID~\citep{wang2024instantid} is a control-based method that adds ControlNet~\citep{zhang2023adding} to further control the pose and facial expression. MoA~\citep{ostashev2024moa} proposes a mixture-of-attention architecture to better fuse the vision reference and the text prompts. Imagine Yourself~\citep{meta24memu} proposed a full parallel model architecture,
a multi-stage finetuning strategy, and a novel synthetic paired data generation mechanism for better identity preservation, prompt alignment, and visual quality.

\textbf{Personalized Video Generation.}
While the aforementioned works have shown promising results in personalized image generation, adding personalization capability to video generation remains a challenging and unsolved problem. There are a few novel challenges in the area of personalized video generation: 1) compared to personalized image generation, personalized video generation needs to support more diverse and complex modifications on the reference image, \eg, turning the head, changing poses, and camera motion movements,
2) personalized video generation increases the expected quality threshold on expression and motion naturalness due to its temporal nature, and 3) finetuning a video model is much more costly than finetuning an image model, given the larger model and input sizes.

One direction of personalized video generation utilizes pose to control the video generation. Both GAN-based methods~\citep{chan2019everybody, yoon2021pose} and diffusion-based method~\citep{wang2024disco, wang2024disco, hu2024animate, xu2024magicanimate} have been proposed to generate videos following reference poses. These models are designed to animate the reference image towards the target motion, and are good for motions like singing and dancing. However, these models
require a pose sequence as reference, usually extracted from a real video, limiting its usage to a broader scope of scenearios. Also, these models often introduce occlusion and unnatural motion due to the non-ideal pose extraction and control.

On the other hand, preliminary works have been proposed to turn a personalized image model to a personalized video model. Magic Me~\citep{ma2024magic} uses identity-specific finetuning to inject identity into a video generation model. ID-Animator~\citep{he2024id} leverages a face adapter to extract identity information from single reference facial image for personalized video generation. DreamVideo~\citep{wei2024dreamvideo} and Still-Moving~\citep{chefer2024still} combine an identity adapter and a motion adapter for flexible video customization. CustomCrafter~\citep{wu2024customcrafter} proposes a
plug-and-play module for subject concept injection. Our work also focuses on using identity extraction and control methodology, as it is more generalizable to different users.

\subsection{Instruction-Guided Video Editing}
\label{subsec:edit_background}
In the task of text-based video editing\footnote{For brevity, we refer to text or instruction-based video editing simply as video editing.} the user provides the model with a video (either real or generated) along with an editing instruction text that specifies how they would like to alter the video.
The model is then expected to precisely modify the input video according to the given instruction, changing only the specified elements while preserving those that should remain intact.
The main challenge in developing a high-performing video editing model arises from the difficulty in collecting supervised data for this task.

As a result of this challenge, most prior work relies on training-free approaches~\citep{meng2021sdedit,Geyer2023TokenFlowCD,vid2vid-zero,text2video-zero,li2023vidtome,Ceylan2023Pix2VideoVE,2312.04524,yang2023rerender}, which can be applied to any \textToV model without requiring additional training.
In contrast to training-free methods, some approaches train models to generate videos by providing additional features of the video as input (\eg, depth or segmentation maps)~\citep{esser2023structure,Liang2023FlowVidTI,Yan2023MotionConditionedIA}.
However, these methods are inherently limited, as they cannot control features that were not incorporated during training.
For example, preserving the identity of an object or subject is impossible when conditioning only on the depth maps of a video.
Overall, both training-free methods and feature-based approaches tend to be imprecise and have been shown to perform worse than methods that explicitly adapt model parameters to process the entire video input during training~\citep{EVE,Qin2023InstructVid2VidCV}.

The current state-of-the-art approach for video editing, Emu Video Edit (EVE)~\citep{EVE}, employs two training stages to develop a video editing model.
First, it trains dedicated adapters for text-image-to-video generation and image editing on top of a shared \textToI model.
Next, it performs Factorized Diffusion Distillation (FDD) to align the adapters towards video editing.
In each training step of FDD, the model first generates an edited video through multiple diffusion steps.
Then, the edited video is given as input to two adversarial losses and two knowledge distillation losses, which provide supervision for the quality of the edited video.
Finally, this supervision is backpropagated through both the different losses and the entire generation chain.

We identify several key differences when comparing our approach to EVE.
First, we initialize training from a \textToV model and perform full model training (\Cref{subsec:edit_stage_1}), rather than training adapters for \textToV and image editing on top of a shared \textToI model.
Additionally, EVE's FDD backpropagates supervision through multiple forward passes with the model during generation, and an additional forward pass using each of the models it uses to provide supervision.
This makes FDD an order of magnitude more memory demanding than our approach, which limits the scalability of FDD.
By applying our approach to the \OursVideo (see~\Cref{sec:image_video_model}, we demonstrate that we can significantly surpass the reported results by EVE and set new state-of-the-art results in video editing.

\subsection{Audio Generation}\label{sec:related_audio}

\textbf{Video-to-audio generation.}
There are many recent studies exploring video-to-audio generation. Most of them are based on latent diffusion models similar to ours~\citep{luo2024diff, xu2024video, xing2024seeing, zhang2024foleycrafter} with a few exceptions being token-based language models~\citep{kondratyuk2023videopoet, mei2023foleygen}. Diff-Foley~\citep{luo2024diff} and VTA-LDM~\citep{xu2024video} are the standard latent diffusion models with U-Net architectures conditioned on video features, extracted from a pre-trained contrastive audio-video encoder (CAVP~\citep{luo2024diff}) and video-text encoder (CLIP4CLIP~\citep{luo2022clip4clip}), respectively. Seeing-and-hearing (S\&H)~\citep{xing2024seeing} proposes a training-free method that uses ImageBind as a classifier guidance to guide a pre-trained diffusion-based text-to-audio (TTA) model (AudioLDM~\citep{liu2023audioldm}) to generate video aligned audio. FoleyCrafter~\citep{zhang2024foleycrafter} uses the adaptor-based approach~\citep{zhang2023adding} to finetune a pre-trained TTA model to add video control. To enhance the temporal alignment between audio and video, it further conditions on timestamps to inform the model which segments are sounded/silence, which are predicted from the video during inference.

Most of these models are directly trained on, or built on TTA models trained on solely in-the-wild datasets, such as VGGSound (550 hours)~\citep{vggsound} or AudioSet (5K hours)~\citep{gemmeke2017audio}.
These datasets have several limitations. In terms of quality, many of the videos are recorded with non-professional devices like smartphones or low-end cameras, and hence both the video and the audio quality are subpar.
In terms of sound design, most of the videos are uploaded by amateur creators who had done none or minimal post-processing and post-production, which contain only diegetic sounds. Compared to professionally created films, such videos contain an abundance of distracting sounds (\eg, irrelevant off-screen speech, wind noise, high ambient noise), do not emphasize main sound events (\eg, exaggerated breathing sound, Foley sounds for footsteps and object clatters), and lack carefully designed non-diegetic sounds that are critical for a cinematic feeling (\eg, use of riser, braam, underscore music, action scoring).
Training on such datasets inevitably prevents the resulting model from generating cinematic soundtracks including both music and sound effects.

On the other hand, the size of prior video-to-audio generation models are relatively small, typically ranging from 300M to 1.3B parameters~\citep{xing2024seeing, luo2024diff, mei2023foleygen}. Combined with the data size, the scale also limits the performance of these models, as we demonstrated in the ablation study (see~\cref{sec:audio_ablation}) that scaling to 13B significantly improves both quality and video alignment.
Compared with the prior works that also offer text control, we additionally provide quality control and fine-grained music control, which alleviates the quality issue when training on mixed-quality data, and improves soundtrack design flexibility.

There are a few products offering video-to-audio capabilities, including PikaLabs\footnote{\url{https://pika.art/}} and ElevenLabs.\footnote{\url{https://github.com/elevenlabs/elevenlabs-examples/tree/main/examples/sound-effects/video-to-sfx}}, but neither can really generate motion-aligned sound effects or cinematic soundtracks with both music and sound effects. PikaLabs supports sound effect generation with video and optionally text prompts; however it will generate audio longer than the video where a user needs to select an audio segment to use. This implies under the hood it may be an audio generation model conditioned on a fixed number of key image frames. The maximum audio length is capped at 15 seconds without joint music generation and audio extension capabilities, preventing its application to soundtrack creation for long-form videos.
ElevenLabs leverages GPT-4o to create a sound prompt given four image frames extracted from the video (one second apart), and then generates audio using a TTA model with that prompt.
Lastly, Google released a research blog\footnote{\url{https://deepmind.google/discover/blog/generating-audio-for-video/}} describing their video-to-audio generation models that also provide text control. Based on the video samples, the model is capable of sound effects, speech, and music generation. However, the details (model size, training data characterization) about the model and the number of samples (13 samples with 11 distinct videos) are very limited, and no API is provided. It is difficult to conclude further details other than the model is diffusion-based and that the maximum audio length may be limited as the longest sample showcased is less than 15 seconds.

\textbf{Video to music generation.}
Many studies often focus on symbolic music (MIDI) generation~\citep{Di2021VideoBM, Zhuo2022VideoBM, Kang2023Video2MusicSM} for piano or other instruments, as MIDI is easier to predict compared to raw audio. Compared to end-to-end modeling, such a paradigm imposes many restrictions. First, MIDI is a form of music transcription which cannot capture all the details from the original music. Hence, the generated music from such systems tend to sound more monotonic. Second, it requires MIDI annotation or a high quality music transcription model, which limits the sources of training data one can consider. Lastly, such models cannot learn the relationship between music and other audio components like sound effects and speech, which are not trivial in cinematic films.

Most of these works, along those directly predicting audio~\citep{Zhu2022QuantizedGF}, extract low-level music-related features from videos, such as human motion, scene change timing, and tempo, for conditioning. These features along with the training data (\eg, dancing videos) are rather domain specific, which cannot generalize to general videos.

Our work is most related to \citet{Su2023V2MeowMT} and \citet{Tian2024VidMuseAS}. Both prior works extract general video features (CLIP, flow, image tokens) and predict general audio representations (EnCodec~\citep{Defossez2022HighFN}, Soundstream~\citep{Zeghidour2022SoundStreamAE}, w2vBERT~\citep{Chung2021w2vBERTCC}). In addition to our novel scaling, the main differences in our work are twofold: first we aim for joint non-diegetic music and sound effect generation while these studies only focus on non-diegetic music; second we adopt diffusion modeling which is free of tokenization information loss and hence enjoys the other benefits described in previous sections, while \citet{Tian2024VidMuseAS} points out explicitly that their quality is suffer from the audio codec limitation.

\section{Conclusion}

The \OURS cast of foundation models represents a significant improvement in \textToV generation, video personalization, video editing, as well as sound effect and music generation.
We show that training such models by scaling data, training compute, and model size together leads to such significant improvements.
We focus on curating high quality large scale data for \pretraining and relatively smaller scale but even higher quality data for finetuning.
This general recipe works well for improving the quality of image, video, and audio generation.
We introduce a new approach for equipping strong video foundation models with state-of-the-art video editing capabilities without relying on supervised video editing data. This is achieved through multi-task training on image editing and video generation, followed by two short fine-tuning stages: one on synthetic multi-frame editing data, and another on video editing via backtranslation.

Despite these improvements, we observe that video generation models still suffer from issues -- artifacts in generated or edited videos around complex geometry, manipulation of objects, object physics, state transformations \etc
Generated audio is sometimes out of synchronization when motions are dense (\eg, tap dance), visually small or occluded (\eg, footsteps), or when it requires finer-grained visual understanding (\eg, recognizing the guitar chords).
It currently does not support voice generation either due to our design choices.
Reliable benchmarking of media generation models is important for identifying such shortcomings and for future research.
Having access to a few cherry picked generations or black box systems without clear details on model or data makes reliable comparisons hard.
Along with details on models, data, and inference, we additionally release multiple non cherry picked generations and prompt sets to enable easy and reliable comparisons for future work.
We also note that defining objective criteria evaluating model generations using human evaluations remains challenging and thus human evaluations can be influenced by a number of other factors such as personal biases, backgrounds \etc.
While our models are trained separately for video and audio generation, developing models that can generate these modalities jointly is an important area of research.

\par \noindent \textbf{Safety considerations.}
The \OURS cast of foundation models were developed for research purposes and need multiple improvements before deploying them.
We consider a few risks from a safety viewpoint.
Any real world usage of these models requires considering such aspects.
Our models learn to associate text and optionally additional inputs like video to output modalities like image, video and audio.
It is also likely that our models can learn unintentional associations between these spaces.
Moreover, generative models can also learn biases present in individual modalities, \eg, visual biases present in the video training data or the language used in the text prompts.
Our study in this paper is limited to text inputs in the English language.
Finally, when we do deploy these models, we will incorporate safety models that can reject input prompts or generations that violate our policies to prevent misuse.

\section*{Contributors and Acknowledgements}
A large number of people at Meta worked to create \OURS.
We list \textbf{core contributors} (people who worked on \OURS for at least $\nicefrac{2}{3}$rd of the runtime of the project), and \textbf{contributors} (people who worked on \OURS for at least $\nicefrac{1}{3}$rd of the runtime of the project).
We list all contributors in alphabetical order of the first name.

\subsection*{Core Contributors}
Adam Polyak, Amit Zohar, Andrew Brown, Andros Tjandra, Animesh Sinha, Ann Lee, Apoorv Vyas, Bowen Shi, Chih-Yao Ma, Ching-Yao Chuang, David Yan, Dhruv Choudhary, Dingkang Wang, Geet Sethi, Guan Pang, Haoyu Ma, Ishan Misra, Ji Hou, Jialiang Wang, Kiran Jagadeesh, Kunpeng Li, Luxin Zhang, Mannat Singh, Mary Williamson, Matt Le, Matthew Yu, Mitesh Kumar Singh, Peizhao Zhang, Peter Vajda, Quentin Duval, Rohit Girdhar, Roshan Sumbaly, Sai Saketh Rambhatla, Sam Tsai, Samaneh Azadi, Samyak Datta, Sanyuan Chen, Sean Bell, Sharadh Ramaswamy, Shelly Sheynin, Siddharth Bhattacharya, Simran Motwani, Tao Xu, Tianhe Li, Tingbo Hou, Wei-Ning Hsu, Xi Yin, Xiaoliang Dai, Yaniv Taigman, Yaqiao Luo, Yen-Cheng Liu, Yi-Chiao Wu, Yue Zhao, Yuval Kirstain, Zecheng He, Zijian He

\subsection*{Contributors}
Albert Pumarola, Ali Thabet, Artsiom Sanakoyeu, Arun Mallya, Baishan Guo, Boris Araya, Breena Kerr, Carleigh Wood, Ce Liu, Cen Peng, Dimitry Vengertsev, Edgar Sch\"onfeld, Elliot Blanchard, Felix Juefei-Xu, Fraylie Nord, Jeff Liang, John Hoffman, Jonas Kohler, Kaolin Fire, Karthik Sivakumar, Lawrence Chen, Licheng Yu, Luya Gao, Markos Georgopoulos, Rashel Moritz, Sara K. Sampson, Shikai Li, Simone Parmeggiani, Steve Fine, Tara Fowler, Vladan Petrovic, Yuming Du

\subsection*{Acknowledgements}
Ahmad Al-Dahle, Ahnaf Siddiqui, Ahuva Goldstand, Ajay Ladsaria, Akash Jaiswal, Akio Kodaira, Andrew Treadway, Andrés Alvarado, Antoine Toisoul, Baishan Guo, Bernie Huang, Brandon Wu, Brian Chiang, Brian Ellis, Chao Zhou, Chen Fan, Chen Kovacs, Ching-Feng Yeh, Chris Moghbel, Connor Hayes, Daniel Ho, Daniel Lee, Daniel Li, Danny Trinh, David Kant, David Novotny, Delia David, Dong Li, Ellen Tan, Emory Lin, Gabriella Schwarz, Gael Le Lan, Jake Lately, Jeff Wang, Jeremy Teboul, Jiabo Hu, Jianyu Huang, Jiecao Yu, Jiemin Zhang, Jinho Hwang, Joelle Pineau, Jongsoo Park, Junjiao Tian, Kathryn Stadler, Laurence Rouesnel, Lindsey Kishline, Manohar Paluri, Matt Setzler, Max Raphael, Mengyi Shan, Munish Bansal, Nick Zacharov, Pasan Hapuarachchi, Peter Bi, Peter Carras, Philip Woods, Prash Jain, Prashant Ratanchandani, Ragavan Srinivasan, Rebecca Kogen, Ricky T. Q. Chen, Robbie Adkins, Rod Duenes, Roman Shapovalov, Ruihan Shan, Russ Maschmeyer, Shankar Regunathan, Shaun Lindsay, Sreeram R Chakrovorthy, Sudarshan Govindaprasad, Thai Quach, Tiantu Xu, Tom Monnier, Ty Toledano, Uriel Singer, Vlad Shubin, Wei Jiang, Will Seyfer, Xide Xia, Xinyue Zhang, Yael Yungster, Yang Liu, Yang Shu, Yangyang Shi, Yaron Lipman, Yash Mehta, Ye Jia, Zhaoheng Ni

\phantomsection
\addcontentsline{toc}{section}{References}
\bibliographystyle{assets/plainnat}
\bibliography{refs}

\newpage
\beginappendix
\phantomsection
\addcontentsline{toc}{section}{Appendix}
\renewcommand{\thesection}{\Alph{section}}
\renewcommand{\thesubsection}{\Alph{section}.\arabic{subsection}}

\section{Additional Model Training Details}

\subsection{Text encoders}
As described in~\cref{sec:t2v_text_encoder} we use text embeddings from three text encoders.
We provide details on the text encoders and how they are used for visual-text generation.

\noindent \textbf{Long-prompt MetaCLIP training.}
We train our own CLIP-style~\citep{xu2023demystifying} model that can process long text prompts up to $256$ text tokens.
We utilized synthetic image captions generated by an image captioning model~\citep{llama3} to finetune the MetaCLIP~\citep{xu2023demystifying} text encoder.
We expanded the position embedding in the text encoder to $256$ tokens, allowing it to handle longer input sequences.
However, we found that finetuning all parameters in the text encoder led to rapid overfitting and suboptimal text encoding performance.
We hypothesize that this is due to the uniform text style in the synthetic long image captions.
To address this issue, we froze both the image and text encoders and finetuned the position embedding and all biases in the text encoder using a mix of long synthetic, short synthetic, and human-annotated captions.

\par \noindent \textbf{Visual text generation.}
We implemented several key modifications to the input and prompt rewriting, model architecture, and data processing steps to enable visual text generation.
Specifically, our approach consists of three main components.
First, we assume that text enclosed within quotation marks (\textbf{`` ''}) in the input text prompt needs to be generated as visual-text.
We also employ prompt rewriting (see~\cref{sec:t2v_prompt_rewrite}) to automatically identify such text at inference time.
Second, we use the character-level ByT5 encoder for encoding the text within the quotation marks.
Finally, we ensure that our \pretraining data has a good mix of visual text that covers 10-50\% of the image area by leveraging OCR detection models.

\subsection{Model scaling and training efficiency}
\label{app:t2v_impl_scaling_additional}

For memory capacity constrained models like \OursVideo with a context length of 73K, sub-optimal performance may be achieved through memory-saving techniques like activation checkpointing (AC).
AC reduces peak memory demand by trading off FLOPs and memory, which can be sometimes be more effective than fully-optimized parallelisms due to physical training system constraints, \eg, the FLOPs/sec and HBM capacity of each GPU, and the intra- and inter- node GPU-GPU interconnect bandwidths.

\noindent \textbf{Overlapping communication and computation.}

While the parallelism techniques mentioned in \cref{sec:t2v_impl_scaling}—which aim to partition training FLOP and memory demands across GPUs at the cost of added communication—are successful at enabling the 0-to-1 ability to train such large sequence transformer models, their direct implementation and composition come with overheads and inefficiencies with respect to memory and communication. As a result, under such overheads optimal \textit{realized} performance for a model which is memory capacity constrained, as-is \OursVideo with a context length of 73K, may actually be achieved through the use of other memory-saving techniques in addition to, or even in place of, the above parallelisms.

Given that AC can never provide strong scaling due to strictly increasing the amount of work needed to compute a training step, we focused our efforts on improving the scaling characteristics of model parallelism techniques. Specifically, we aimed to have a final model parallelism implementation which achieved as close to strong scaling as possible for both: 1) activation memory size (to reduce the usage of AC), and 2) forward/backward step time.

As the four existing model parallel techniques discussed in~\cref{sec:t2v_impl_scaling} (TP, SP, CP, FSDP) all already achieve strong theoretical FLOP scaling, we began by: 1) determining the scaling characteristics of the activation sizes of our model under these techniques as well as 2) building an analytical framework to model the change and interdependence of the compute and communication times of our model execution. Using both, we: 1) identified all occurrences of duplicated activations, and 2) identified which inter-GPU communications are required as well as exposed.

From this we designed and implemented a model parallel solution for the \OursVideo backbone based upon the principles of TP, SP, and CP, which: 1) achieves strong activation memory scaling, and 2) minimizes exposed communication time for our training cluster. This enabled us to achieve close-to-strong scaling even with model parallelism widths spanning multiple nodes. We achieved such performance through a custom implementation defining the execution of both the forward and backward paths of the \OursVideo backbone block. Forgoing the use and ease of chaining smaller autograd operators together allowed us to precisely partition, control, and overlap compute and communication execution. This also enabled us to transparently apply techniques such as selective recomputation without any performance loss.

Our implementation is fully written in PyTorch and compiles into CUDAGraphs supporting the execution of variable sized inputs at the start of every training or inference job, dynamically based on the specified model and system configurations.

\noindent \textbf{Sharding plan generation and selection.}
We expanded the performance modeling tools developed above to further estimate the memory utilization and latency of end-to-end model execution (\eg, backbone blocks, text encoders, \taeShort). We then utilized this to generate multiple sharding and parallelism plans, which exhibit similar theoretical and estimated latencies, and are deemed valid as they are estimated to fit within the available GPU High-Bandwidth Memory (HBM).

This development: 1) enabled us to generate sharding plans in which different components and stages of the end-to-end model execution can be sharded by different strategies; and 2) allowed us to empirically identify training parallelism configurations which have an approximately neutral batch-size to step time scaling relationship.

The ability to accomplish the latter was important for the successful training of \OursVideo due to the relationship of model parallel size to the global batch size of its corresponding training step. Specifically, both the number of GPUs over which parameters are disjointly sharded (TP) and the number of GPUs over which the input sequence of a single sample are disjointly sharded (SP, CP), proportionally reduce the effective global batch size—and impact the wall latency of—the corresponding training step. Identifying groups of sharding plans which have neutral batch-size to step time scaling allows us to effectively scale the number of optimizer steps, while holding the number of GPUs and total training data size constant. Although the training data throughput and end-to-end efficiency of such groups of sharding plans are similar, the number of optimizer steps taken to process varying amounts of training data across various stages of training can significantly impact the final model's quality.

The final sharding plan used during the the most expensive stage of \OursVideo training, processing 768px video inputs with a per-sample token sequence length of 73K, was the following:
\begin{itemize}
  \item \textbf{Text encoders.} Due to their relatively small size and weights being frozen, ByT5 and Long-Prompt MetaCLIP were replicated on all GPUs. UL2 however has significantly more parameters and memory overhead, yet it is still relatively small in terms of end-to-end execution latency, and was sharded FSDP-only across the DP-group of each TP-rank.
  \item \textbf{\taeShort.} Although containing a relatively small number of frozen parameters, the size of intermediate activations of the \taeShort can become prohibitively expensive as the input size grows. Furthermore, unlike the text encoders, the latency of the \taeShort is non-trivial with respect to the end-to-end step time, and unlike the backbone, it is non-trivial to efficiently partition and model parallelize the \taeShort's execution. These limitations resulted in us performing a data pre-processing step where the latents for high resolution video inputs were pre-computed and cached prior to their ingestion in the \OursVideo backbone training pipeline.
  \item \textbf{\OursVideo} \textbf{backbone.} While a transformer block at its core, the \OursVideo backbone has additional learned components, such as factorized positional embeddings and per-context embedders, each with not only their own memory and compute requirements but also their own input and output activation connections. This results in the backbone parameter sharding and input and output activation and gradient flow changing as the model moves through its different stages. The final backbone contained interconnected sections sharded: FSDP-only (\eg, patchifier), FSDP+TP (\eg, context embedders), FSDP+TP+SP (\eg, cross-attention), and FSDP+TP+SP+CP (\eg, self-attention).
\end{itemize}

\section{Additional Data Details}

\subsection{Video Data Curation Thresholds}  \label{appendix:data_curation_thresholds}
In this section, we share more details about models used in data curation and the corresponding thresholds used.

\textbf{OCR model.} Our internal OCR model samples frames adaptively, detects words within those sampled frames, and then recognizes the text of those detected words.
We only retained videos where the word detection score multiplied by the word recognition score was below 0.6 for all sampled frames.

\textbf{Border detection.} We noticed that the presence of borders in training videos resulted in generated videos having black borders around them. This issue is particularly common in portrait-mode videos. We removed such videos by writing a simple border detector based on first order derivative calculations. We first detect pixels with large vertical and horizontal deltas and then apply a scanning line algorithm to find the borders.

\textbf{Clip sampling.} With an average duration of 28 seconds, our raw videos needed to be clipped into shorter segments to meet our \OursVideo model's training requirements of 4-16 second clips. However, we noticed that randomly sampling clips without considering scene boundaries leads to generating videos with frequent and abrupt scene changes.
Thus, we used FFmpeg~\citep{ffmpeg} to detect scene changes and sampled 1-2 scenes with duration exceeding 16 seconds from each video. We then randomly extracted a single clip from each scene, with a duration ranging from 4-16 seconds, to use as a training clip.  More than 50\% of our training clips have duration ranging from 15 seconds to 16 seconds.

\textbf{Aesthetic filtering.} We removed clips with poor aesthetic quality such as blurry or compressed clips by applying the public LAION aesthetics image model~\citep{schuhmann2022laion} on the middle frame of each clip.
We removed all clips with an aesthetic score less than 4, ensuring to have high-quality clips for training. We also calculated average aesthetic scores across multiple frames in a clip and observed that multi-frame aesthetic score didn't lead to a significant increase in the recall of poor-quality clips.

\textbf{Jittery motion detection.} FFmpeg~\citep{ffmpeg} motion scores and motion vectors struggle to detect videos with frequent, jittery camera movements,
which ultimately leads to our model generating videos with a jittery quality.
We noticed that the Shot Boundary Detection (SBD) from PySceneDetect~\citep{pyscenedetect} breaks down jittery videos into numerous false-positive shots.
To identify and remove jittery videos, we used the number of shots detected per second, removing clips with a rate exceeding 0.85 shots per second.

\textbf{Data volume per filter.} We analyzed the data volume drop at each filtering step when using our most strict curation thresholds in~\cref{tab:pt-volume-drop}.
These thresholds were used to curate our high resolution set.

\begin{table}
    \centering
    \adjustbox{max width=\textwidth}{
    \begin{tabular}{cccc}
        \toprule
        Curation Step& Thresholds & Remaining volume in \% \\
        \midrule
        Duration & 4s $\leq$ duration $\leq$ 120s & 100 \\
        Resolution & width $\geq$ 768 and height $\geq$ 768 & 25 \\
        Aspect Ratio & width $\geq$ height & 7\\
        No Text & No sampled frame with
        word detection score * word recognition score $\geq$ 0.6
         & 1.94 \\
        No Border & No videos with borders around them
        & 1.87 \\
        No Scene Change & 1 clip of duration 12s to 16s sampled from 1 scene in a video
        & 1.78 \\
        Aesthetics &  aesthetic score on middle frame of clip $\geq$ 4.0
        & 1.57 \\
        No Slow Motion &  motion score $>$ 2.0, motion vector average $>$ 0.5,
        motion vector average $<$ 7
        & 1.32 \\
        No Jittery Motion &  Number of shots per second $<$ 0.85
        & 1.22 \\
        No Content Duplicates &  Embedding cosine similarity $<$ 0.99
        & 1.15 \\
        Concept Resampling &  Volume per cluster: 1/sqrt(cluster size)
        & 0.94 \\
        \bottomrule
    \end{tabular}}
    \caption{\textbf{Volume drop during high resolution set curation.}
    Note that our data acceptance rate with these thresholds is less than 1\%.
    }
    \label{tab:pt-volume-drop}
\end{table}

\subsection{Camera Motion Control types} \label{appendix:data_camera_motion}
Here, we explain in more detail the different kinds of camera motion control that we train our model for.
As described in \cref{sec:pt-data} in the technical report, to enable cinematic camera motion control, we train a camera motion classifier to predict 16 different camera motion types. The predictions from this classifier are prefixed to the training captions.
The 16 camera motion control types are: zoom in, zoom out, push in, pull out, pan right, pan left, truck right, truck left, tilt up, tilt down, pedestal up, pedestal down, arc shot, tracking shot, static shot, and handheld shot.
As detailed in~\cref{sec:t2v_post_training}, during supervised finetuning, we label 6 additional camera motion and position types in the finetuning set.
These include: wide angle, close-up, aerial, low angle, over the shoulder, and first person view.

\section{Additional Evaluation Details}

\subsection{Annotation Variance Study on Text-to-Video Evaluation} \label{appendix_sec:variance}
To ensure trustworthy human annotation results and assess the significance of the winning or losing outcomes, we analyze the annotation variance across each evaluation axis.
Specifically, we repeated the same annotation tasks four times using a subset of 381 prompts for text-faithfulness, quality, and realness \& aesthetics A/B tests.
We calculate the standard deviation of the net win rate (win\% - lose\%) for each evaluation axis.
This estimation is detailed in ~\cref{tab:eval_variance}.
In~\cref{sec:t2v_main_results} we use these standard deviations to gauge the statistical significance of the results.

As shown in ~\cref{tab:eval_variance}, the overall quality axis exhibits higher variance than text-faithfulness, mainly due to the subjectivity introduced by combining different evaluation signals within the overall quality axis.
Among the quality axes, frame consistency displays a higher variance than the others, as determining which video has greater distortion is more challenging than judging which has larger or more natural overall motion.
Furthermore, realness demonstrates less variance than aesthetics, as it is generally more objective to identify \textit{generated looking} content (for the realness axis) than to align on a universally pleasing aesthetic definition (for the aesthetic axis).

  \begin{table}[h]
      \centering
      \resizebox{\linewidth}{!}{%
      \begin{tabular}{ccccccc}
          \toprule
          Text Faithfulness & Overall & Frame consistency & Motion Completeness & Motion Naturalness & Realness & Aesthetics \\
          \midrule
          3.74\% & 5.07\% & 4.08\% & 3.98\% & 1.68\% & 2.52\% & 4.84\%\\
           \bottomrule
      \end{tabular}
      }
      \caption{\textbf{Evaluation Variance.}
        Each number is acquired by sending the same A/B testing four times and compute standard deviation on the net win rate of each evaluation metric.
      }
      \label{tab:eval_variance}
  \end{table}

\subsection{T2V comparison to prior work} \label{appendix_prior_work_details}
In~\cref{sec:t2v_main_results} in the main paper, we compare to prior works for text-to-video generation.
Here, we provide extra details on how we obtain generated videos for each method and on how we post-process our generated videos to ensure fair comparison and reduce annotator bias.
The size parameters for the videos from prior work that we use for comparion are shown in~\cref{tab:t2v_competitors}.
We assume that many of the black box industry models that we compare to are being updated and improved over time.
Hence, we include the dates on which we collected the videos from website.

\begin{table}[h!]
  \centering
  \adjustbox{max width=\textwidth}{%
  \begin{tabular}{cccccc}
      \toprule
       & \multicolumn{5}{c}{Generation Specs} \\
        & \RunwayGen & \LumaLabs & \Sora & \Kling & \OURS \\
       \midrule
       Resolution  & 1280$\times$768 & 1360$\times$752 & 1920$\times$1080 & 1920$\times$1080 & 1920$\times$1080 \\
       Duration  & 10s & 5s & 10s & 10.43s & 10.67s \\
       \#Video frames  & 256 & 121 &300 & 313 & 256 \\
       FPS  & 24 & 24 & 30 & 30 & 24 \\
       \bottomrule
  \end{tabular}}
  \caption{
      We compare with the above generation specification with external works.
      \RunwayGen videos are collected on September 23th, 2024.
      \LumaLabs videos are collected on September 23th, 2024.
      \Kling videos are collected on September 25th, 2024.
      The videos released from \Sora are at a variety of different resolutions and durations.
      We note that the max resolutions of the released videos is 1920$\times$1080, whereas the majority of videos are at 1280$\times$720 and 10s in duration.
  }
  \label{tab:t2v_competitors}
\end{table}

\textbf{\Sora. }
Our only option for comparing to \Sora is by using the prompts and videos from their publicly released website (158 videos in total).
We note that for these closed source methods, the only videos released publicly on their website are likely to only represent their ``best'' samples, obtained through some unknown amount of cherry picking.
As discussed in~\cref{sec:t2v_main_results} in the main paper, for fair comparison to \Sora we hence also select samples from \OursVideo using what we consider a modest amount of cherry picking.
Specifically, for each prompt, we generate 5 different videos from \OursVideo using different random seeds.
From each of these 5 videos, we manually pick the ``best''.
The videos released by \Sora are at a variety of different resolutions.
Specifically, a small number of videos are 1080p HD, whilst the majority are at 1280$\times$720, whereas all videos from \OursVideo are 1080p HD.
For fair comparison, and to reduce annotator bias in the human evaluation, we adaptively spatially downsample our generated videos such that they are the same resolution as the corresponding \Sora video for each prompt (we do this for all prior work comparisons).
The videos released by \Sora are at a variety of different durations, with the majority being 10s, whereas all videos from \OursVideo are 10.66s.
For fair comparison, we adaptively temporally center crop either our or \Sora's videos such that they are the same duration for each prompt, whilst retaining the original frame rate.
For the \Sora comparisons, we sampled from \OursVideo using 500 linear steps.

To allow for easy comparison to \OursVideo for future work, we release non cherry picked generated videos from \OursVideo on the \textToVBenchmarkName (see~\cref{sec:t2v_eval_benchmark} in the main paper).

\subsection{Correlations between audio-based objective and subjective metrics}\label{sec:app_audio_metric_corr}
\label{sec:app-audio-corr}

We show the relationship between the objective metrics and subjective metrics presented in Section \ref{sec:audiobox_metrics}. Using around $10,000$ annotations over 53 evaluations for each of the Video-Audio Alignment and Audio Quality tasks, we explore (1) how well we can ascertain system-level net win rate on a subjective metric from the objective metrics, and (2) on an item level, how subjective scores depend on objective metrics.

For brevity we choose to focus on a single aspect of both tasks: ``Overall'' audio quality for the Audio Quality pairwise task and the ``Correctness'' aspect of the Video-SFX alignment task. For Audio quality, we observe a high degree of correlation between the aspects: ``overall'', ``professional'' and ``naturalness'' have Pearson correlation coefficients of 0.9 or higher. For the Video-SFX alignment task, the correctness and synchronization aspects have similarly high
observed correlation of 0.76.

For the Text-Alignment task, we combine annotated precision and recall into an $F_1$ score; the 1-5 scores for both are mapped to (20\%, 40\%, 60\%, 80\% and 100\%) for both precision and recall. The number of systems evaluated for the Text-Alignment task is small, so we do not consider system-level correlations.

\subsubsection{System-level correlations}
\label{sec:app-audio-system-level-corr}
\begin{figure}
    \centering
    \includegraphics[width=0.8\textwidth]{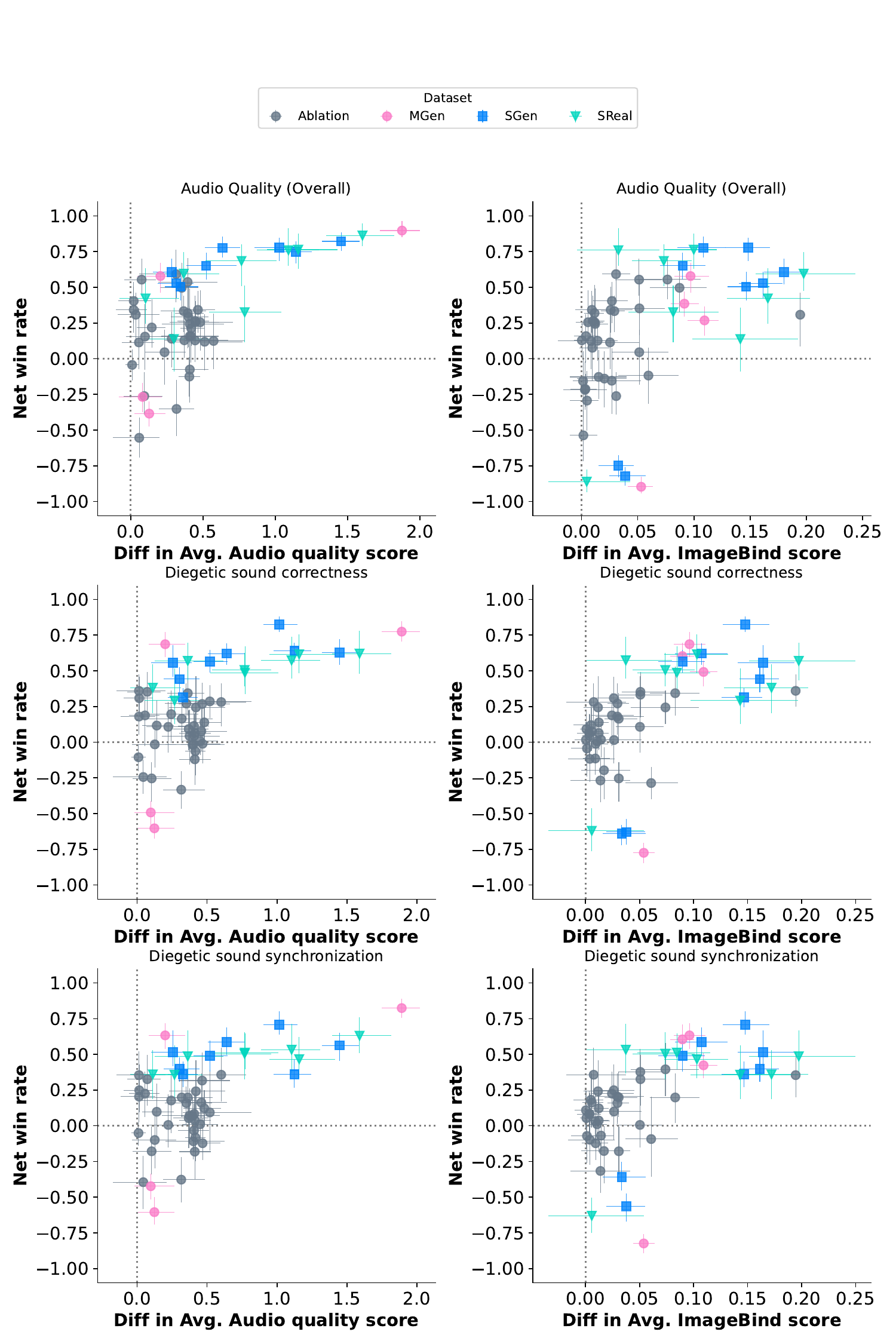}
    \caption{\label{fig:audiobox_system_corr} Comparing subjective and objective metrics at the system level. Each scatter point is a pair of models evaluated on a given dataset. The $x$-axis shows the absolute value of the mean item-level difference in objective metric and the $y$ axis shows the net win rate for the model with the higher mean item-level difference in objective scores. Grey scatter points show model ablation comparisons for \OursAudio, other points show pairwise comparisons
    between \OursAudio and an external baseline.}
\end{figure}

Figure \ref{fig:audiobox_system_corr} shows how system-level pairwise performance based on subjective evaluations (using net win rate) compares to the mean difference of the item-level objective metrics.

We find that objective metrics are predictive of subjective measures but with caveats. For instance, there is a significant amount of model-specific bias, meaning two model pairs with the same mean objective score difference may have different or opposing net win rates, which means relying solely on differences in objective metrics to make superiority claims is risky.
Pairwise comparisons of \OursAudio with external baselines (non-ablation ones) show larger net win rates on Audio Quality at a given difference in audio quality score,
which may indicate that \OursAudio improves aspects of perceived audio quality not captured by audio quality score alone.

If we rely on ablations alone and consider model pairs (33 evaluations) where differences in objective scores are statistically significant,
we find that significant differences in audio quality score correctly predict overall audio quality preferences in 21 out of 24 comparisons (87.5\% of the time, with a 95\% CI: 71.7 - 96\%).
\IB score was similarly predictive of overall audio quality preference but differences in \IB score were smaller, so only 17 out of 33 pairs had statistically significant differences in \IB scores,
and of these 82.3\% (95\% CI: 63.6\% - 94.5\%) correctly predicted preferences on overall audio quality.

\begin{figure}
    \centering
    \includegraphics[width=0.9\linewidth]{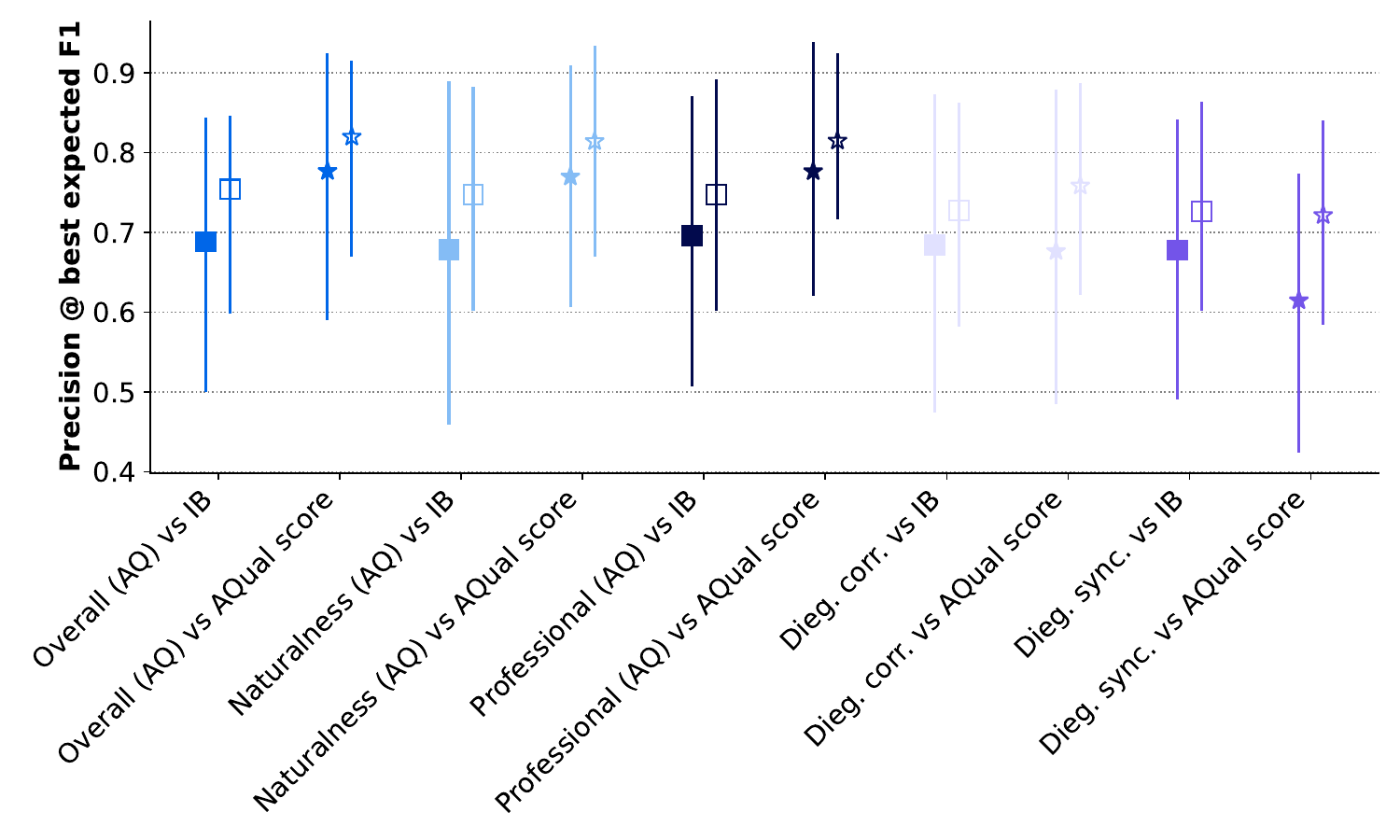}
    \caption{\label{fig:audiobox_precision} Precision at the system level: fraction of model pairs for which a given objective metric correctly predicts the human preference. 95\% CI obtained via bootstrap resampling of both items and model pairs. Filled markers represent only considering model ablations, and unfilled markers consider both model ablations and comparisons between \OursAudio and external baselines.}
\end{figure}

Figure \ref{fig:audiobox_precision} shows the precision (the fraction of model pairs where the average difference between the objective metric correctly predicts the subjective preference) for the best expected F1 score for each (objective metric, subjective metric) pairing. We also show 95\% confidence intervals obtained from bootstrap resampling of both items and model pairs. We find due to the limited sample that precision estimates are quite uncertain but that (1) audio quality score
tends to be a better predictor of audio quality preference than \IB, while both \IB and audio quality score seem to be comparably good predictors of Video-SFX Alignment aspects, and (2) both metrics are more predictive for the sample of model pairs comparing \OursAudio and an external model.

\begin{figure}
    \centering
    \includegraphics[width=0.8\textwidth]{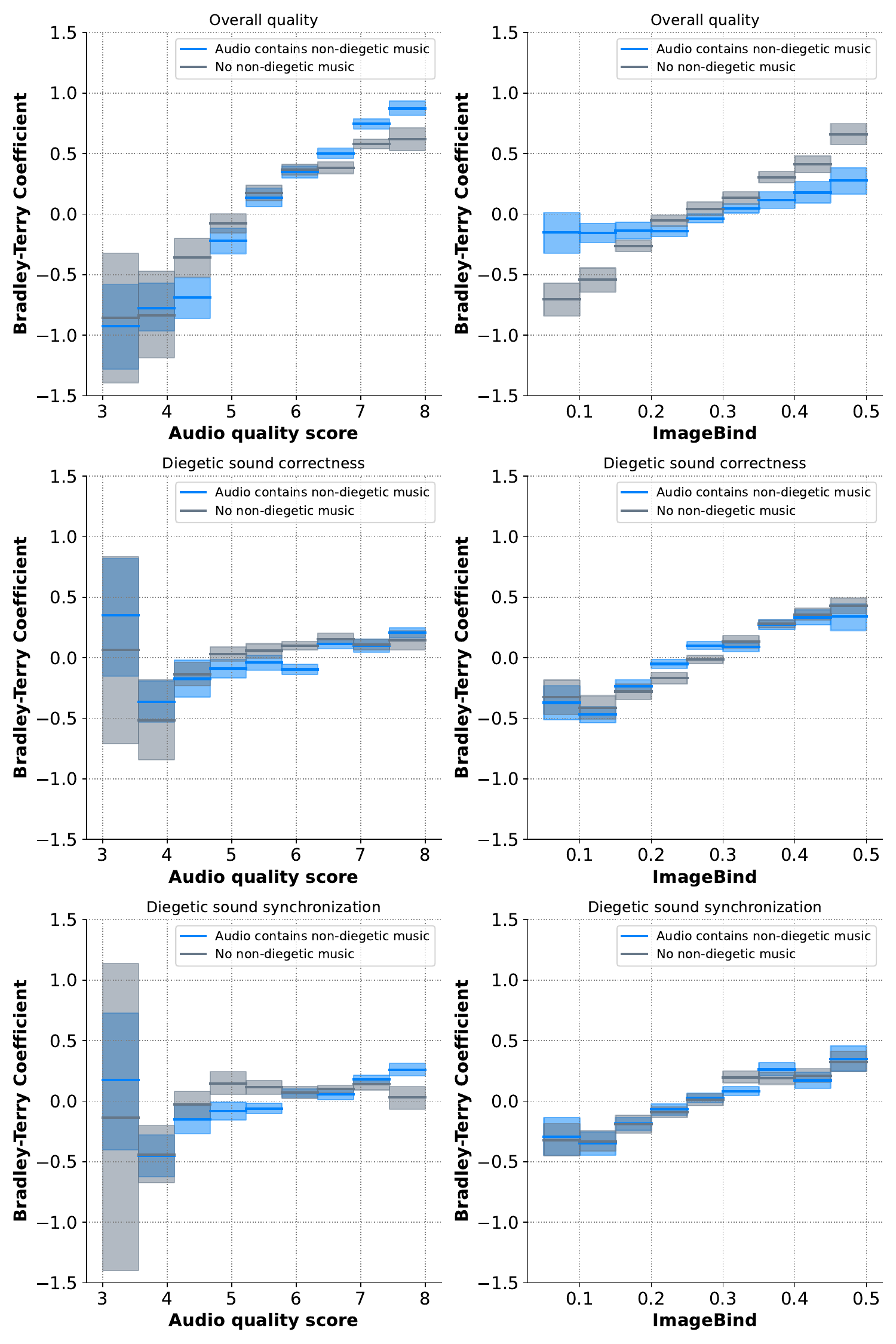}
    \caption{\label{fig:audiobox_bt_impact} We show a regression coefficient for a given subjective preference metric as a function of objective scores. The difference between the coefficients of two items indicates the log-odds of the item being preferred on a given subjective aspect.}
\end{figure}

\begin{figure}
\centering
    \includegraphics[width=0.5\textwidth]{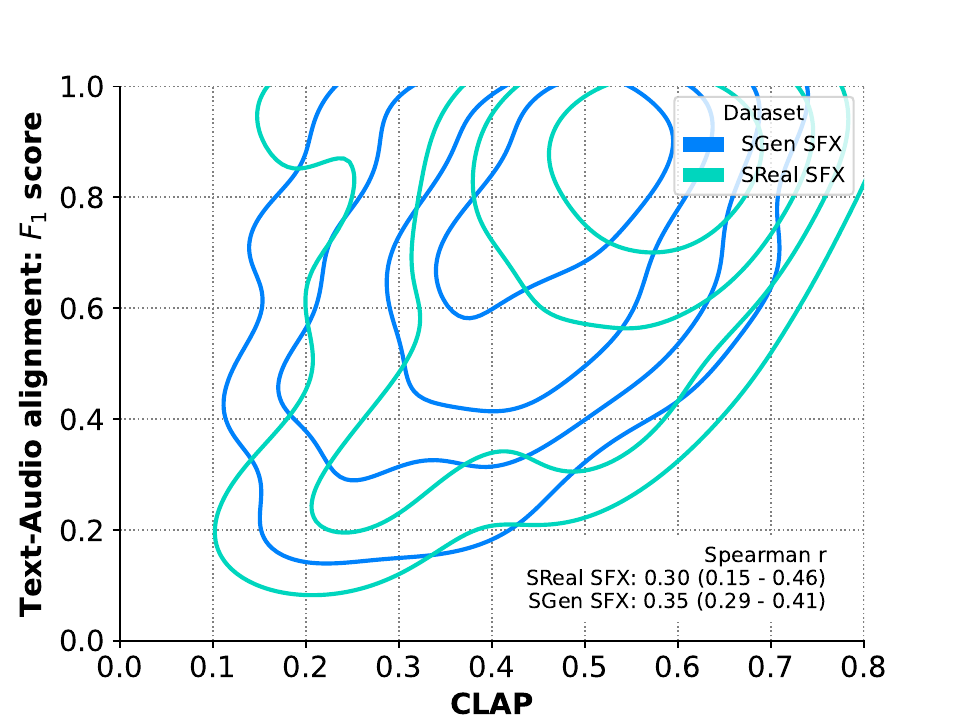}
    \caption{\label{fig:audiobox_ta_align} Above shows the correlation between consensus subjective evaluations of Text-Audio alignment and CLAP. We find moderate correlation (Spearman $r$ of $\sim 0.3$ for single-scene (real) videos and $\sim0.35$ for generated single-scene videos).}
\end{figure}

\subsubsection{Item-level correlations}
\label{sec:app-audio-item-level-corr}
Figure \ref{fig:audiobox_bt_impact} shows how the (log) odds of a model generation being preferred depend on various objective metrics. The value shown on the $y$-axis is obtained by fitting a linear model with a Bradley-Terry likelihood \citep{bradleyterry1952} on the observed subjective evaluations. We model the latent quality vector for model $m$ and item $i$ $z_{mi} = \beta^{(0)}_m + \sum_r \beta_{r,g_i} x_{mir}$ where $\beta^{(0)}_m$ is an offset parameter for each evaluated model, $g_i$
indicates the group of the $i$-th item (in this case the presence or absence of non-diegetic music), and $r$
is the index of the regression variable -- in this case the $r$-th bin of the objective metric, and $x_{mir} = 1$ when the item $i$ for model $m$ has an objective metric that falls in the $r$-th bin and $0$ otherwise. We then show $\beta_{r,g_i}$ in the $y$ axis of Figure \ref{fig:audiobox_bt_impact}. Note that for each subjective measure, we only regress on one objective metric for these visualizations.

The relative difference in the latent quality corresponds to the log-odds of one item being preferred to the other.

We make several observations

\begin{itemize}
    \item Both audio quality score and \IB score are predictive of overall audio quality, with a generation with the best observed \IB score having $\sim 4-4.5$ to 1 odds in favor of being preferred to a generation with the lowest observed \IB score.
    \item However, when non-diegetic music is present, the \IB score becomes less predictive of preferences for overall audio quality.
    \item A bad ($<$ 4) audio quality score does indicate a slightly lower odds of an item showing stronger Video-SFX alignment (``correctness'' aspect) compared to an item with a very high audio quality score ($\sim$1.5 to 1), but the impact of audio quality score improving past a $\sim4-5$ has little to no further impact.
    \item Gains in overall audio quality tend to saturate at audio quality scores above $\sim$ 6.5, but only for items without non-diegetic music. Items with non-diegetic music continue to see higher preferences for audio quality scores beyond the $\sim6.5$ point, which is also reflected in direct tests shown in Table \ref{tab:audio_abl_qual_sbj}.
    \item Increases in \IB score tend to have a stronger relative impact on overall audio quality than on Video-SFX alignment.
    \item \IB scores do tend to monotonically increase Video-SFX correctness without saturating.
    \item Diegetic synchronization does seem to be correlated with \IB like diegetic correctness, however this is likely not causal based on findings that IB score is insensitive to shifts in audio. Instead, model changes that improve correctness likely also improve synchronization, leading to a spurious correlation between synchronization and IB score.
\end{itemize}

For the Text-Audio alignment subjective evaluations, where we measure both recall and precision in 20\% increments, we report correlations with CLAP \citep{wu2023clap} scores in Figure \ref{fig:audiobox_ta_align}. We find that on the item level, there is moderate correlation. To compute confidence intervals via bootstrap we utilize a two-stage approach to also account for annotation error; we first resample evaluated items, then resample the individual annotations for each item
and compute consensus score on the bootstrap sample. We find nominally that single-scene generated videos have a higher Spearman correlation between CLAP and human evaluations compared to single-scene real videos, however this difference is not statistically significant.

\section{Additional Results}

\subsection{Additional Audio Generation Results}

\subsubsection{Additional Metrics for Main Results on Sound Effect Generation}\label{sec:app_audio_main_sfx}
We present results on objective metrics and additional subjective metrics for the ``sound effect generation'' experiments in \ref{sec:main_res}.
Table~\ref{tab:app_eval_audio_main_ss_sfx} should be compared against \cref{tab:eval_audio_main_ss_sfx}.

\begin{table}[h]
    \centering
    \adjustbox{max width=\textwidth}{%
    \begin{tabular}{lll cc ccc}
    \toprule
    \multirow{4}{*}{Dataset} & \multirow{4}{*}{Baseline} & \multirow{4}{*}{Type} & \multicolumn{2}{c}{Subjective} & \multicolumn{3}{c}{Objective} \\
    \cmidrule(lr){4-5} \cmidrule(lr){6-8}
    & & & \multicolumn{2}{c}{TA Align} & Quality & VA Align & TA Align \\
    & & & Recall (\%) $\uparrow$ & Precision (\%) $\uparrow$ & AQual $\uparrow$ & IB $\uparrow$ & CLAP $\uparrow$ \\
    \midrule
    \multirow{9}{*}{\shortstack[c]{\OursAudioBenchSReal SFX}}
    & Diff-Foley        & V2A  & - & - & 5.68 & 0.28 & 0.36* \\
    & FoleyCraft        & V2A  & - & - & 6.03 & 0.31 & 0.35* \\
    & VTA-LDM           & V2A  & - & - & 5.97 & 0.30 & 0.35* \\
    & Seeing\&Hearing   & V2A  & - & - & 5.21 & 0.38 & 0.33* \\
    & Seeing\&Hearing   & TV2A & \meanci{63.0}{6.7} & \meanci{56.6}{6.5} & 5.70 & 0.34 & 0.47\\
    & PikaLabs          & V2A  & - & - & 6.45 & 0.18 & 0.28* \\
    & PikaLabs          & TV2A & \meanci{64.5}{7.7} & \meanci{62.4}{8.3} & 6.69 & 0.21 & 0.43\\
    & ElevenLabs        & T2A  & \meanci{82.1}{5.9} & \meanci{78.8}{6.6} & 6.53 & 0.24 & 0.54\\
    \cmidrule{2-8}
    & \OursAudio        & TV2A & \meanci{84.7}{4.8} & \meanci{75.5}{5.0} & 6.76 & 0.38 & 0.51\\
    \midrule
    \multirow{9}{*}{\shortstack[c]{\OursAudioBenchSGen SFX}}
    & Diff-Foley        & V2A  & - & - & 5.48 & 0.18 & 0.17*\\
    & FoleyCraft        & V2A  & - & - & 6.03 & 0.24 & 0.25*\\
    & VTA-LDM           & V2A  & - & - & 6.03 & 0.22 & 0.25*\\
    & Seeing\&Hearing   & V2A  & - & - & 5.21 & 0.38 & 0.27*\\
    & Seeing\&Hearing   & TV2A & \meanci{72.3}{3.8} & \meanci{67.2}{3.8} & 5.49 & 0.37 & 0.42 \\
    & PikaLabs          & V2A  & - & - & 6.18 & 0.16 & 0.26*\\
    & PikaLabs          & TV2A & \meanci{63.0}{4.6} & \meanci{68.7}{5.1} & 6.20 & 0.18 & 0.41 \\
    & ElevenLabs        & T2A  & \meanci{71.0}{3.8} & \meanci{68.7}{3.9} & 6.39 & 0.19 & 0.45 \\
    \cmidrule{2-8}
    & \OursAudio        & TV2A & \meanci{73.9}{3.7} & \meanci{71.8}{3.8} & 6.47 & 0.34 & 0.46 \\
    \bottomrule
    \end{tabular}}
    \caption{
        \textbf{Additional sound effect generation subjective and objective evaluations.} This table reports additional metrics on the \OursAudioBenchSReal and \OursAudioBenchSGen SFX benchmarks. TA and VA are abbreviations for text-audio and video-audio respectively. We put ``*'' on CLAP results when text is not used at inference.
    }
    \label{tab:app_eval_audio_main_ss_sfx}
\end{table}

\begin{table}[h]
    \centering
    \adjustbox{max width=\textwidth}{%
    \begin{tabular}{lll ccc}
    \toprule
    \multirow{4}{*}{Dataset} & \multirow{4}{*}{Baseline} & \multirow{4}{*}{Type} & \multicolumn{3}{c}{Objective} \\
    \cmidrule(lr){4-6}
    & & & Quality & VA Align & TA Align \\
    & & & AQual $\uparrow$ & IB $\uparrow$ & CLAP $\uparrow$ \\
    \midrule
    \multirow{7}{*}{\shortstack[c]{\OursAudioBench SFX}}
    & Diff-Foley        & V2A  & 5.26 & 0.21 & 0.19*\\
    & FoleyCraft        & V2A  & 5.73 & 0.25 & 0.24*\\
    & VTA-LDM           & V2A  & 5.74 & 0.22 & 0.17*\\
    & Seeing\&Hearing   & V2A  & 5.21 & 0.37 & 0.25*\\
    & Seeing\&Hearing   & TV2A & 5.52 & 0.33 & 0.43 \\
    & ElevenLabs        & T2A  & 6.17 & 0.20 & 0.47 \\
    \cmidrule{2-6}
    & \OursAudio        & TV2A & 6.39 & 0.36 & 0.45 \\
    \bottomrule
    \end{tabular}}
    \caption{
        \textbf{Sound effect generation objective evaluation on \OursAudioBench.} TA and VA are abbreviations for text-audio and video-audio respectively. We put ``*'' on CLAP results when text is not used at inference. Similar trends as in \cref{tab:eval_audio_main_ss_sfx,tab:app_eval_audio_main_ss_sfx} are observed.
    }
    \label{tab:app_eval_audio_main_ss_sfx_audiobench}
\end{table}

We first note that on text-audio alignment, \OursAudio outperforms all baselines supporting text input on recall (how many sound effects described in the text caption are generated),
which is the main metric we concern for the text-audio alignment axis. It also outperforms most baselines on precision and is on par with ElevenLabs on the real videos.
It should also be noted that CLAP score between TV2A are similar, which does not correlate strongly with the relative ranking.
However, CLAP is much higher for TV2A models compared to V2A models,
which shows its discriminative power when the delta is large enough,
as we expect V2A models to generate audio that has much worse alignment with text not they are not conditioned on.

For audio quality, we find reasonable correlation between AQual and subjective pairwise comparison.
On pairwise subjective tests \OursAudio outperforms all baselines,
and PikaLabs and ElevenLabs have the smallest gaps.
On the objective metrics, \OursAudio also has the highest score, followed by those two models.

On video-SFX alignment, the correlation between objective and subjective is much weaker.
We first note that Seeing\&Hearing directly use ImageBind for classifier-guidance which maximizes that.
Therefore, both the V2A and TV2A variant show IB score on par with \OursAudio, despite that the subjective tests reveal that the video-SFX alignment is much worse compared to \OursAudio.
Next, we observe that \OursAudio achieves the higher IB score compared to other baselines, and also outperforms them on subjective evaluations.
However the ranking for the baselines in terms of relative performance to \OursAudio on subjective tests does not correlate highly with IB ranking.

Objective results on \OursAudioBench for sound effect generation are shown in \cref{tab:app_eval_audio_main_ss_sfx_audiobench}. We note that similar trends are observed, where \OursAudio leads in AQual and IB (except when compared to Seeing\&Hearing which optimizes IB in inference), and on par with baselines on CLAP.

\subsubsection{Control audio quality through text prompts}\label{app:qual_control}
\cref{fig:audio_ablation_audio_quality_sfx} and \cref{fig:audio_ablation_audio_quality_sfx_music} present examples for audio quality control via text prompts for sound effect generation and joint music and sound effect generation respectively.
When conditioned on prompts indicating low quality (\ie, quality score of 5.0), \OursAudio generates natural but low quality audio that contains for example wind noises or music with broken bass.

\begin{table}[h]
    \centering
    \captionsetup{type=figure}
    \setlength{\tabcolsep}{1pt}
    \adjustbox{max width=0.95\textwidth}{%
    \centering
    \begin{tabular}{cccccc}
        \rowcolor{blue300}
        \multicolumn{6}{c}{\textbf{Ablation: SFX audio generation with different audio quality scores prompts (10s, \OursVideo)}} \\
        \includegraphics[width=0.17\linewidth]{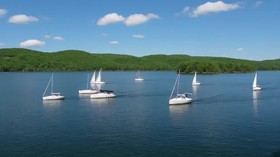}&
        \includegraphics[width=0.17\linewidth]{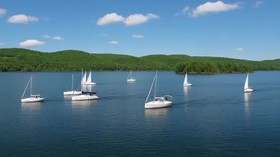}&
        \includegraphics[width=0.17\linewidth]{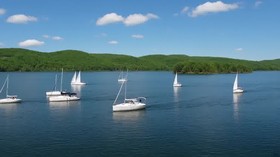}&
        \includegraphics[width=0.17\linewidth]{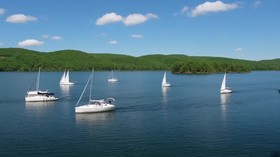}&
        \includegraphics[width=0.17\linewidth]{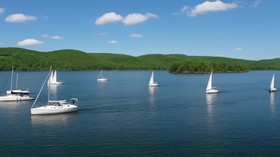}&
        \includegraphics[width=0.17\linewidth]{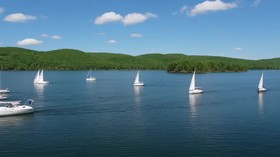}\\
        \multicolumn{6}{c}{{(a) Audio quality score = 5}} \\
        \multicolumn{6}{c}{
            \includegraphics[width=\linewidth, height=0.1\linewidth]
            {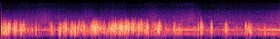}
        } \\
        \multicolumn{6}{c}{{(b) Audio quality score = 6}} \\
        \multicolumn{6}{c}{
            \includegraphics[width=\linewidth, height=0.1\linewidth]
            {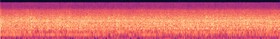}
        } \\
        \multicolumn{6}{c}{{(c) Audio quality score = 7}} \\
        \multicolumn{6}{c}{
            \includegraphics[width=\linewidth, height=0.1\linewidth]
            {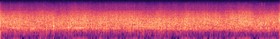}
        } \\
        \multicolumn{6}{c}{{(d) Audio quality score = 8}} \\
        \multicolumn{6}{c}{
            \includegraphics[width=\linewidth, height=0.1\linewidth]
            {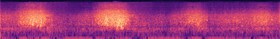}
        } \\
        \multicolumn{6}{c}{{(e) Audio quality score = 9}} \\
        \multicolumn{6}{c}{
            \includegraphics[width=\linewidth, height=0.1\linewidth]
            {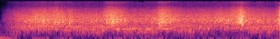}
        } \\
    \end{tabular}}
\vspace{-3mm}
\caption{\textbf{Examples of generated audio with different audio quality score in the text prompt with \OursAudio.}
For these samples, higher audio quality scores tends to produced audio without wind noises compared to lower audio quality scores.
Videos in this Figure found at \url{https://go.fb.me/MovieGen-Figure40}.
}
\label{fig:audio_ablation_audio_quality_sfx}
\end{table}

\begin{table}[h]
    \centering
    \captionsetup{type=figure}
    \setlength{\tabcolsep}{1pt}
    \adjustbox{max width=0.95\textwidth}{%
    \centering
    \begin{tabular}{cccccc}
        \rowcolor{blue300}
        \multicolumn{6}{c}{\textbf{Ablation: SFX+music audio generation with different audio quality scores prompts (10s, \OursVideo)}} \\
        \includegraphics[width=0.17\linewidth]{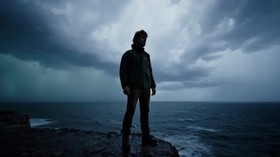}& 
        \includegraphics[width=0.17\linewidth]{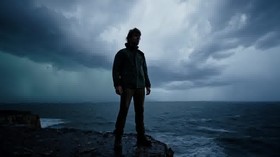}& 
        \includegraphics[width=0.17\linewidth]{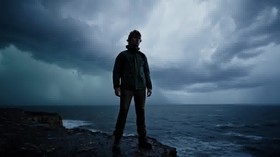}& 
        \includegraphics[width=0.17\linewidth]{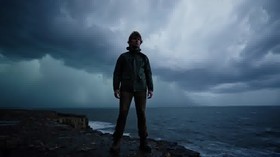}& 
        \includegraphics[width=0.17\linewidth]{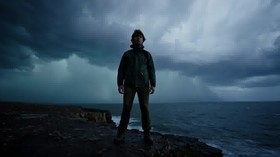}& 
        \includegraphics[width=0.17\linewidth]{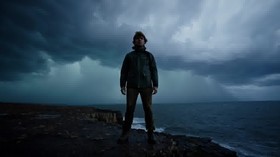}\\
        \multicolumn{6}{c}{{(a) Audio quality score = 5}} \\
        \multicolumn{6}{c}{
            \includegraphics[width=\linewidth, height=0.1\linewidth]
            {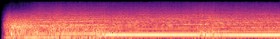}
        } \\
        \multicolumn{6}{c}{{(b) Audio quality score = 6}} \\
        \multicolumn{6}{c}{
            \includegraphics[width=\linewidth, height=0.1\linewidth]
            {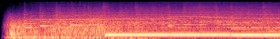}
        } \\
        \multicolumn{6}{c}{{(c) Audio quality score = 7}} \\
        \multicolumn{6}{c}{
            \includegraphics[width=\linewidth, height=0.1\linewidth]
            {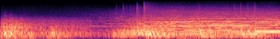}
        } \\
        \multicolumn{6}{c}{{(d) Audio quality score = 8}} \\
        \multicolumn{6}{c}{
            \includegraphics[width=\linewidth, height=0.1\linewidth]
            {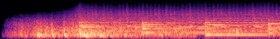}
        } \\
        \multicolumn{6}{c}{{(e) Audio quality score = 9}} \\
        \multicolumn{6}{c}{
            \includegraphics[width=\linewidth, height=0.1\linewidth]
            {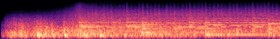}
        } \\
    \end{tabular}}

    \vspace{-3mm}

    \caption{\textbf{Examples of generated audio with different audio quality score in the text prompt with \OursAudio.}
    \OursAudio controls the generated audio output quality using the input text prompts (refer to \ref{tab:audio_caption}). As we can observed in the spectrogram plot, lower quality scores contains some high frequency noises and by gradually increasing the quality scores, we got a cleaner audio spectrogram.
    Videos in this Figure found at \url{https://go.fb.me/MovieGen-Figure41}.
}
\label{fig:audio_ablation_audio_quality_sfx_music}
\end{table}

\subsubsection{Control music style through text prompts}\label{app:music_control}
\cref{fig:audio_ablation_text_music} shows an example of varying music captions for the same video.

\begin{table}[h]
    \centering
    \captionsetup{type=figure}
    \setlength{\tabcolsep}{1pt}
    \adjustbox{max width=0.95\textwidth}{%
    \centering
    \begin{tabular}{cccccc}
        \rowcolor{blue300}
        \multicolumn{6}{c}{\textbf{Ablation: SFX+music audio generation with different text prompts (10s, \OursVideo)}} \\
        \includegraphics[width=0.17\linewidth]{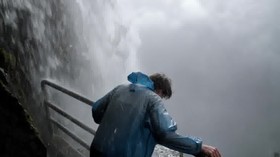}& 
        \includegraphics[width=0.17\linewidth]{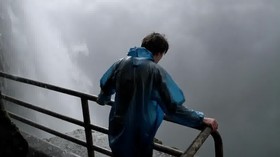}& 
        \includegraphics[width=0.17\linewidth]{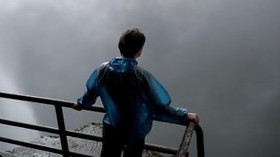}& 
        \includegraphics[width=0.17\linewidth]{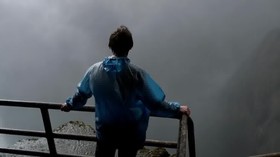}& 
        \includegraphics[width=0.17\linewidth]{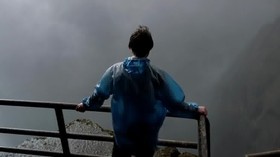}& 
        \includegraphics[width=0.17\linewidth]{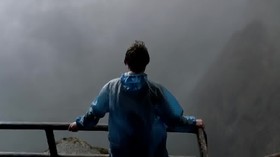}\\
        \multicolumn{6}{c}{{(a) Music style caption: A beautiful and emotional classical orchestral track with a cinematic feel.}} \\
        \multicolumn{6}{c}{
            \includegraphics[width=\linewidth, height=0.05\linewidth]
            {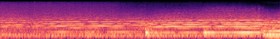}
        } \\
        \multicolumn{6}{c}{{(b) Music style caption: A dark and mysterious electronic track with a haunting melody.}} \\
        \multicolumn{6}{c}{
            \includegraphics[width=\linewidth, height=0.05\linewidth]
            {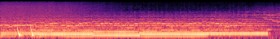}
        } \\
        \multicolumn{6}{c}{{(c) Music style caption: A slow, emotional, and passionate country song.}} \\
        \multicolumn{6}{c}{
            \includegraphics[width=\linewidth, height=0.05\linewidth]
            {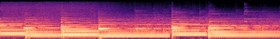}
        } \\
    \end{tabular}}
\vspace{-3mm}
\caption{\textbf{Examples of generated audio with different music description in the text prompt with \OursAudio.}
Videos in this Figure found at \url{https://go.fb.me/MovieGen-Figure42}.
}
\label{fig:audio_ablation_text_music}
\end{table}

\subsubsection{Additional Audio Samples}\label{app:audio_samples}
\cref{fig:audio_more_sfx} and \cref{fig:audio_more_sfx_music} presents additional samples of sound effect generation and joint SFX + music generation, respectively.

\begin{table}[h]
    \centering
    \captionsetup{type=figure}
    \setlength{\tabcolsep}{1pt}
    \adjustbox{max width=0.95\textwidth}{%
    \centering
    \begin{tabular}{cccc}
        \rowcolor{blue300}
        \multicolumn{4}{c}{\textbf{Appendix: additional sound effect samples}} \\

        \includegraphics[width=0.25\linewidth]{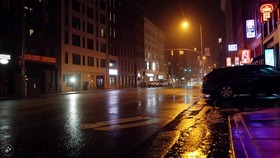} &
        \includegraphics[width=0.25\linewidth]{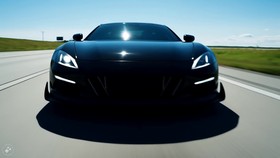} &
        \includegraphics[width=0.25\linewidth]{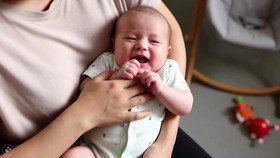} &
        \includegraphics[width=0.25\linewidth]{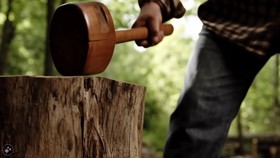} \\
        
        \multicolumn{1}{c}{
            \includegraphics[width=0.25\linewidth, height=0.05\linewidth]{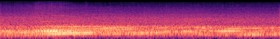}
        }  &
        \multicolumn{1}{c}{
            \includegraphics[width=0.25\linewidth, height=0.05\linewidth]{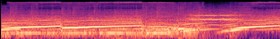}
        }  &
        \multicolumn{1}{c}{
            \includegraphics[width=0.25\linewidth, height=0.05\linewidth]{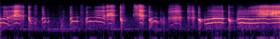}
        }  &
        \multicolumn{1}{c}{
            \includegraphics[width=0.25\linewidth, height=0.05\linewidth]{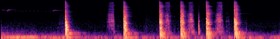}
        }  \\
        
        \includegraphics[width=0.25\linewidth]{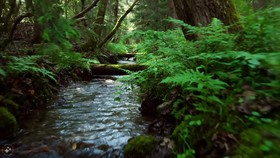} &
        \includegraphics[width=0.25\linewidth]{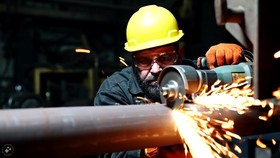} &
        \includegraphics[width=0.25\linewidth]{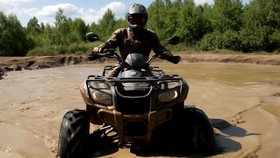} &
        \includegraphics[width=0.25\linewidth]{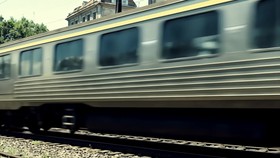} \\
        
        \multicolumn{1}{c}{
            \includegraphics[width=0.25\linewidth, height=0.05\linewidth]{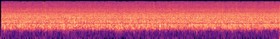}
        }  &
        \multicolumn{1}{c}{
            \includegraphics[width=0.25\linewidth, height=0.05\linewidth]{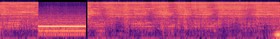}
        }  &
        \multicolumn{1}{c}{
            \includegraphics[width=0.25\linewidth, height=0.05\linewidth]{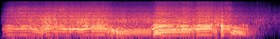}
        }  &
        \multicolumn{1}{c}{
            \includegraphics[width=0.25\linewidth, height=0.05\linewidth]{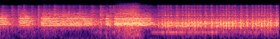}
        }  \\
        
        \includegraphics[width=0.25\linewidth]{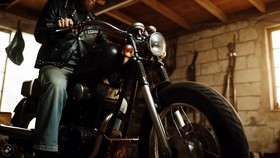} &
        \includegraphics[width=0.25\linewidth]{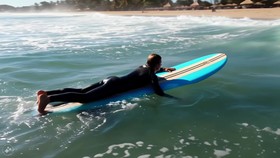} &
        \includegraphics[width=0.25\linewidth]{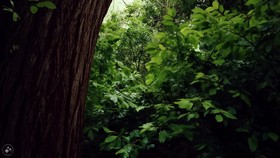} &
        \includegraphics[width=0.25\linewidth]{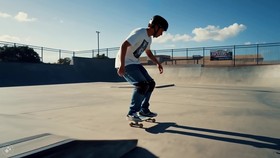} \\
        
        \multicolumn{1}{c}{
            \includegraphics[width=0.25\linewidth, height=0.05\linewidth]{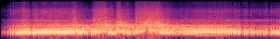}
        }  &
        \multicolumn{1}{c}{
            \includegraphics[width=0.25\linewidth, height=0.05\linewidth]{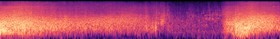}
        }  &
        \multicolumn{1}{c}{
            \includegraphics[width=0.25\linewidth, height=0.05\linewidth]{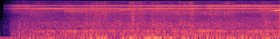}
        }  &
        \multicolumn{1}{c}{
            \includegraphics[width=0.25\linewidth, height=0.05\linewidth]{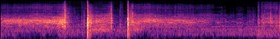}
        }  \\
        
        \includegraphics[width=0.25\linewidth]{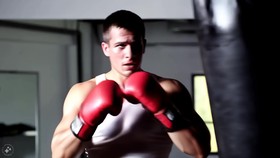} &
        \includegraphics[width=0.25\linewidth]{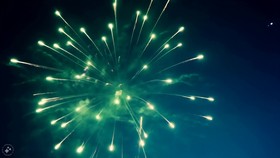} \\
        
        \multicolumn{1}{c}{
            \includegraphics[width=0.25\linewidth, height=0.05\linewidth]{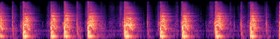}
        }  &
        \multicolumn{1}{c}{
            \includegraphics[width=0.25\linewidth, height=0.05\linewidth]{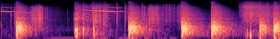}
        }  \\
    \end{tabular}}
\vspace{-3mm}
\caption{\textbf{Additional \OursAudio sound effect samples.}
Videos in this Figure found at \url{https://go.fb.me/MovieGen-Figure43}.
}
\label{fig:audio_more_sfx}
\end{table}

\begin{table}[h]
    \centering
    \captionsetup{type=figure}
    \setlength{\tabcolsep}{1pt}
    \adjustbox{max width=0.95\textwidth}{%
    \centering
    \begin{tabular}{cccc}
        \rowcolor{blue300}
        \multicolumn{4}{c}{\textbf{Appendix: additional sound effect + music samples}} \\
        
        \includegraphics[width=0.25\linewidth]{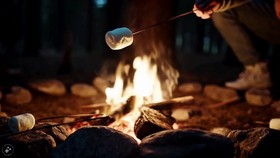} &
        \includegraphics[width=0.25\linewidth]{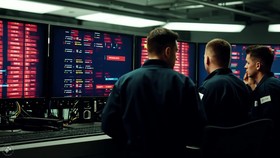} &
        \includegraphics[width=0.25\linewidth]{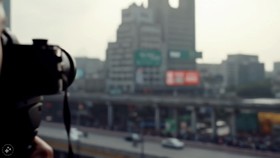} &
        \includegraphics[width=0.25\linewidth]{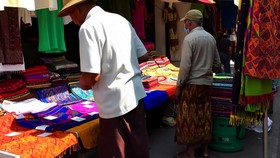}\\
        
        \multicolumn{1}{c}{
            \includegraphics[width=0.25\linewidth, height=0.05\linewidth]{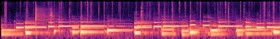}
        }  &
        \multicolumn{1}{c}{
            \includegraphics[width=0.25\linewidth, height=0.05\linewidth]{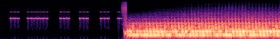}
        }  &
        \multicolumn{1}{c}{
            \includegraphics[width=0.25\linewidth, height=0.05\linewidth]{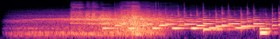}
        }  &
        \multicolumn{1}{c}{
            \includegraphics[width=0.25\linewidth, height=0.05\linewidth]{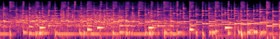}
        } \\
        
        \includegraphics[width=0.25\linewidth]{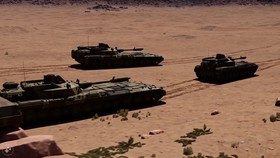} &
        \includegraphics[width=0.25\linewidth]{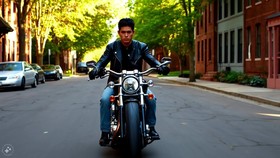} &
        \includegraphics[width=0.25\linewidth]{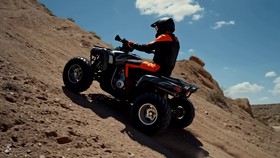} &
        \includegraphics[width=0.25\linewidth]{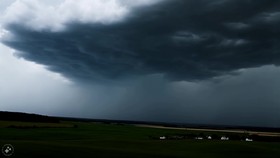}\\
        
        \multicolumn{1}{c}{
            \includegraphics[width=0.25\linewidth, height=0.05\linewidth]{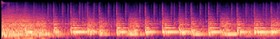}
        }  &
        \multicolumn{1}{c}{
            \includegraphics[width=0.25\linewidth, height=0.05\linewidth]{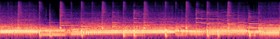}
        }  &
        \multicolumn{1}{c}{
            \includegraphics[width=0.25\linewidth, height=0.05\linewidth]{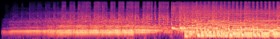}
        }  &
        \multicolumn{1}{c}{
            \includegraphics[width=0.25\linewidth, height=0.05\linewidth]{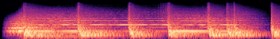}
        }  \\
        
        \includegraphics[width=0.25\linewidth]{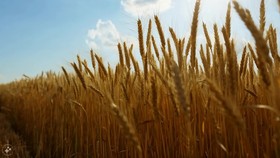} &
        \includegraphics[width=0.25\linewidth]{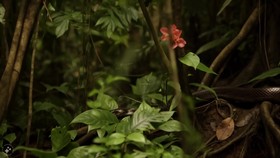} &
        \includegraphics[width=0.25\linewidth]{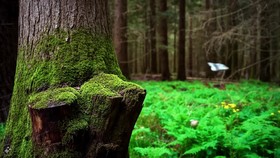} &
        \includegraphics[width=0.25\linewidth]{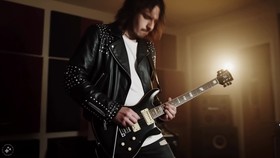}\\
        
        \multicolumn{1}{c}{
            \includegraphics[width=0.25\linewidth, height=0.05\linewidth]{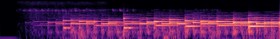}
        }  &
        \multicolumn{1}{c}{
            \includegraphics[width=0.25\linewidth, height=0.05\linewidth]{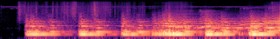}
        }  &
        \multicolumn{1}{c}{
            \includegraphics[width=0.25\linewidth, height=0.05\linewidth]{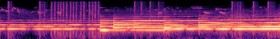}
        }  &
        \multicolumn{1}{c}{
            \includegraphics[width=0.25\linewidth, height=0.05\linewidth]{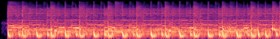}
        } \\
        
        \includegraphics[width=0.25\linewidth]{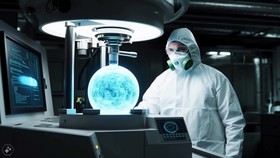} &
        \includegraphics[width=0.25\linewidth]{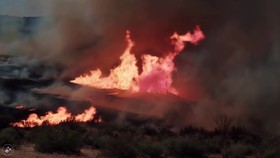} &
        \includegraphics[width=0.25\linewidth]{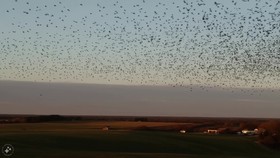}\\
        
        \multicolumn{1}{c}{
            \includegraphics[width=0.25\linewidth, height=0.05\linewidth]{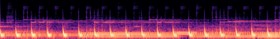}
        }  &
        \multicolumn{1}{c}{
            \includegraphics[width=0.25\linewidth, height=0.05\linewidth]{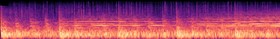}
        }  &
        \multicolumn{1}{c}{
            \includegraphics[width=0.25\linewidth, height=0.05\linewidth]{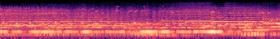}
        } \\

    \end{tabular}}
\vspace{-3mm}
\caption{\textbf{Additional \OursAudio sound effect and music samples.}
Videos in this Figure found at \url{https://go.fb.me/MovieGen-Figure44}.
}
\label{fig:audio_more_sfx_music}
\end{table}

\clearpage
\subsection{Additional Results from \OursVideo}
Here, we include some further example generations from \OursVideo in~\cref{fig:t2v_qual_ours_figure_app1,fig:t2v_qual_ours_figure_app2}

\begin{center}
    \centering
    \captionsetup{type=figure}
    \setlength{\tabcolsep}{1pt}
\adjustbox{max width=\textwidth}{%
\centering
\begin{tabular}{cccc}
    \multicolumn{4}{c}{\textit{Prompt}: A person harvesting clouds from a field, placing them in a basket.} \\
    \includegraphics[width=0.25\linewidth]{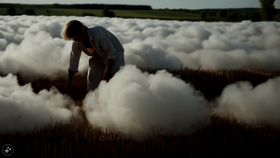} &
    \includegraphics[width=0.25\linewidth]{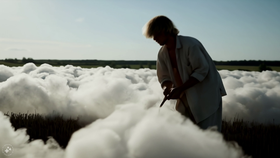} &
    \includegraphics[width=0.25\linewidth]{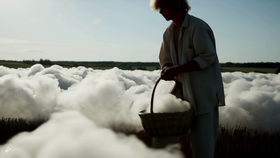} &
    \includegraphics[width=0.25\linewidth]{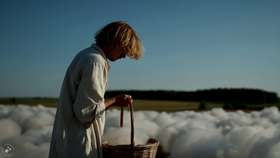} \\

    \multicolumn{4}{c}{\textit{Prompt}: cinematic trailer for a group of samoyed puppies learning to become chefs.} \\
    \includegraphics[width=0.25\linewidth]{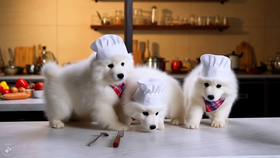} &
    \includegraphics[width=0.25\linewidth]{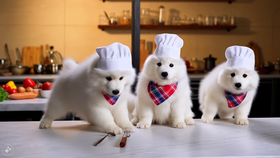} &
    \includegraphics[width=0.25\linewidth]{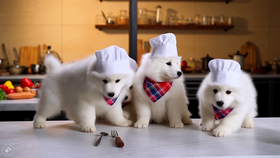} &
    \includegraphics[width=0.25\linewidth]{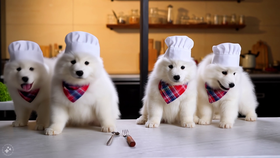} \\

    \multicolumn{4}{c}{\textit{Prompt}: A monk meditating in a temple carved into the cliffs of Bhutan.} \\
    \includegraphics[width=0.25\linewidth]{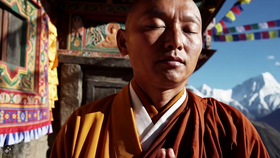} &
    \includegraphics[width=0.25\linewidth]{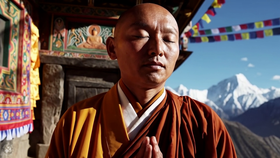} &
    \includegraphics[width=0.25\linewidth]{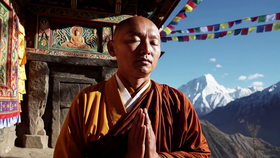} &
    \includegraphics[width=0.25\linewidth]{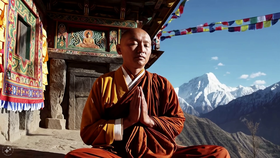} \\

    \multicolumn{4}{c}{\textit{Prompt}: Rivers of lava flow through a landscape of ice and snow, with steam rising into the air.} \\
    \includegraphics[width=0.25\linewidth]{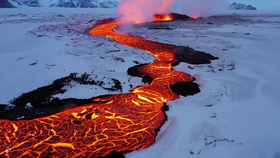} &
    \includegraphics[width=0.25\linewidth]{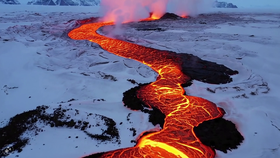} &
    \includegraphics[width=0.25\linewidth]{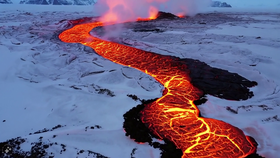} &
    \includegraphics[width=0.25\linewidth]{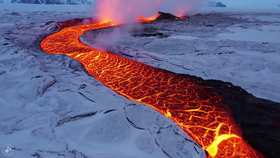} \\

    \multicolumn{4}{c}{\textit{Prompt}: A young explorer who discovers a cave filled with glowing crystals.} \\
    \includegraphics[width=0.25\linewidth]{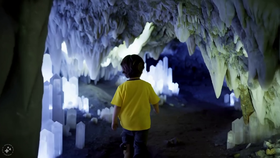} &
    \includegraphics[width=0.25\linewidth]{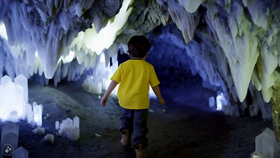} &
    \includegraphics[width=0.25\linewidth]{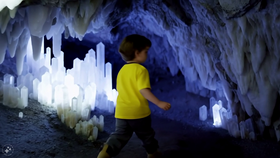} &
    \includegraphics[width=0.25\linewidth]{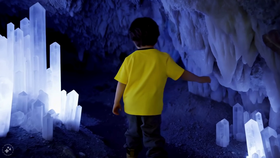} \\

    \multicolumn{4}{c}{\textit{Prompt}: A potter crafting ceramics using volcanic ash in Hawaii.} \\
    \includegraphics[width=0.25\linewidth]{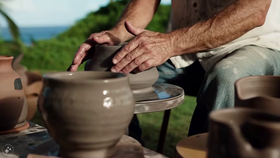} &
    \includegraphics[width=0.25\linewidth]{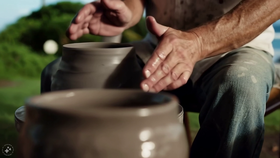} &
    \includegraphics[width=0.25\linewidth]{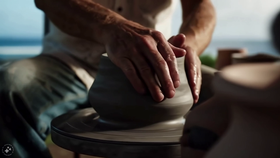} &
    \includegraphics[width=0.25\linewidth]{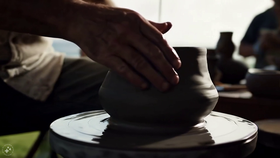} \\

\end{tabular}}

    \vspace{-3mm}
    \caption{\textbf{Generated videos from \OursVideo.} 
    Videos in this Figure found at \url{https://go.fb.me/MovieGen-Figure45}.}
    \label{fig:t2v_qual_ours_figure_app1}
\end{center}%

\begin{center}
    \centering
    \captionsetup{type=figure}
    \setlength{\tabcolsep}{1pt}
\adjustbox{max width=\textwidth}{%
\centering
\begin{tabular}{cccc}
    \multicolumn{4}{c}{\textit{Prompt}: A dense jungle pathway is illuminated by oversized, bioluminescent mushrooms.} \\
    \includegraphics[width=0.25\linewidth]{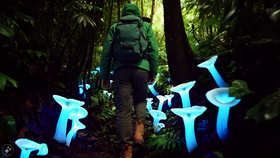} &
    \includegraphics[width=0.25\linewidth]{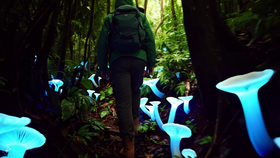} &
    \includegraphics[width=0.25\linewidth]{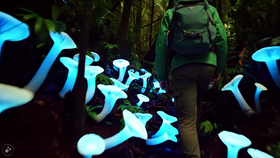} &
    \includegraphics[width=0.25\linewidth]{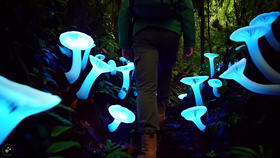} \\

    \multicolumn{4}{c}{\textit{Prompt}: A turtle in a racing suit, riding a skateboard down a steep hill.} \\
    \includegraphics[width=0.25\linewidth]{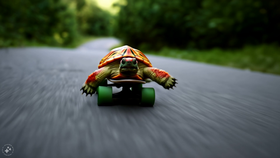} &
    \includegraphics[width=0.25\linewidth]{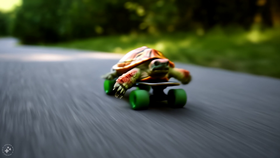} &
    \includegraphics[width=0.25\linewidth]{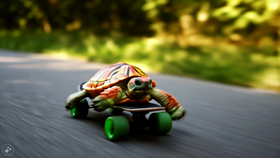} &
    \includegraphics[width=0.25\linewidth]{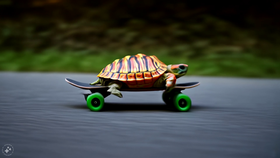} \\

    \multicolumn{4}{c}{\textit{Prompt}:a woman eating ice scream.} \\
    \includegraphics[width=0.25\linewidth]{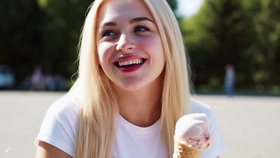} &
    \includegraphics[width=0.25\linewidth]{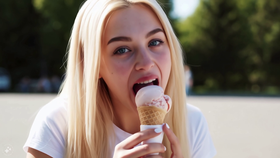} &
    \includegraphics[width=0.25\linewidth]{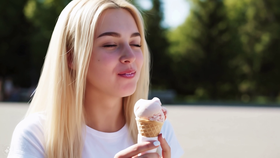} &
    \includegraphics[width=0.25\linewidth]{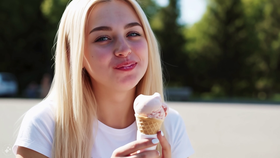} \\

    \multicolumn{4}{c}{\textit{Prompt}: Sailboat sailing through the crystal-clear waters of Bora Bora. Camera aerial shot.} \\
    \includegraphics[width=0.25\linewidth]{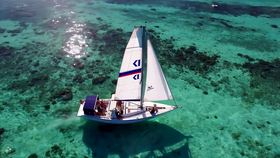} &
    \includegraphics[width=0.25\linewidth]{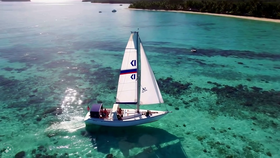} &
    \includegraphics[width=0.25\linewidth]{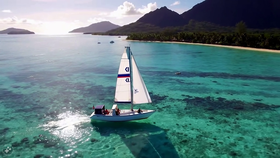} &
    \includegraphics[width=0.25\linewidth]{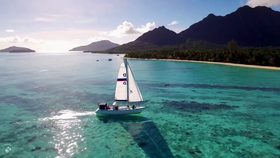} \\

    \multicolumn{4}{c}{\textit{Prompt}: A young detective who solves the case of the glowing plants.} \\
    \includegraphics[width=0.25\linewidth]{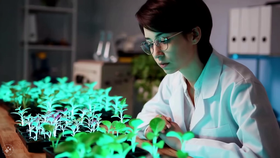} &
    \includegraphics[width=0.25\linewidth]{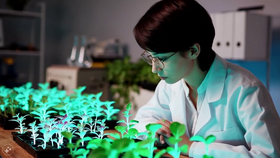} &
    \includegraphics[width=0.25\linewidth]{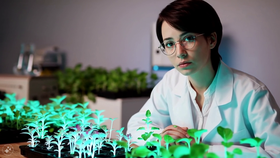} &
    \includegraphics[width=0.25\linewidth]{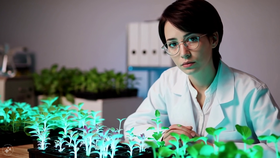} \\

    \multicolumn{4}{c}{\textit{Prompt}: Giant panda riding a bike through the streets of Beijing. Camera tracking shot.} \\
    \includegraphics[width=0.25\linewidth]{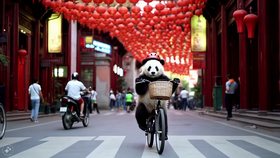} &
    \includegraphics[width=0.25\linewidth]{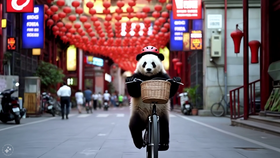} &
    \includegraphics[width=0.25\linewidth]{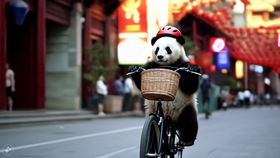} &
    \includegraphics[width=0.25\linewidth]{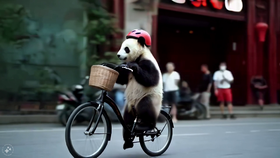} \\

\end{tabular}}

    \vspace{-3mm}
    \caption{\textbf{Generated videos from \OursVideo.} 
    Videos in this Figure found at \url{https://go.fb.me/MovieGen-Figure46}.}
    \label{fig:t2v_qual_ours_figure_app2}
\end{center}%

\end{document}